\documentclass[twoside,11pt]{article}

\usepackage{jmlr2e}
\usepackage{listings}
\usepackage{subcaption}

\usepackage{rotating}

\usepackage{amsmath}
\usepackage{float}

\ewrlheading{14}{2018}{October 2018, Lille, France}{P. Henderson, J. Romoff, \& J. Pineau}

\ShortHeadings{Where did my optimum go?}{Peter Henderson, Joshua Romoff, Joelle Pineau}
\firstpageno{1}

\begin{document}

\title{Where Did My Optimum Go?: An Empirical Analysis of Gradient Descent Optimization in Policy Gradient Methods}

\author{\name Peter Henderson
\email phend@stanford.edu\\
\addr Stanford University
\AND
\name Joshua Romoff
\email joshua.romoff@mail.mcgill.ca\\
\addr MILA, McGill University, Facebook
\AND
\name Joelle Pineau
\email jpineau@cs.mcgill.ca\\
\addr MILA, McGill University, Facebook
}

\maketitle

\begin{abstract}%
Recent analyses of certain gradient descent optimization methods have shown that performance can degrade in some settings -- such as with stochasticity or implicit momentum. In deep reinforcement learning (Deep RL), such optimization methods are often used for training neural networks via the temporal difference error or policy gradient. As an agent improves over time, the optimization target changes and thus the loss landscape (and local optima) change. Due to 
the failure modes of those methods,
the ideal choice of optimizer for Deep RL remains unclear.
As such, we provide an empirical analysis of the effects that a wide range of gradient descent optimizers and their hyperparameters have on policy gradient methods, a subset of Deep RL algorithms, for benchmark continuous control tasks.
We find that adaptive optimizers have a narrow window of effective learning rates, diverging in other cases, and that the effectiveness of momentum varies depending on the properties of the environment. 
Our analysis suggests that there is significant interplay between the dynamics of the environment and Deep RL algorithm properties which aren't necessarily accounted for by traditional adaptive gradient methods. 
We provide suggestions for optimal settings of current methods and further lines of research based on our findings.
\end{abstract}


\section{Introduction}
\label{sec:intro}
Deep reinforcement learning (Deep RL) algorithms often rely on the same stochastic gradient descent methods as other deep learning techniques to optimize value function approximators and policies~\citep{mnih2015human,PPO}. For example, the optimizer used for Proximal Policy Optimization by \citet{PPO} is Adam~\citep{kingma2014adam}, while the optimizer for Advantage Actor Critic in \citet{PPO} is RMSProp~\citep{RMSProp}. However, it is unclear why different optimizers may have different behaviours in Deep RL algorithms and whether the theoretical properties of optimizers in the supervised learning setting generalize to Deep RL algorithms.

To initially illustrate some of the possible problems with the adaptive gradient descent optimization in Deep RL, we examine recent literature. First, recent work has discovered that in online stochastic settings (as is often the case in Deep RL), there are cases where Adam does not converge to an optimal solution~\citep{amsgrad}.
~\citet{wilson2017marginal} demonstrate similar findings for adaptive methods such as Adam and RMSProp. It is clear from prior work~\citep{henderson2017deep,islam2017reproducibility} that reinforcement learning methods can be highly stochastic and variant in nature. As such the problem formulation may result exactly in the cases described by~\citet{amsgrad} and \citet{wilson2017marginal}.

Furthermore, we note recent work in optimization which suggests that momentum needs to be adjusted to compensate for implicit momentum generated by the system~\citep{mitliagkas2016asynchrony}. That is, in asynchronous optimization itself adds an additional implicit momentum. Once again, a parallel issue can easily be mapped to asynchronous Deep RL methods \citep{a3c} or distributed learning~\citep{heess2017emergence,barth2018distributed}. However, more importantly, it is easy to imagine how this might affect synchronous methods as well. In TD methods especially, there is a staleness to the gradients since the value function is bootstrapping off of its own predictions and updates may be biased toward previous policies and points in a trajectory.

To this extent, we focus on two Deep RL algorithms from the family of policy gradient methods~\citep{sutton2000policy} -- synchronous Advantage Actor Critic (A2C) \citep{PPO,a3c} and Proximal Policy Optimization (PPO)~\citep{PPO} -- to examine empirically what effects certain optimizers and learning rates may have on learning on policy gradient methods and what effect momentum may have in learning (suggesting possible sources of implicit momentum).

\section{Background}

\subsection{Deep RL Algorithms}
We focus on two Deep RL algorithms from the class of policy gradient methods: PPO and A2C. Each has a unique set of properties, but are generally very similar. In both cases, a value function approximator is learned via a temporal difference (TD) update loss~\citep{sutton1988learning}: $\mathcal{L}(\theta_V) =  \mathbb{E} \left[ \left(Y_t - V_\gamma^\pi (s_t; \theta_V) \right)^2 \right]$, where $Y_t = \sum_{n=0}^{N-1} \gamma ^n r_{t+n} + \gamma^N V_\gamma^\pi(s_{t+n+1})$. In the case of A2C, generally several workers use a policy to synchronously collect a small number of samples in the environment before updating a stochastic parameterized policy ($\pi_\theta(a | s)$). The policy reuses the TD error (the advantage) through a policy gradient update. PPO is similar, except that generally it uses longer Monte Carlo rollouts as in REINFORCE~\citep{williams1992simple}, relying on the value function for its variance reduction baseline properties rather than an estimate of the value in the gradient update. Furthermore, PPO imposes an adaptive trust region on the policy update to prevent the updated policy from straying too far from the prior one. In our case, we use a clipping objective which is proposed by~\citet{PPO} such that the policy is update via the loss $L^{CLIP} (\theta) = \hat{\mathbb{E}} \left[ \min ( r_t(\theta) \hat{A_t}, \text{clip} (r_t (\theta), 1- \epsilon, 1+\epsilon)\hat{A_t} \right],$ where the likelihood ratio is $r_t(\theta) = \frac{\pi_\theta (a_t | s_t)}{\pi_{\theta_{old}}(a_t|s_t)}$, $\hat{A_t}$ is the generalized advantage function~\citep{schulman2015high}, and $\epsilon < 1$ is some small factor to constrain the update.

\subsection{Gradient Optimization Methods}

We consider several gradient descent-based optimization methods in our analysis: stochastic gradient descent (SGD), SGD with Nesterov momentum (SGDNM)~\citep{nesterov1983method}, Averaged Stochastic Gradient Descent (ASGD)~\citep{polyak1992acceleration}, Adagrad~\citep{duchi2011adaptive}, Adadelta~\citep{zeiler2012adadelta}, RMSProp~\citep{RMSProp}, Adam~\citep{kingma2014adam}, AMSGrad~\citep{amsgrad}, Adamax~\citep{kingma2014adam}, and YellowFin~\citep{zhang2017yellowfin}. 

In brief, all of these algorithms represent some variation of SGD with differing updates to the learning rate and modifications of the gradient. \citet{ruder2016overview} provides a more detailed overview of these methods, but we briefly describe the variations on SGD here. Momentum can be thought of as accelerating an update in the relevant direction with Nesterov momentum correcting the acceleration upon reaching the next step. ASGD keeps a running average of the parameters $\theta$ and then uses the averaged value for the final result. 
Overall, the family of adaptive algorithms (Adagrad, Adadelta, RMSProp, Adam, Adamax, AMSGrad, and Adamax) can be thought of as performing a diagonal rescaling of the gradient updates.
Adagrad uses a per parameter learning rate instead of a single learning rate, adapting each such that sparse parameters see an increased learning rate by accumulating past gradients. Adadelta is a similar algorithm which attempts to compensate for the monotonically decreasing learning rate resulting from Adagrad's gradient accumulation via a window of gradient accumulation. 
RMSProp was similarly developed to account for the same issue in Adagrad. 
Adam also uses per parameter adaptive learning rates, but additionally keeps an exponentially decaying average of prior gradient updates. It uses this in a similar fashion to momentum. As~\citet{heusel2017gans} state, Adam can be described as a ``Heavy Ball with Friction'' such that it ``typically overshoots small local minima that correspond
to mode collapse and can find flat minima which
generalize well''\footnote{We note that this property is particularly relevant to RL. In online learning and with a shifting loss landscape particularly during exploration-heavy phases, a flat minimum seems unlikely in RL settings.}. Adamax is another variant of Adam which uses the $L_\infty$ norm instead of the $L_2$ norm when scaling gradients. AMSGrad is an update to Adam which claims to improve convergence properties in certain stochastic settings by using the maximum of the past gradients rather than the average. YellowFin attempts to solve the notion of implicit momentum by using an active controller to tune the hyperparameters of momentum SGD. 

\section{Analysis}

We investigate the effects of gradient descent optimization on the family of policy gradient Deep RL algorithms, focusing on the benchmark suite of continuous control tasks (environments) provided by OpenAI Gym~\citep{brockman2016openai}. 
These tasks yield different dynamics as described by~\citet{henderson2017deep}, while also being small enough to run a large suite of experiments efficiently.
We use implementations from~\citet{pytorchrl} for A2C and PPO, modifying the codebase to replace the optimizer. We use the default set of hyperparameters provided by each optimizer except for varying the learning rate as discussed in Section~\ref{sec:lrs} and momentum in Section~\ref{sec:momemtum}. We run 10 random seeds for all experiments and further describe our codebase, setup, and hyperparameters in Appendix~\ref{app:setup}.

\subsection{Learning Rates and Performance}
\label{sec:lrs}

\begin{figure}
    \centering
        \includegraphics[width=.49\textwidth]{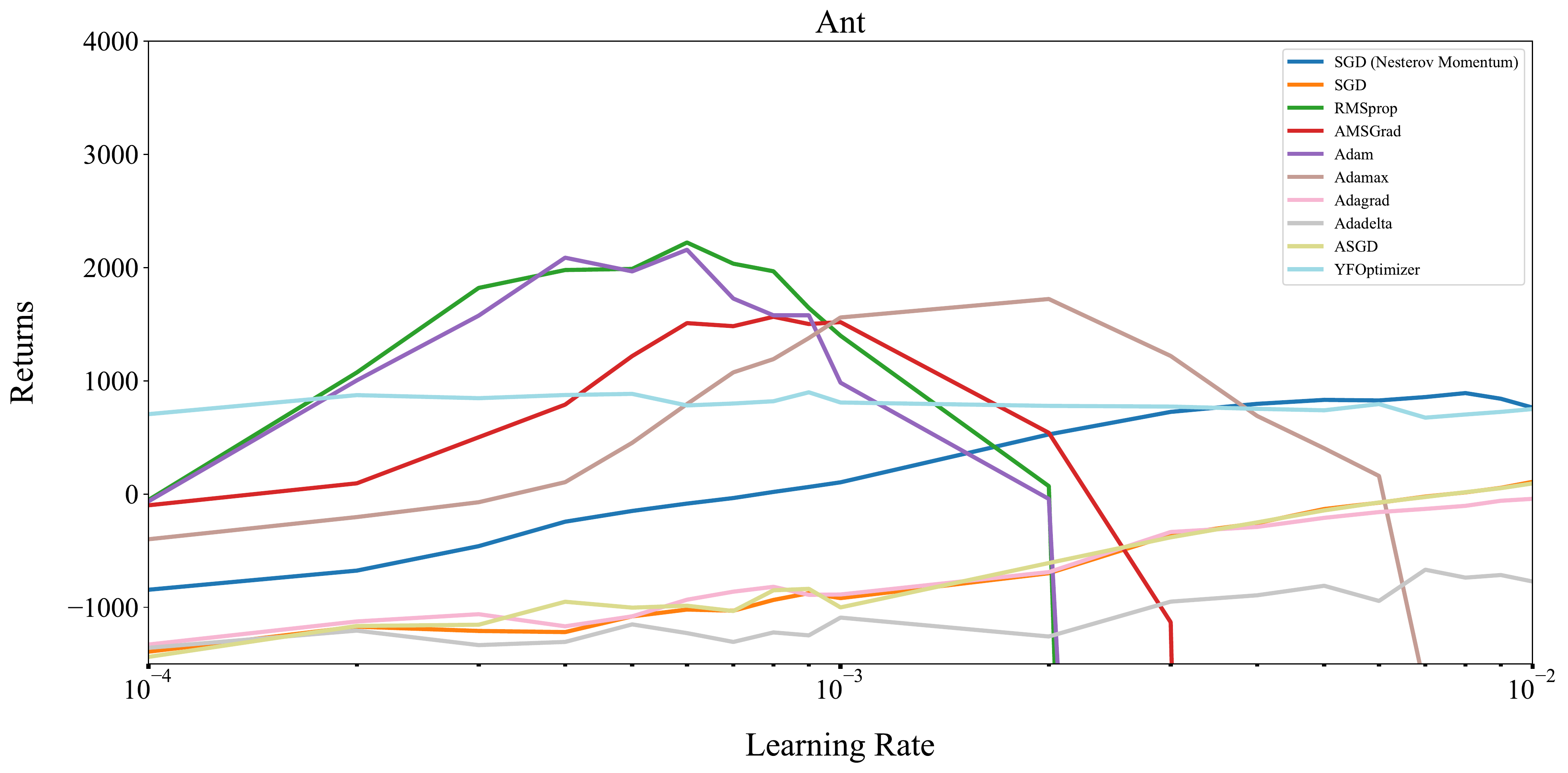}
        \includegraphics[width=.49\textwidth]{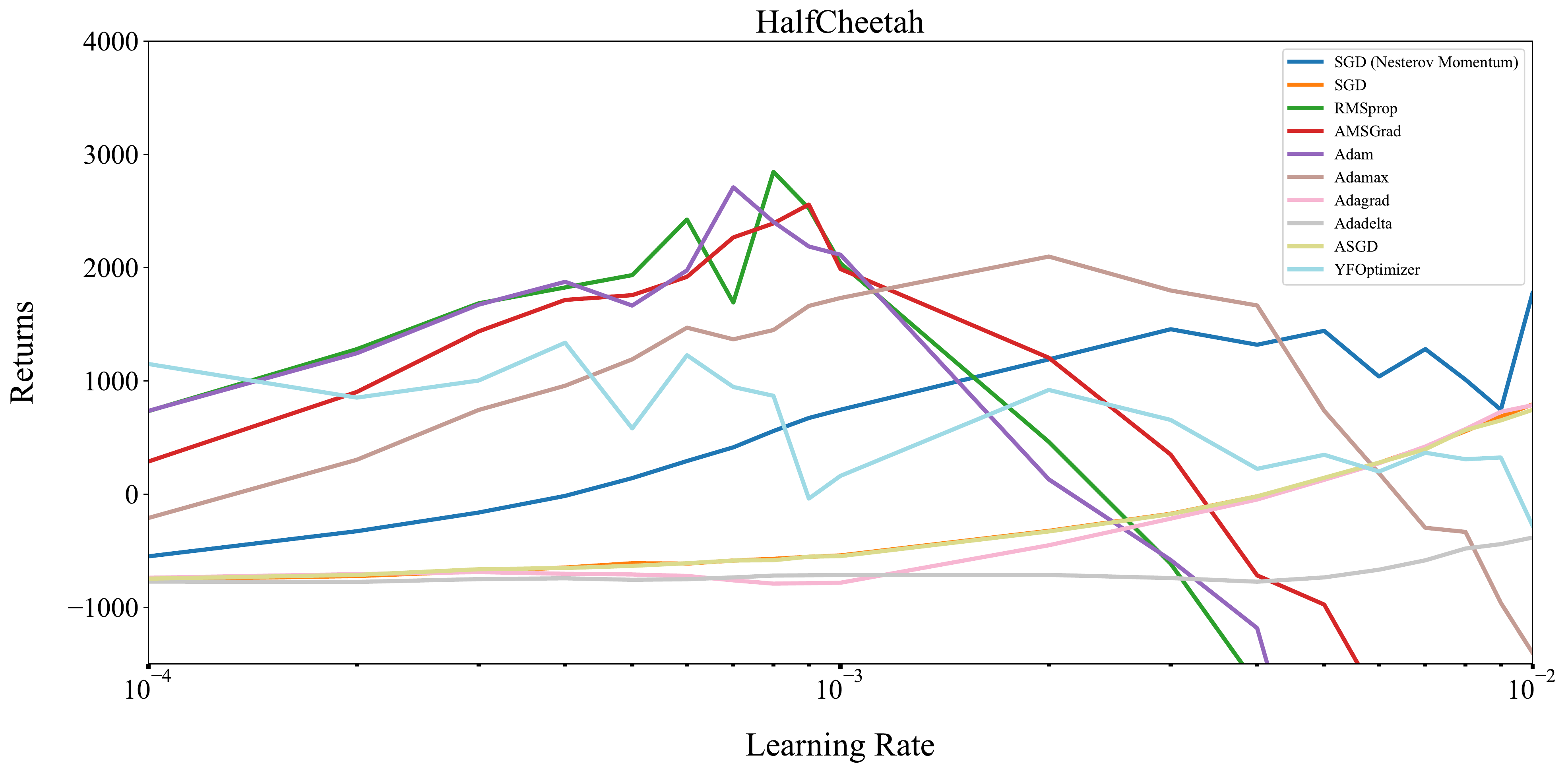}

    \includegraphics[width=.49\textwidth]{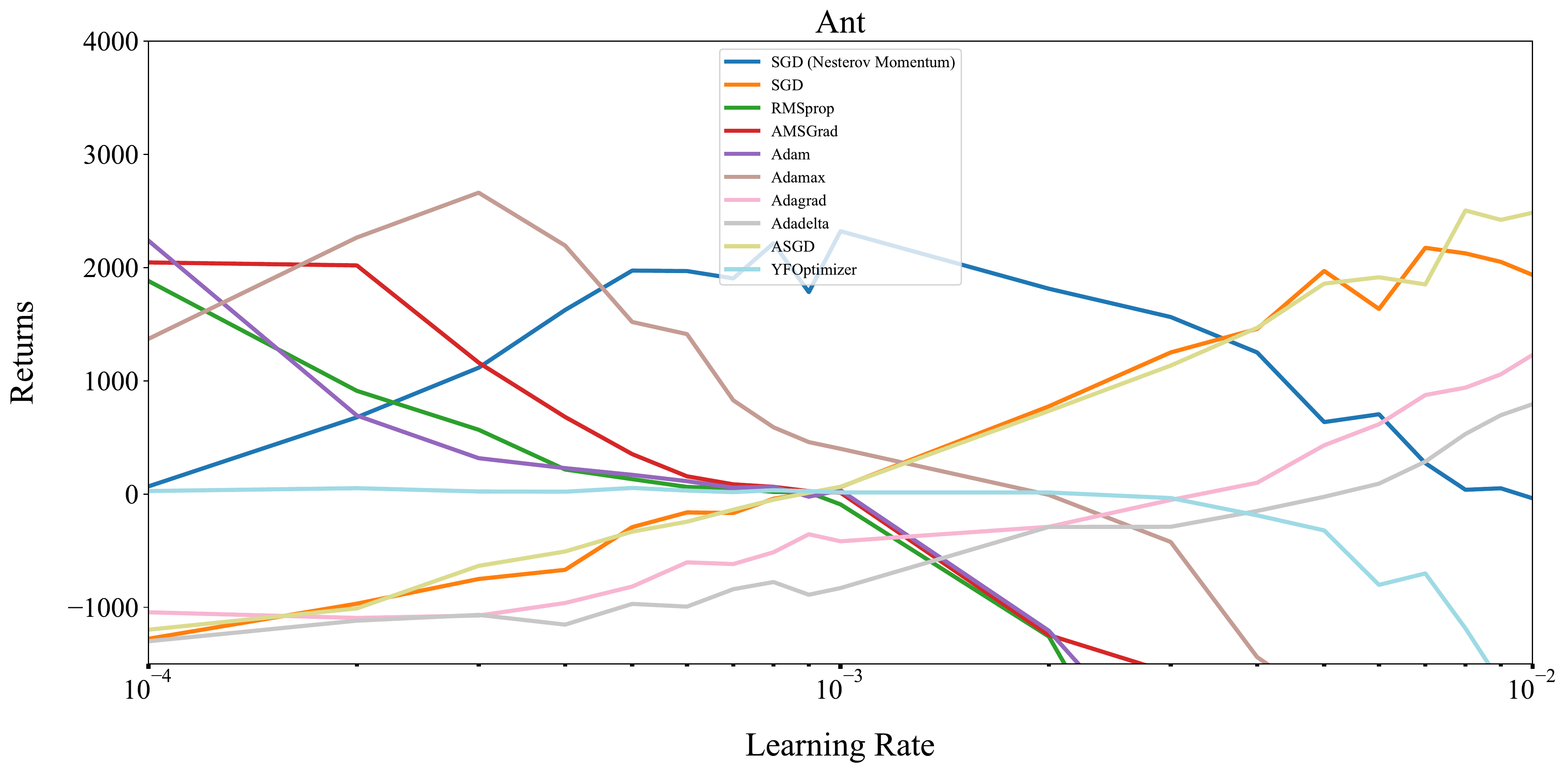}
    \includegraphics[width=.49\textwidth]{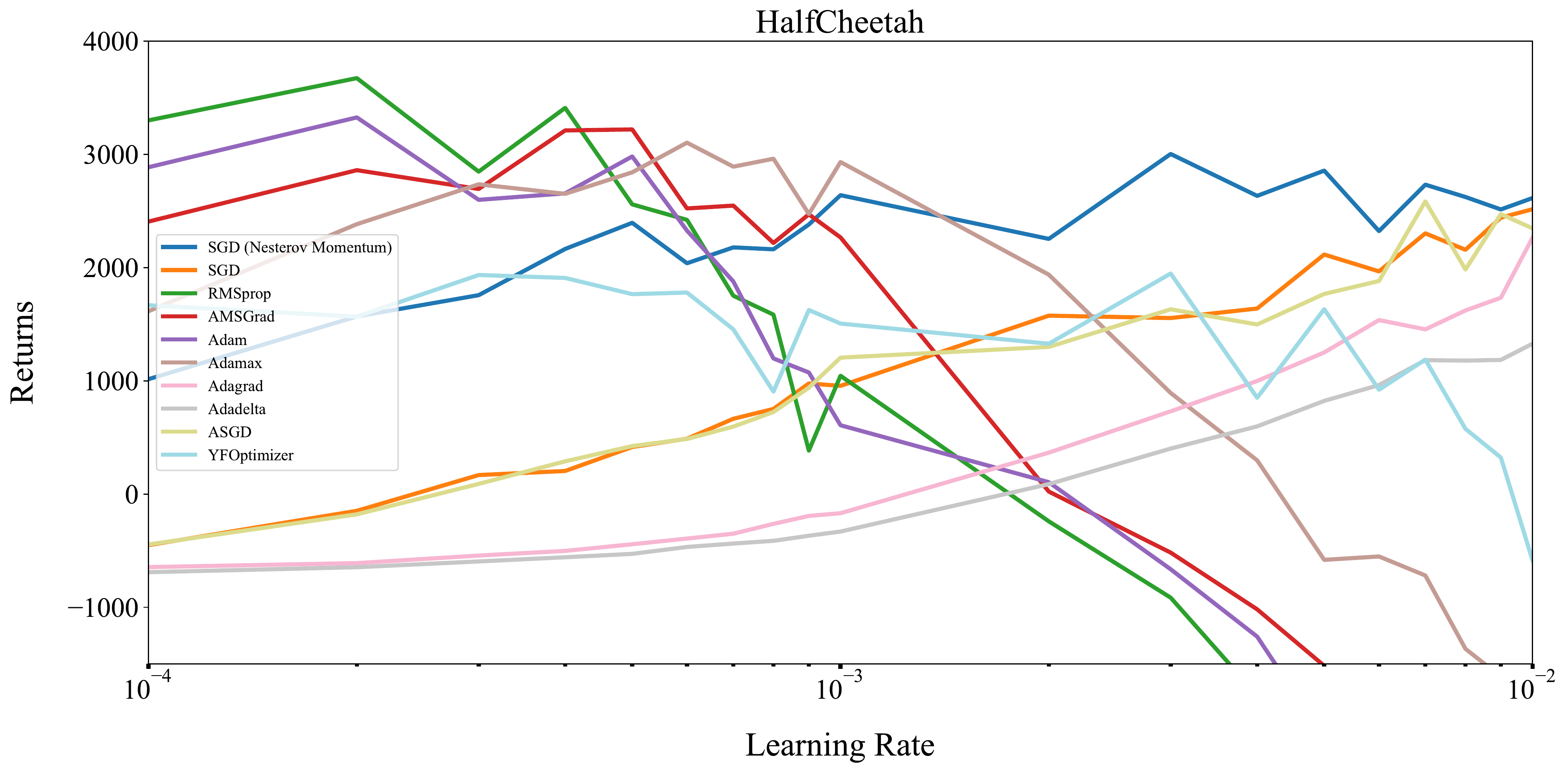}
    \caption{A2C (top row) and PPO (bottom row) performance at different learning rates on Ant and HalfCheetah tasks (left and right, respectively). Equivalent to $\alpha$ plots by~\citet{sutton1998reinforcement}.}
    \label{fig:lrs}
\end{figure}

First, we investigate the effect of the learning rate on optimizer performance. The results can be seen in Figure~\ref{fig:lrs} (with further information and results in Appendix~\ref{app:lrs}). We observe that SGDNM results in performance that is more stable across a variety of learning rates whereas adaptive methods such as Adam and RMSProp diverge entirely at higher learning rates with a small window of well-performing values in certain domains. 
However, it is worth noting that, for A2C, adaptive methods do find better optima with a well-tuned learning rate than we were able to find for SGDNM.

We further describe several interesting observations from our experimentation on learning rate. As is further seen in Appendix~\ref{app:lrs} and in particular Appendix~\ref{app:lrs_alpha}, there are few settings where AMSGrad significantly improves performance despite the suggestion that it should in theory help in stochastic settings such as RL as suggested in~\citep{amsgrad}. 
SGD and ASGD generally perform equally across all learning rates. YellowFin is much more stable across learning rates than most algorithms, but this is likely due to the control-based tuning of SGD hyperparameters. It unfortunately does not yield better performance than other well-tuned algorithms. We find that the algorithm distributions across performance can be grouped into 2 categories generally (with the exception of YellowFin and SGDNM). The first set of optimizers, as seen in Figure~\ref{fig:lrs} (RMSProp, AMSGrad, Adam, AdaMax) have a small window of peak performance at lower learning rates and diverge as the learning rates approach $.01$. The second group (Adadelta, ASGD, Adagrad, SGD) has the opposite distribution, showing poor performance at low learning rates while gaining performance in higher learning rates -- with ASGD and SGD consistently outperforming Adadelta and Adagrad. This is likely due to the similarity in base principles used in implementing these two groups as described in~\citep{ruder2016overview}. While SGDNM can predominantly be categorized with the latter group, in reality its performance falls somewhere in between the two.

\subsection{Momentum}
\label{sec:momemtum}
\begin{figure}
    \centering
        \includegraphics[width=.49\textwidth]{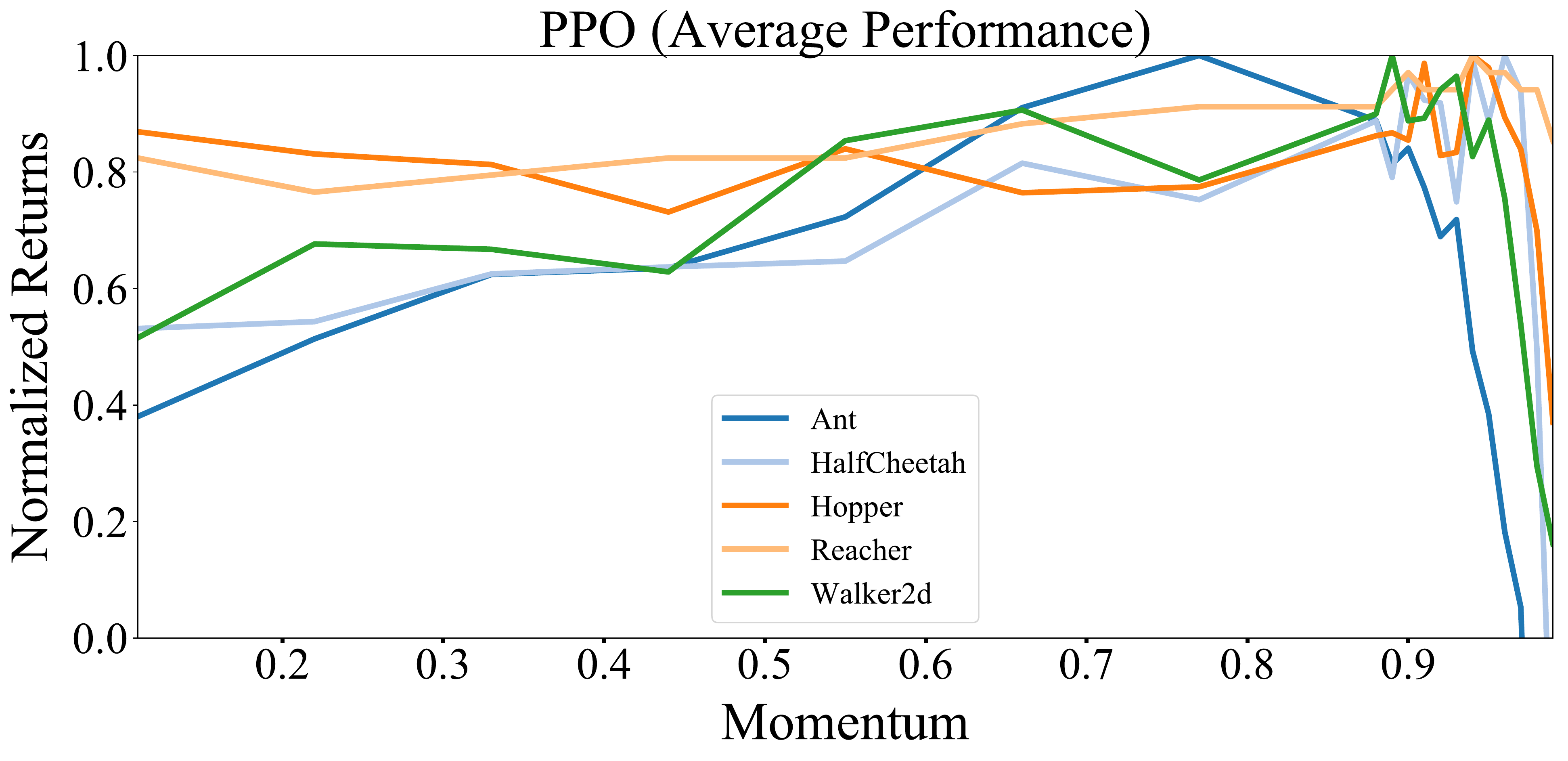}
        \includegraphics[width=.49\textwidth]{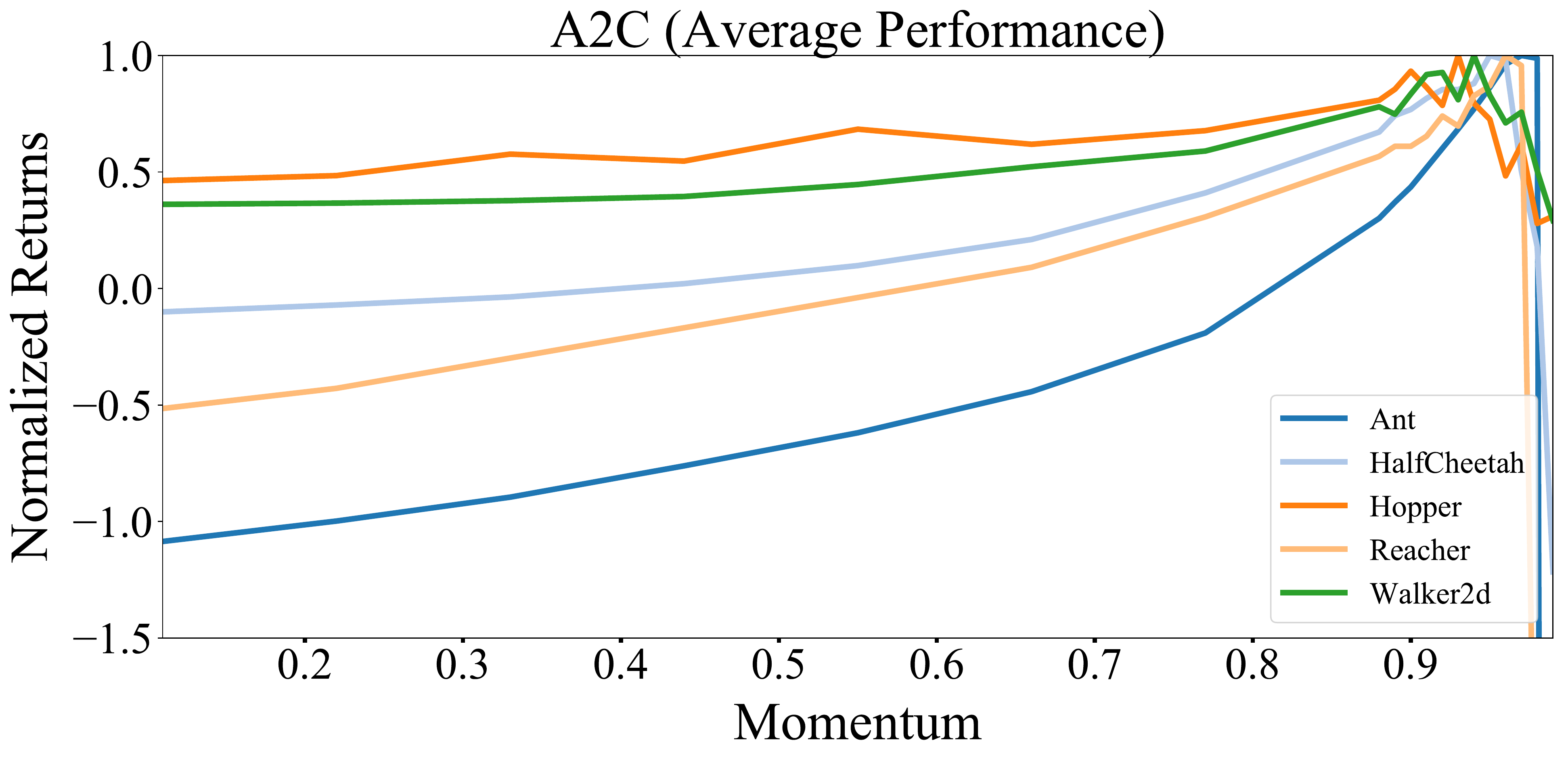}
    \caption{Normalized performance of PPO and A2C across momentum factors in different environments. Normalization is per environment using a random agent policy (see Appendix~\ref{app:random}) such that the Normalized Return corresponds to $\frac{\text{Average Return} - \text{Random Agent}}{\text{Best Average Return} - \text{Random Agent}}$.}
    \label{fig:momentums}
\end{figure}

Next we investigate the effect of momentum on the performance of PPO and A2C. To remove any effects from adaptive aspects of other algorithms we limit the analysis to SGDNM. The goal of these experiments is to probe whether the default momentum hyperparameter of .9 -- that has become the ``industry standard''~\citep{mitliagkas2016asynchrony} -- in fact performs as well as expected. If a lower momentum value is more optimal or yields similar performance as higher momentum values, this may suggest that a sort of implicit momentum is being introduced as described by~\citet{mitliagkas2016asynchrony}, or that the changing loss landscape is negating some of the impact of momentum. We choose the average optimal learning rate of $.003$ for SGDNM as discovered in Section~\ref{sec:lrs} and run a grid of momentum values from $.11$ to $.99$ (with a smaller sub-grid between $.88$ and $.99$ where values tend to diverge).

The final results for the various momentum values can be seen on the normalized average performance across 10 training runs with different random seeds provided in Figure~\ref{fig:momentums} (with additional graphs, tables, and information in Appendix~\ref{app:momentum}). Overall, as momentum approached $1$ from $.9$, in nearly all cases the policy diverged -- as updates change the initial momentum very little, they end up relying on the starting conditions for the update.

We further find that certain environments are less susceptible to momentum in both A2C and PPO. The varying effectiveness of momentum in different environments may be due to the different dynamics in the systems, as also discussed in~\citep{henderson2017deep}. Furthermore, it could be due to high sensitivity of the loss landscape with respect to small changes in the policy. That is, the loss landscape in certain environments might change significantly from episode to episode even with a minor change in the policy (e.g, environments where the agent might fall over and end the episode early). In such cases, momentum would not bear the same positive impact since the optimum might shift in a different direction.

Additionally, we see differences in momentum importance between A2C and PPO. In A2C, the impact of momentum is seen more extensively with values closer to $.9$ yielding large improvements, while in PPO in some environments values remain with 20\% of the optimal momentum returns across all momentum values. The differing behaviours of momentum between PPO and A2C could be attributable to several factors. One effect could be the dissonance caused by constraining the policy while letting the value function update (e.g., the total momentum for the value function is higher than it should be for a constrained policy). Another possible effect could be the nature of long Monte Carlo rollouts by a single worker used by PPO versus small numbers of steps with many workers used in A2C. To probe these effects, we set up another experiment in an attempt to determine if the length of the steps taken in the environment or the number of parallel workers may cause an implicit momentum that should be accounted for. We decreased the number of parallel workers and increased the step size in increments such that the overall number of optimization batch size remained the same and ran a grid of momentum values (see Appendix~\ref{app:steps} for full setup and results). As seen particularly in Appendix~\ref{app:momentum-step-graphs}, there is a slight trend such that lower momentum values see improved performance at higher worker-to-step ratios -- which suggests a source of implicit momentum. However, this trend is noisy and does not generalize to all environments. Overall, while using the default momentum with SGDNM yields close to the optimal results in all cases, the difference in performance improvement at higher momentum values across algorithms and environments provides valuable insights into factors affecting performance in adaptive methods.

\section{Conclusion}
We show that adaptive gradient descent optimization methods in Deep RL are highly sensitive to the choice of learning rate. 
In fact, for PPO, the range of well-performing learning rates for adaptive methods is quite small relative to that of simple SGDNM.
Furthermore, we show that momentum effect can be dependent on the environment and to some extent as well on the number of steps taken and workers used. 
This indicates that there may be sources of implicit momentum or other factors that affect adaptive or momentum-based optimization in certain environments.
There are also other notions that we did not explore that may have effects in the optimization performance which may be explored in future work (examples discussed in Appendix~\ref{app:otherfactors}).
Generally, given current methods, we suggest tuning optimization methods as per the analysis provided here. For example, a small well-tuned learning rate with Adam or RMSProp provides the best performance overall, while relying on SGDNM with momentum equal to $.9$ for PPO yields generally acceptable performance across a wider range of learning rates. 

More importantly though, our findings demonstrate that the use of default values with adaptive optimizers may not be enough for the unique properties of Deep RL.
Both the adaptive methods and non-adaptive methods still require hyperparameter tuning to perform well -- with different optimal settings in each environment.
However, tuning may be difficult in more complex environments or online settings, causing issues of fairness, reproducibility, and efficiency~\citep{henderson2017deep}. Furthermore, recent work proposed that algorithms be evaluated on a small neighbourhood of hyperparameters to determine robustness as this may be a crucial factor for real-world usage \citep{cohen2018distributed}. While tuning may be acceptable in simple settings in the interim, further research is likely needed into developing or using adaptive gradient descent optimization methods which account for changing loss landscapes in different environments and the unique dynamics of Deep RL algorithms.
Such research may allow for evaluation as suggested by~\citet{cohen2018distributed},
yield more reproducible results by avoiding brittle hyperparameters, improve scalability and robustness, and move toward lifelong learning, online, and complex settings where tuning may not be possible or practical. We hope that the analysis and insights we provide here can be used as a foundation for building such Deep RL-specific optimization methods.

\newpage

\appendix

\section{Experimental Setup}
\label{app:setup}
For all experiments we use a modified version of~\citet{pytorchrl} PyTorch implementations of A2C and PPO. We found this to produce the most reliable results close to those reported by the original works. Our modifications simply make it easier to run experiments from a set of configurations files. We provide our modified version of the code along with tools used to generate the graphs here in: \href{https://github.com/facebookresearch/WhereDidMyOptimumGo}{https://github.com/facebookresearch/WhereDidMyOptimumGo}.

We note that the variance and standard deviations in all results below indicate that across 10 different trials with a fixed set of random seeds where we set the seeds to:

\begin{lstlisting}
    { "agent_seed" : 125125, "environment_seed" : 153298},
    { "agent_seed" : 513, "environment_seed" : 623},
    { "agent_seed" : 90135, "environment_seed" : 6412},
    { "agent_seed" : 81212, "environment_seed" : 91753},
    { "agent_seed" : 3523401, "environment_seed" : 52379},
    { "agent_seed" : 15709, "environment_seed" : 17},
    { "agent_seed" : 1, "environment_seed" : 99124},
    { "agent_seed" : 0, "environment_seed" : 772311},
    { "agent_seed" : 8412, "environment_seed" : 19153163},
    { "agent_seed" : 1153780, "environment_seed" : 9231}
\end{lstlisting}

Where ``agent seed'' is the seed provided to all random number generators related to the agent (including network initialization) and ``environment seed'' relates to the seed provided to the environment. In this case, the mean itself provides insight into optimizer performance as all sources of randomness are fixed. The variance instead gives an indication as to an optimizer's ability to find a similar optimum in different conditions. 

For both A2C and PPO, we use the normalized observations and reward as in~\citep{pytorchrl}. We run all on CPU to avoid non-determinism in the GPU. For A2C we use hyperparameters:

\begin{lstlisting}
      "add_timestep" : false, "recurrent_policy" : false,
      "num_frames" : 2e6, "num_steps" : 5,
      "num_processes" : 16, "gamma" : 0.99,
      "tau" : 0.95, "use_gae" : false,
      "value_loss_coef" : 1.0, "entropy_coef" : 0.0,
      "max_grad_norm"  : 0.5, "num_stack" : 1
\end{lstlisting}

\noindent For PPO:

\begin{lstlisting}
      "add_timestep" : false, "recurrent_policy" : false,
      "num_frames" : 2e6, "num_steps" : 2048,
      "num_processes" : 1, "gamma" : 0.99,
      "tau" : 0.95, "use_gae" : true, "num_stack" : 1
      "clip_param" : 0.2, "ppo_epoch" : 10,
      "num_mini_batch" : 32, "value_loss_coef" : 1.0, 
      "entropy_coef" : 0.0, "max_grad_norm"  : 0.5,
      
\end{lstlisting}

When running ablation analysis on optimizers, we use the PyTorch default set of hyperparameters except for some cases which we align with the optimizers used in~\citep{pytorchrl}. They are as follows:

\begin{lstlisting}
Adagrad(params, lr=0.01, lr_decay=0, weight_decay=0,
    initial_accumulator_value=0)
Adam(params, lr=0.001, betas=(0.9, 0.999), eps=1e-5, 
    weight_decay=0, amsgrad=False)
    # amsgrad=True when using AMSGrad
Adamax(params, lr=0.002, betas=(0.9, 0.999), eps=1e-08,
    weight_decay=0)
ASGD(params, lr=0.01, lambd=0.0001, alpha=0.75, t0=1000000.0,
    weight_decay=0)
RMSprop(params, lr=0.01, alpha=0.99, eps=1e-5, weight_decay=0,
    momentum=0, centered=False)
SGD(params, lr=<object object>, momentum=0, dampening=0,
    weight_decay=0, nesterov=False) 
    # momentum=.9 and nesterov=True for SGDNM
\end{lstlisting}

The YellowFin optimizer from: \href{https://github.com/JianGoForIt/YellowFin_Pytorch}{https://github.com/JianGoForIt/YellowFin\_Pytorch} at commit hash 362ed7ada76f3d789aa2c431bc333b33fedc71ea. All default settings were used except for:

\begin{lstlisting}
        "force_non_inc_step": true, "stat_protect_fac" : true
\end{lstlisting}

\noindent We found that otherwise the optimizer would consistently diverge.

Throughout the results we refer to SGDNM which stands for SGD with Nesterov Momentum. A* indicates the Ada family of algorithms. RMS indicates RMSProp and AMS indicates AMSGrad. We also refer to asymptotic performance (which is averaged over the last 50 episodes) and average performance (where the average is over all episodes in the training process). The analysis of average performance of all episodes in training is a similar analysis that of \citet{sutton1998reinforcement} and gives better insight into the effect on both learning speed and final asymptotic performance.

In our momentum experiments, while \citet{mitliagkas2016asynchrony} use negative momentum values, PyTorch (version 0.4) does not accept negative momentum values. We decided not to modify the default SGD optimizer provided by PyTorch as this could yield unknown added effects. As such we do not investigate negative momentum values.

\section{Performance of Random Agent}
\label{app:random}
For some of the results, we rely on the performance of a random agent to normalize the visualizations. These performances were averaged over 100 episodes by uniformly sampling from the action space and can be found in Table~\ref{tab:mujocorandom},

\begin{table}[H]
    \centering
    \begin{tabular}{|c|c|c|c|c|c|}
    \hline
         Env & Ant & Hopper& Walker2d & HalfCheetah & Reacher  \\
         \hline
         Return & -72.58 & 16.97 & 1.54 & -272 & -43.1 \\
         \hline
    \end{tabular}
    \caption{Average return of a random uniform sampling policy on MuJoCo tasks across 100 episodes. }
    \label{tab:mujocorandom}
\end{table}

\section{Learning Rate Experiments}
\label{app:lrs}

\begin{figure}[H]
    \centering
    \includegraphics[width=.32\textwidth]{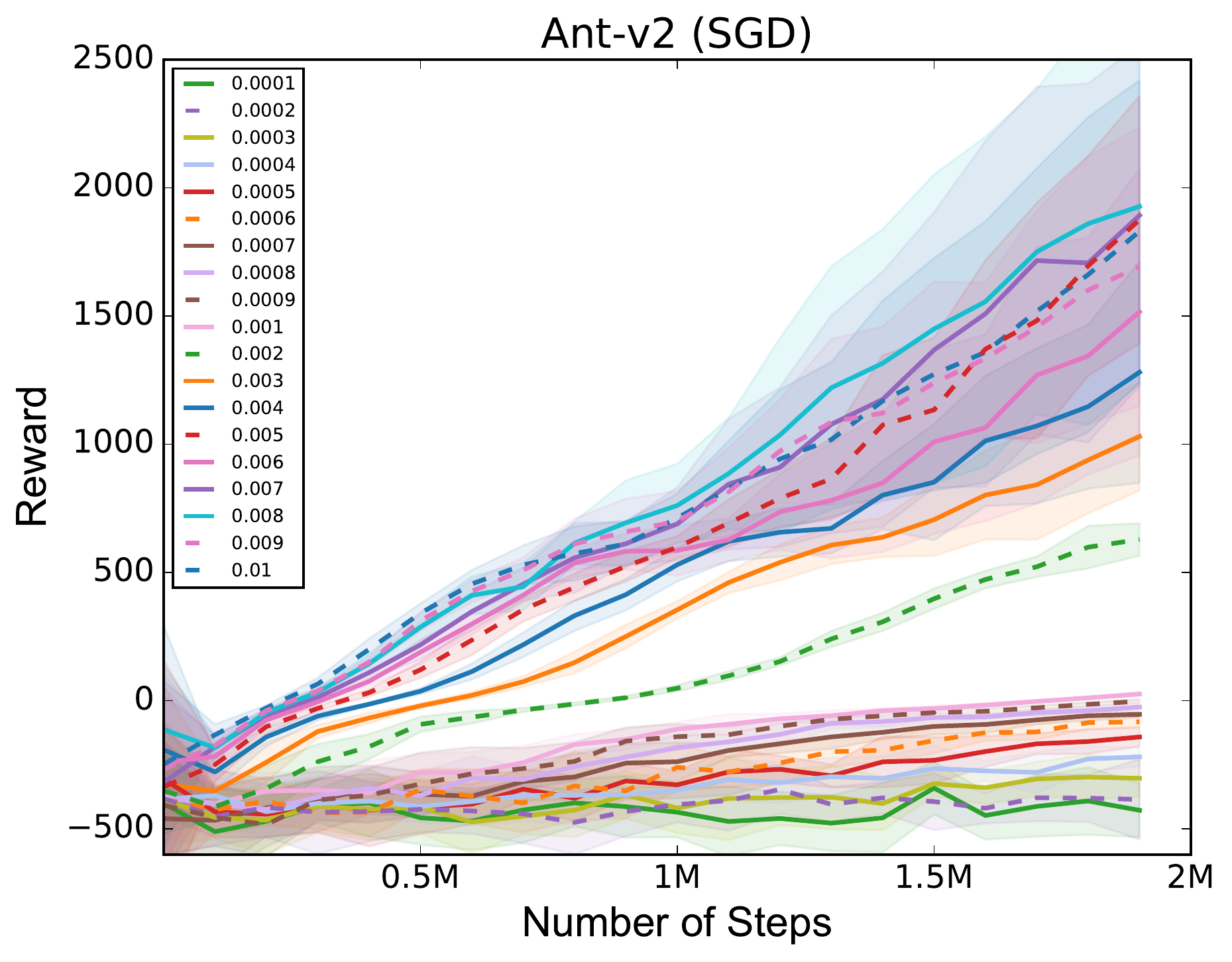}
    \includegraphics[width=.32\textwidth]{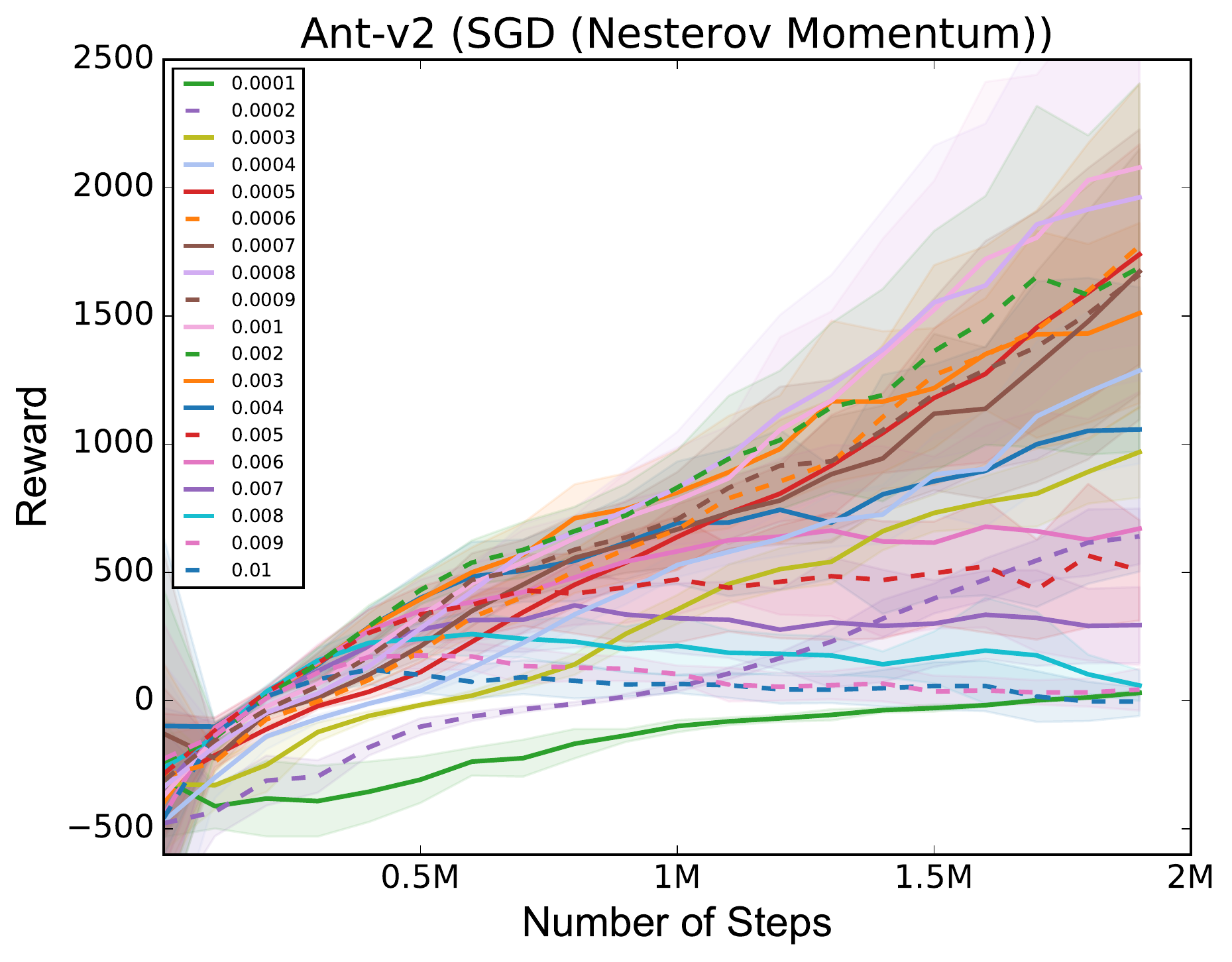}
    \includegraphics[width=.32\textwidth]{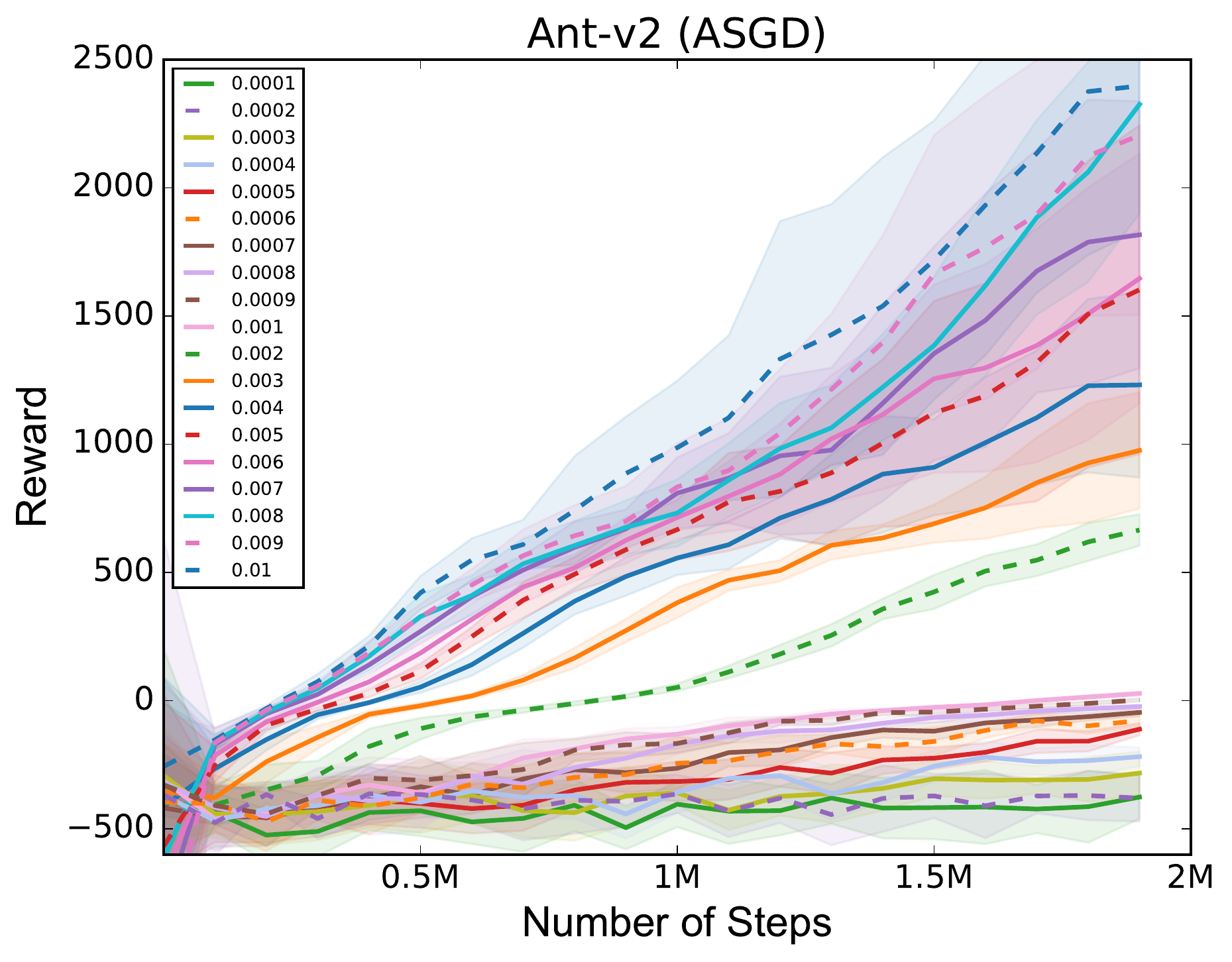}
    \includegraphics[width=.32\textwidth]{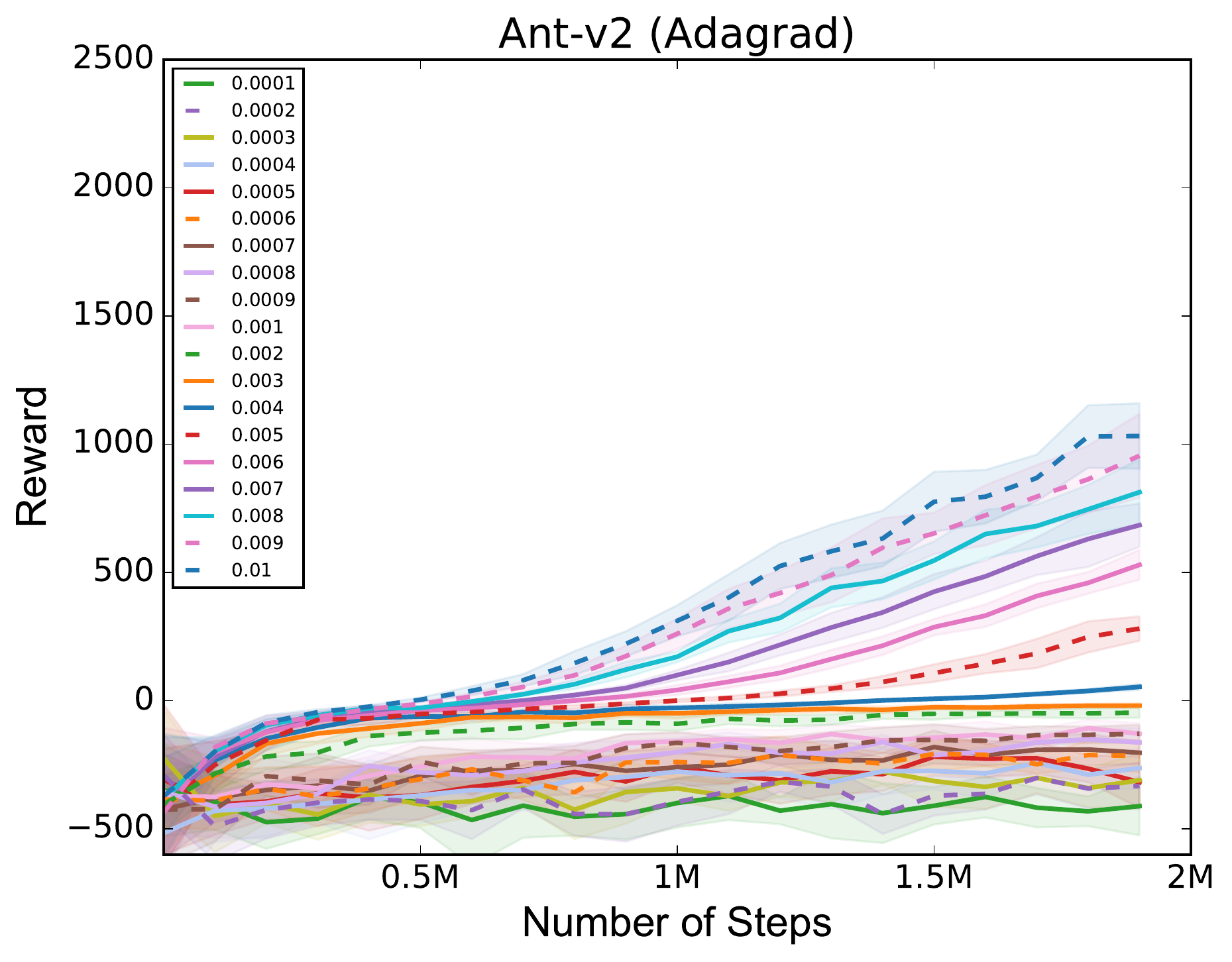}
    \includegraphics[width=.32\textwidth]{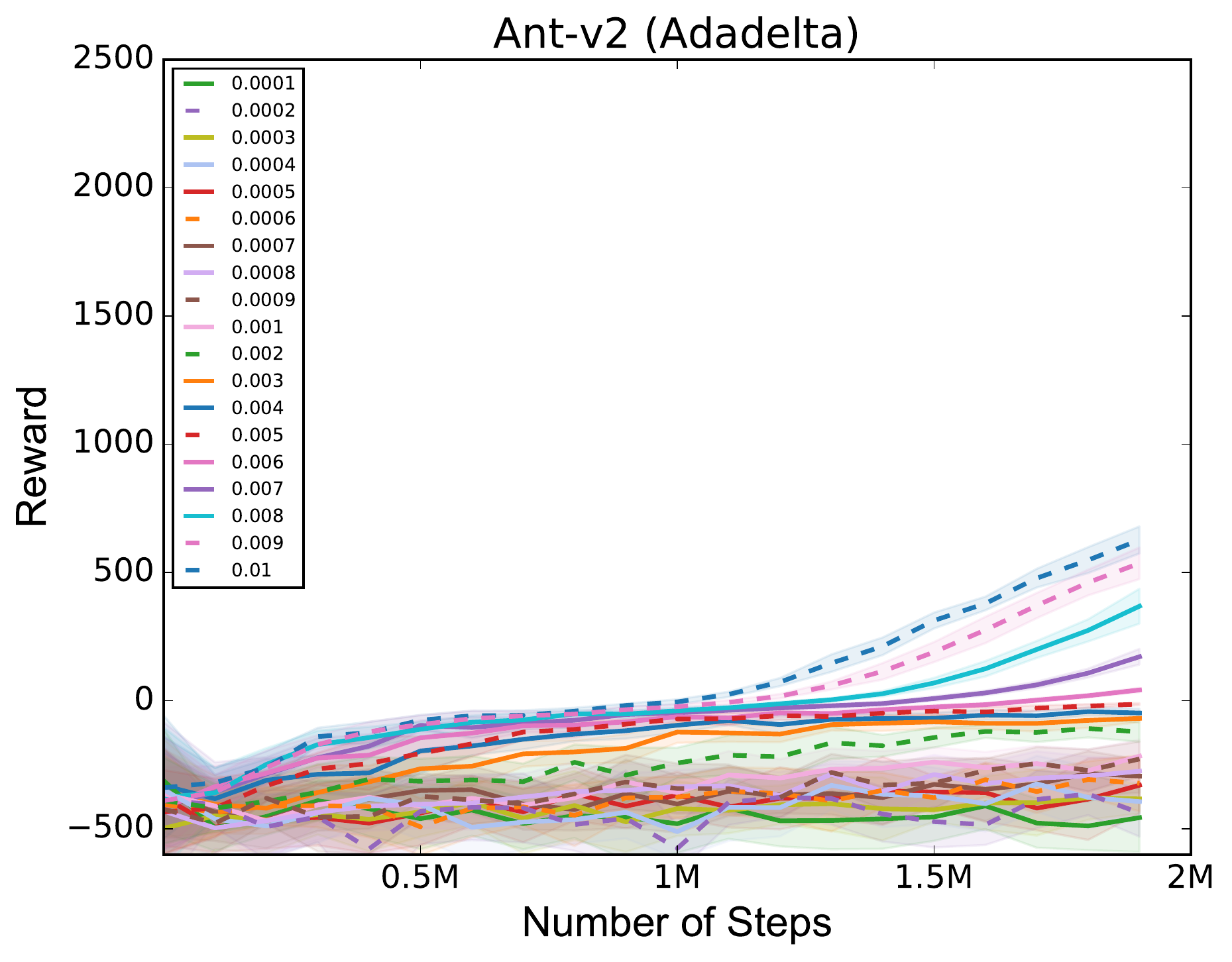}
    \includegraphics[width=.32\textwidth]{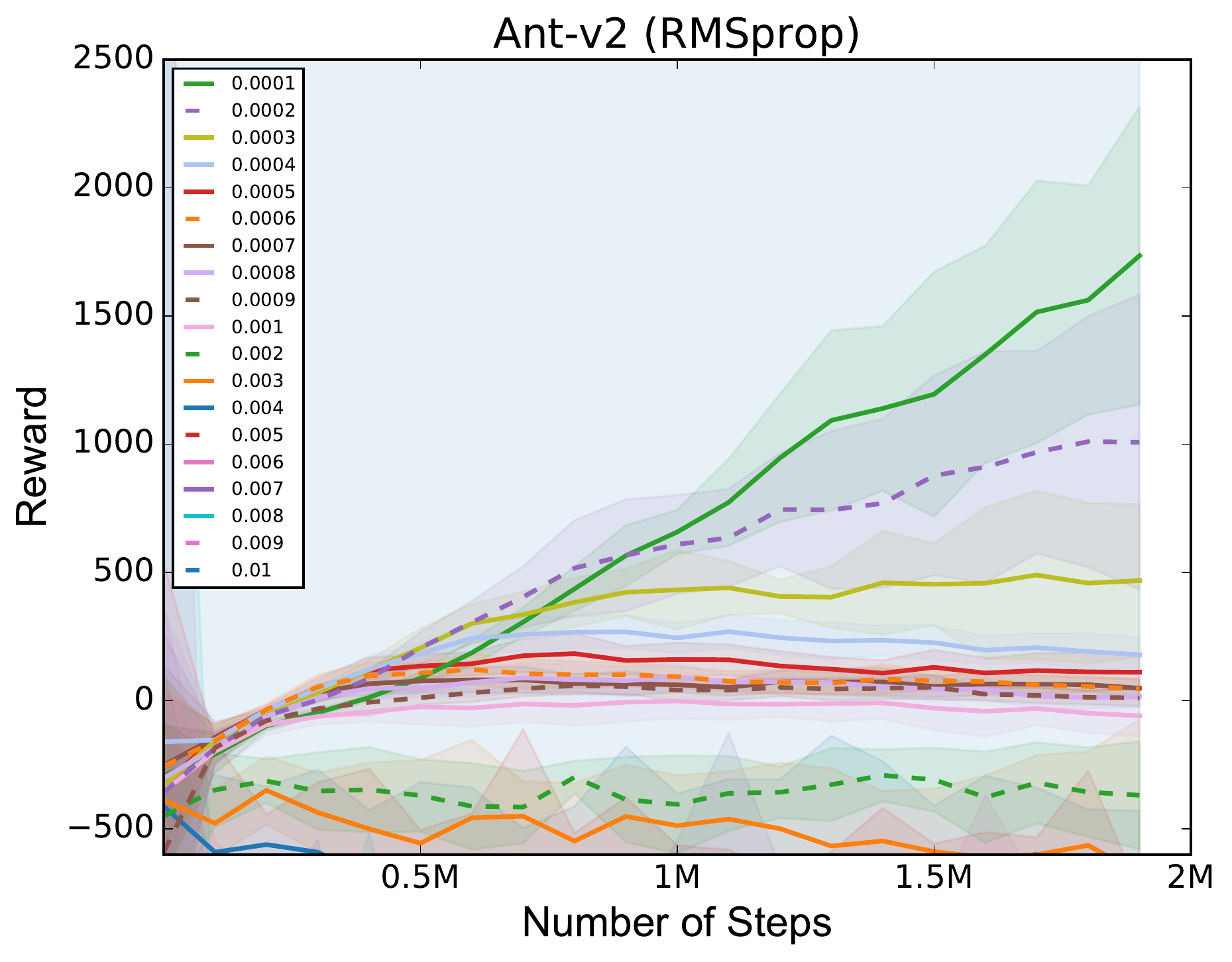}
    \includegraphics[width=.32\textwidth]{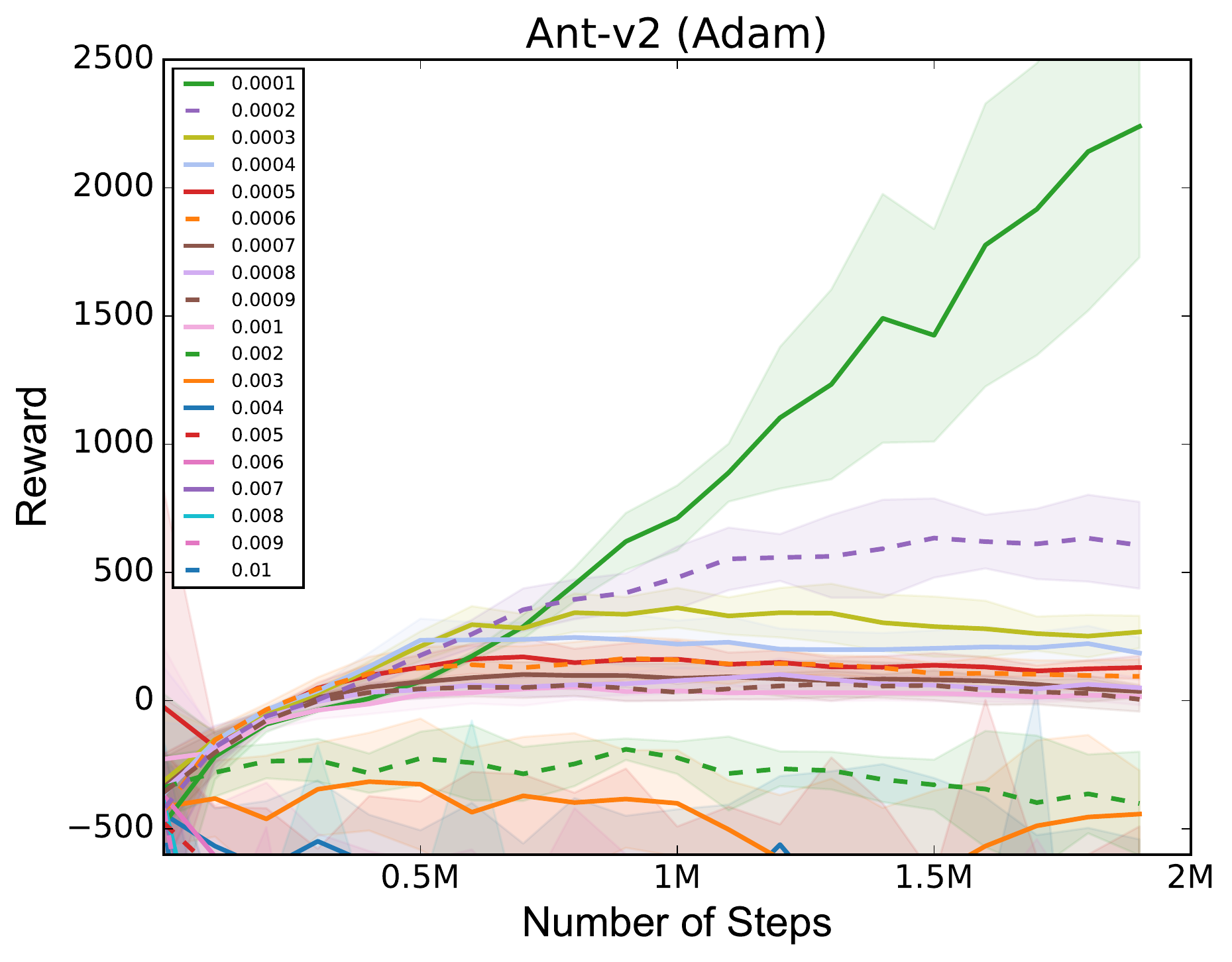}
    \includegraphics[width=.32\textwidth]{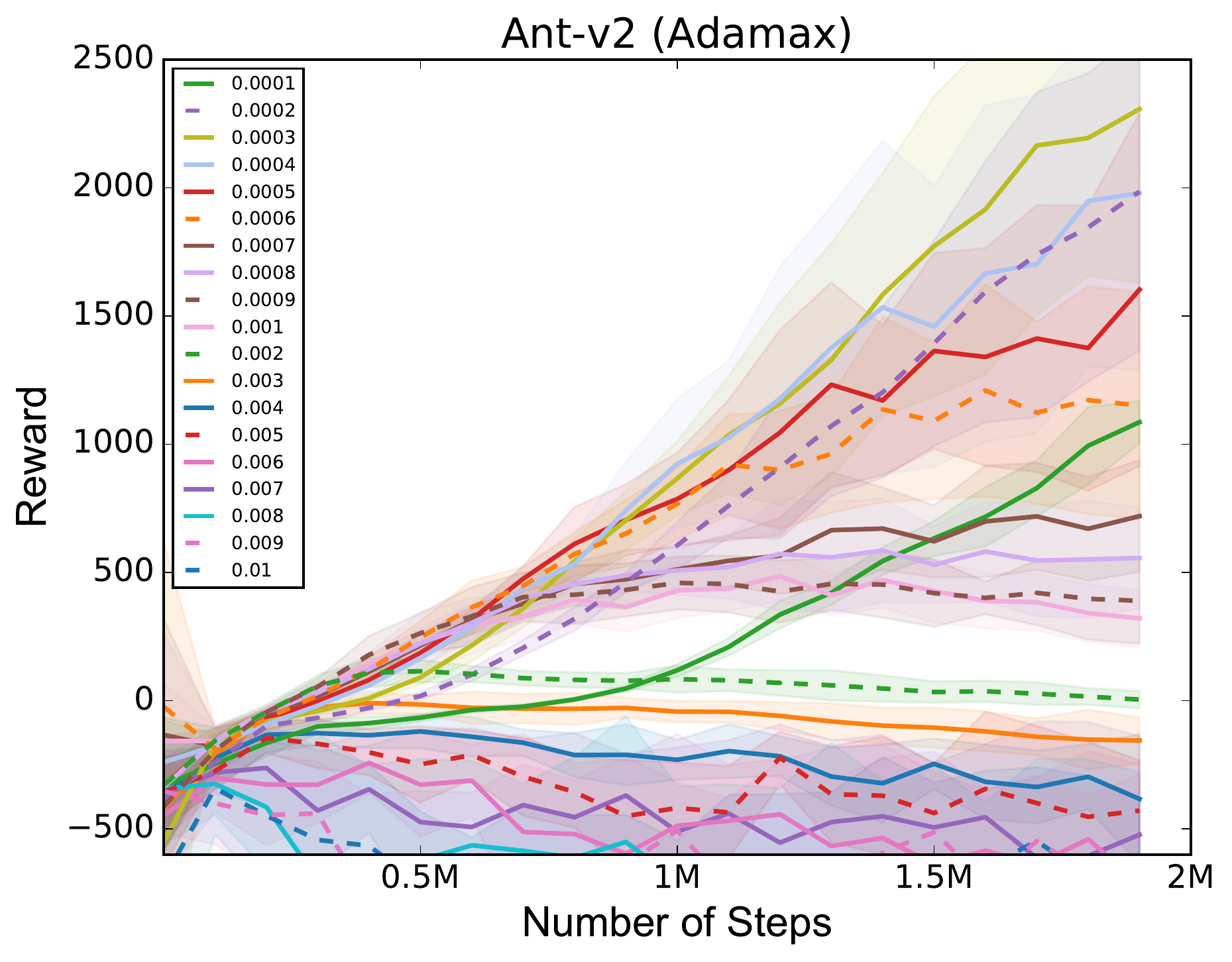}
    \includegraphics[width=.32\textwidth]{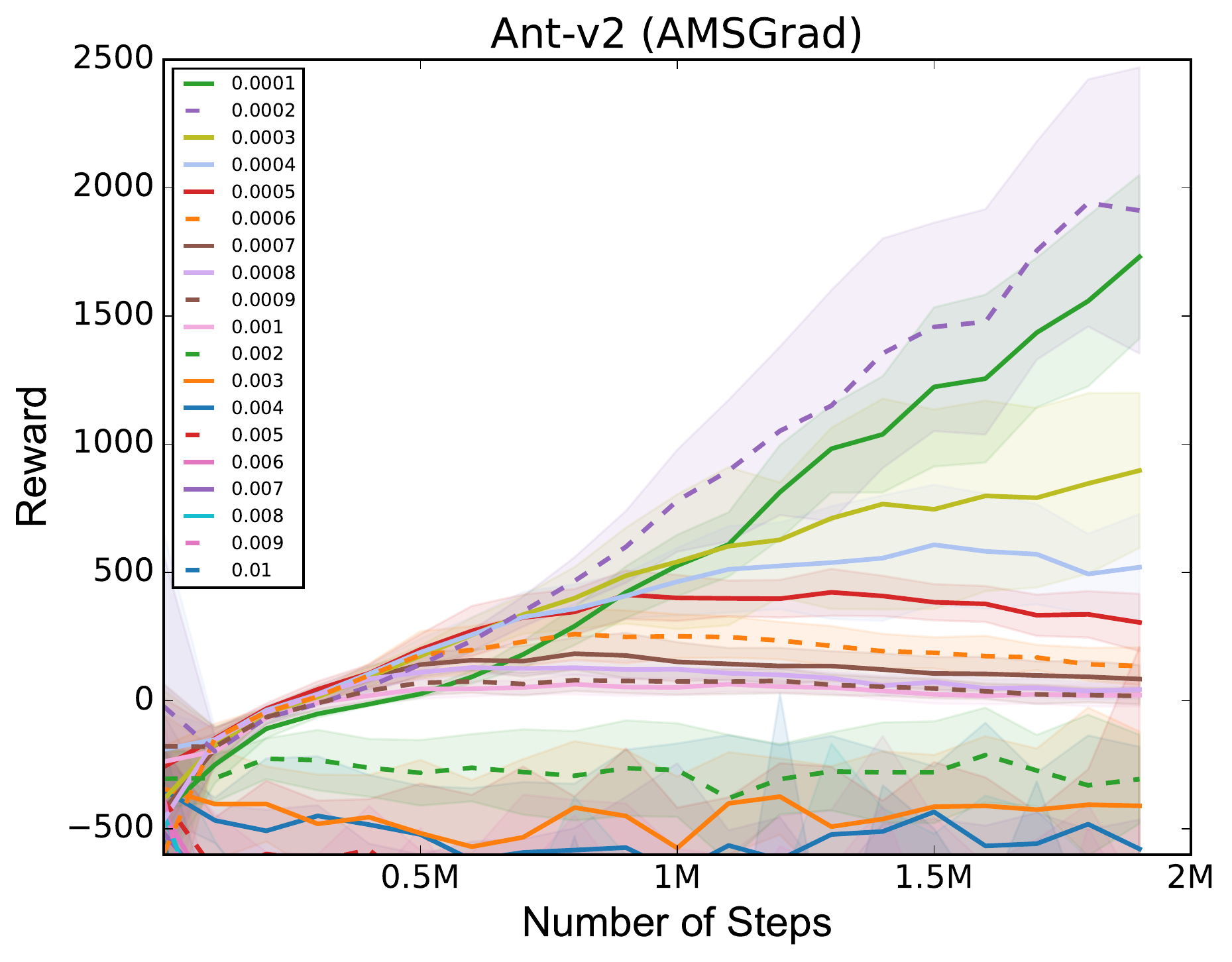}
    \includegraphics[width=.32\textwidth]{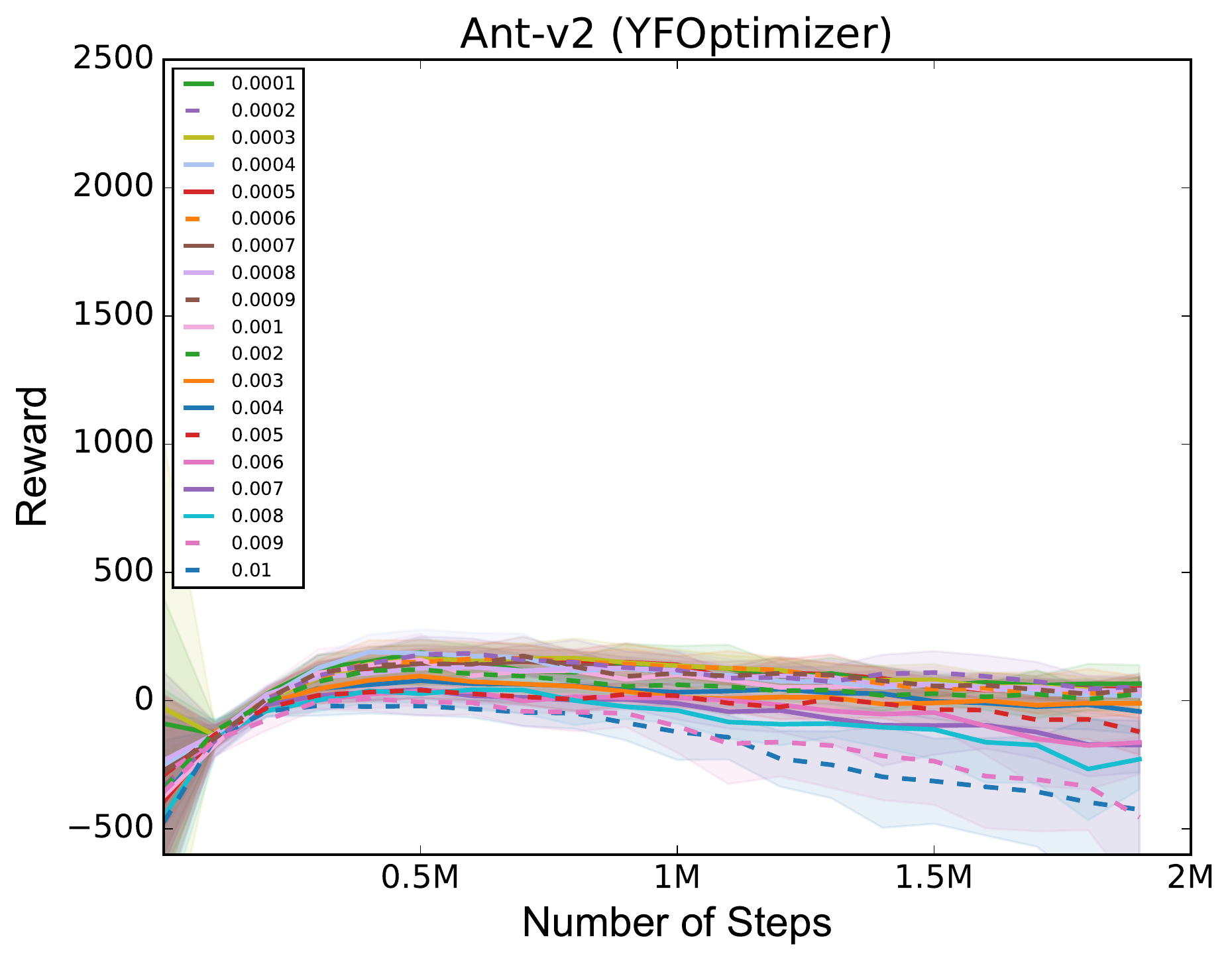}
    \caption{PPO performance across learning rates on the Ant environment.}
    \label{fig:lr1}
\end{figure}

\begin{figure}[H]
    \centering
    \includegraphics[width=.32\textwidth]{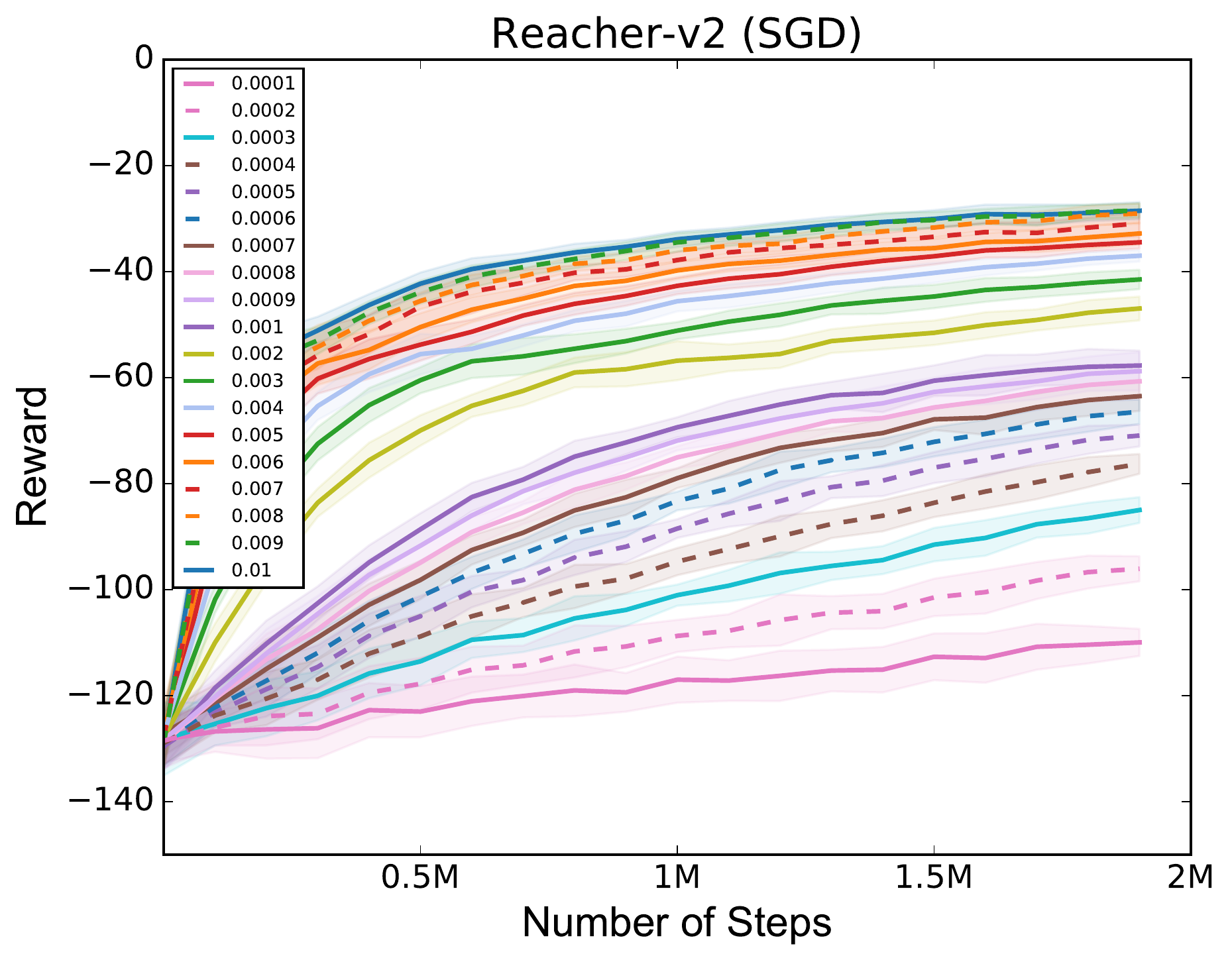}
    \includegraphics[width=.32\textwidth]{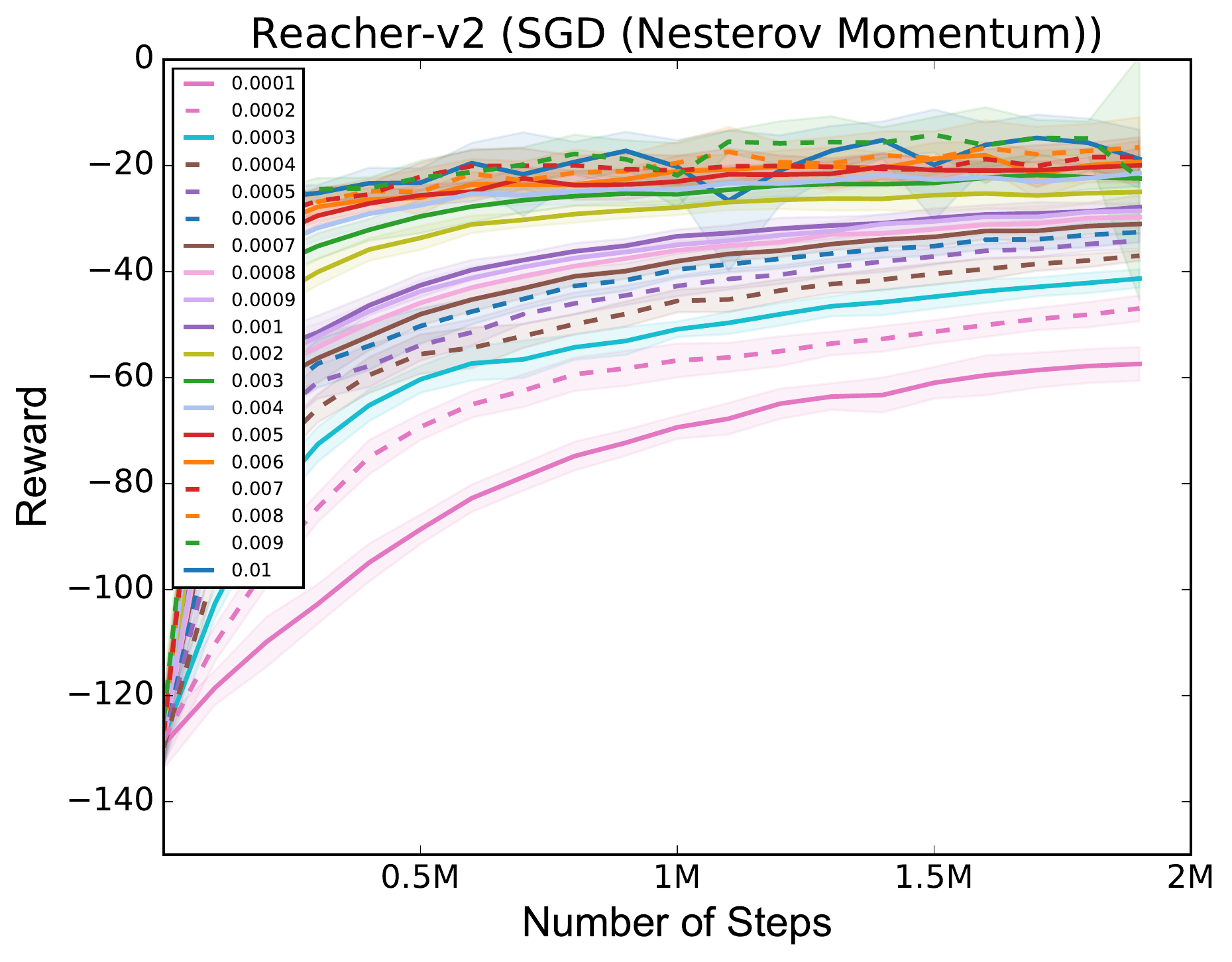}
    \includegraphics[width=.32\textwidth]{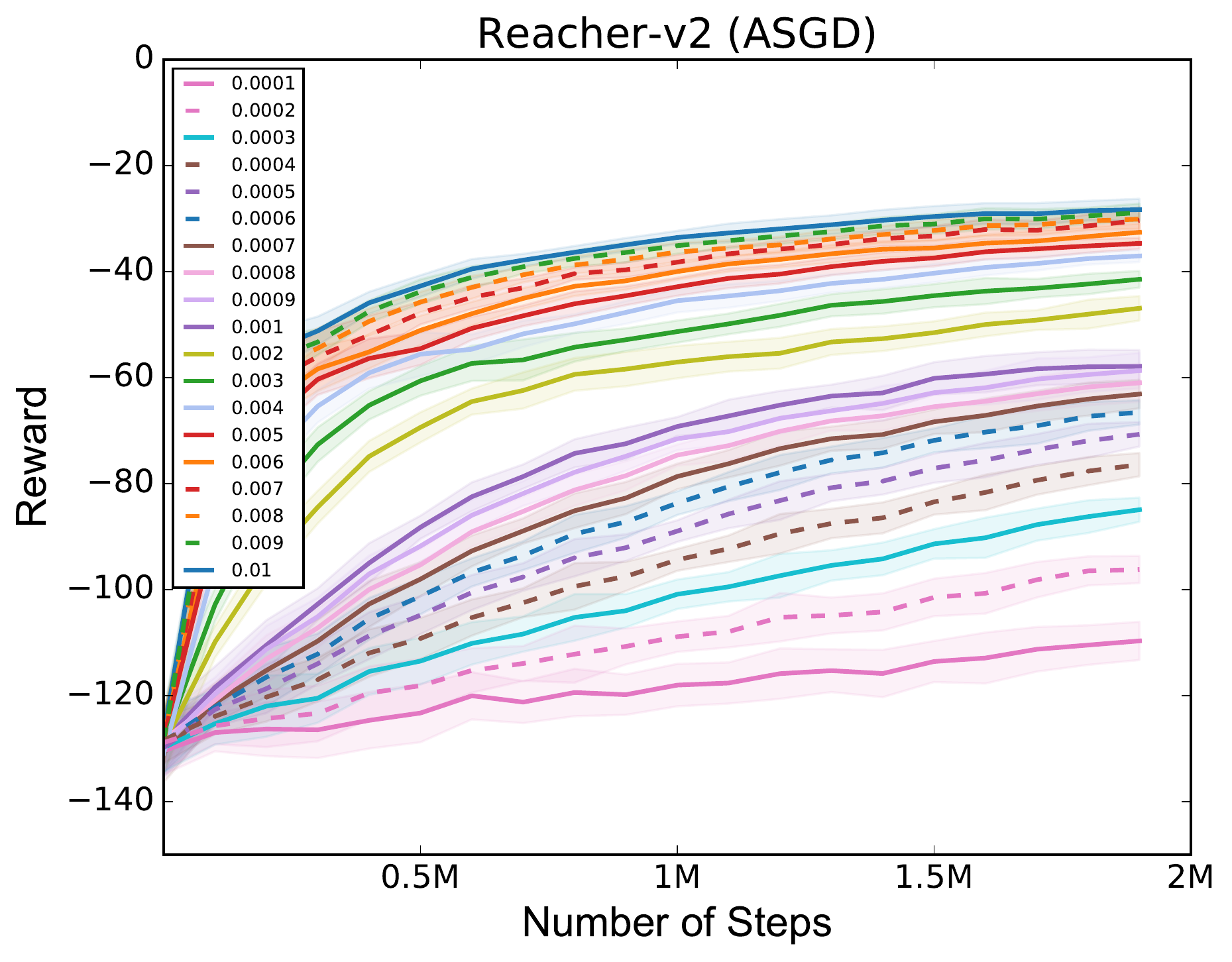}
    \includegraphics[width=.32\textwidth]{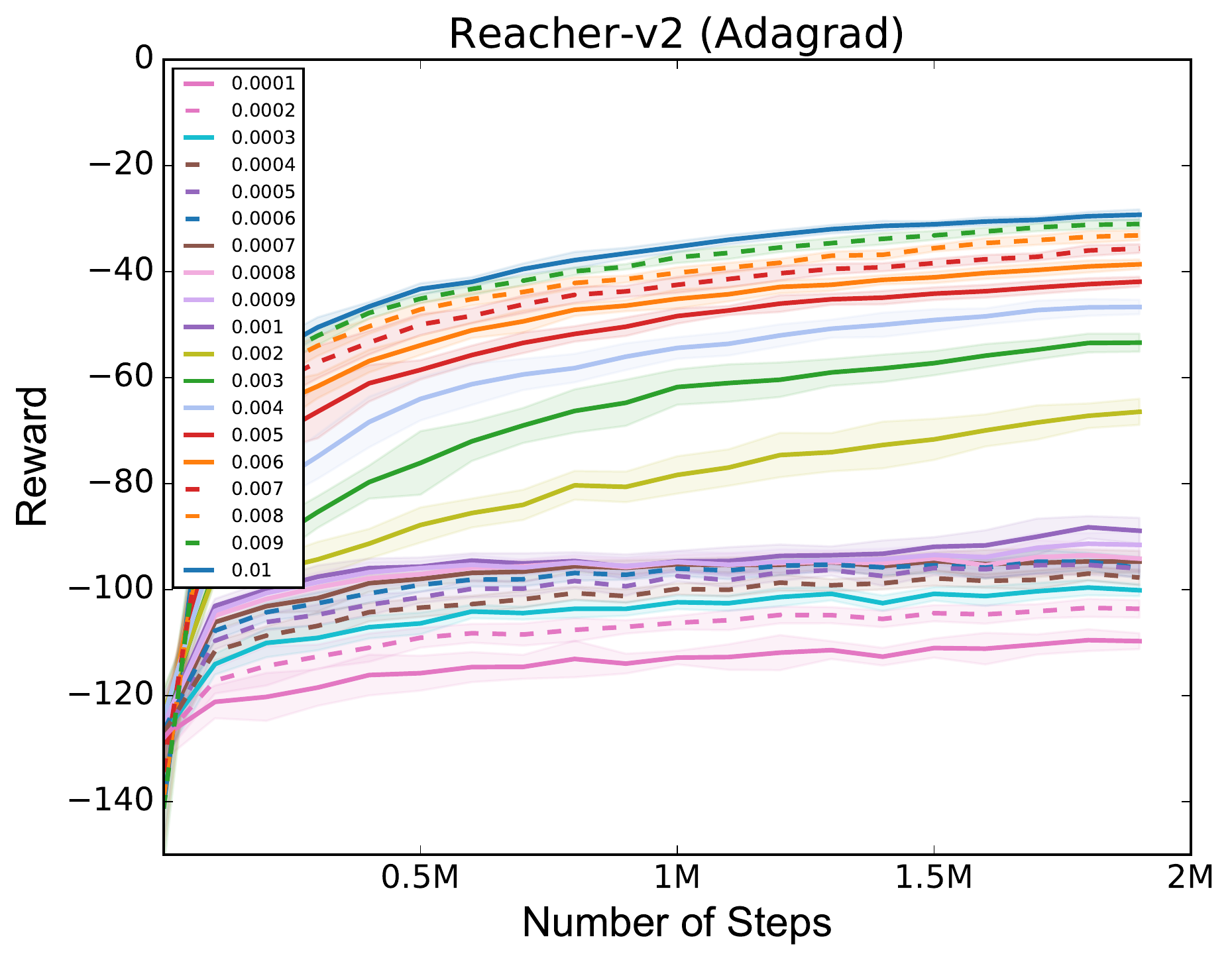}
    \includegraphics[width=.32\textwidth]{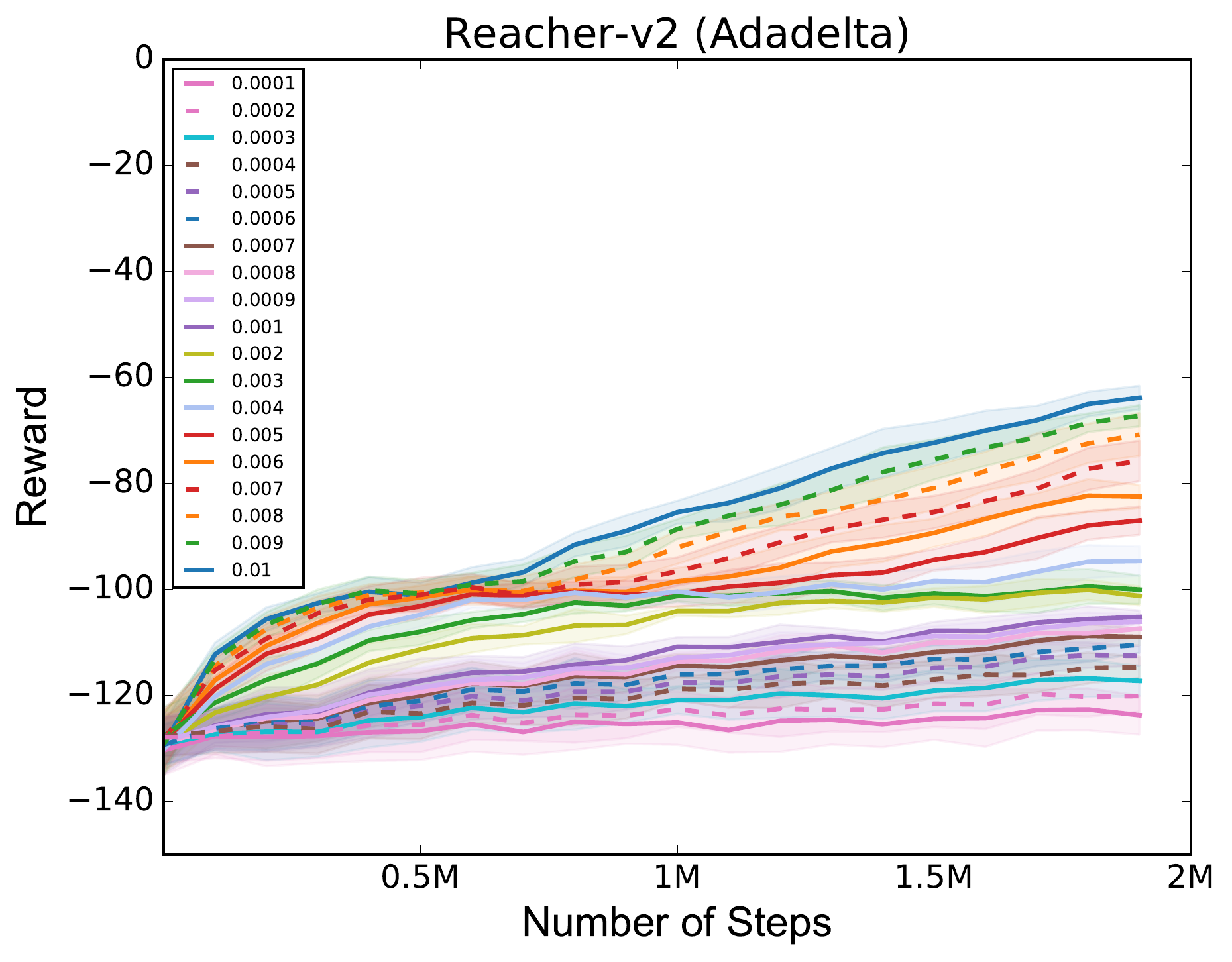}
    \includegraphics[width=.32\textwidth]{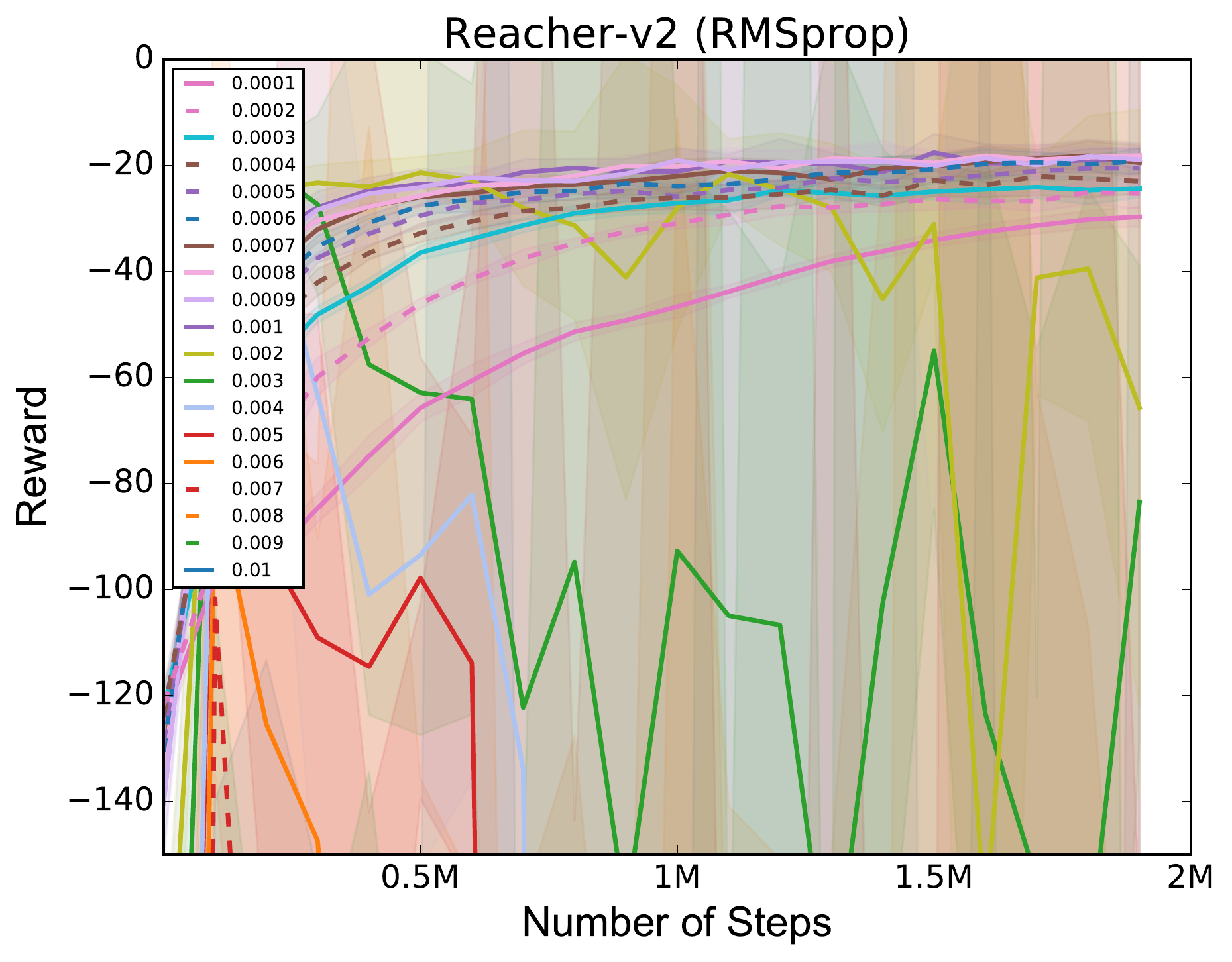}
    \includegraphics[width=.32\textwidth]{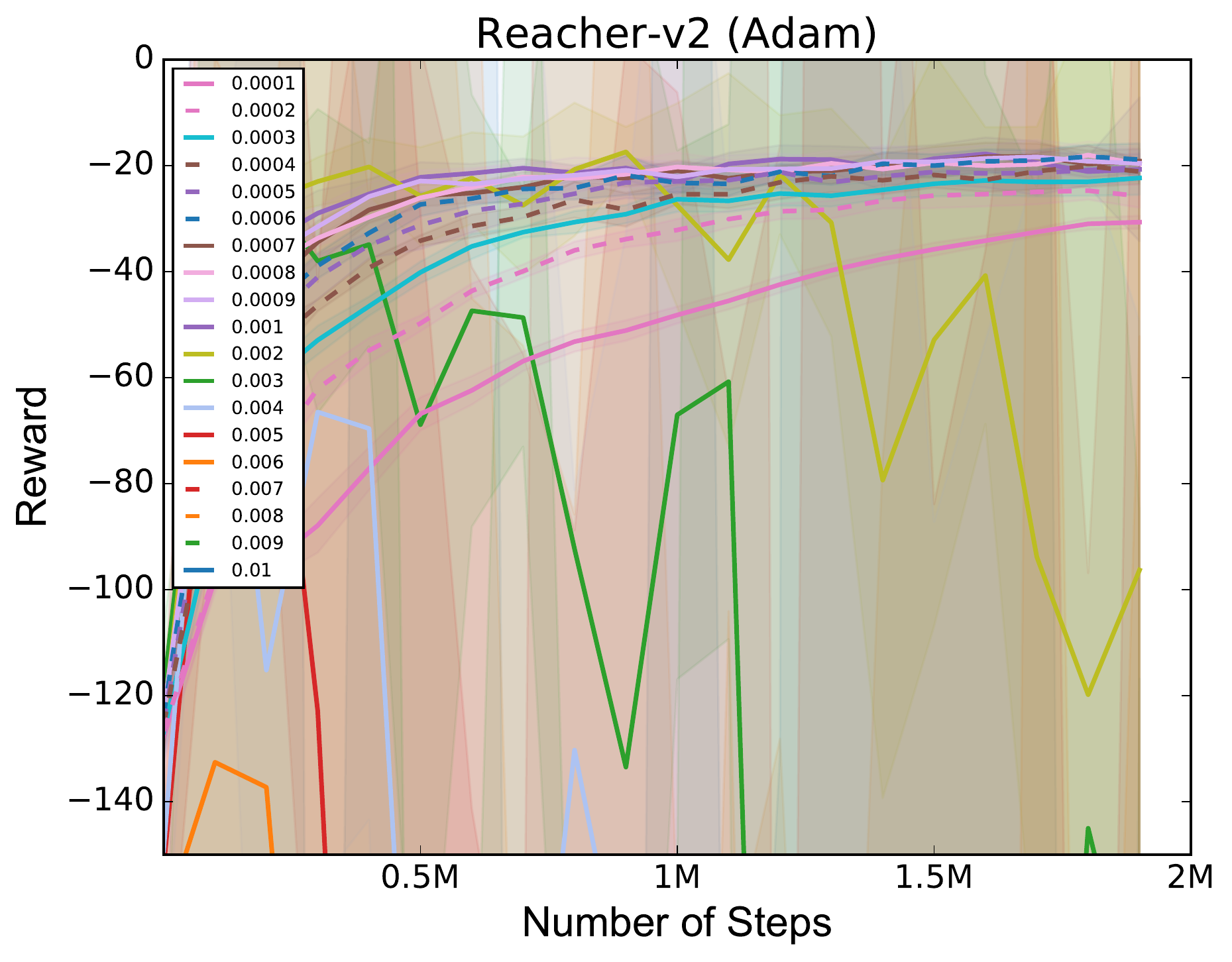}
    \includegraphics[width=.32\textwidth]{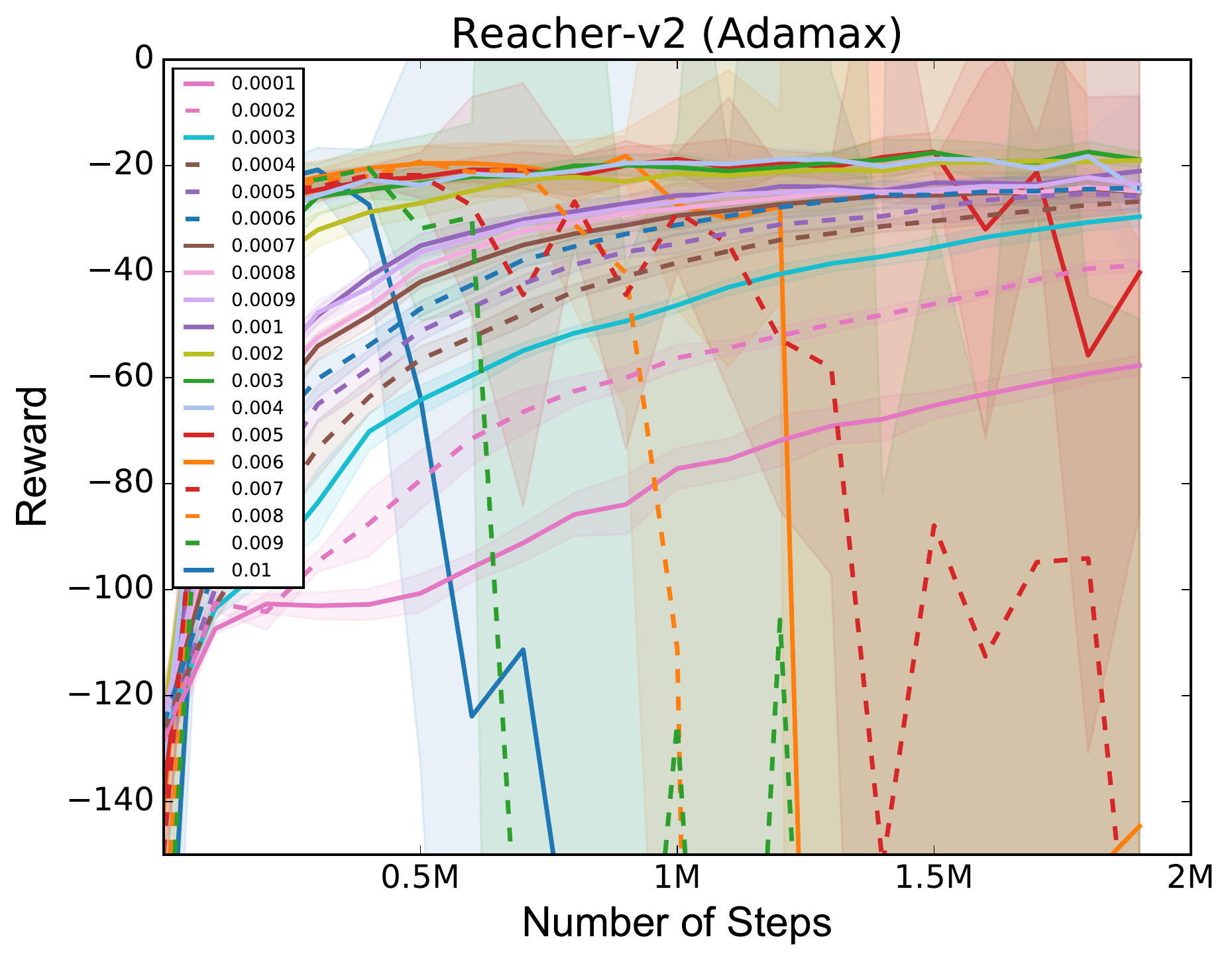}
    \includegraphics[width=.32\textwidth]{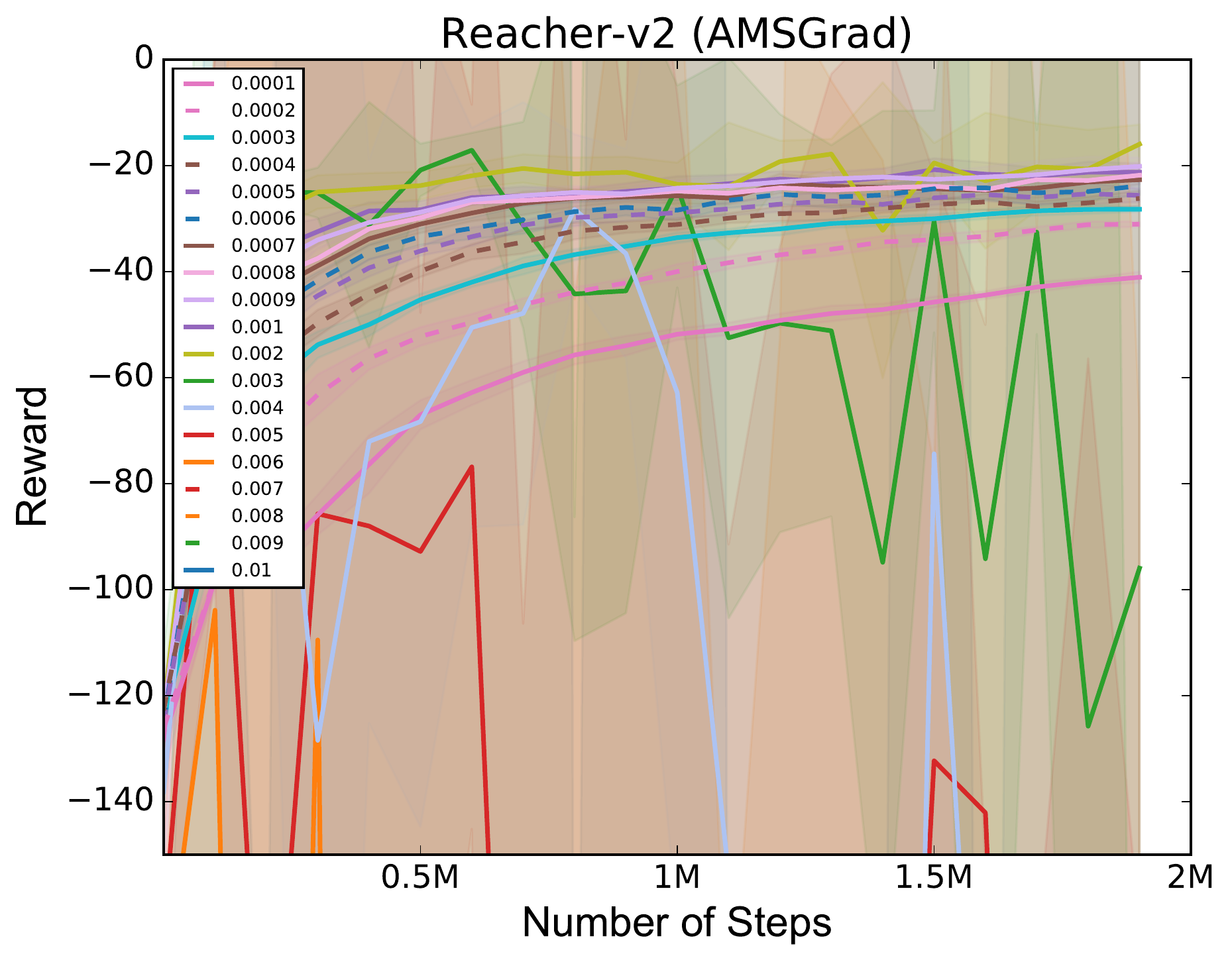}
    \includegraphics[width=.32\textwidth]{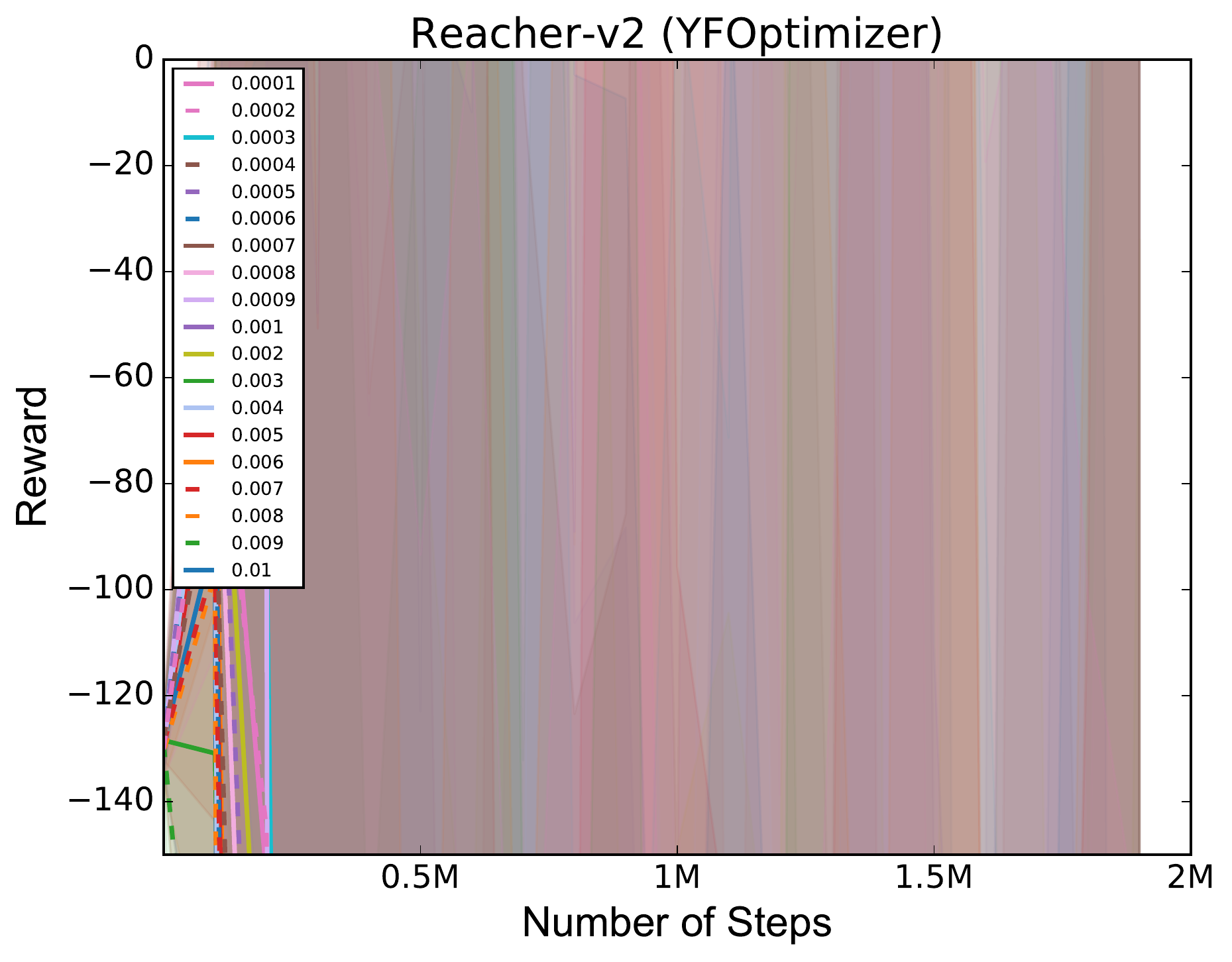}
    \caption{A2C performance across learning rates on the Reacher environment.}
        \label{fig:lr12}
\end{figure}

Figures~\ref{fig:lr1}-\ref{fig:lr12} show the per algorithm performance across learning rates. We note that in some cases where adaptive methods diverge the variance makes the graphs difficult to read. This gives further indication of just how brittle adaptive methods can be at high learning rates. We also note that while YellowFin is generally stable throughout most settings, as seen in Figure~\ref{fig:lr12} we were unable to get convergence at any learning rate on the Reacher environment with A2C.

\begin{figure}[H]
    \centering
    \includegraphics[width=.32\textwidth]{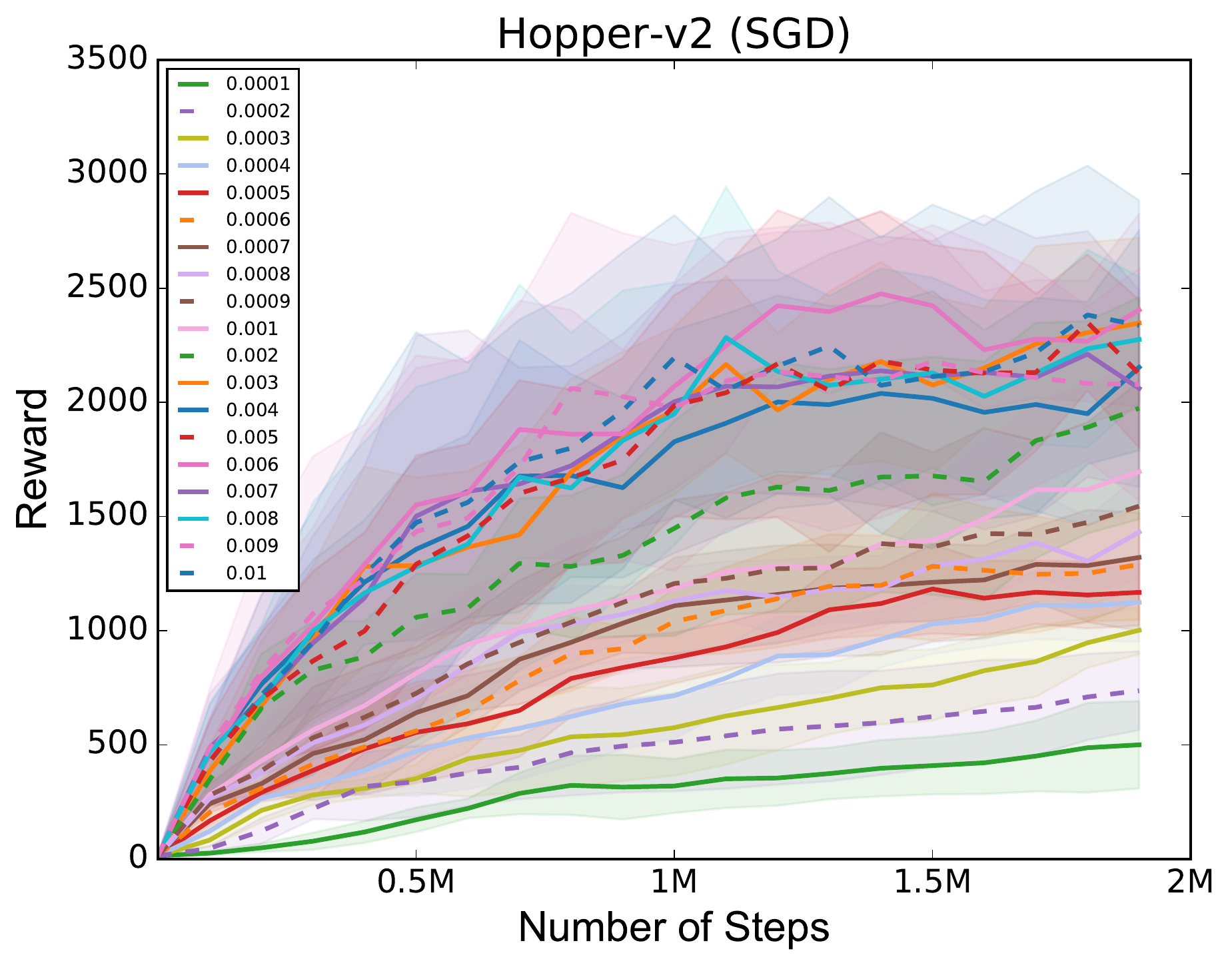}
    \includegraphics[width=.32\textwidth]{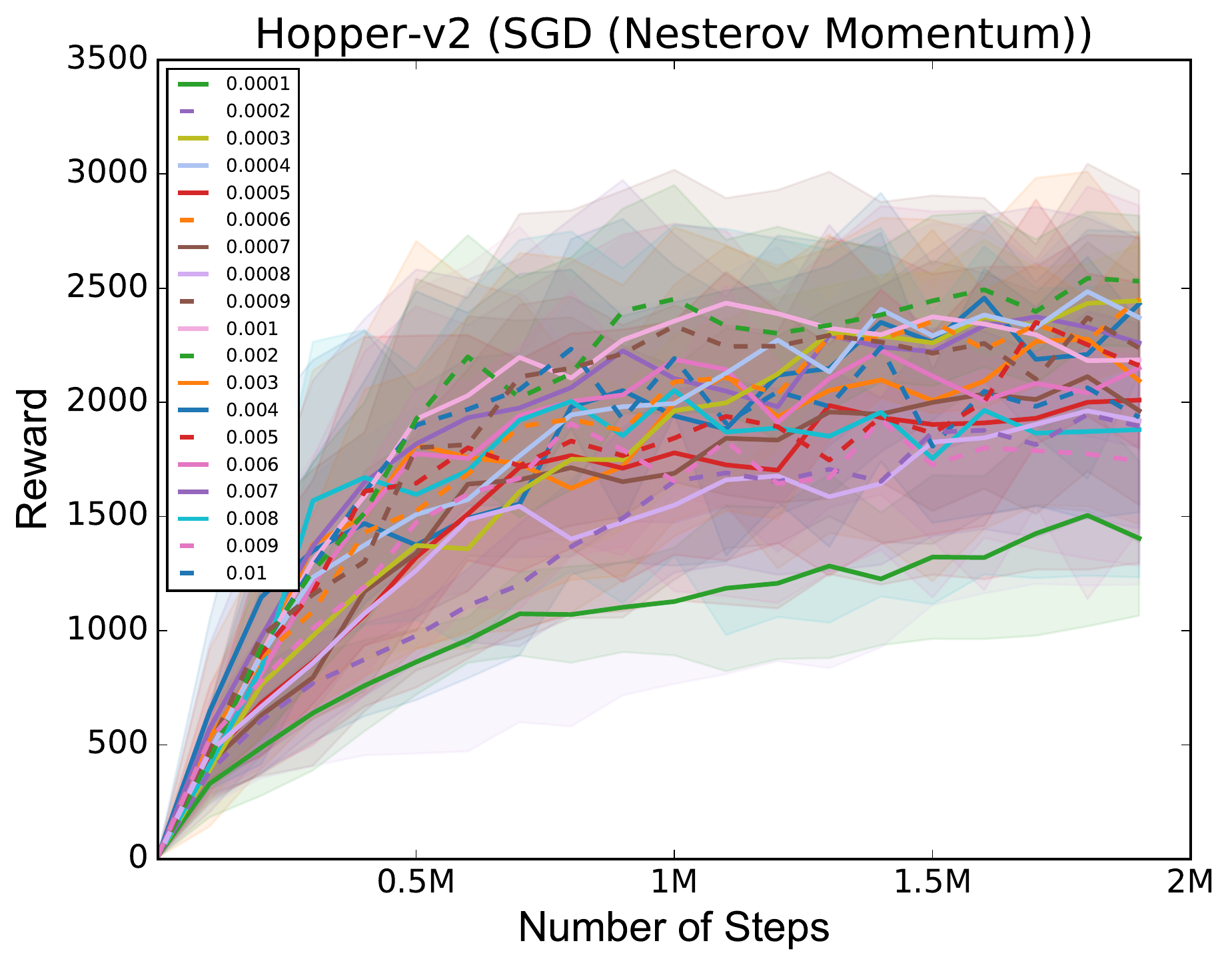}
    \includegraphics[width=.32\textwidth]{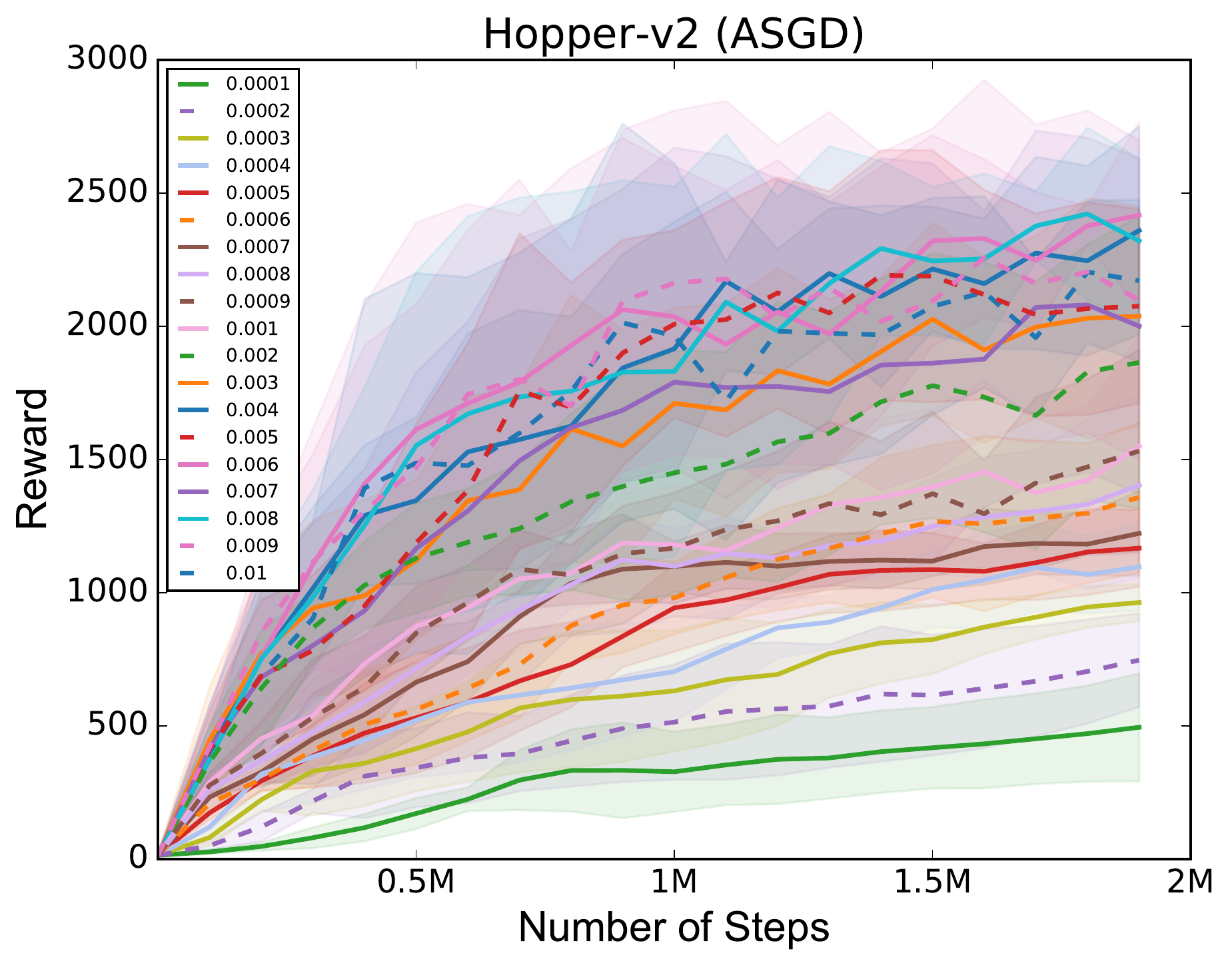}
    \includegraphics[width=.32\textwidth]{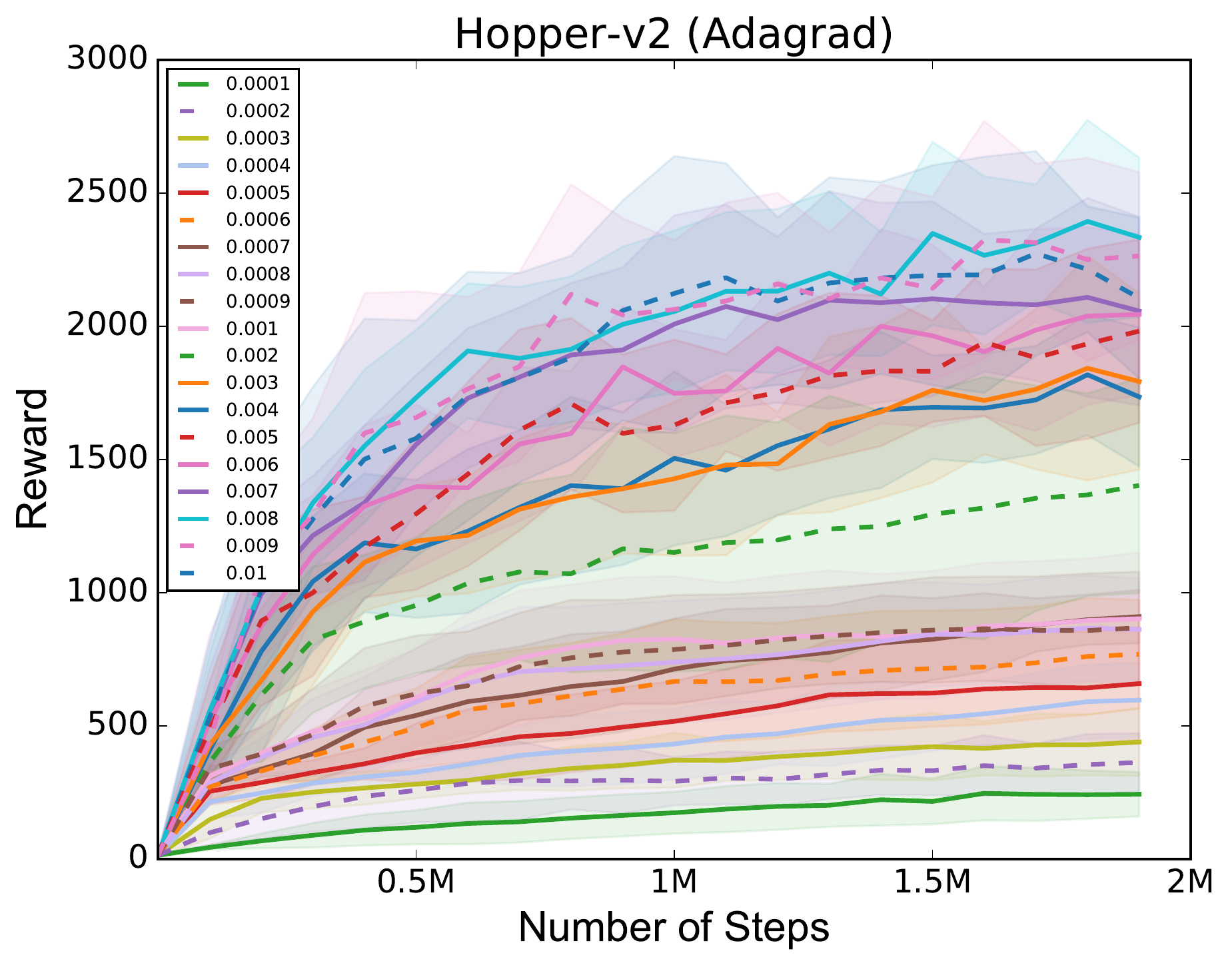}
    \includegraphics[width=.32\textwidth]{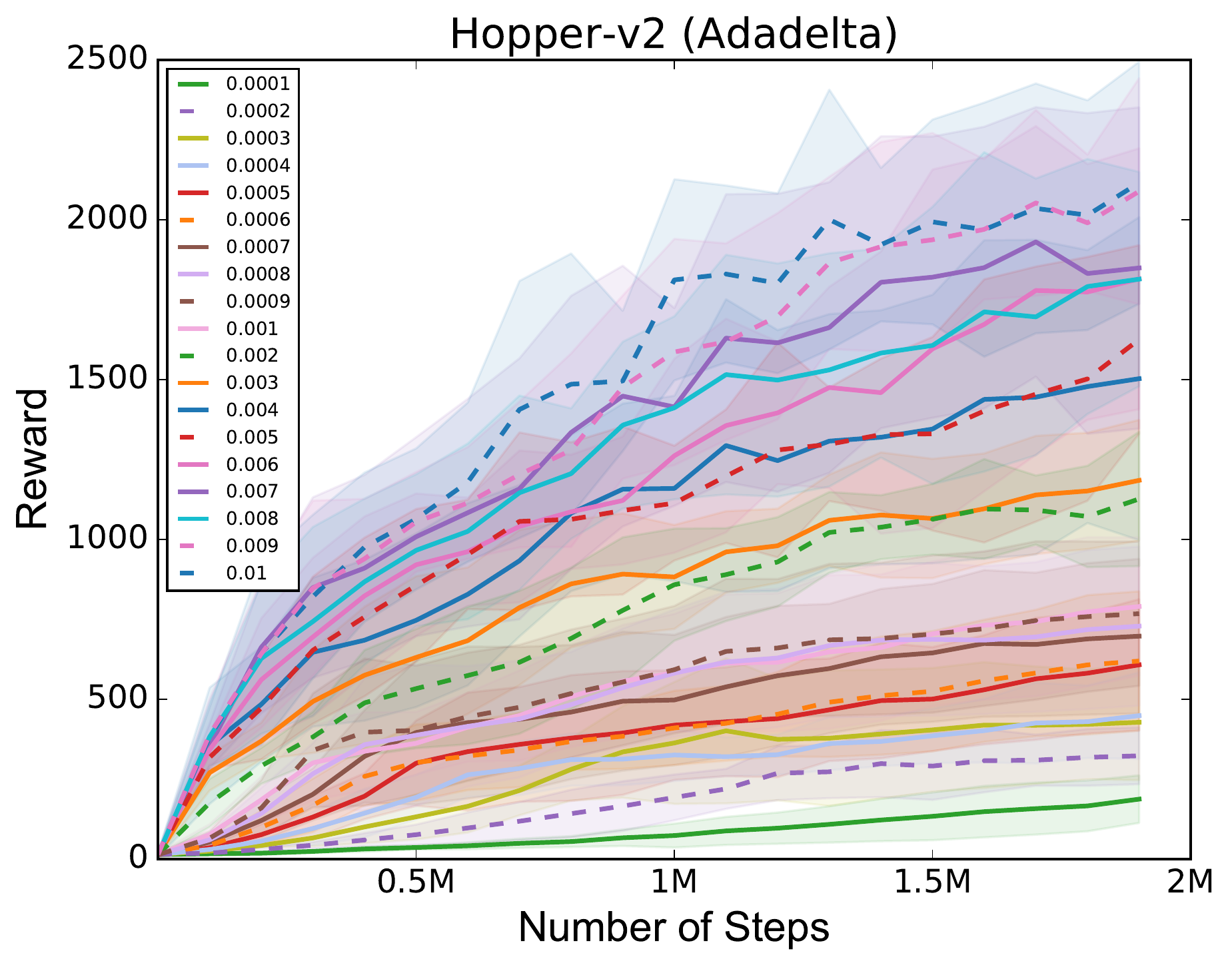}
    \includegraphics[width=.32\textwidth]{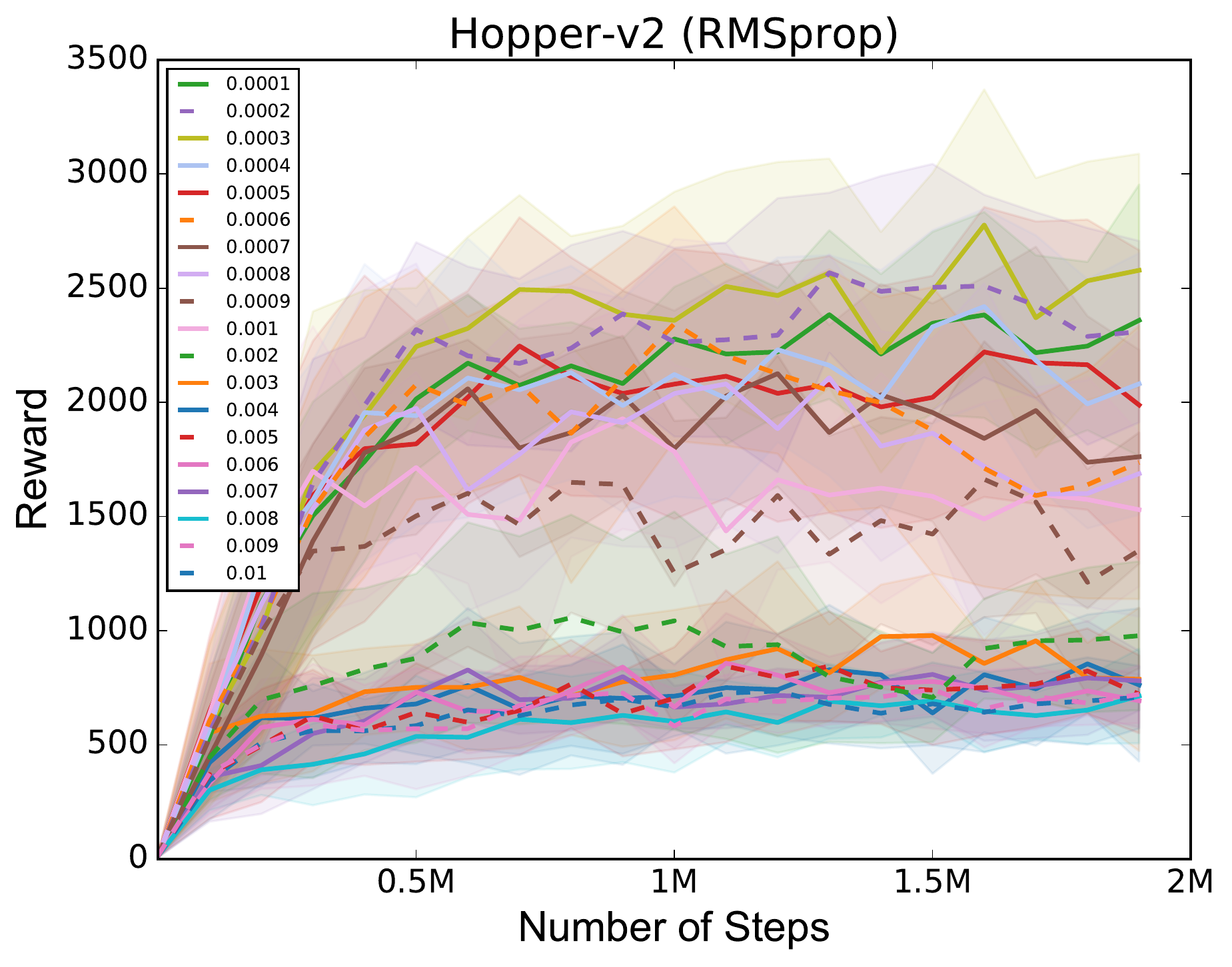}
    \includegraphics[width=.32\textwidth]{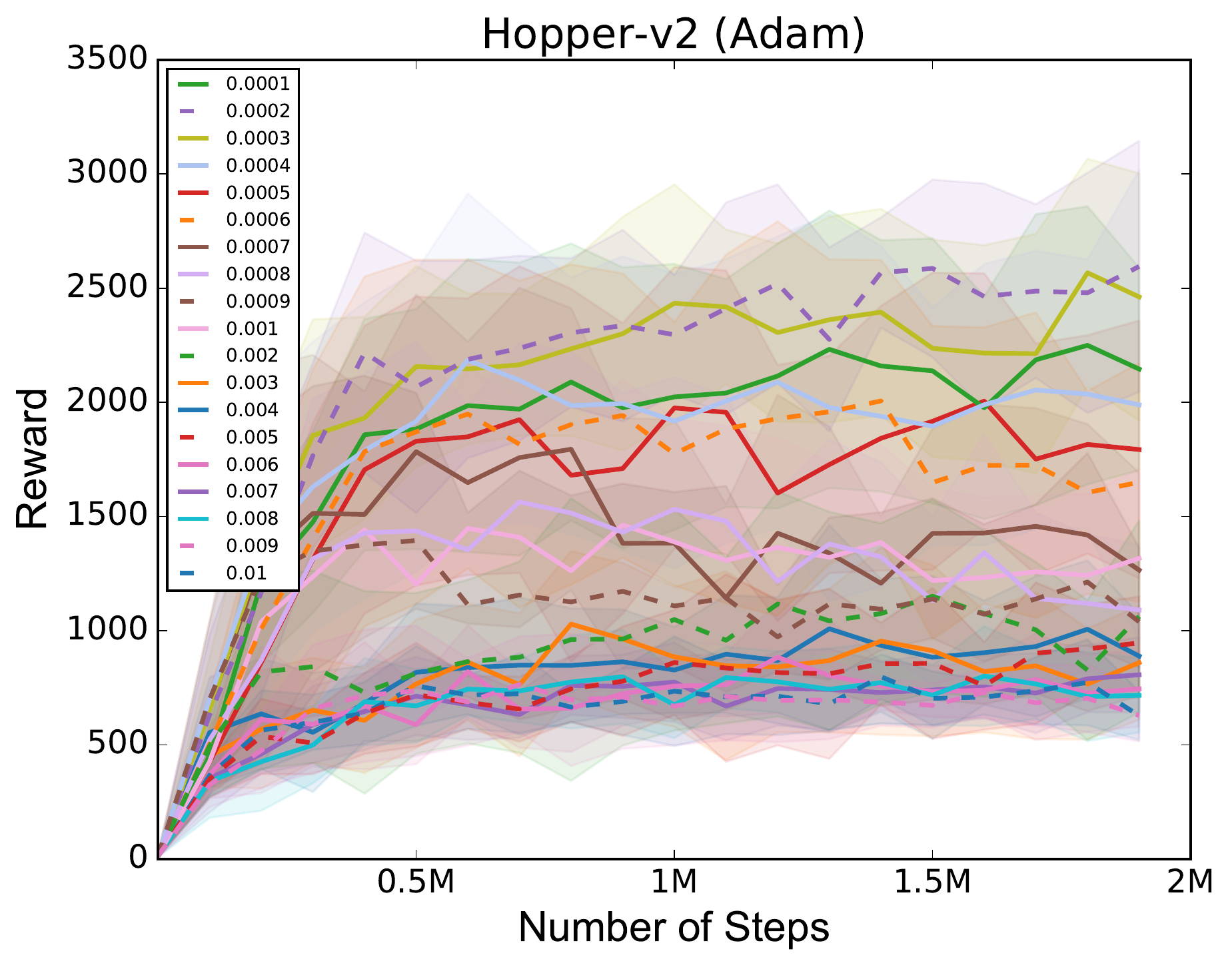}
    \includegraphics[width=.32\textwidth]{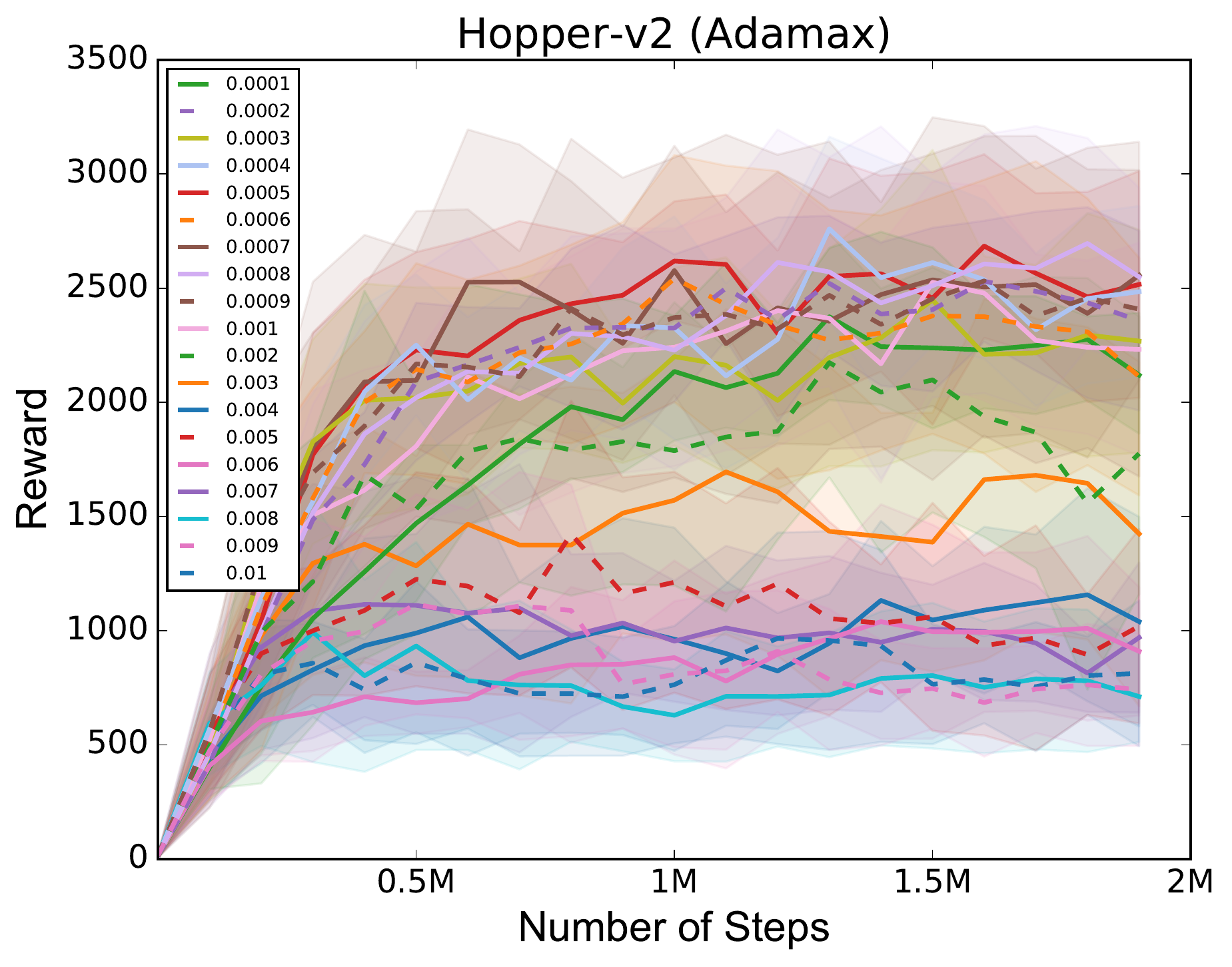}
    \includegraphics[width=.32\textwidth]{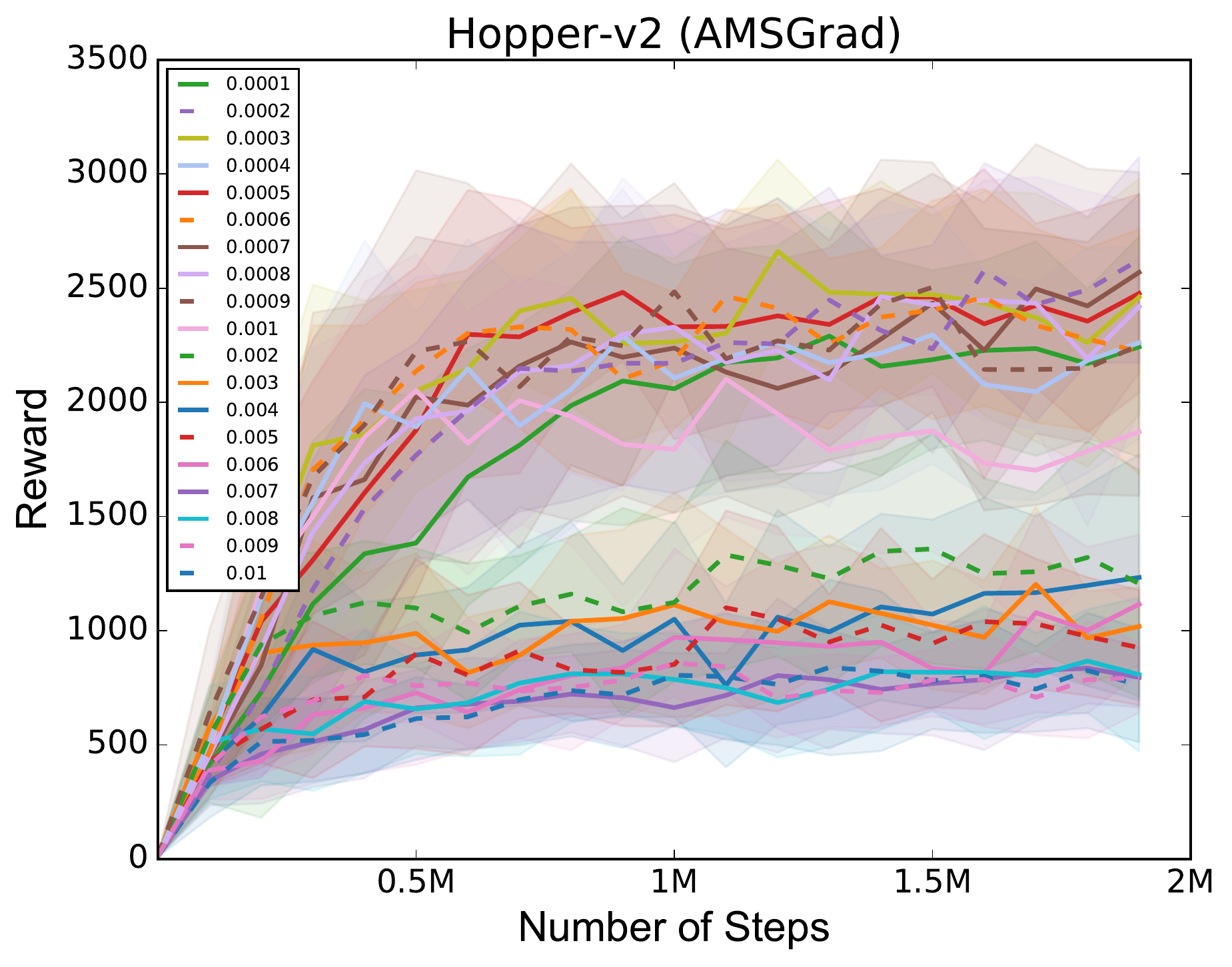}
    \includegraphics[width=.32\textwidth]{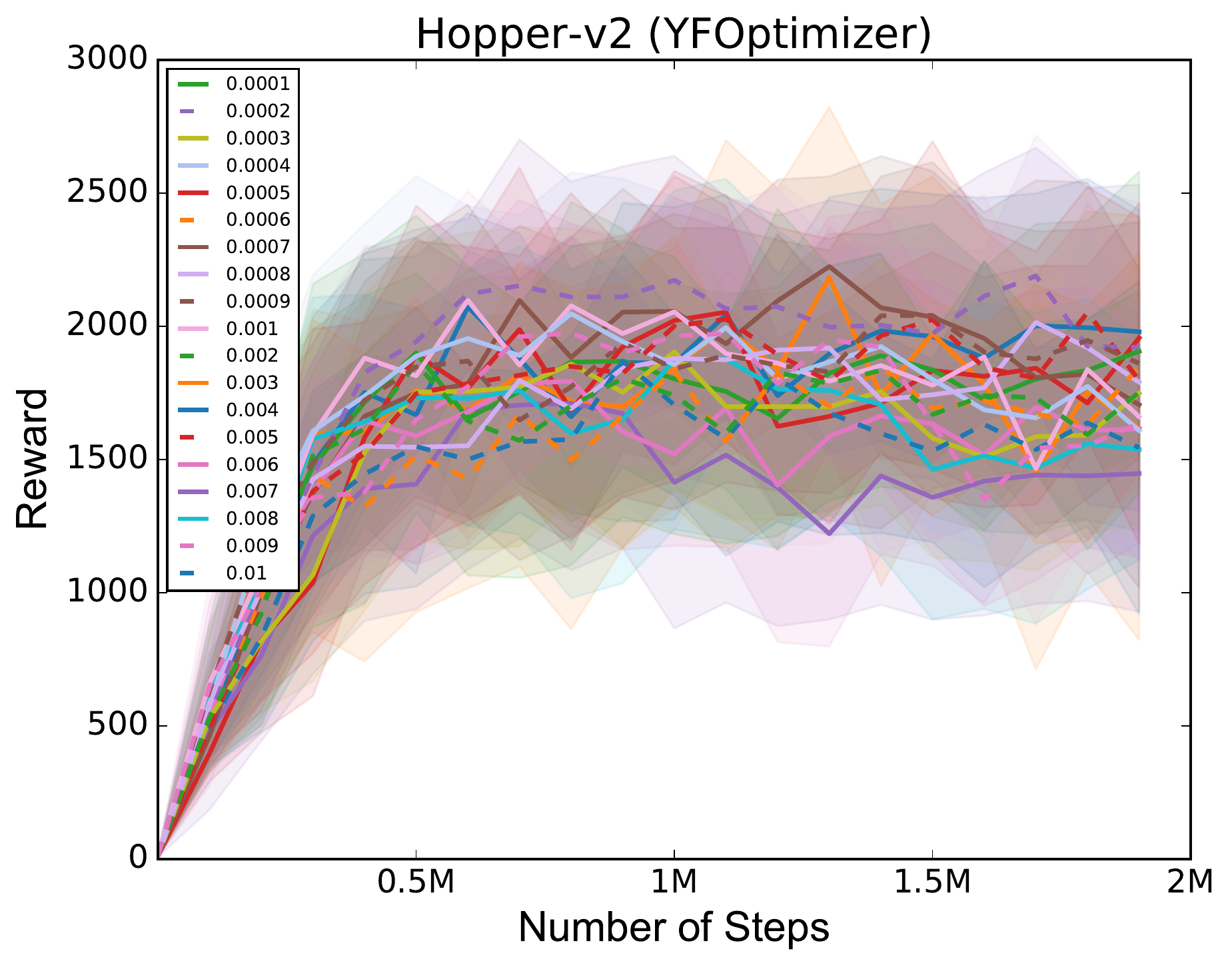}
    \caption{PPO performance across learning rates on the Hopper environment.}
\end{figure}

\begin{figure}[H]
    \centering
    \includegraphics[width=.32\textwidth]{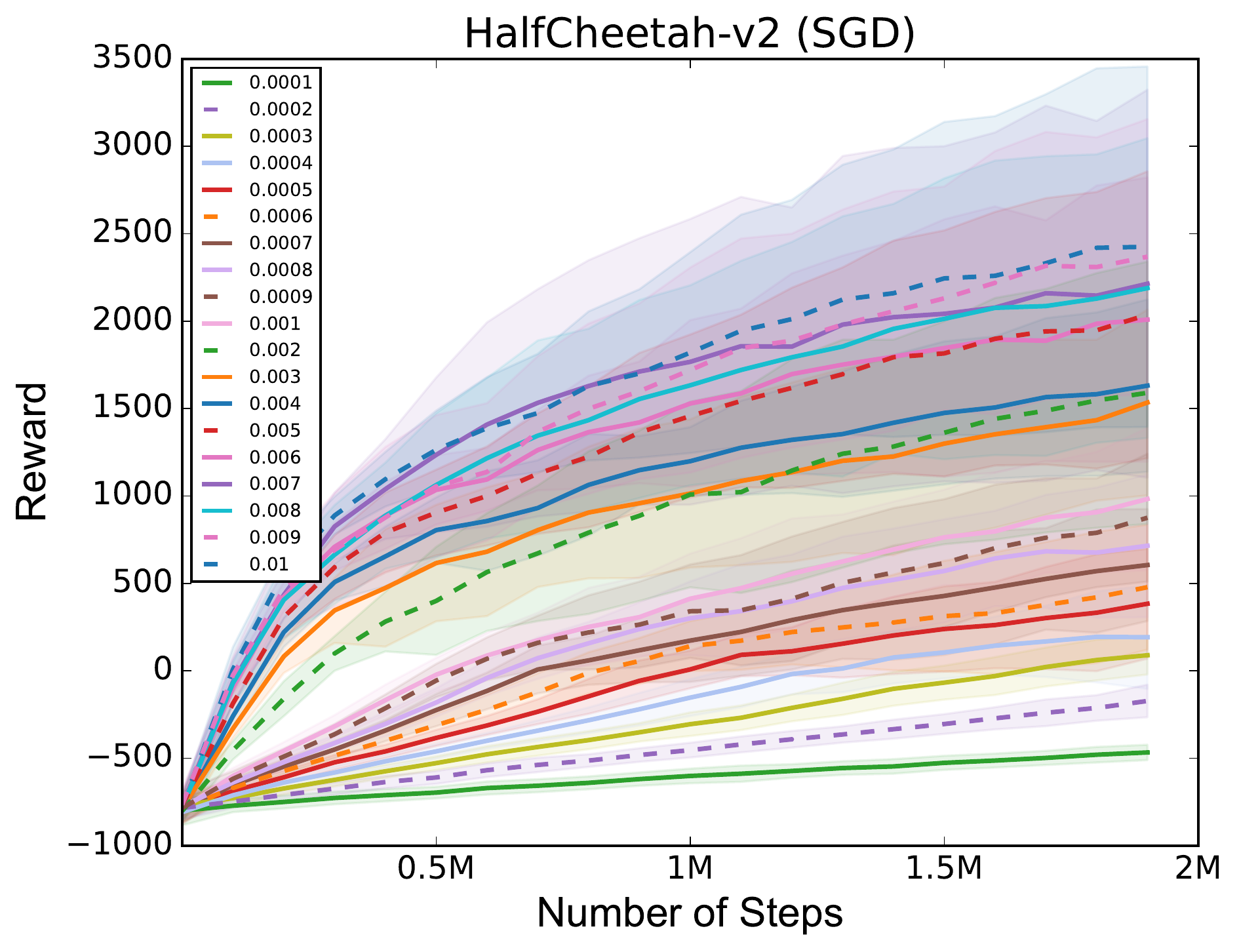}
    \includegraphics[width=.32\textwidth]{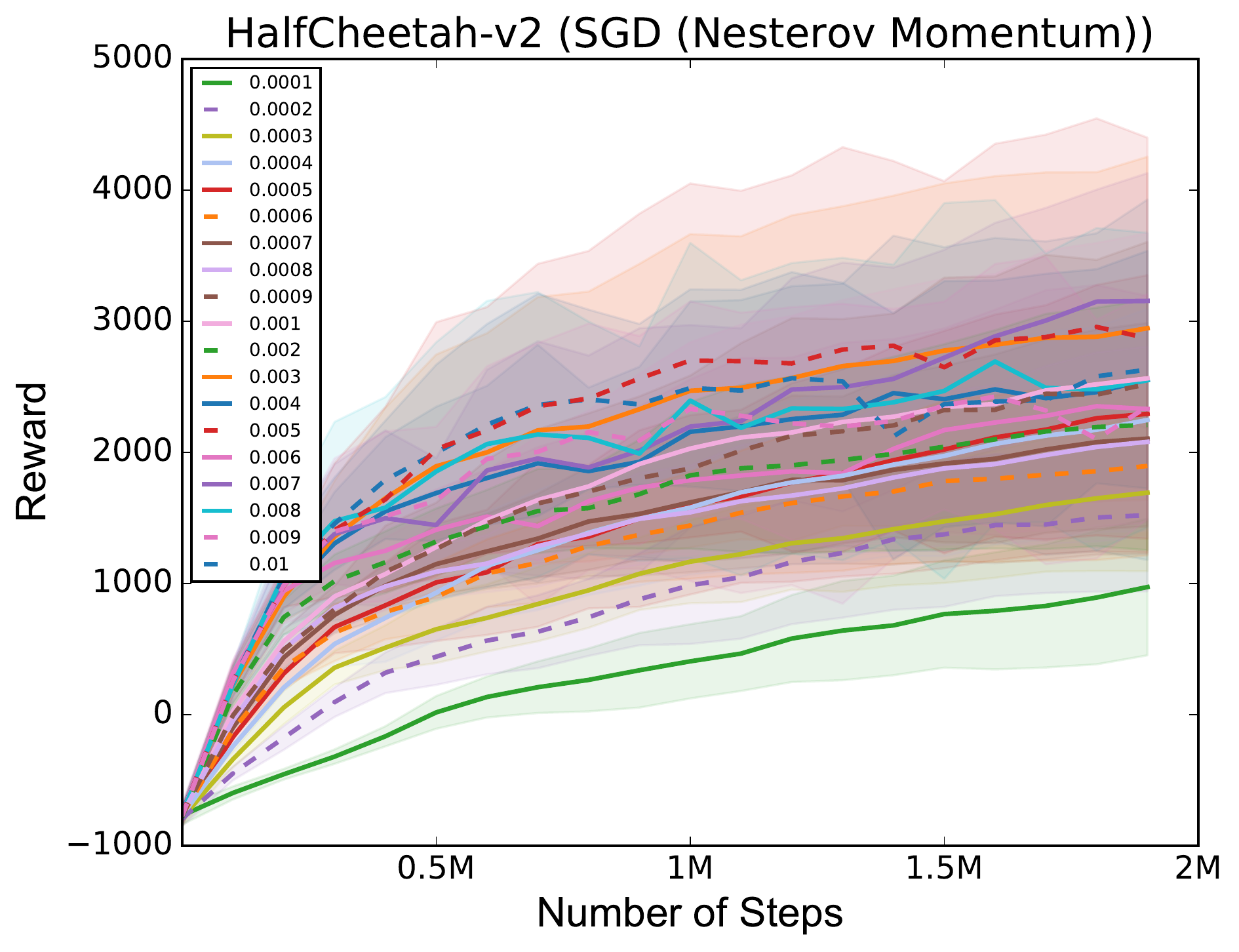}
    \includegraphics[width=.32\textwidth]{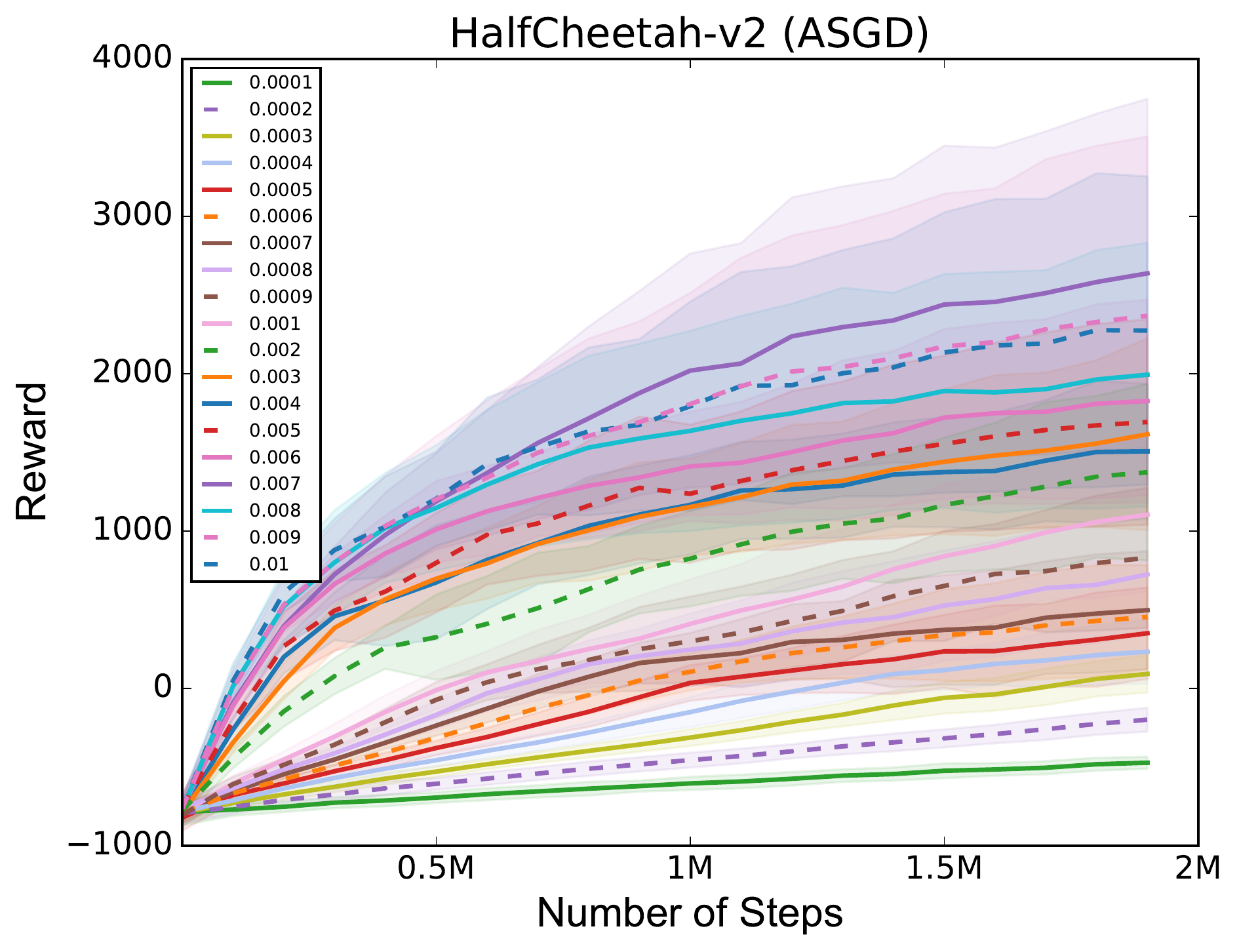}
    \includegraphics[width=.32\textwidth]{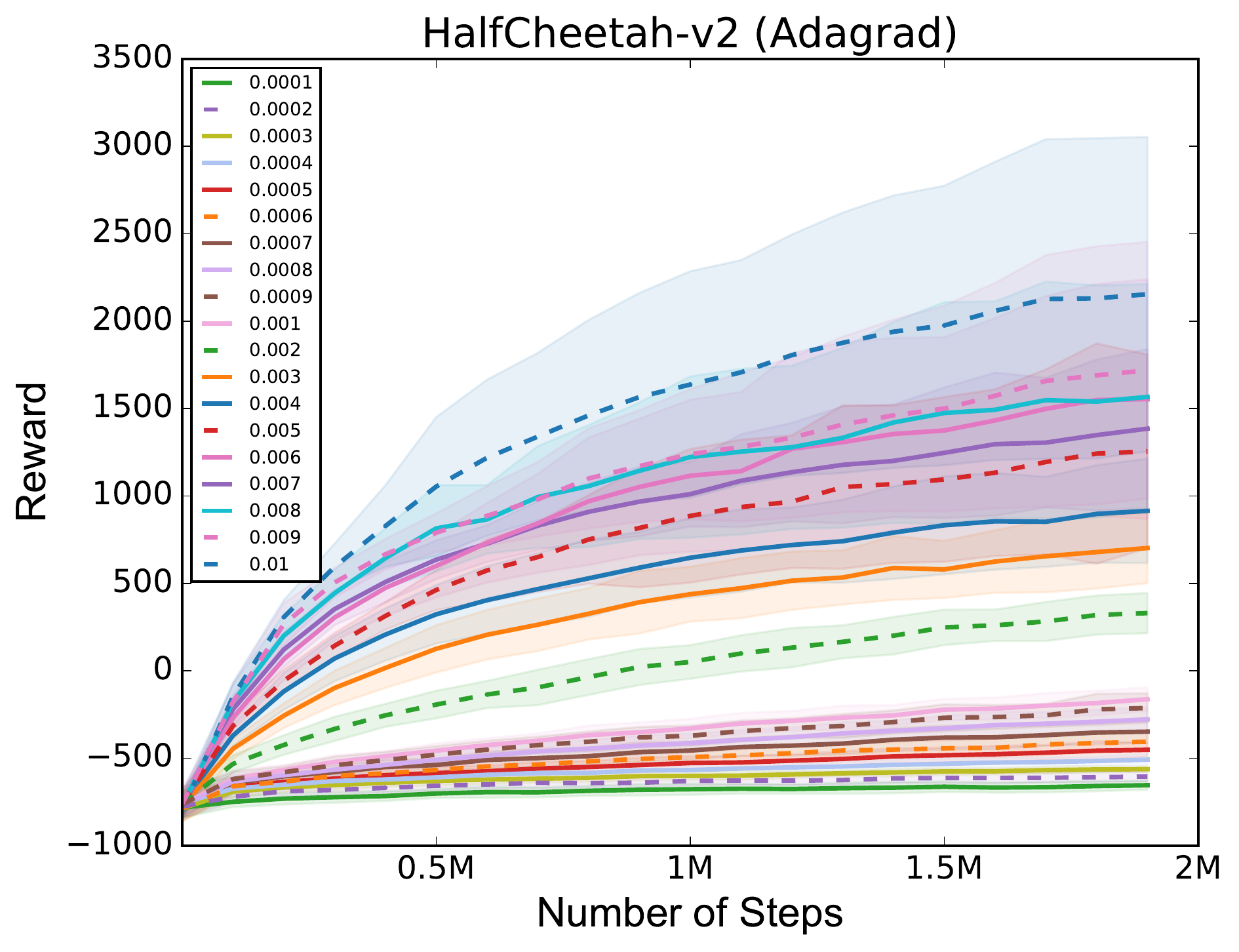}
    \includegraphics[width=.32\textwidth]{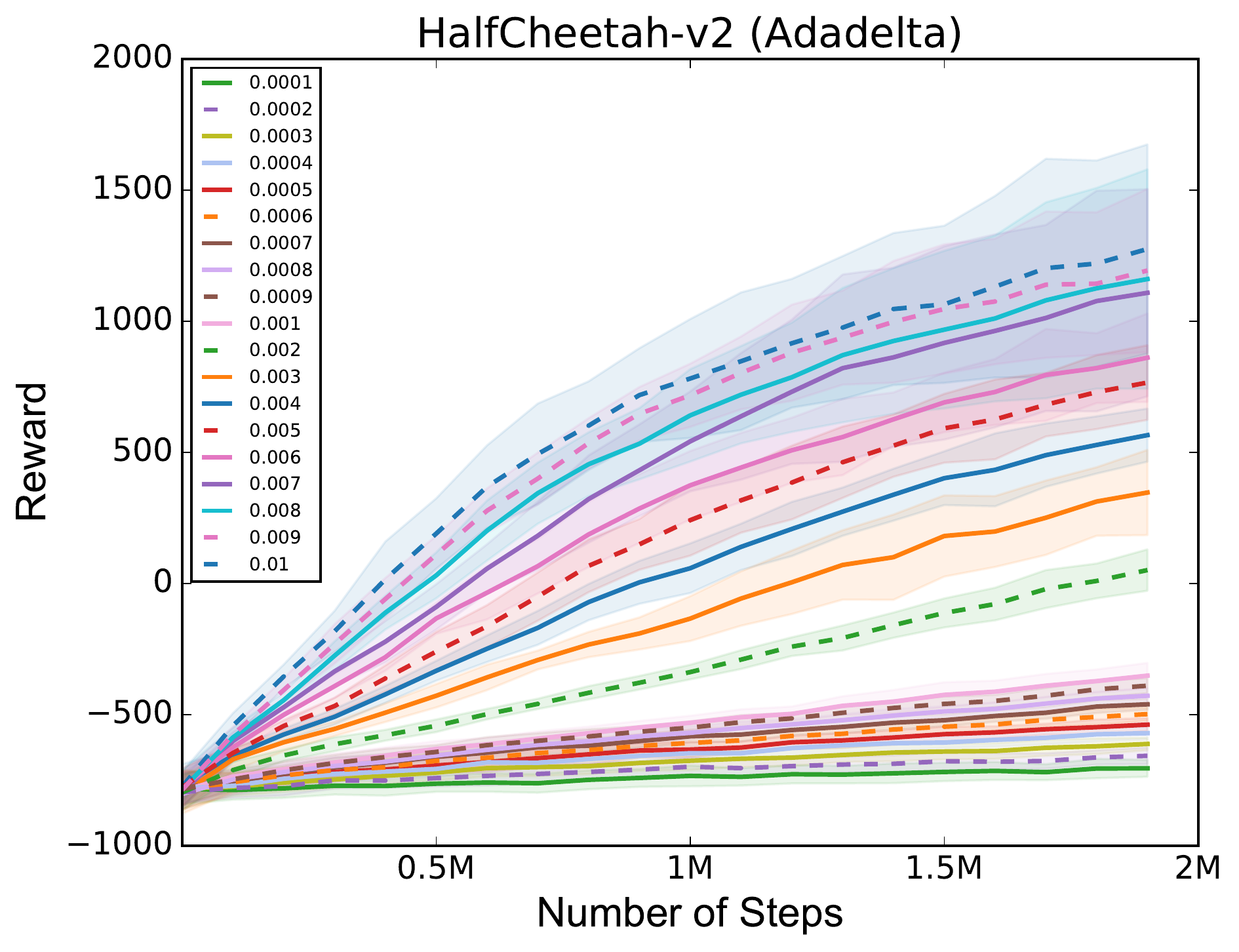}
    \includegraphics[width=.32\textwidth]{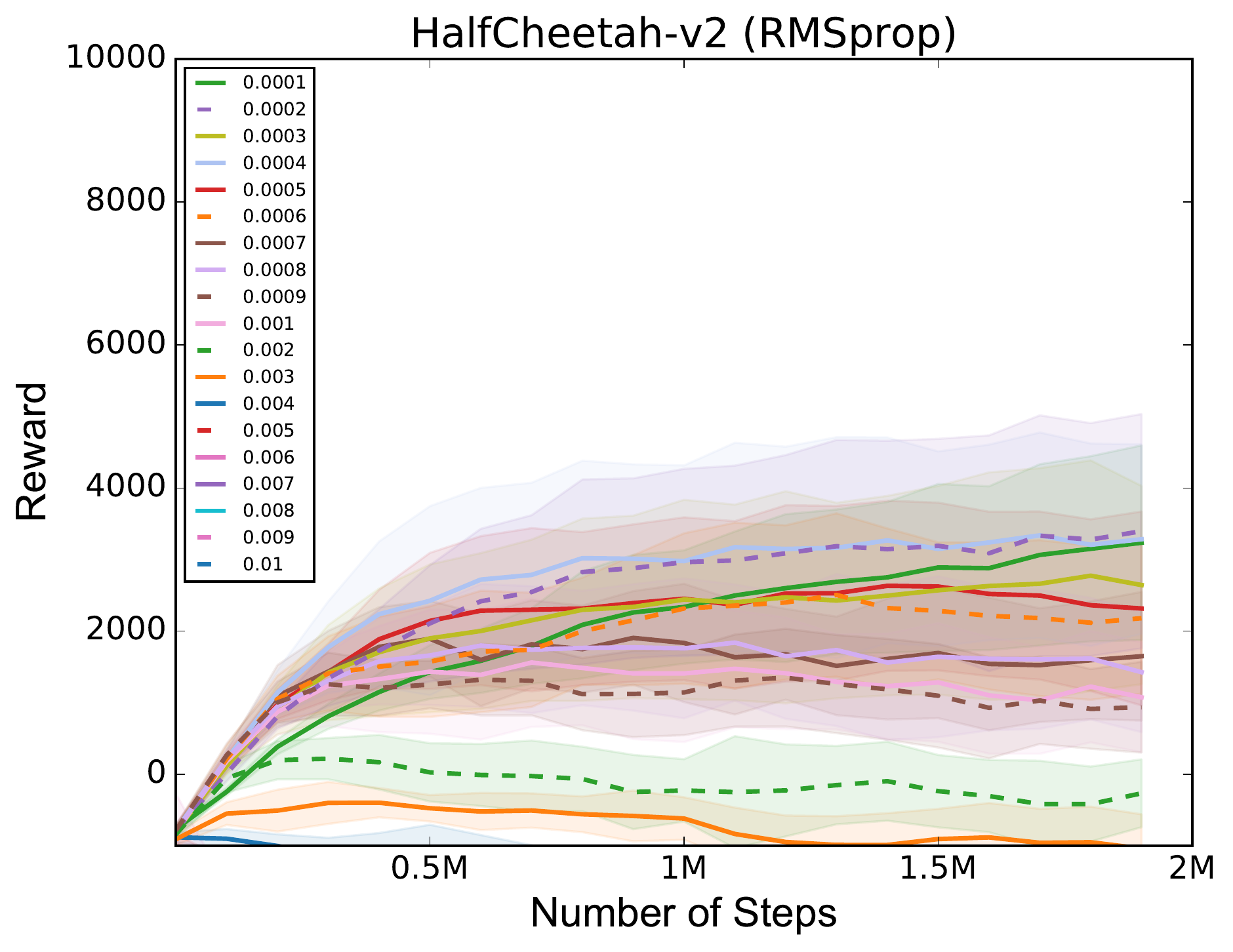}
    \includegraphics[width=.32\textwidth]{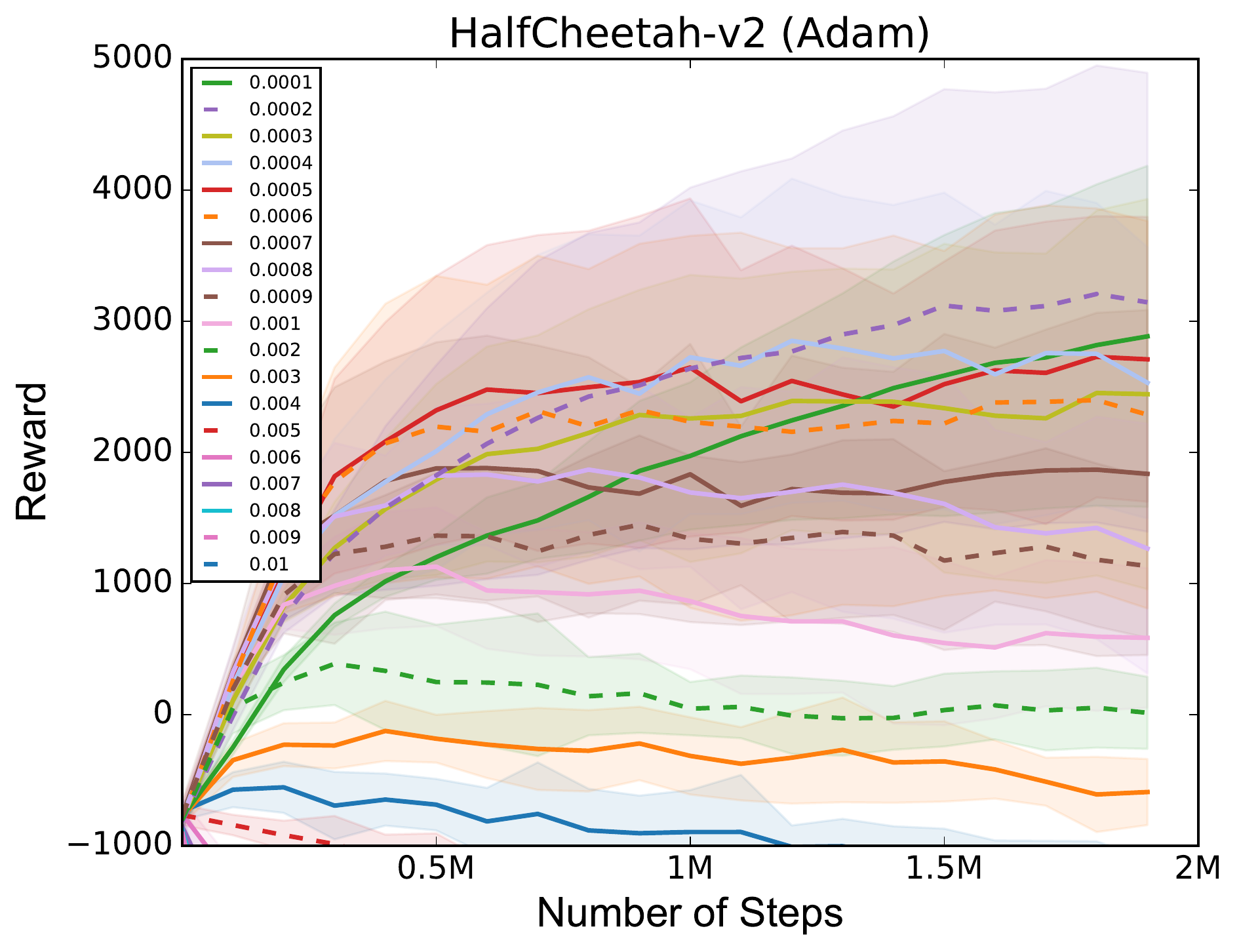}
    \includegraphics[width=.32\textwidth]{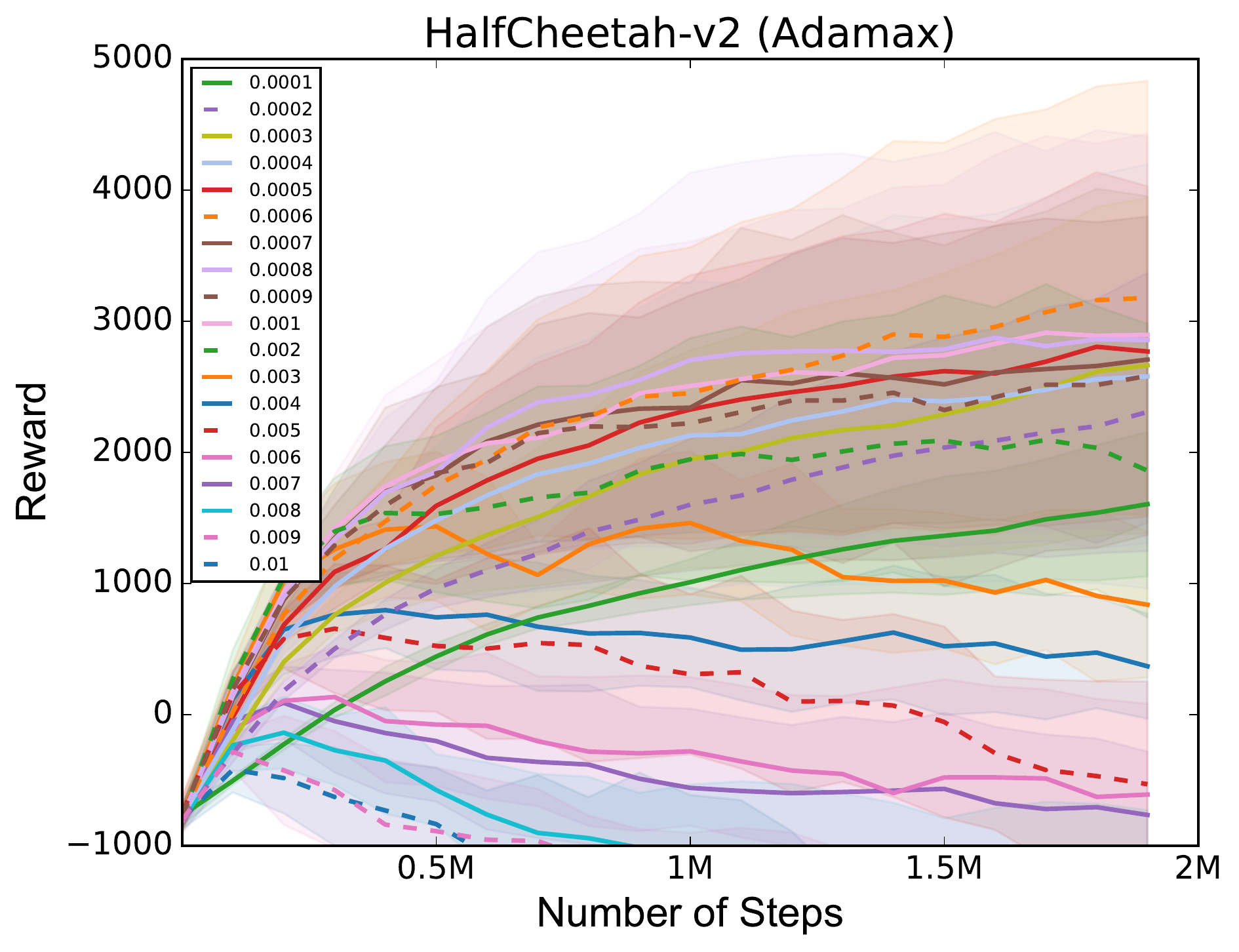}
    \includegraphics[width=.32\textwidth]{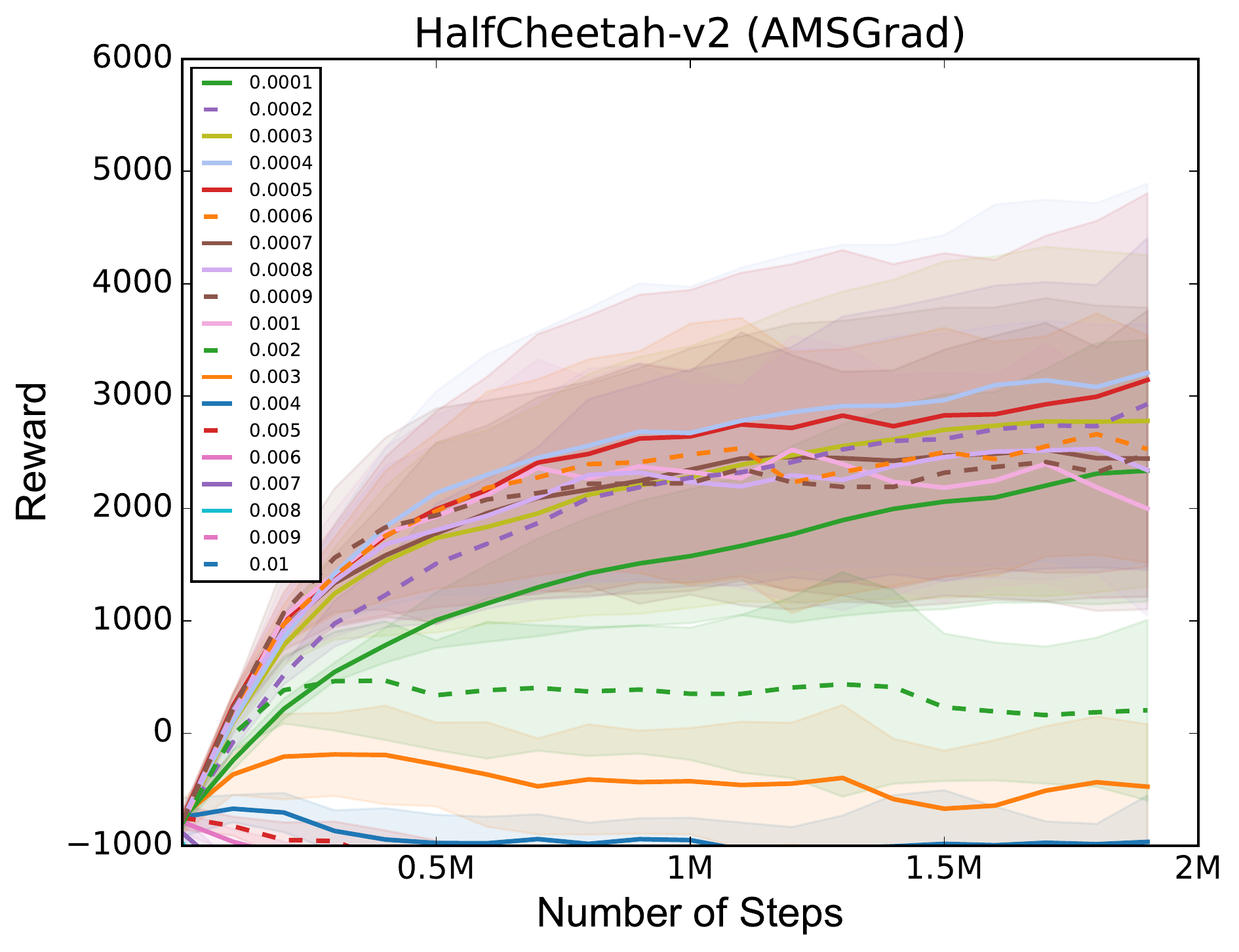}
    \includegraphics[width=.32\textwidth]{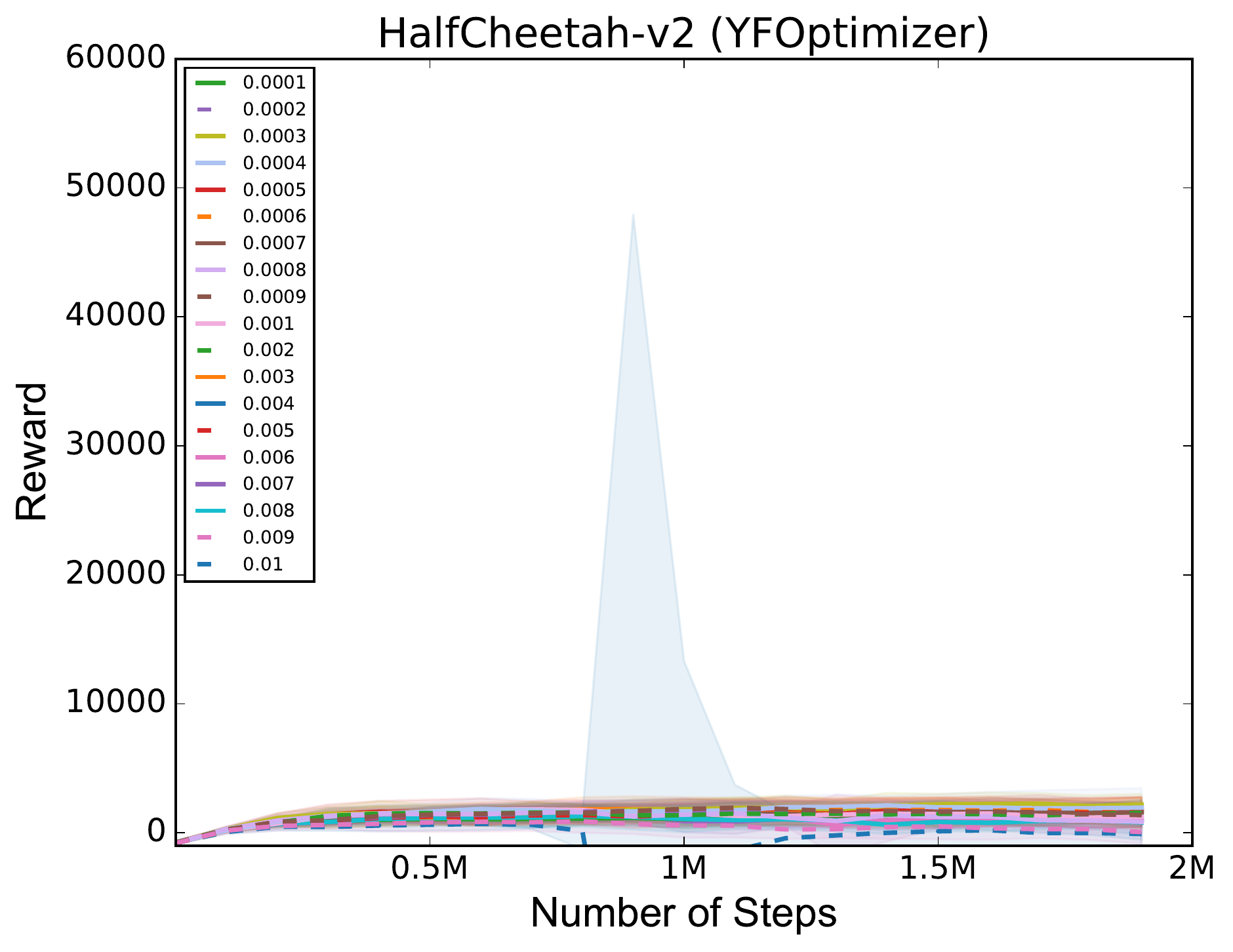}
    \caption{PPO performance across learning rates on the HalfCheetah environment.}
\end{figure}

\begin{figure}[H]
    \centering
    \includegraphics[width=.32\textwidth]{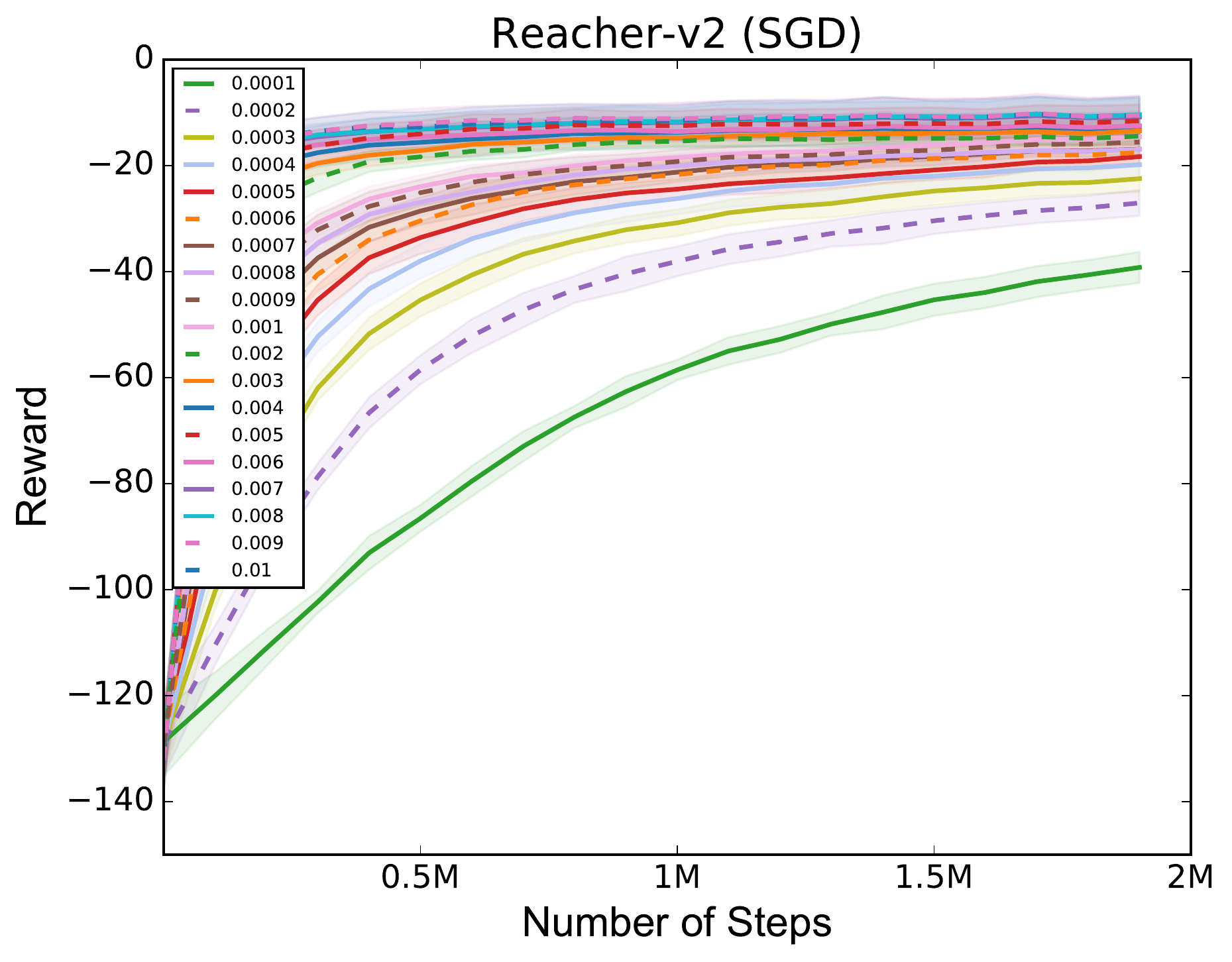}
    \includegraphics[width=.32\textwidth]{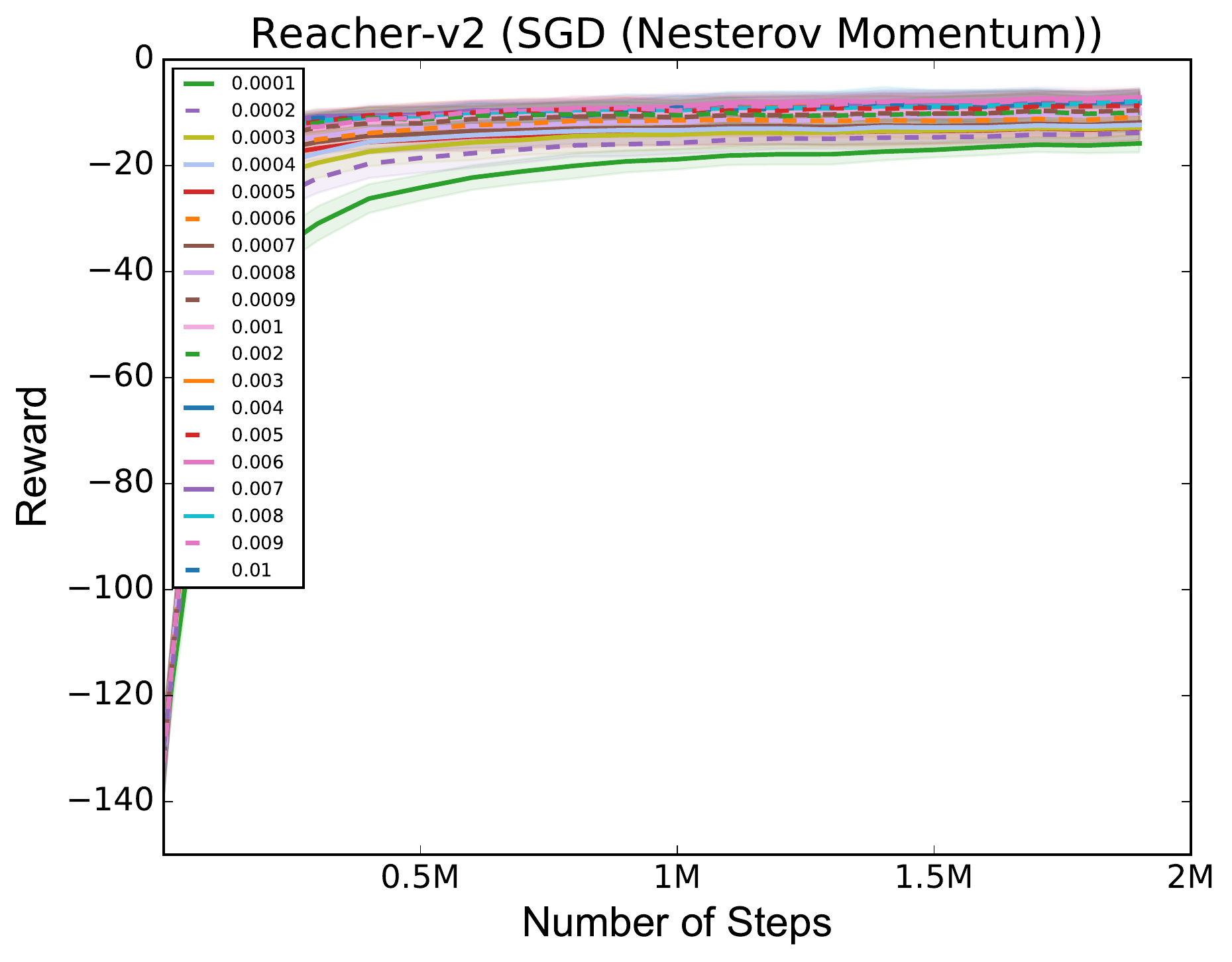}
    \includegraphics[width=.32\textwidth]{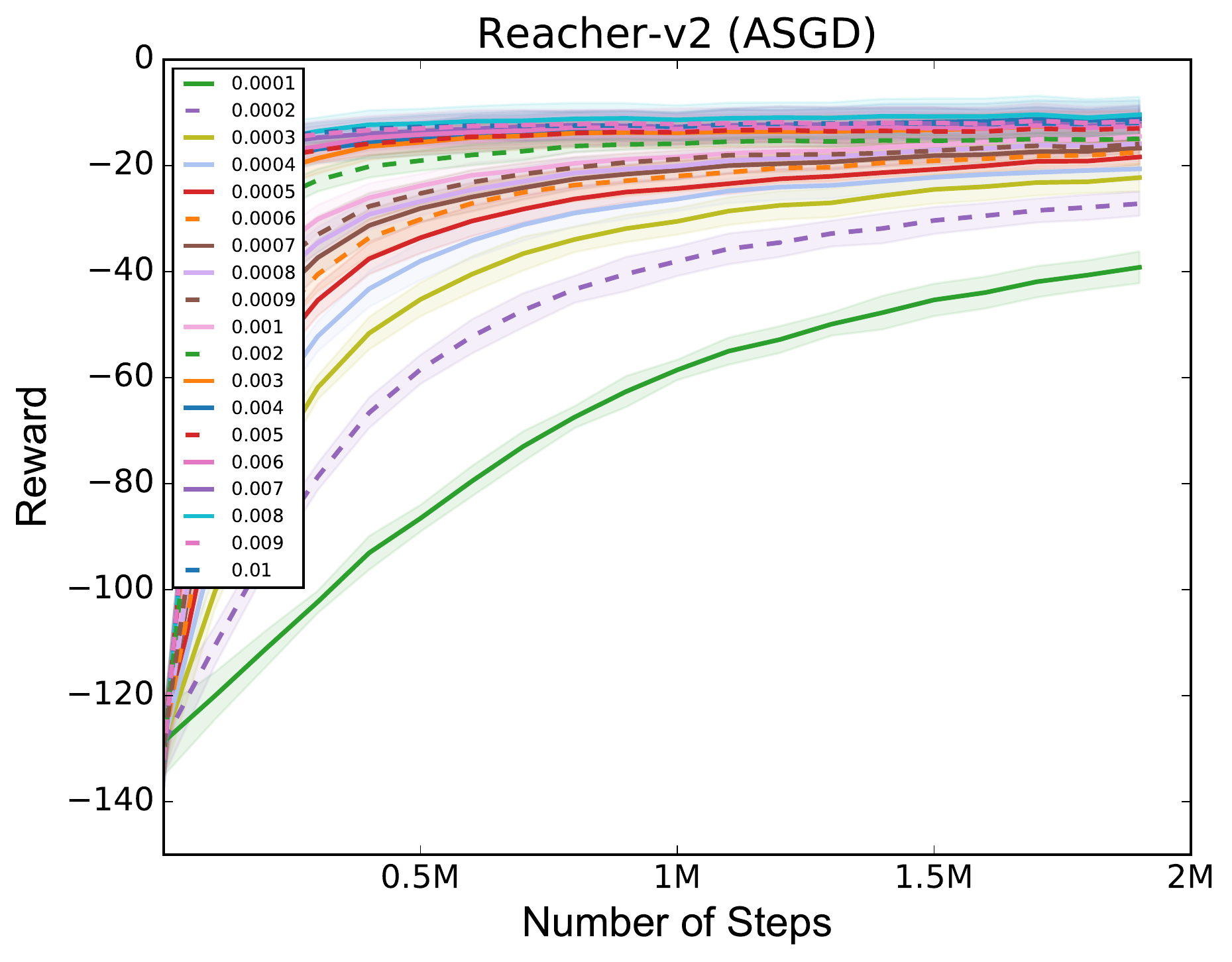}
    \includegraphics[width=.32\textwidth]{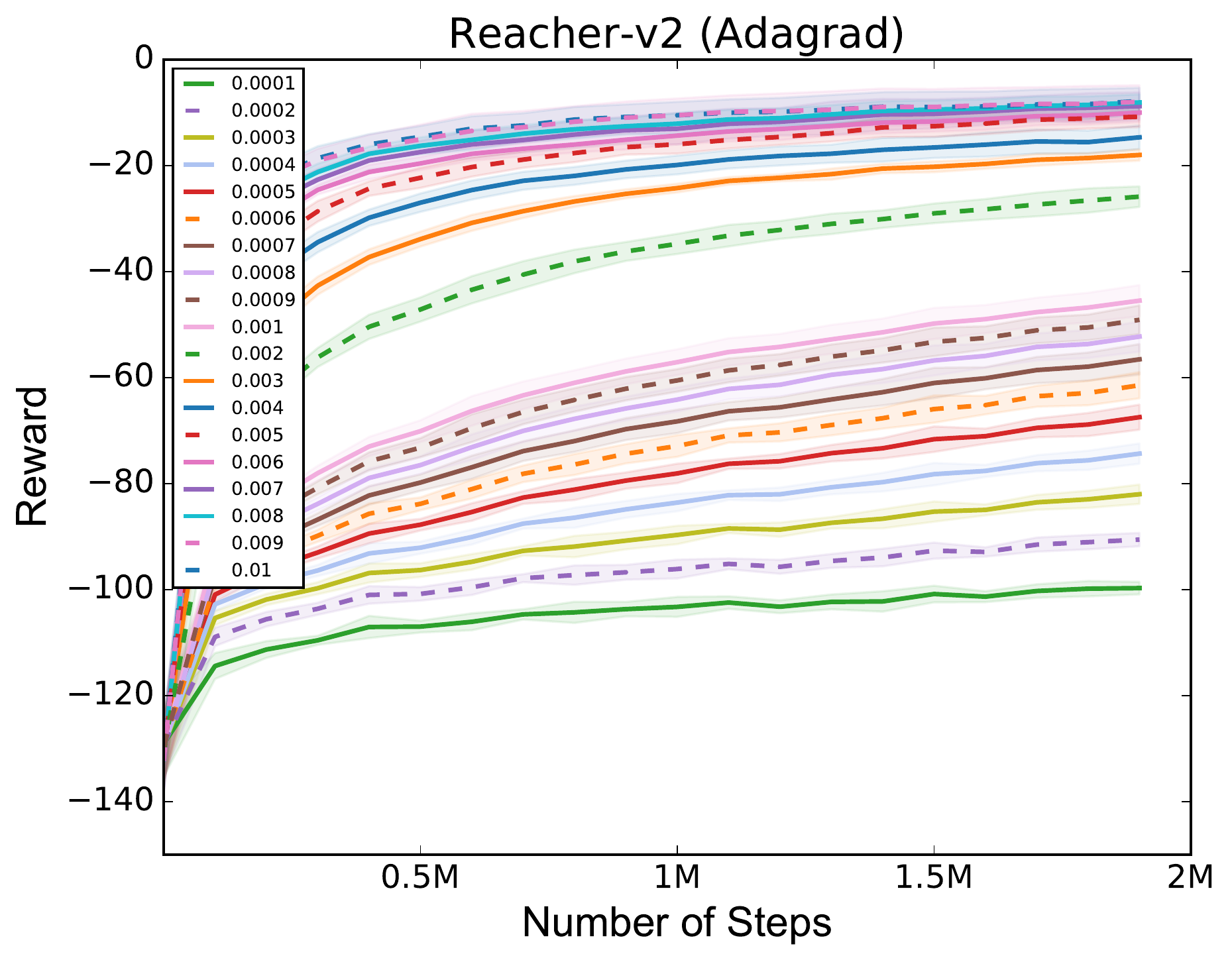}
    \includegraphics[width=.32\textwidth]{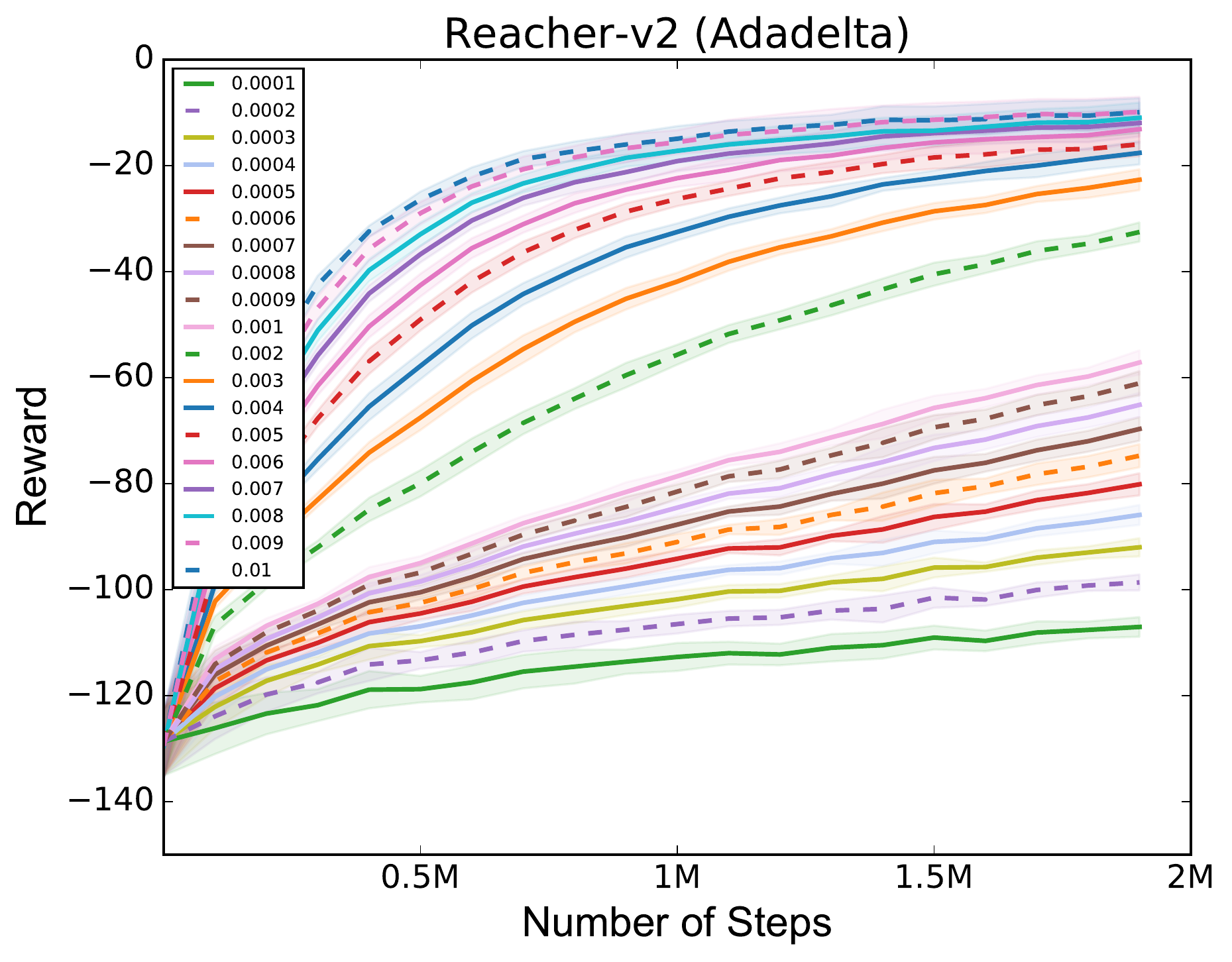}
    \includegraphics[width=.32\textwidth]{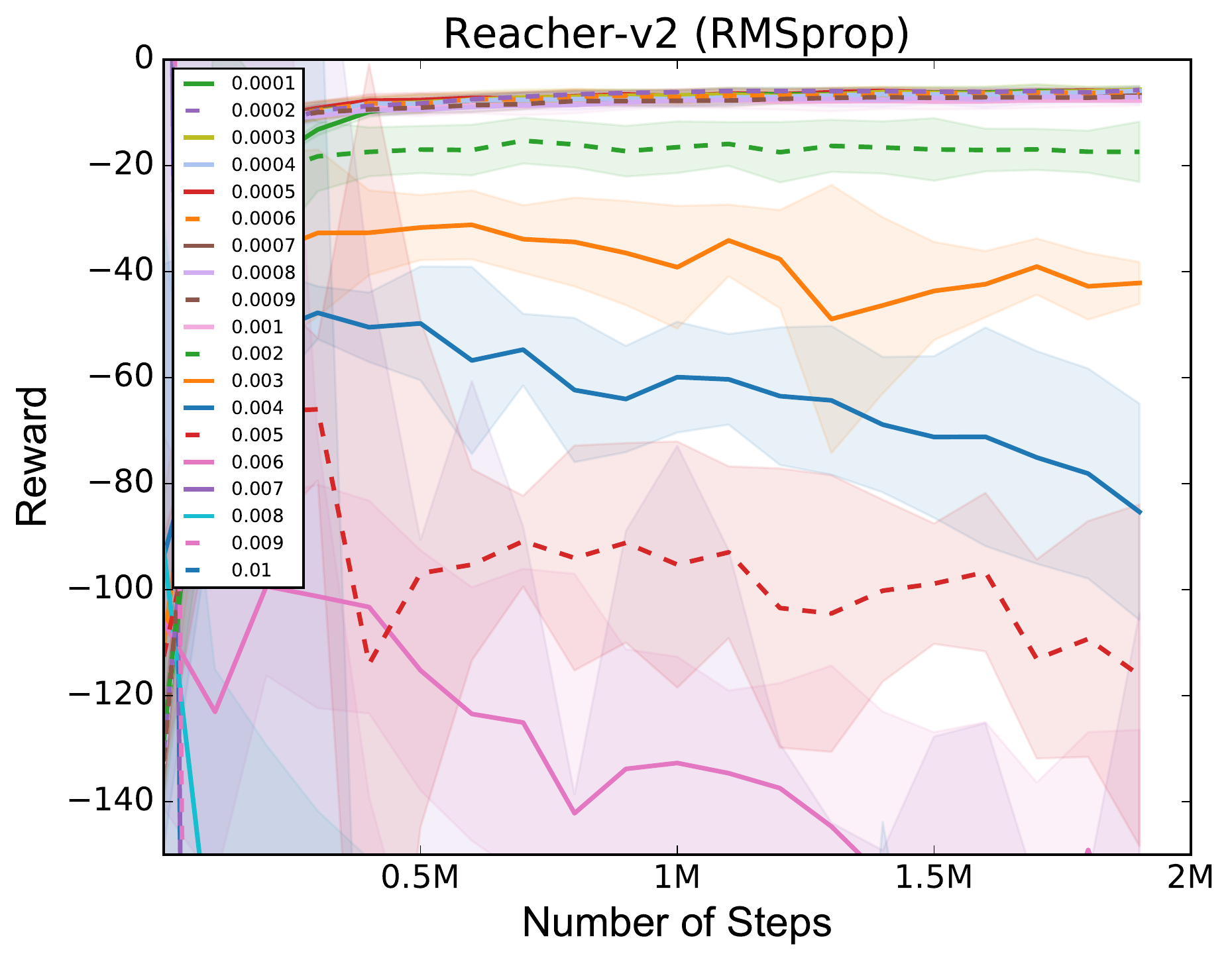}
    \includegraphics[width=.32\textwidth]{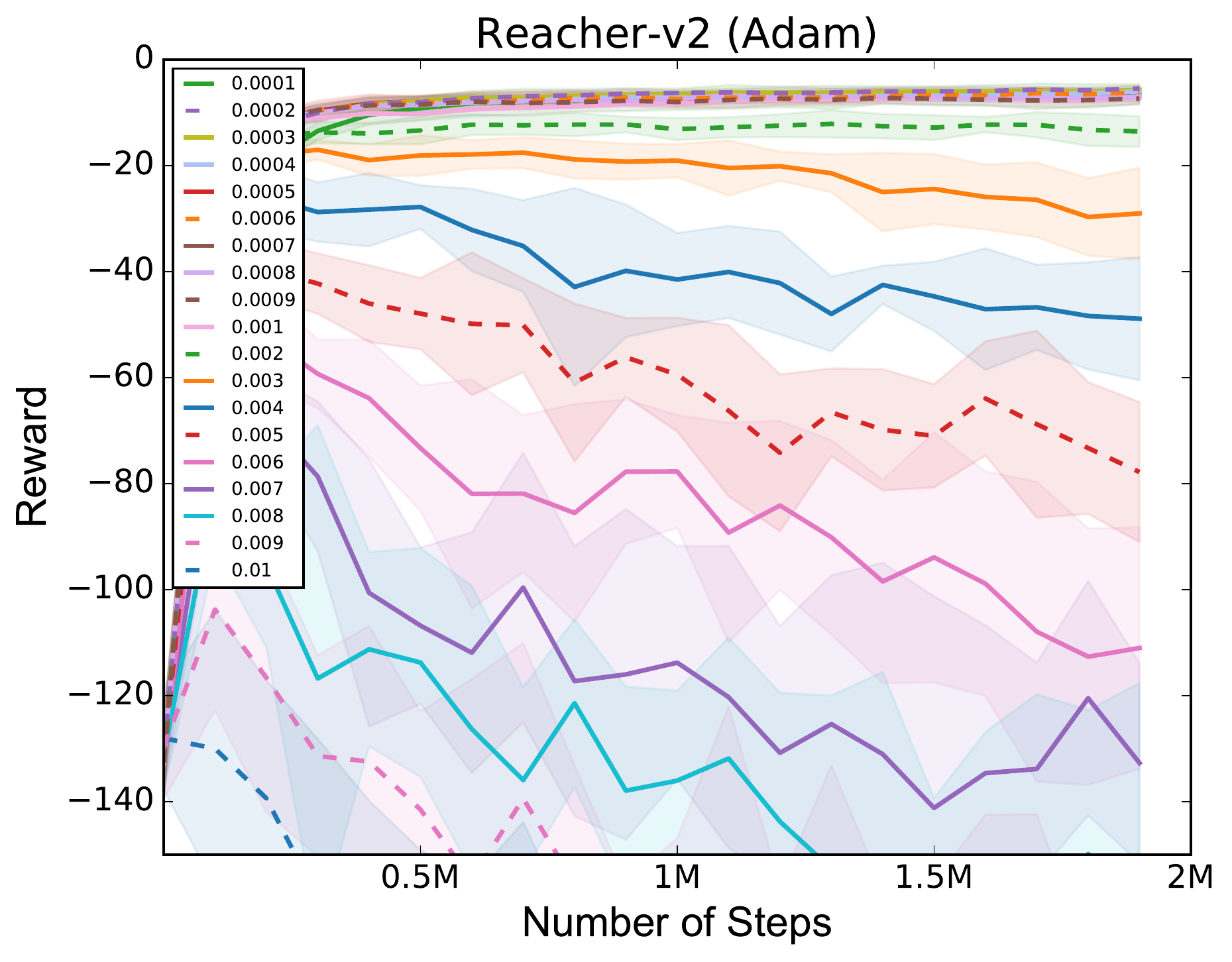}
    \includegraphics[width=.32\textwidth]{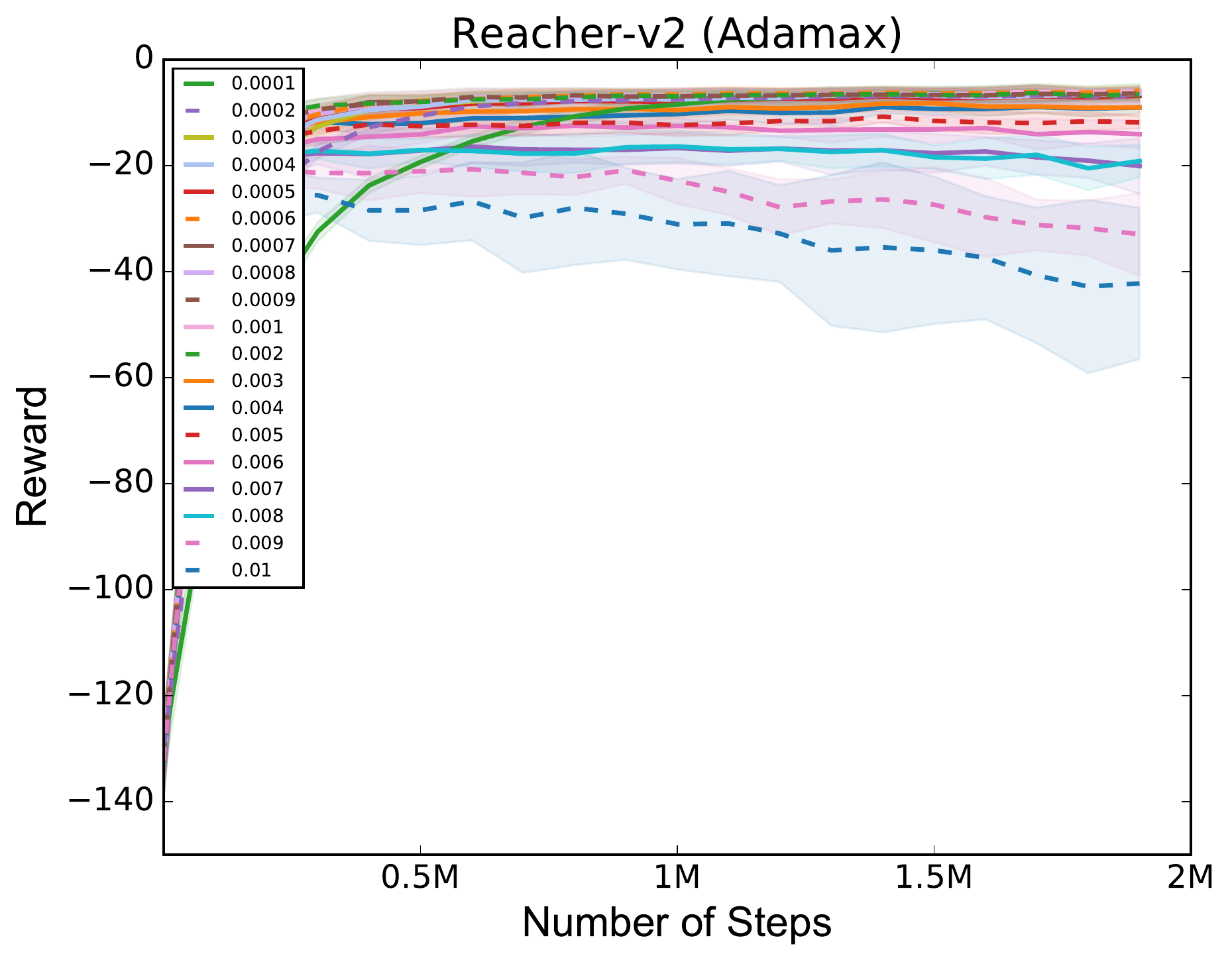}
    \includegraphics[width=.32\textwidth]{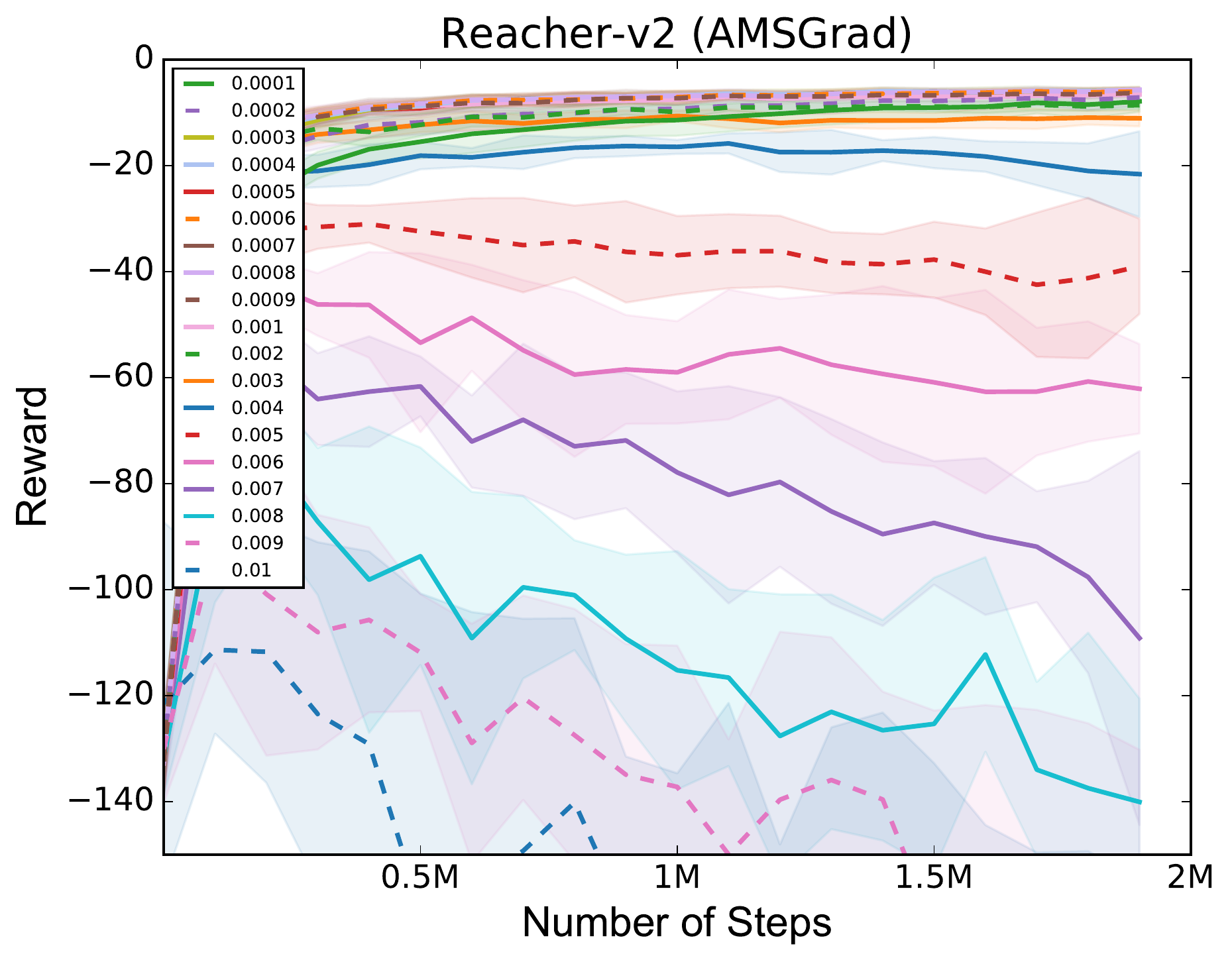}
    \includegraphics[width=.32\textwidth]{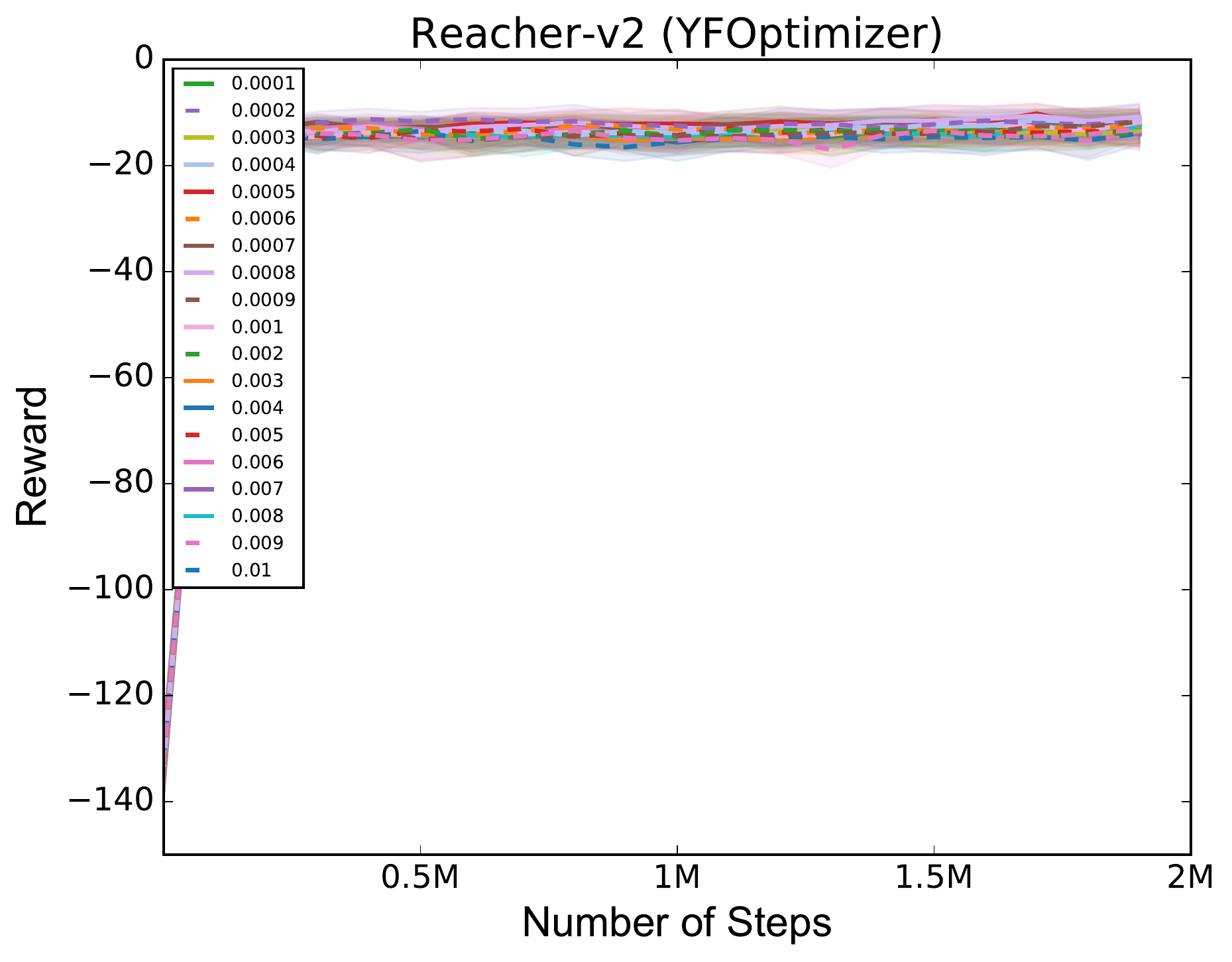}
    \caption{PPO performance across learning rates on the Reacher environment.}
\end{figure}

\begin{figure}[H]
    \centering
    \includegraphics[width=.32\textwidth]{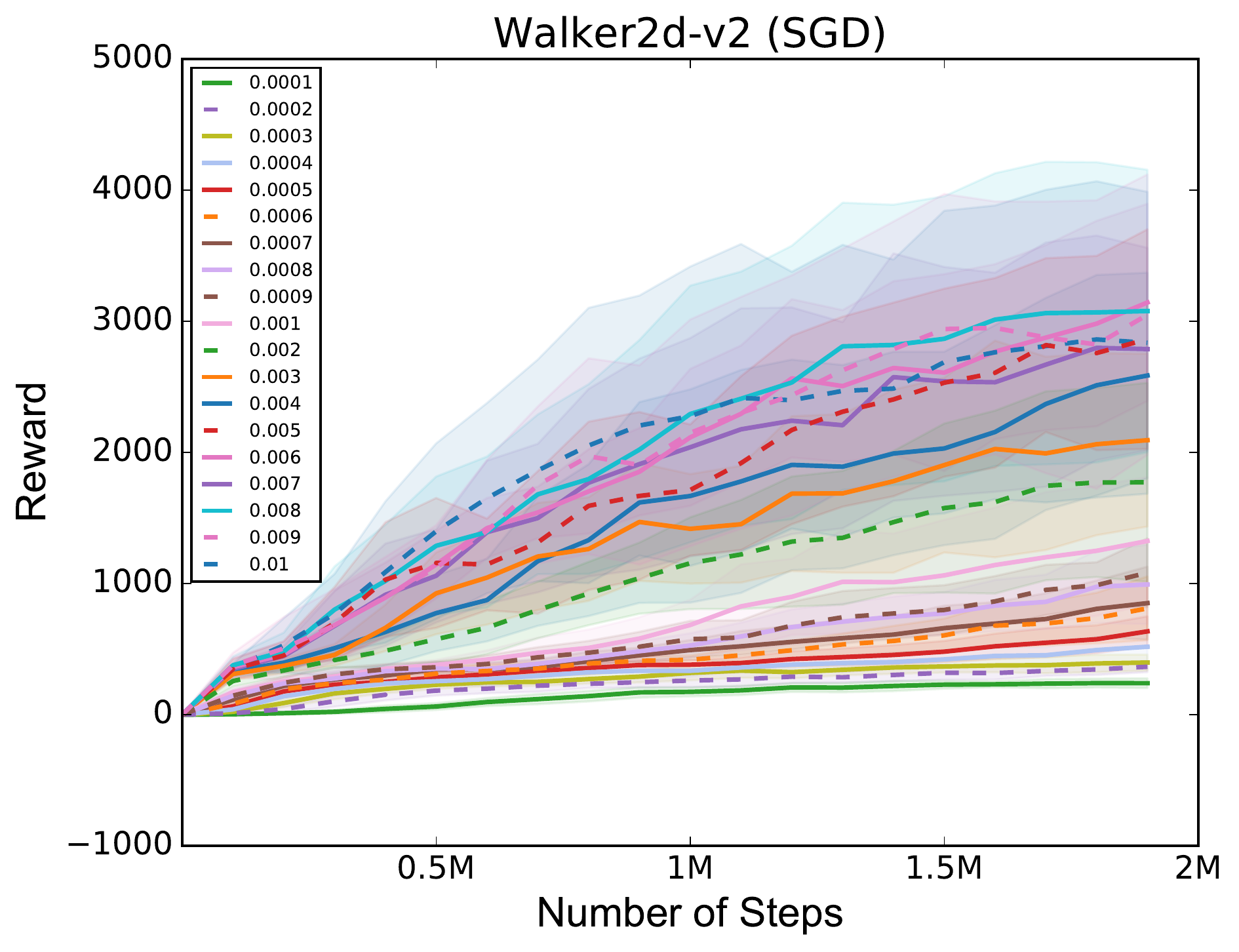}
    \includegraphics[width=.32\textwidth]{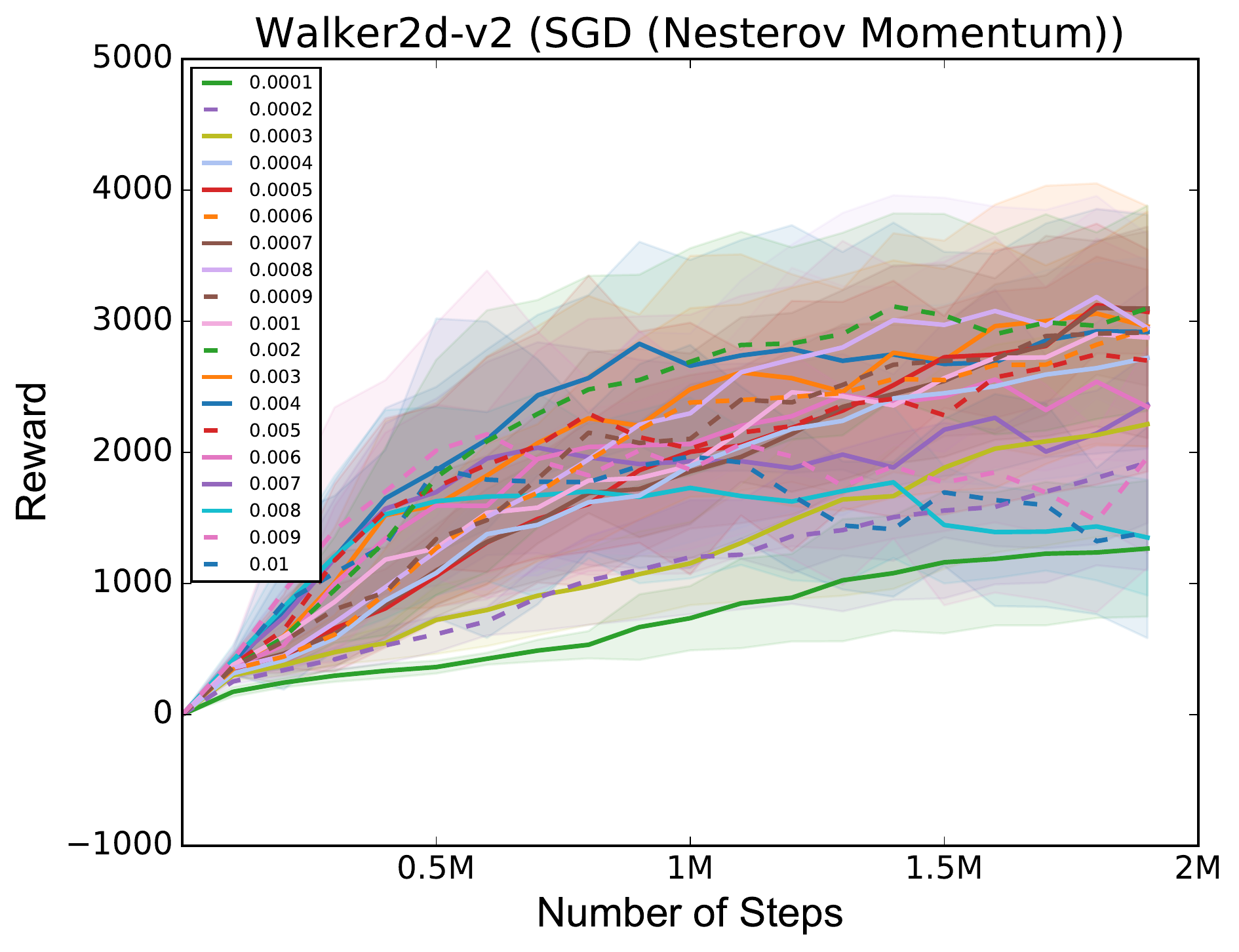}
    \includegraphics[width=.32\textwidth]{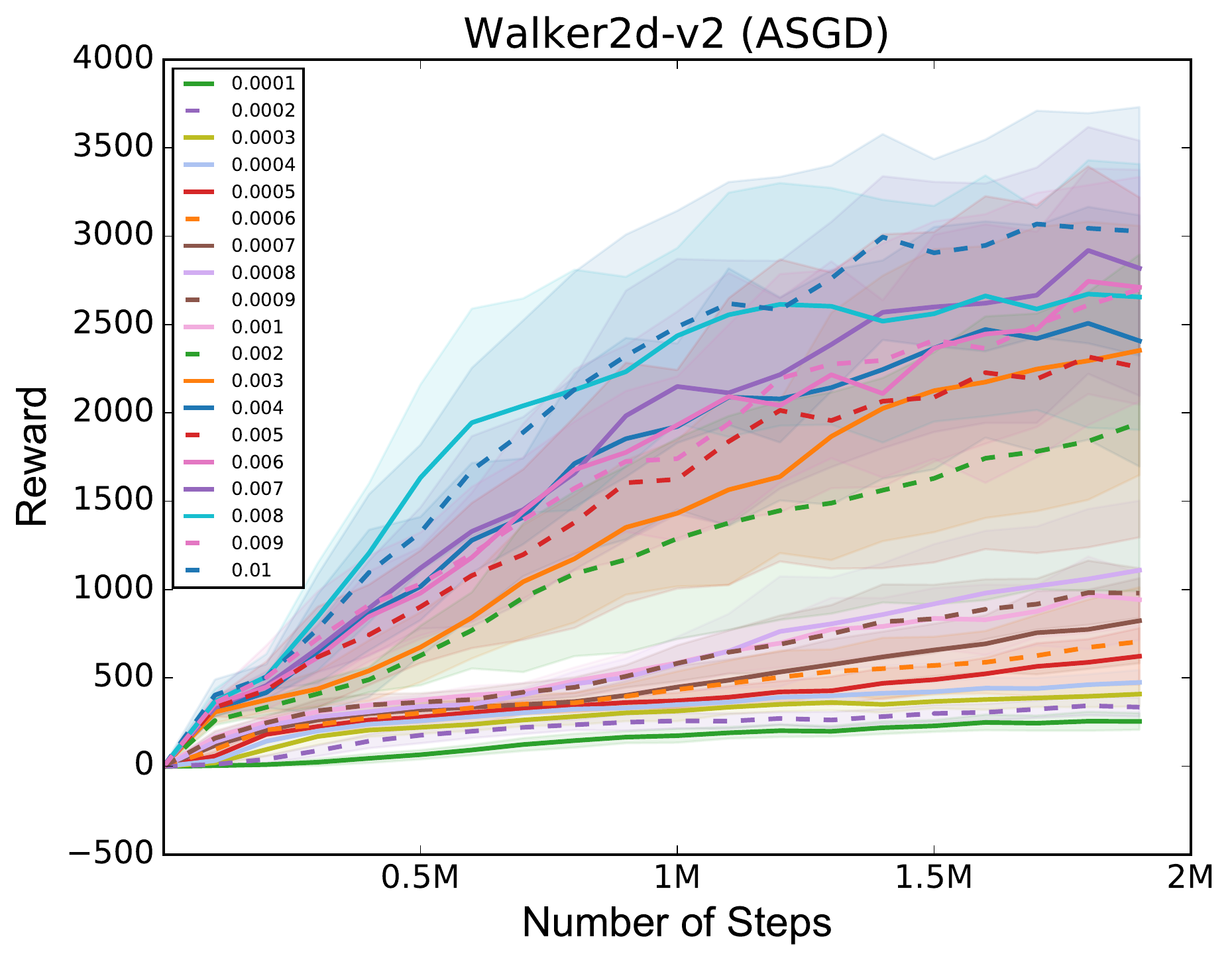}
    \includegraphics[width=.32\textwidth]{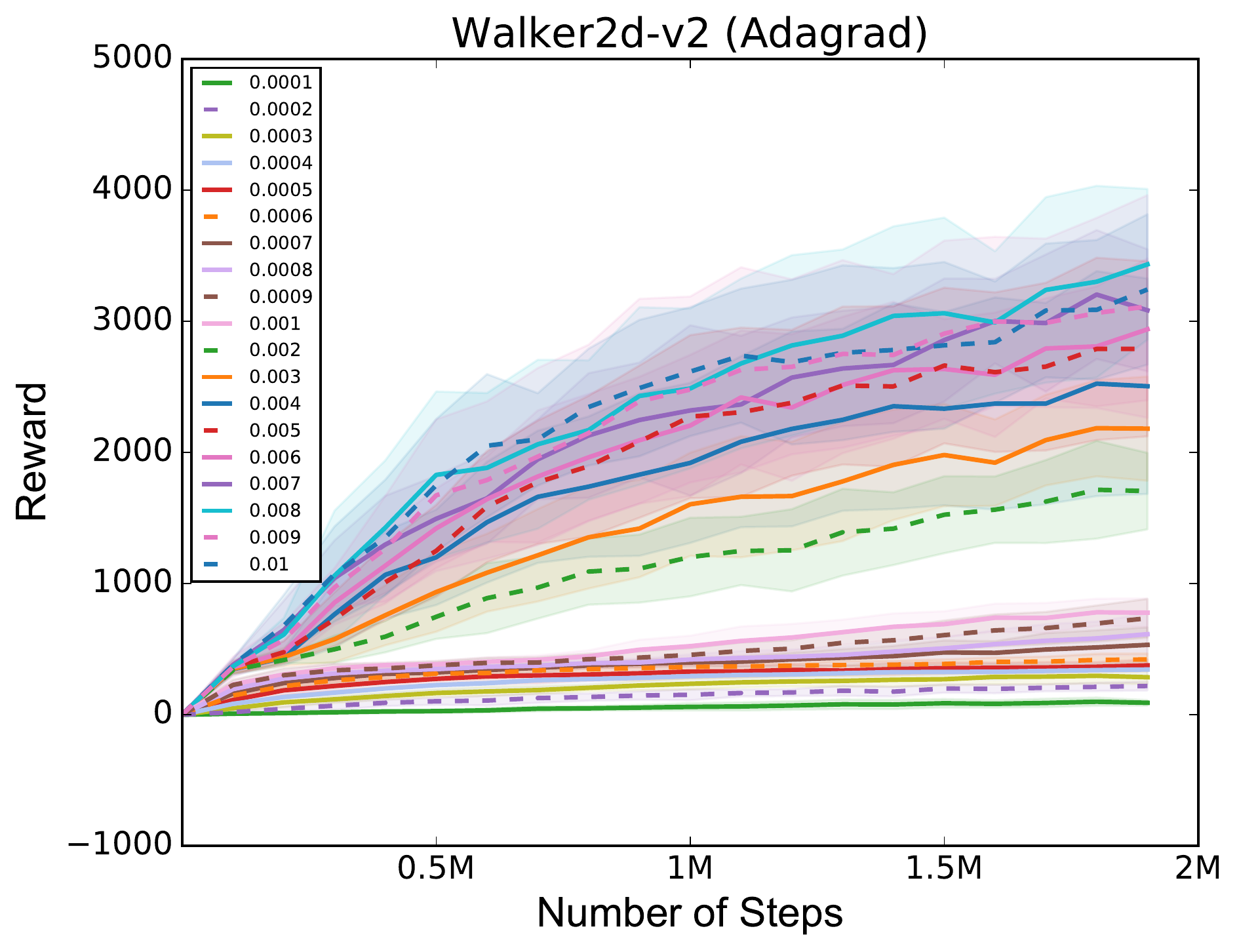}
    \includegraphics[width=.32\textwidth]{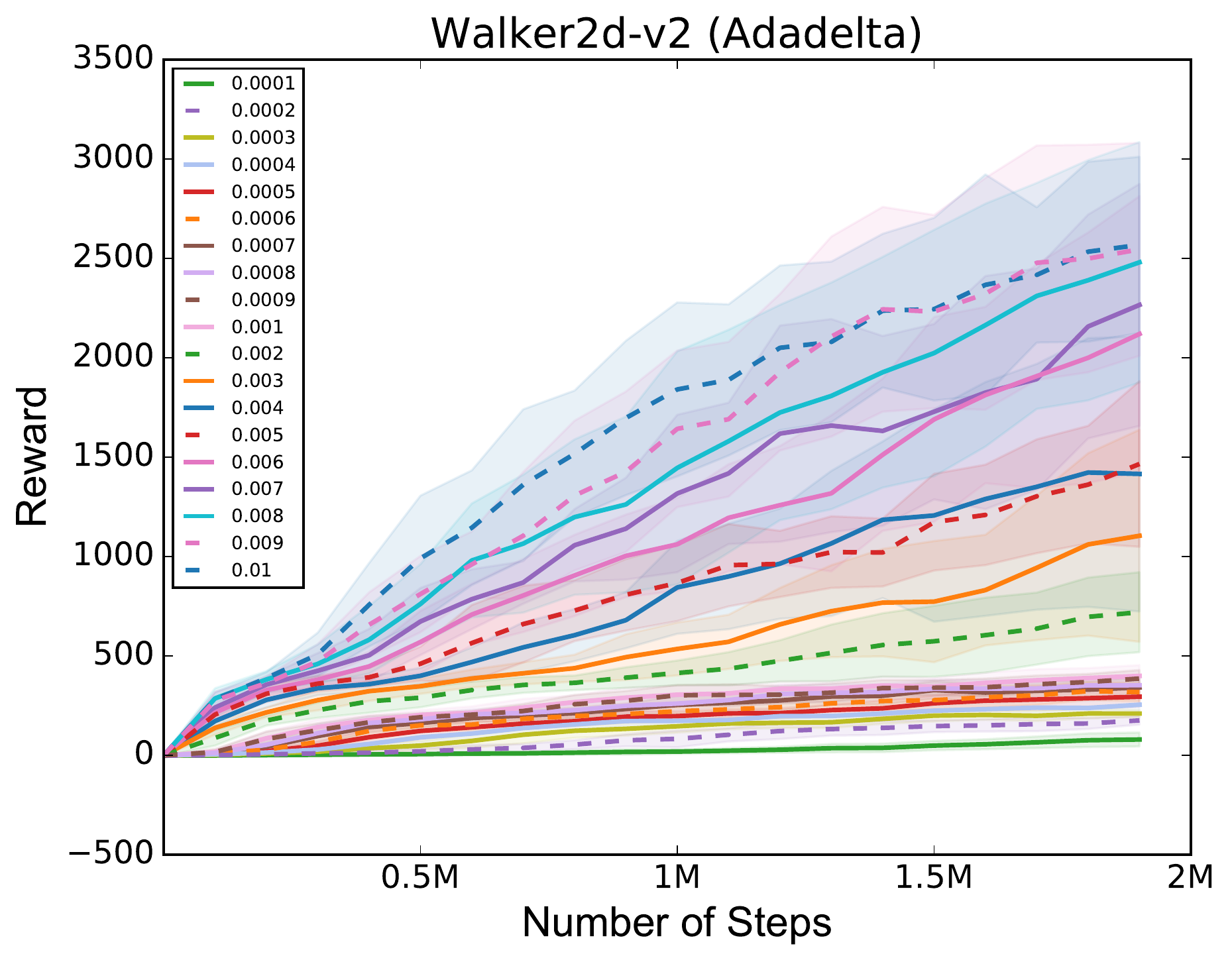}
    \includegraphics[width=.32\textwidth]{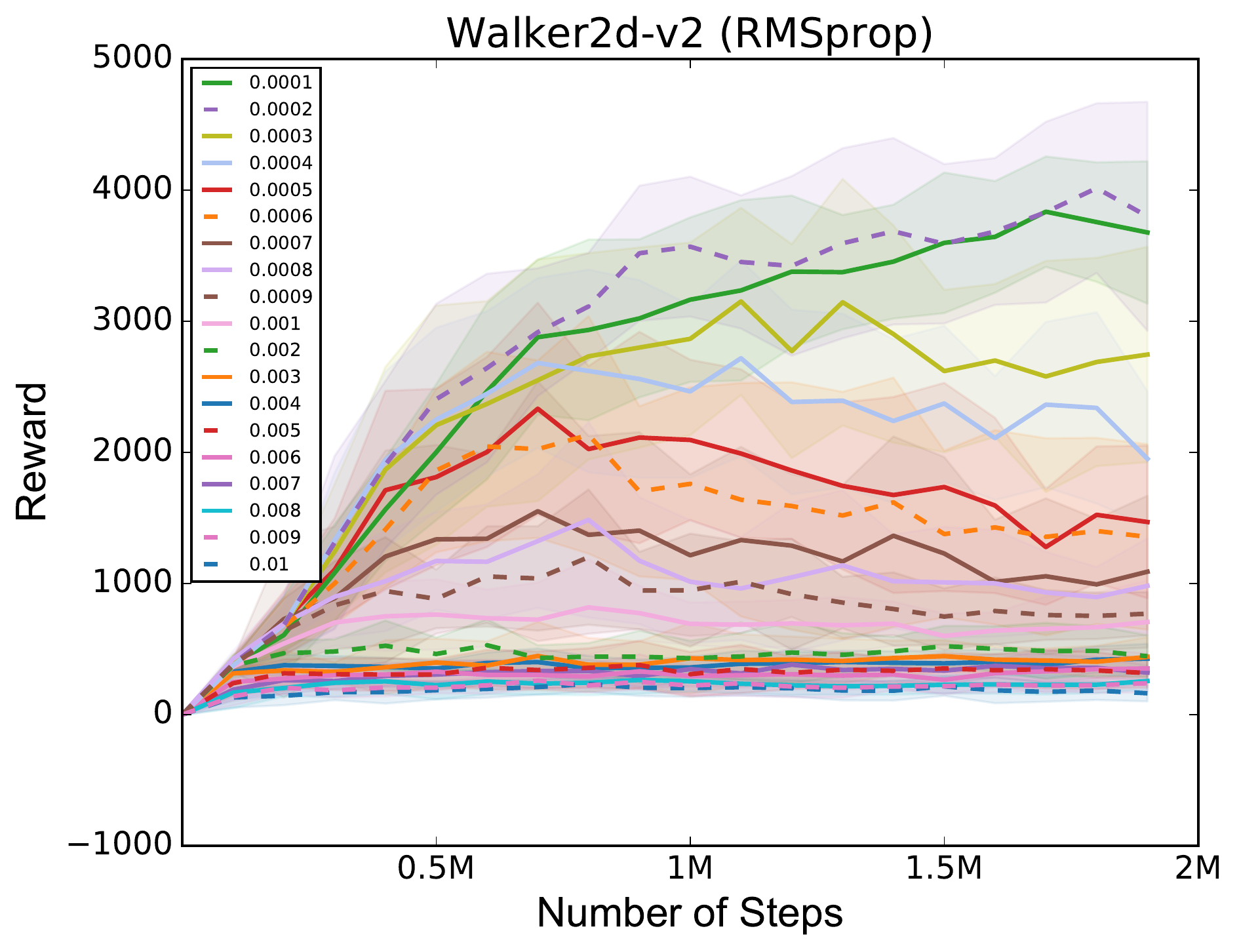}
    \includegraphics[width=.32\textwidth]{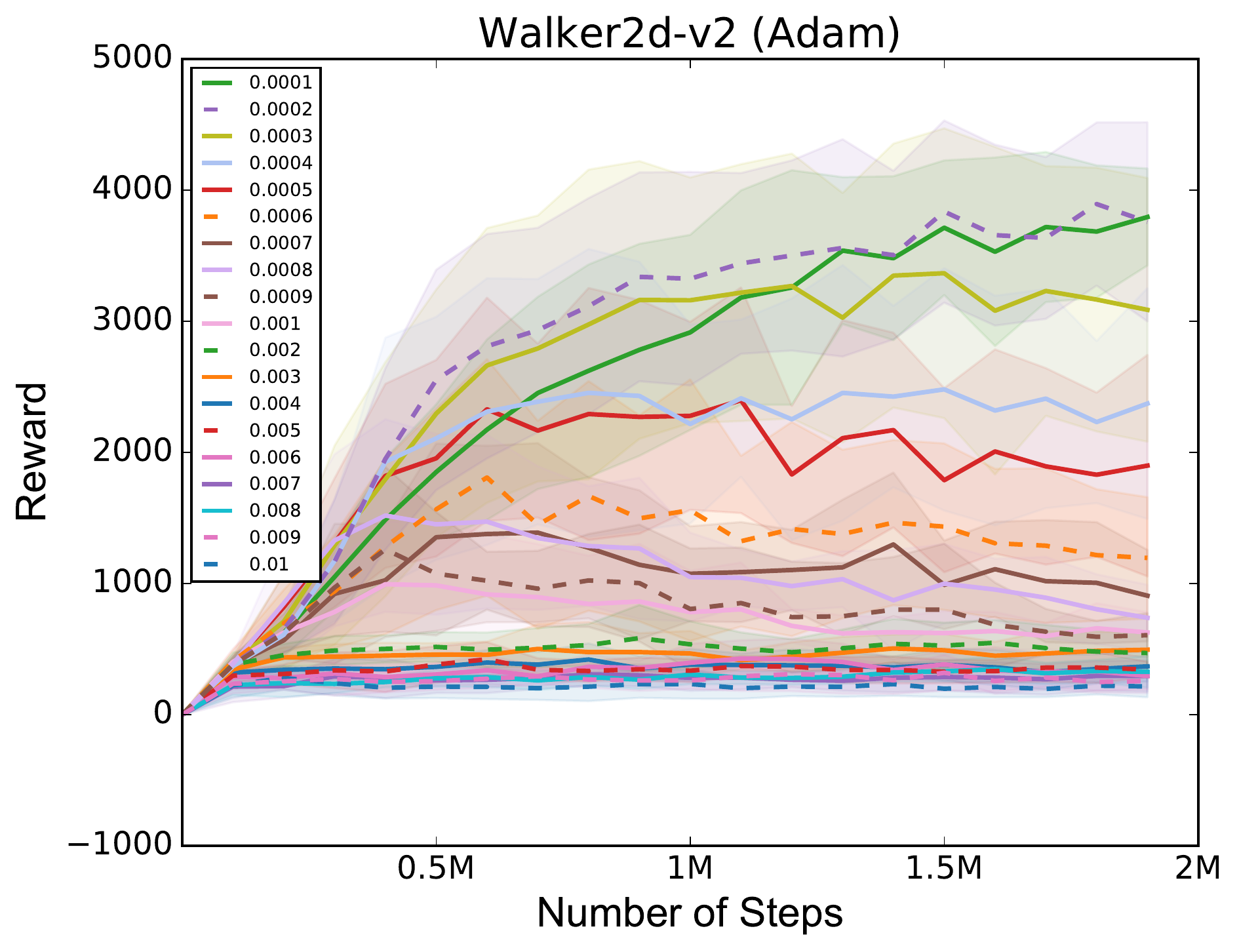}
    \includegraphics[width=.32\textwidth]{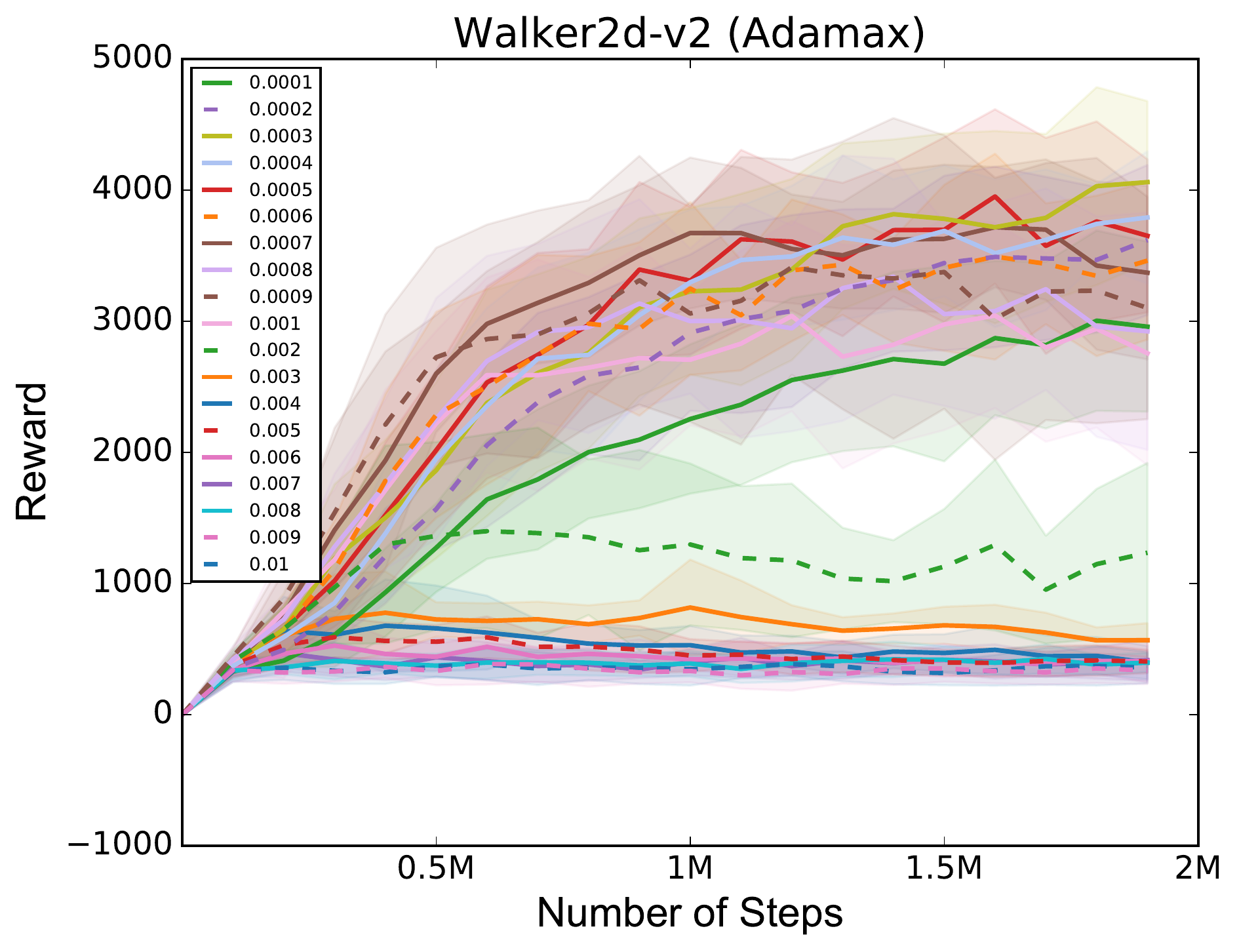}
    \includegraphics[width=.32\textwidth]{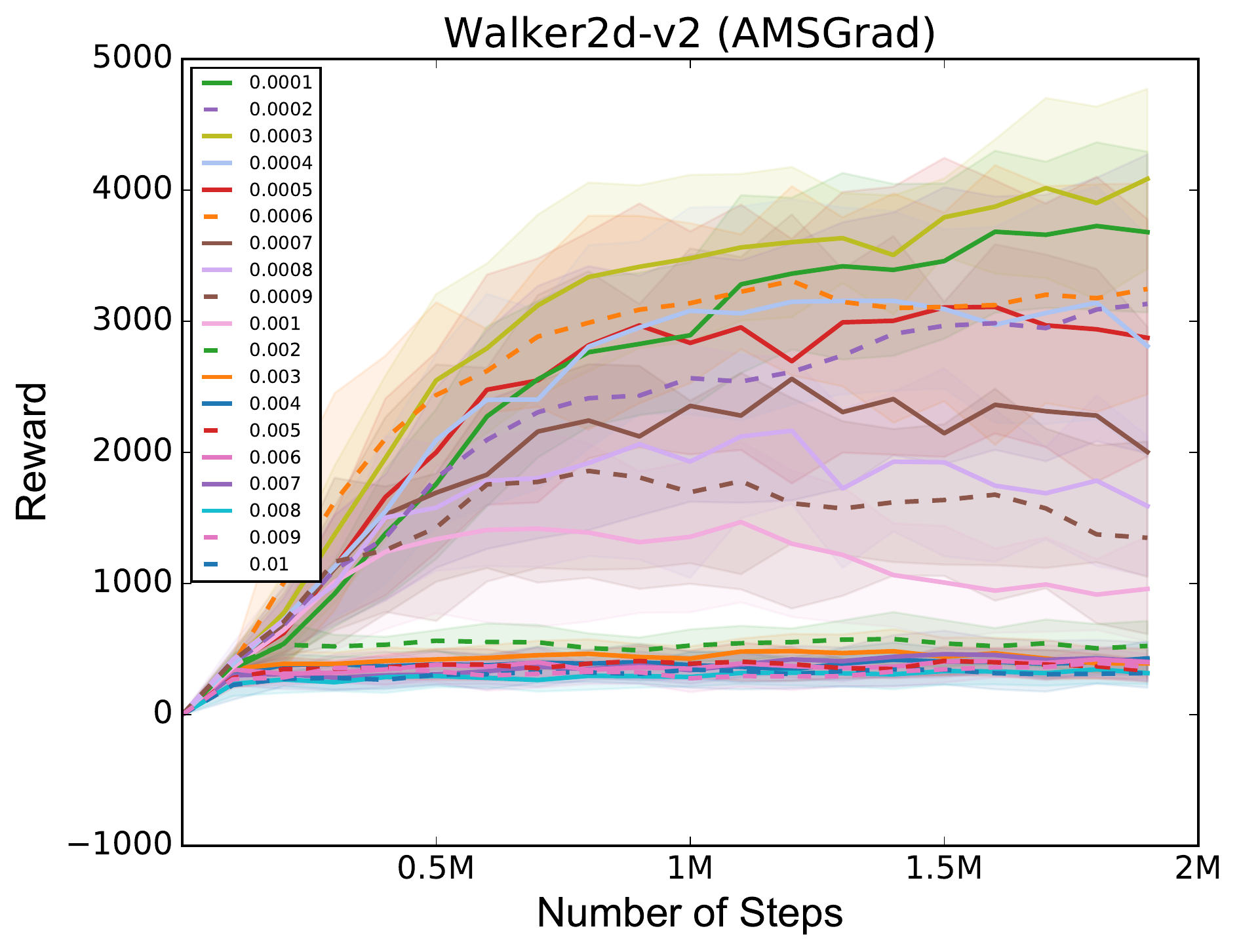}
    \includegraphics[width=.32\textwidth]{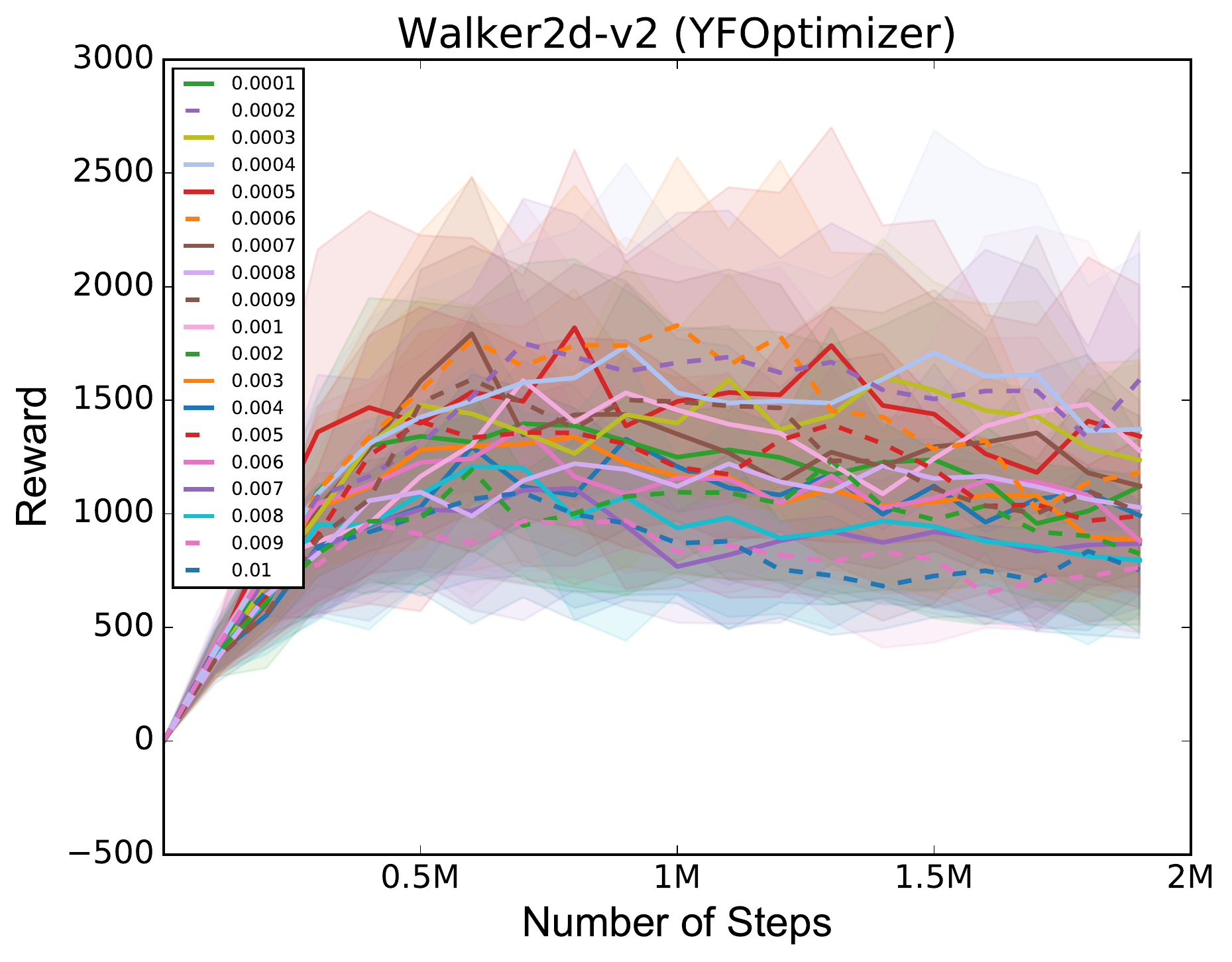}
    \caption{PPO performance across learning rates on the Walker2d environment.}
\end{figure}

\begin{figure}[H]
    \centering
    \includegraphics[width=.32\textwidth]{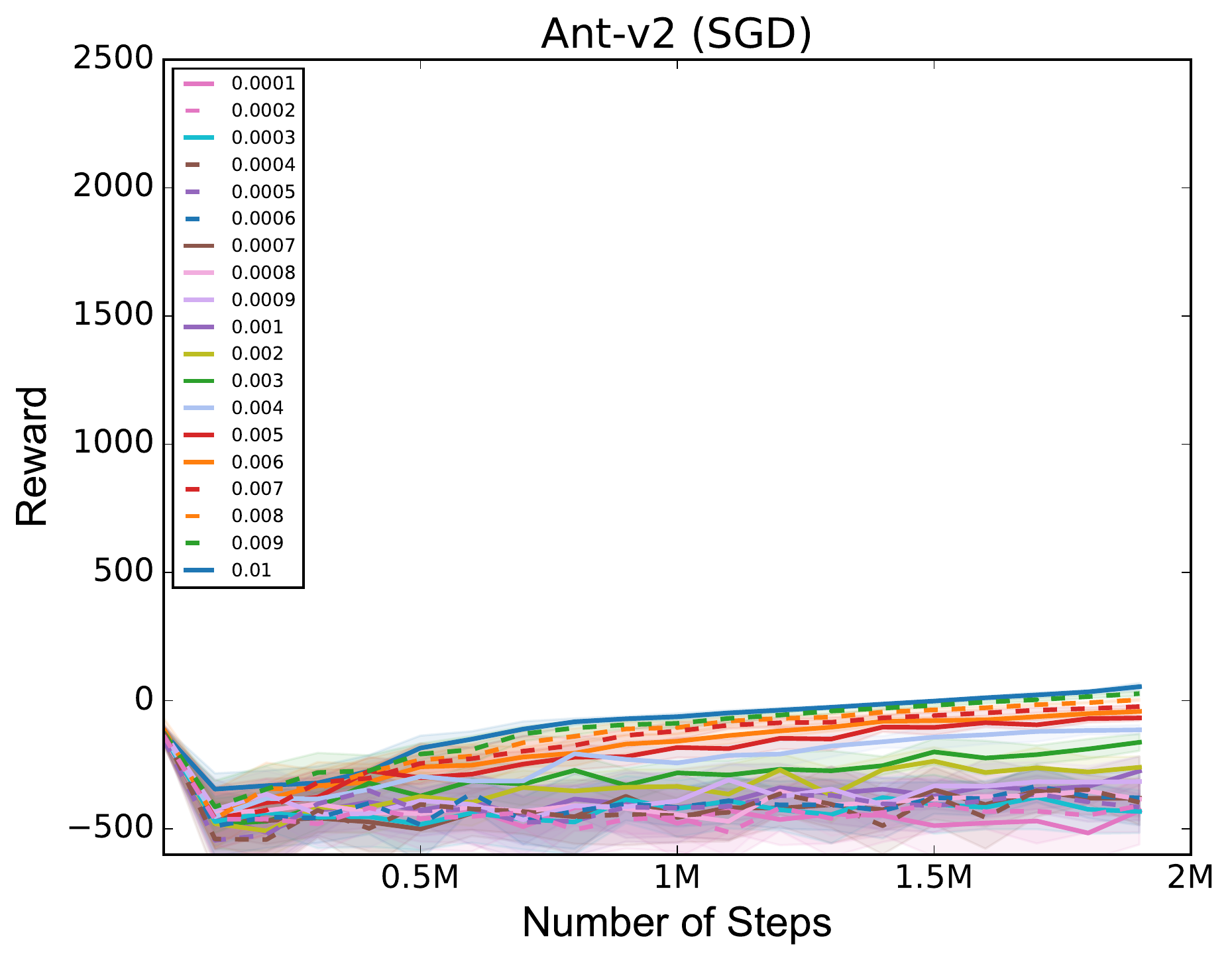}
    \includegraphics[width=.32\textwidth]{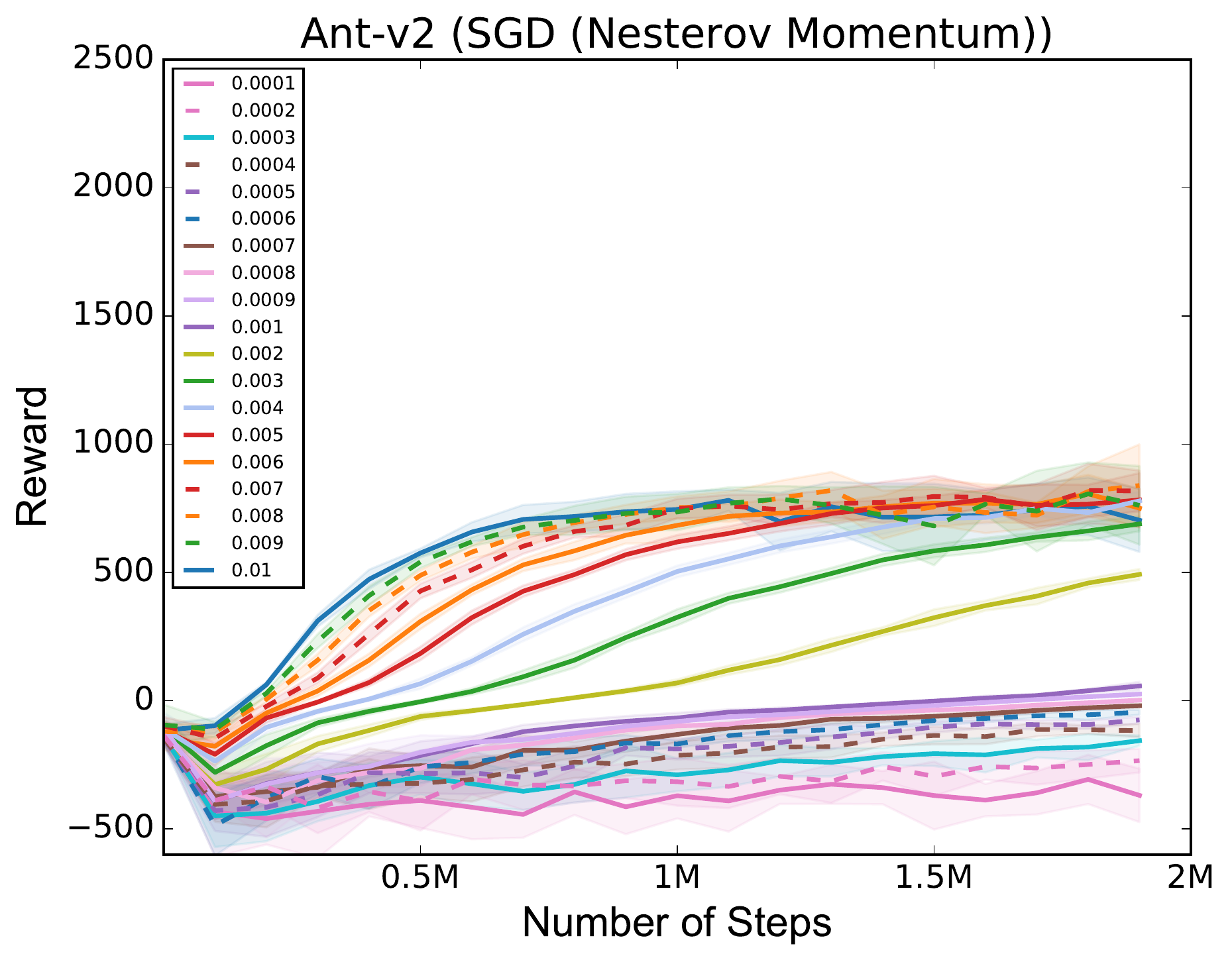}
    \includegraphics[width=.32\textwidth]{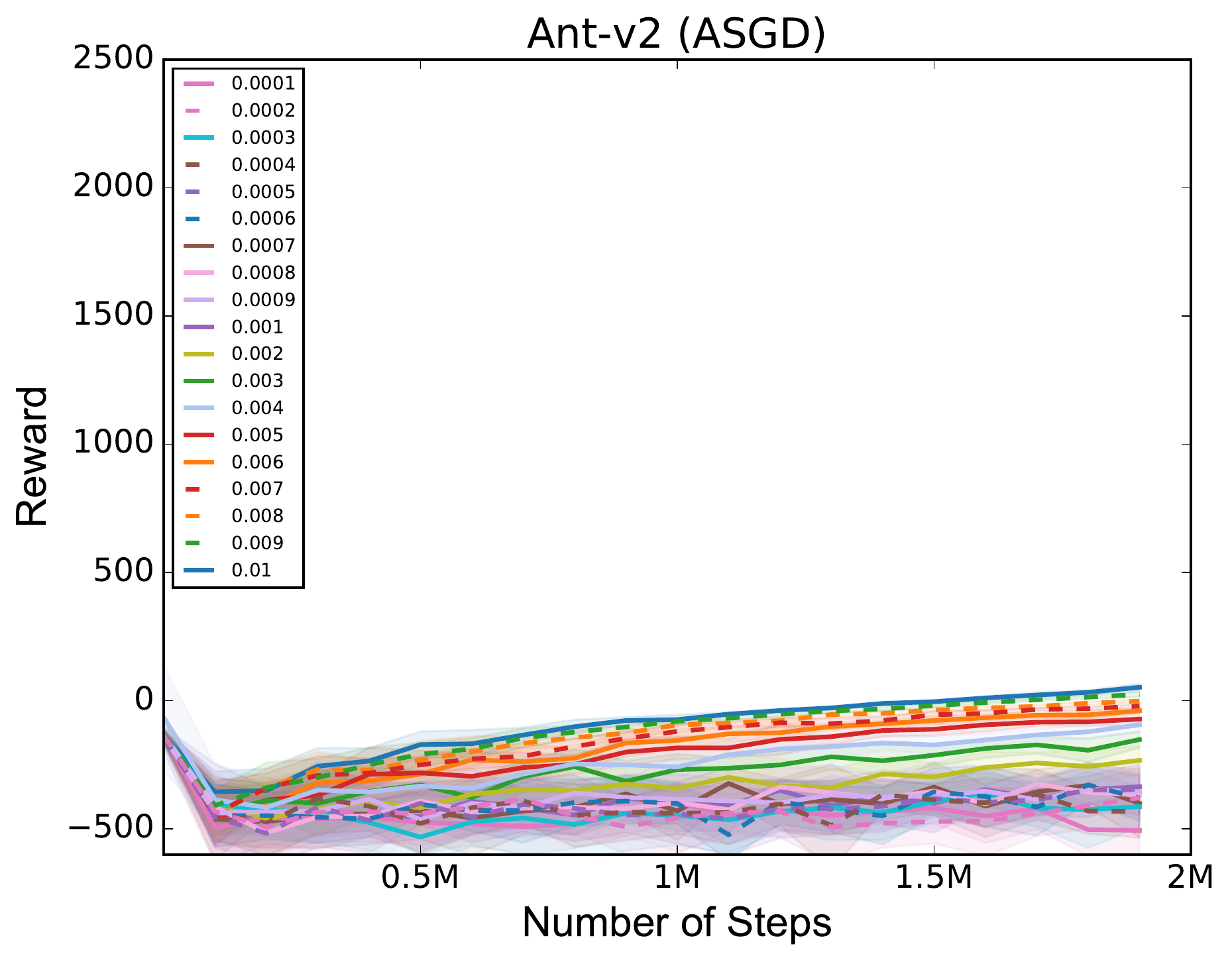}
    \includegraphics[width=.32\textwidth]{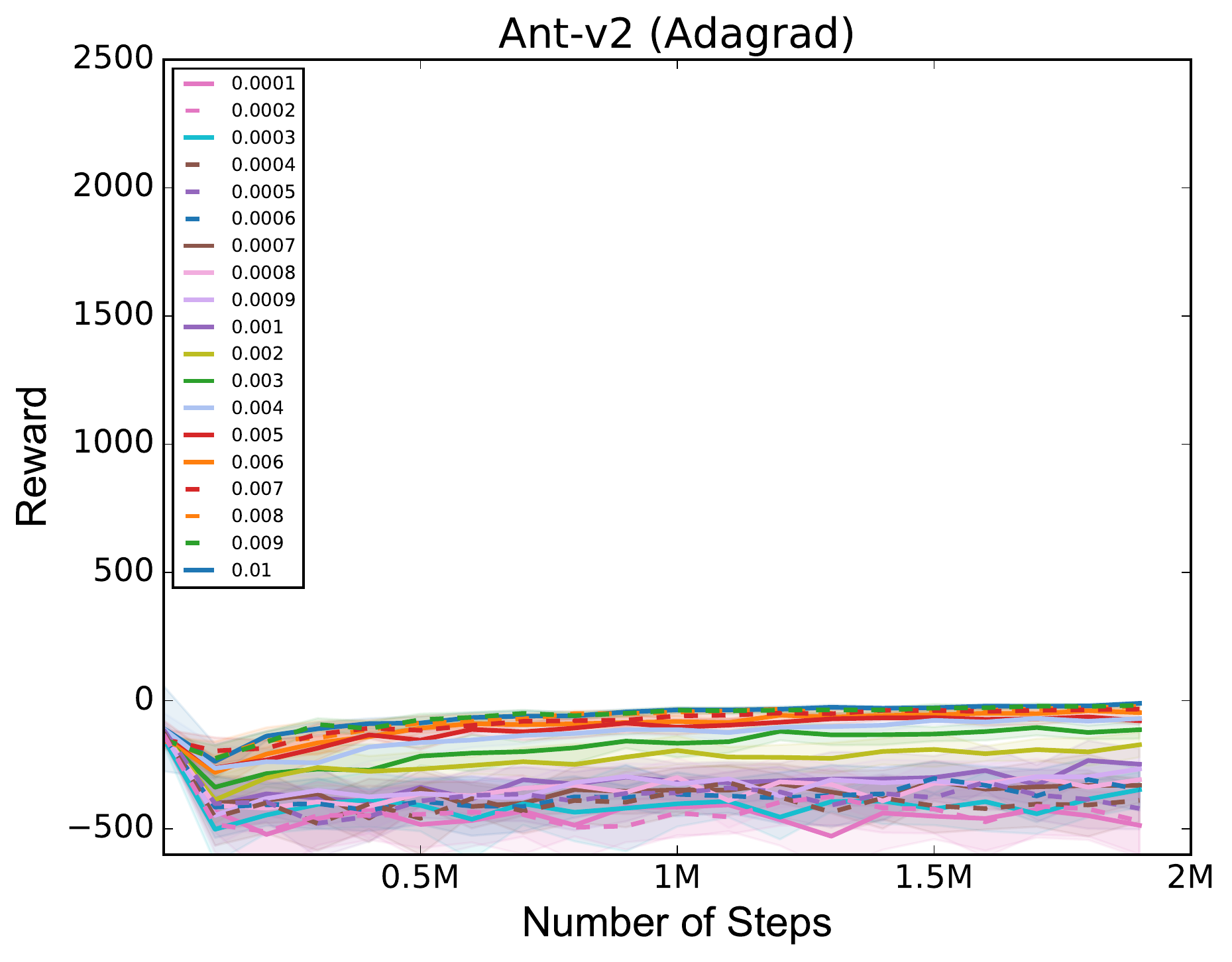}
    \includegraphics[width=.32\textwidth]{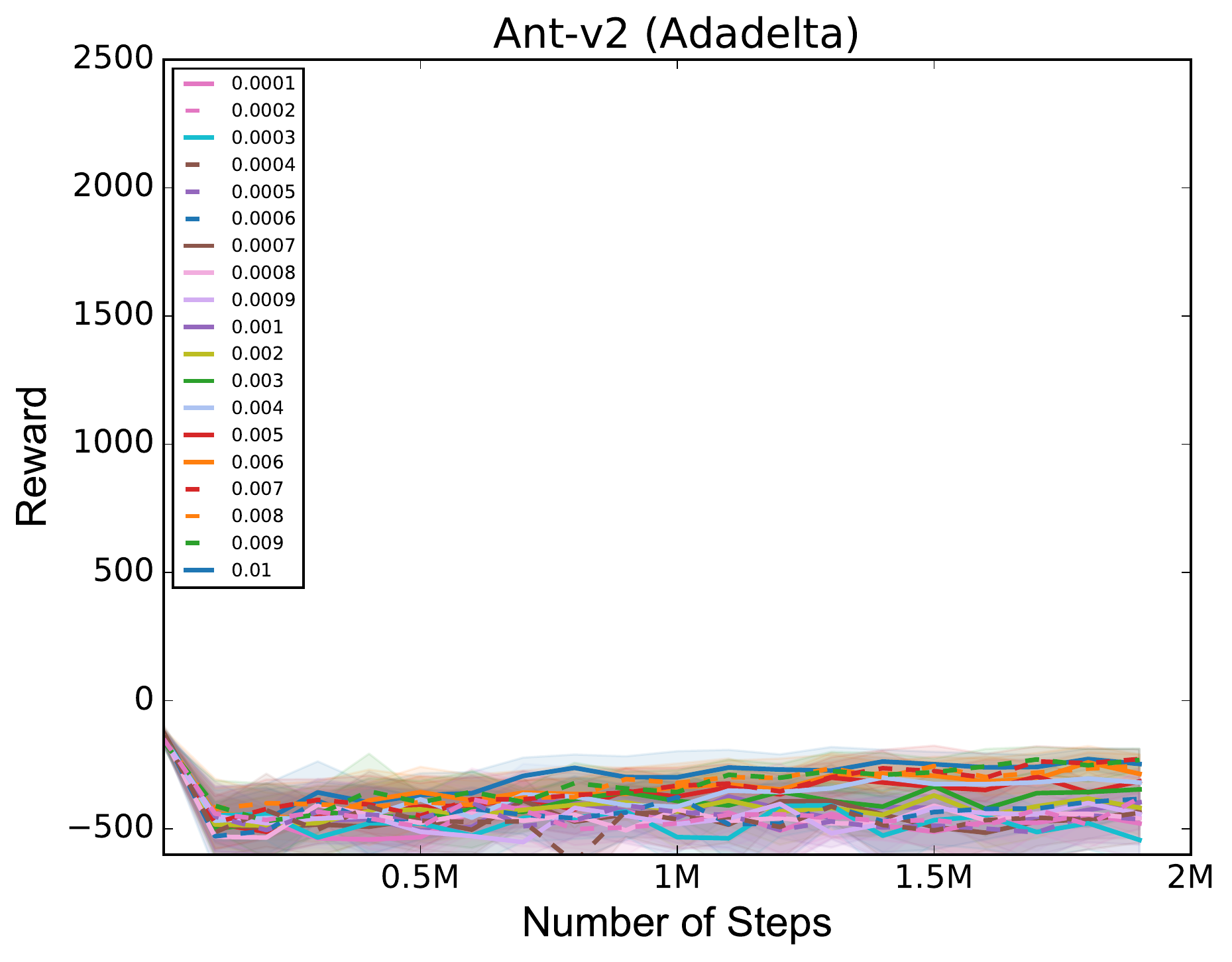}
    \includegraphics[width=.32\textwidth]{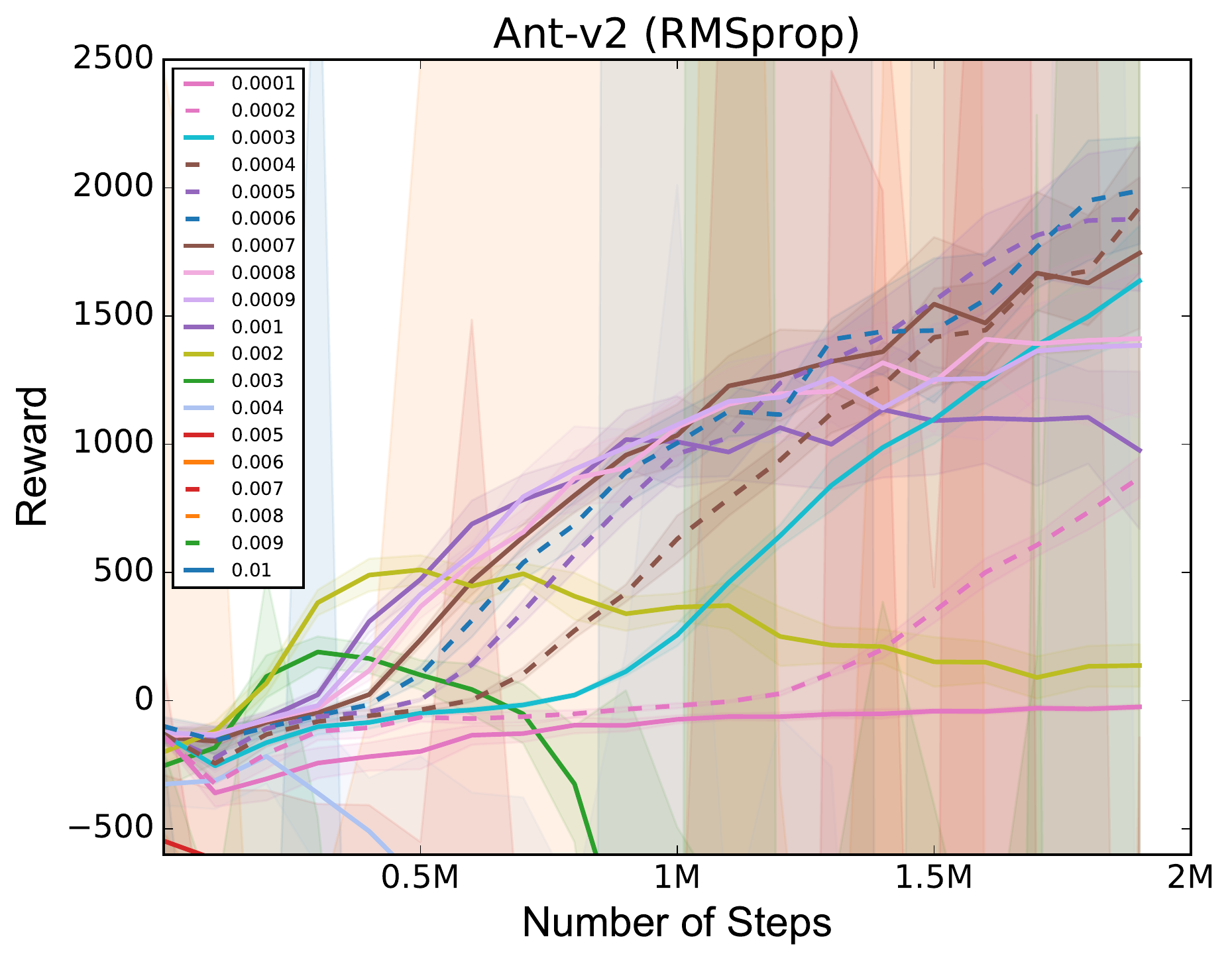}
    \includegraphics[width=.32\textwidth]{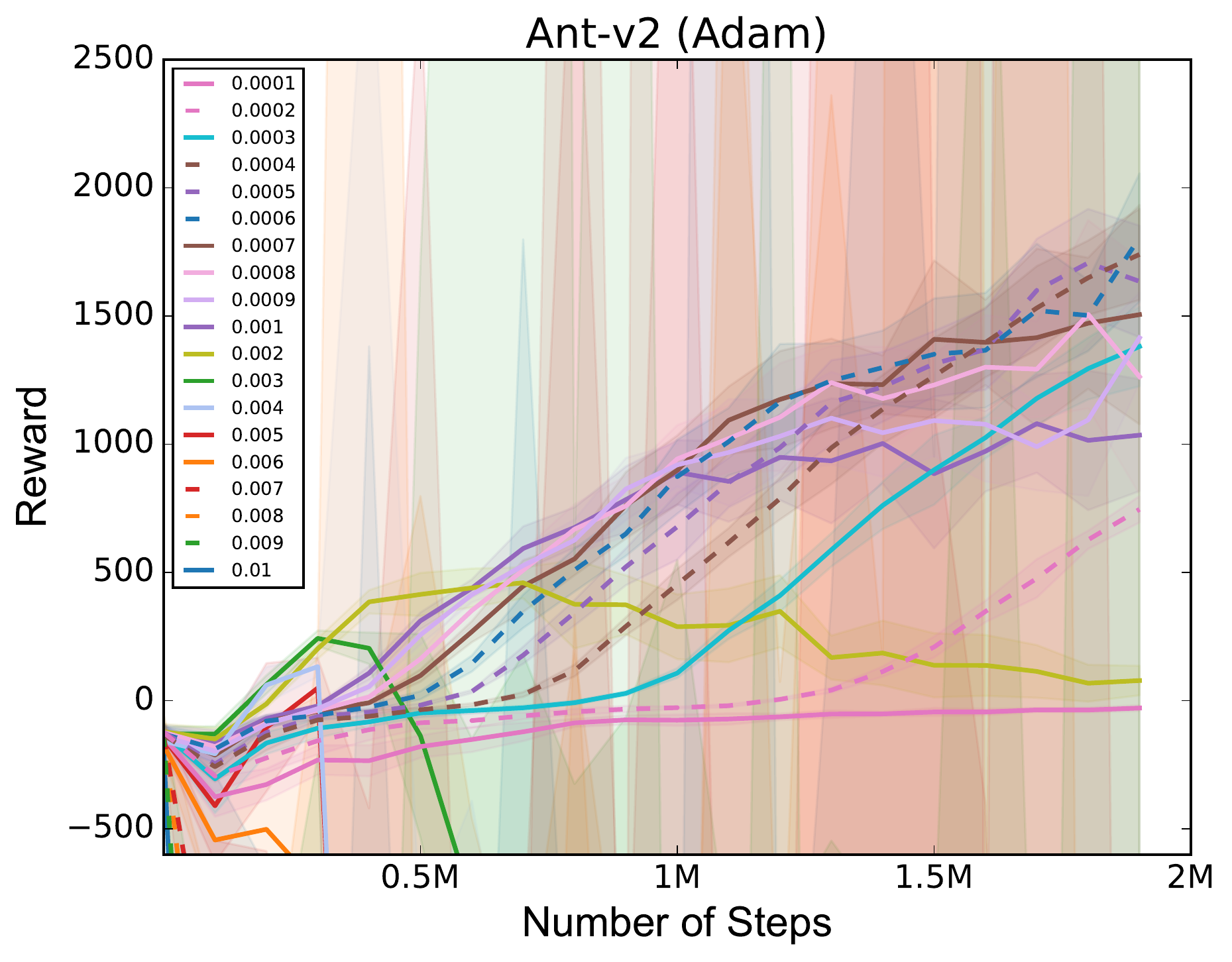}
    \includegraphics[width=.32\textwidth]{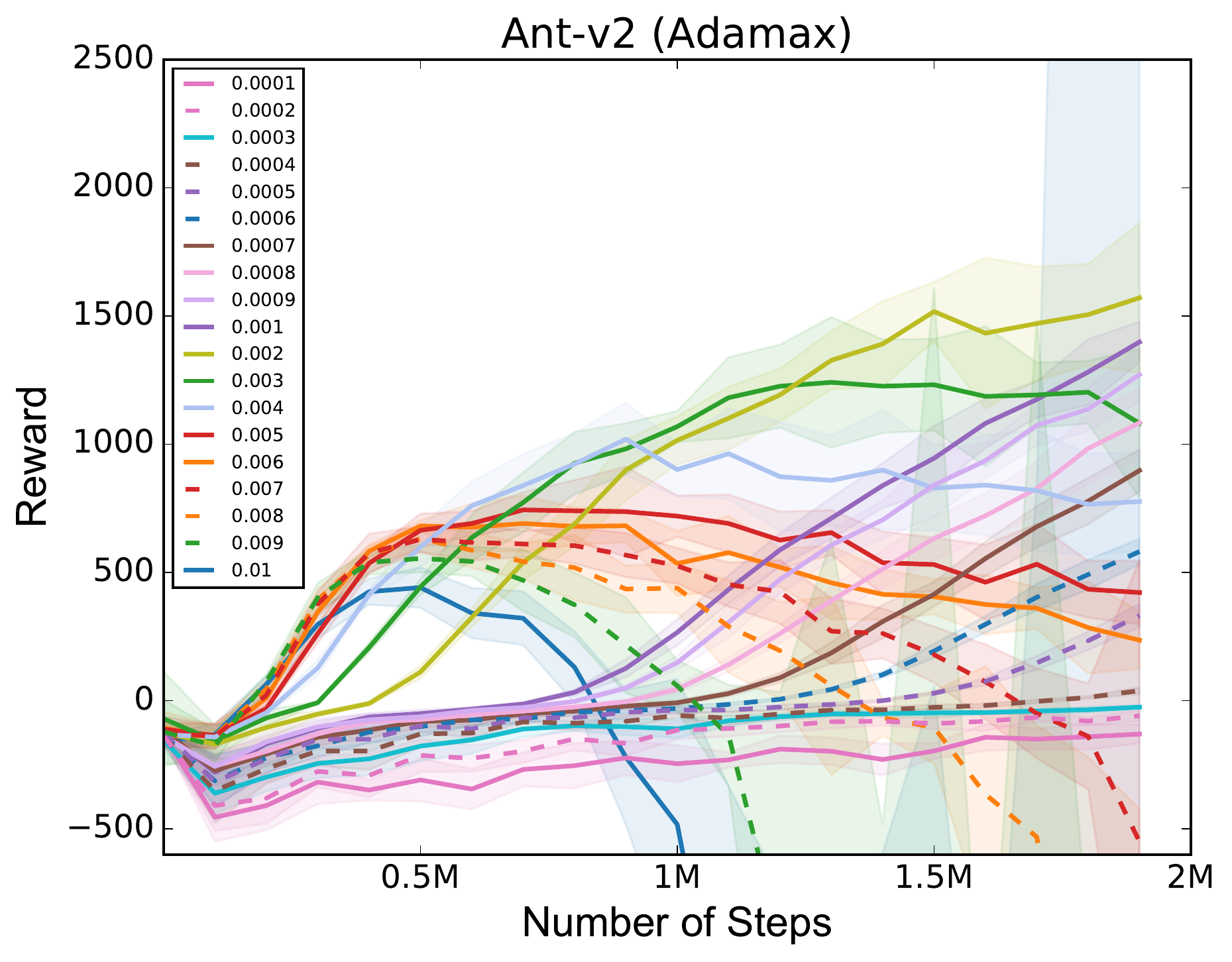}
    \includegraphics[width=.32\textwidth]{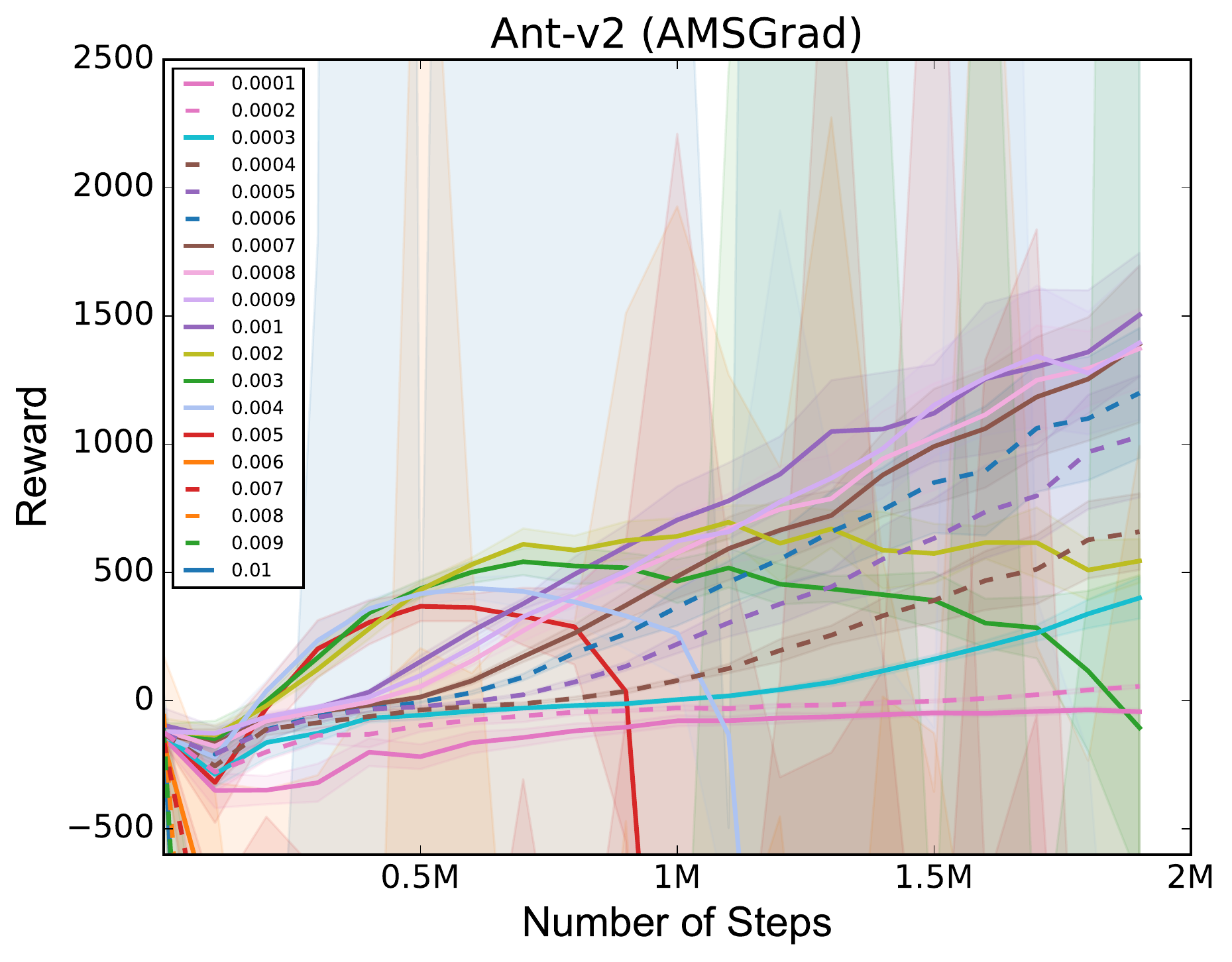}
    \includegraphics[width=.32\textwidth]{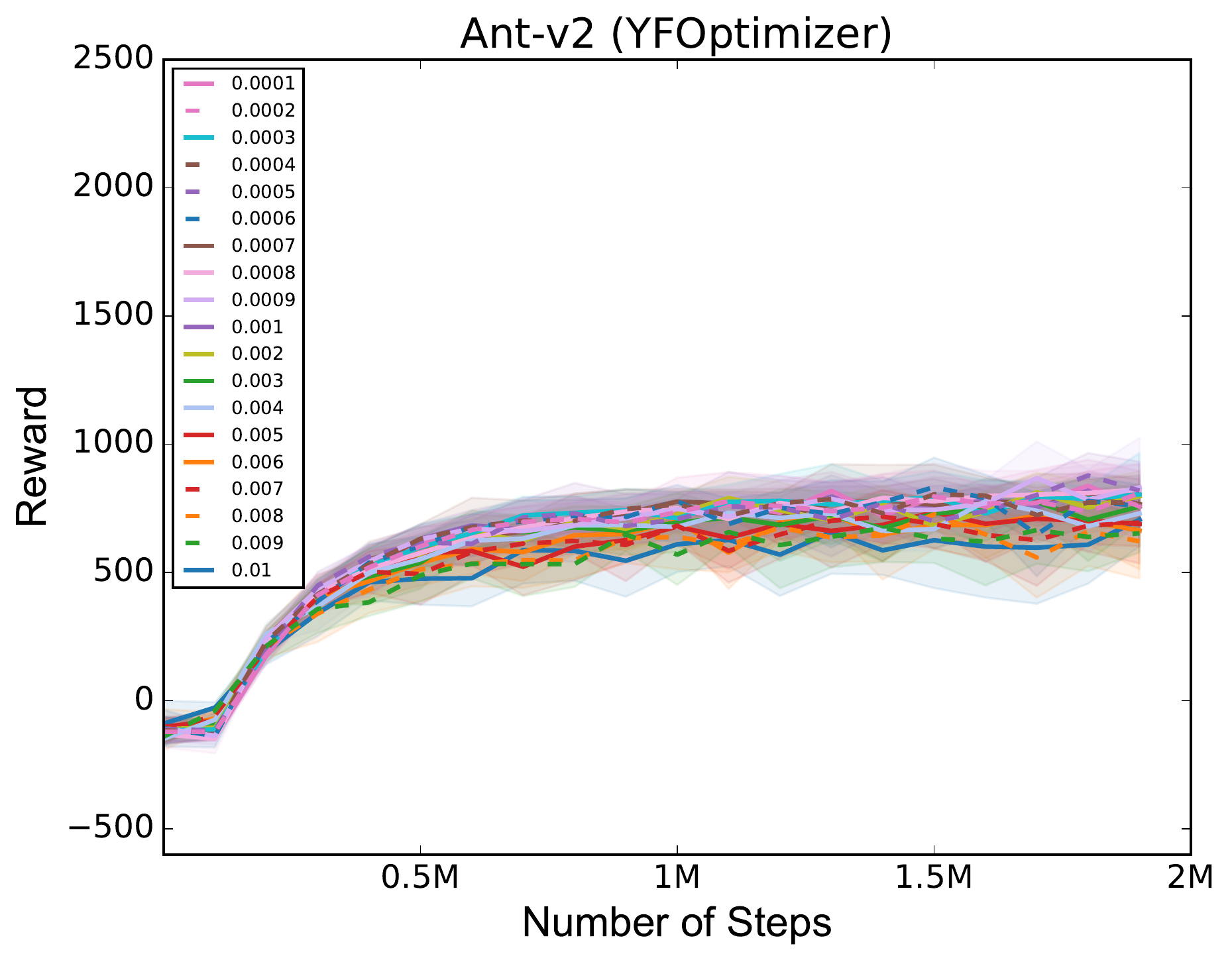}
    \caption{A2C performance across learning rates on the Ant environment.}
\end{figure}

\begin{figure}[H]
    \centering
    \includegraphics[width=.32\textwidth]{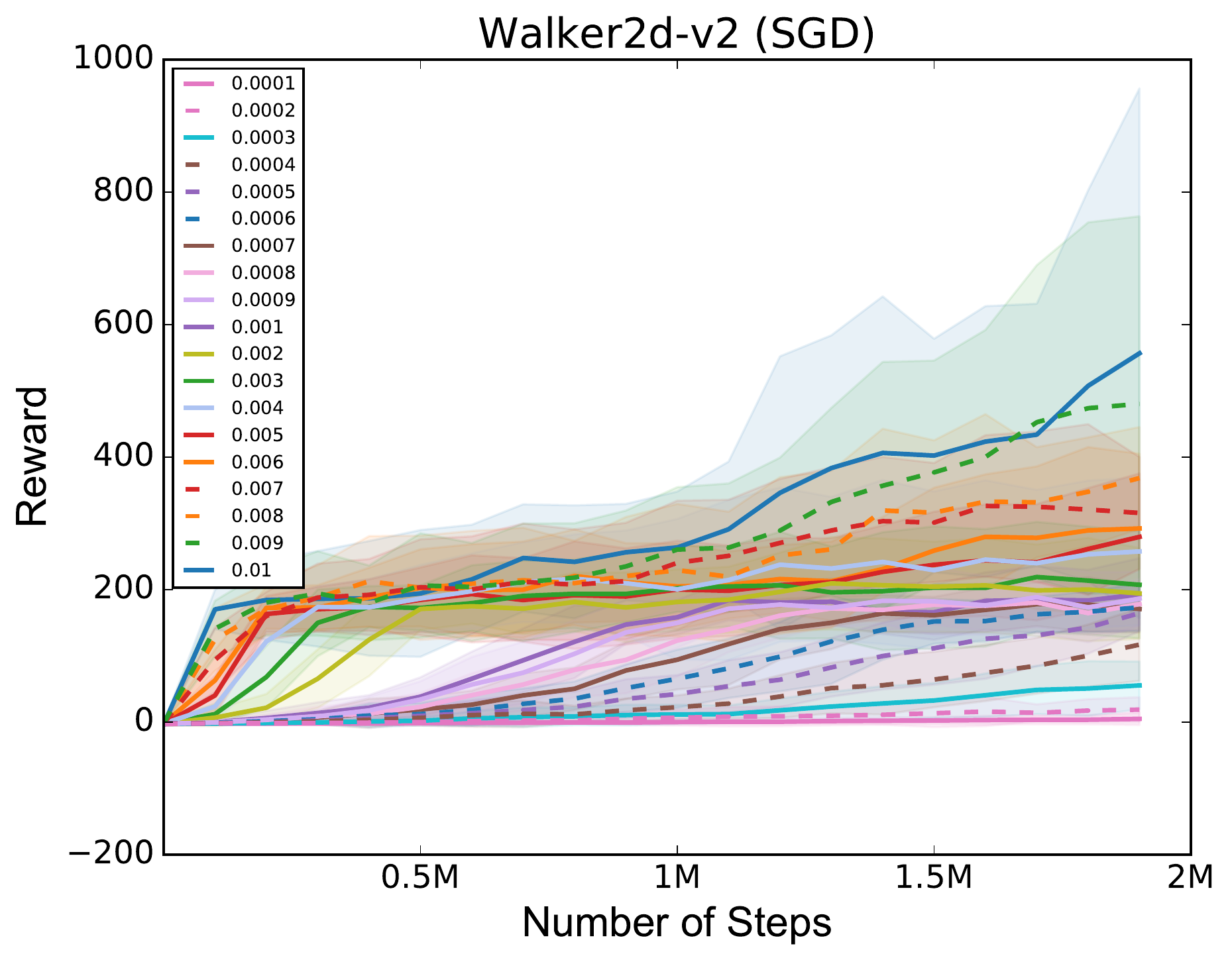}
    \includegraphics[width=.32\textwidth]{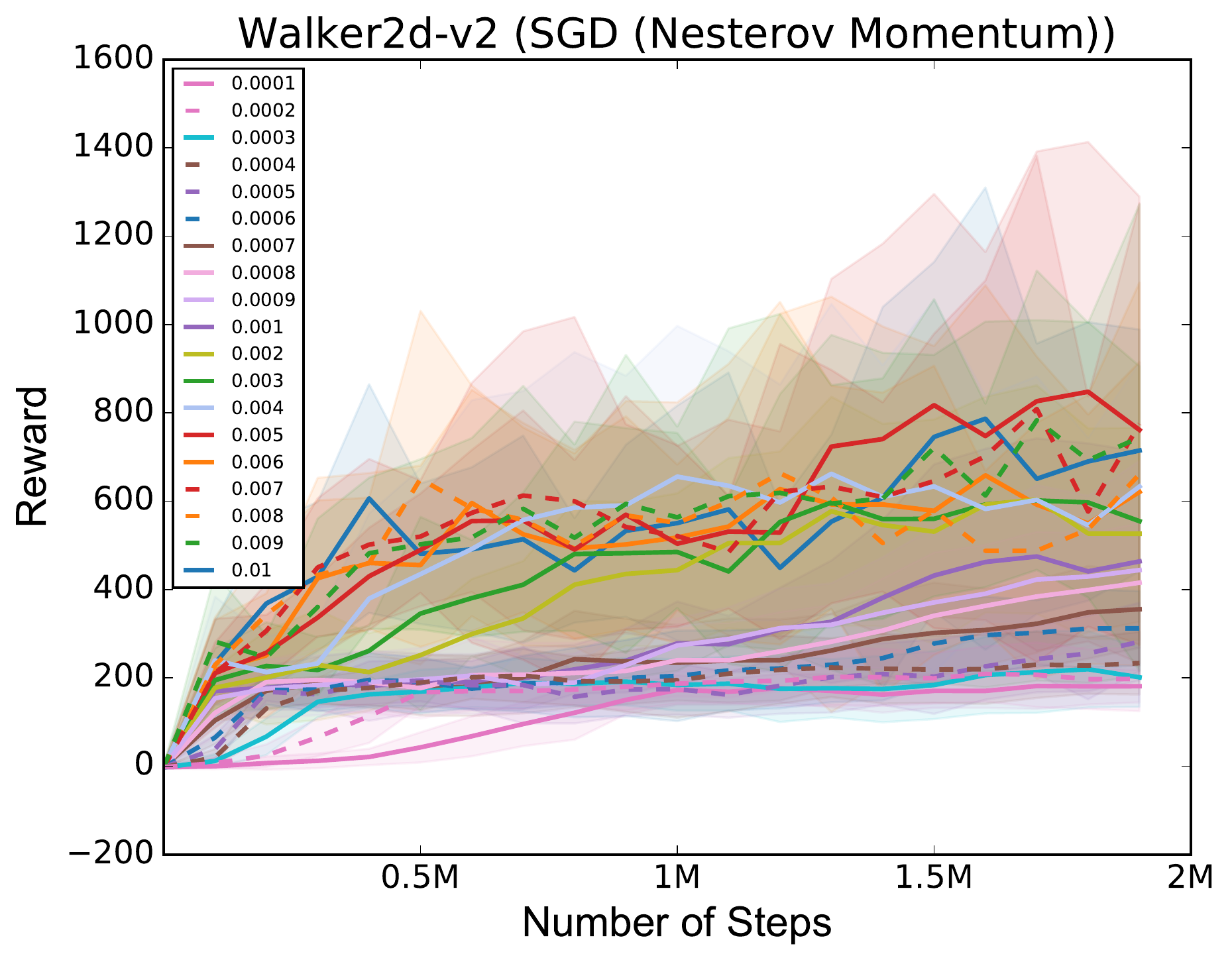}
    \includegraphics[width=.32\textwidth]{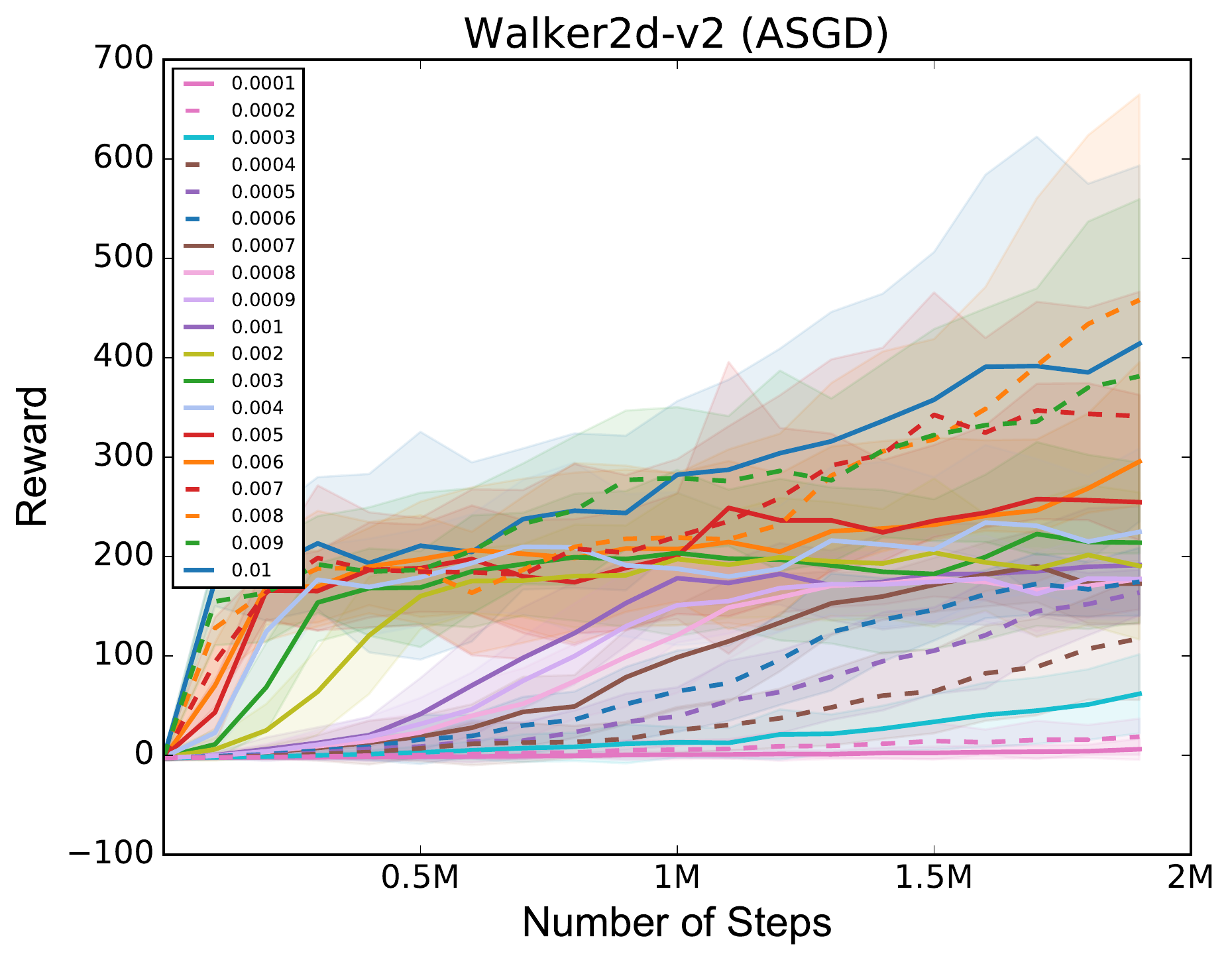}
    \includegraphics[width=.32\textwidth]{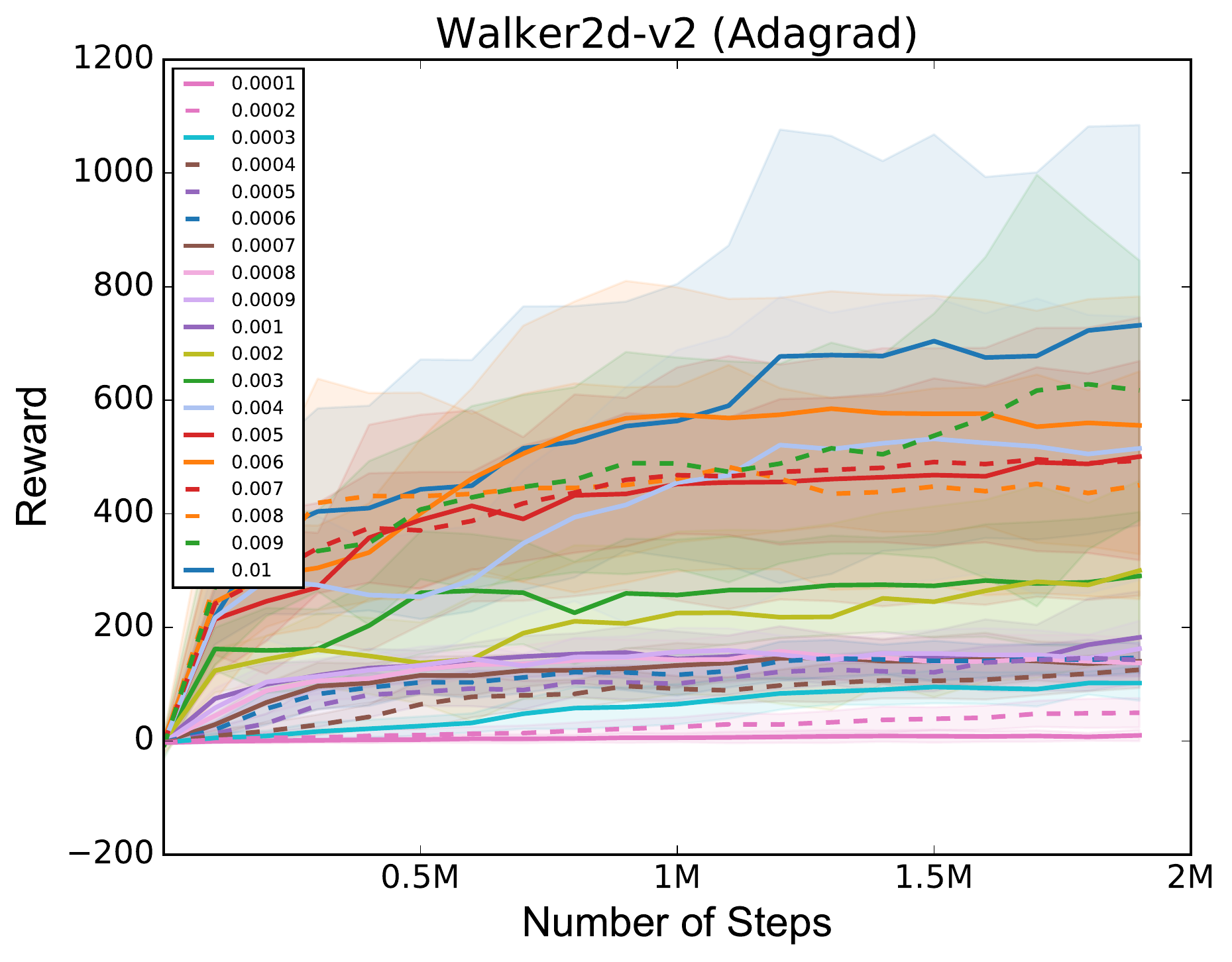}
    \includegraphics[width=.32\textwidth]{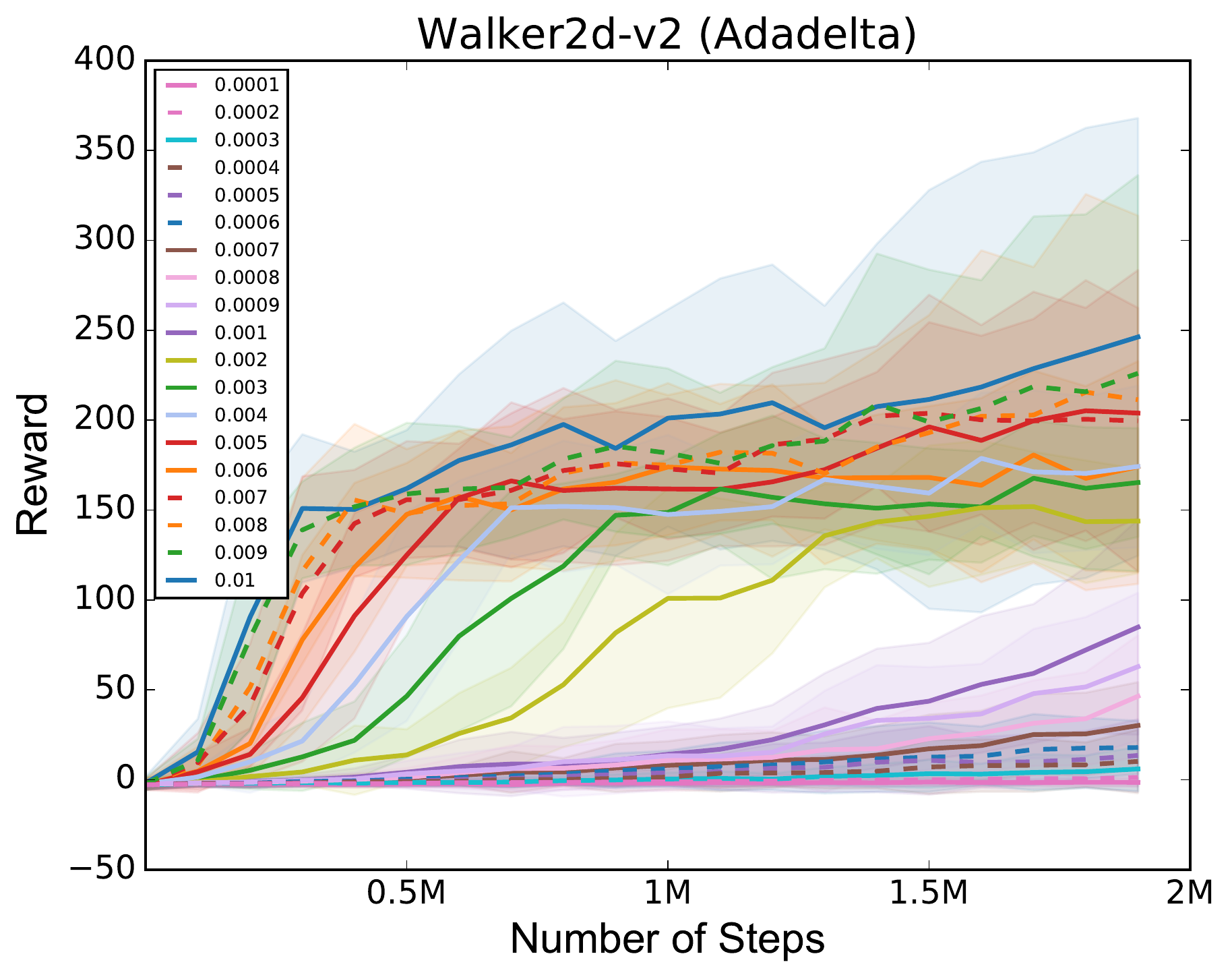}
    \includegraphics[width=.32\textwidth]{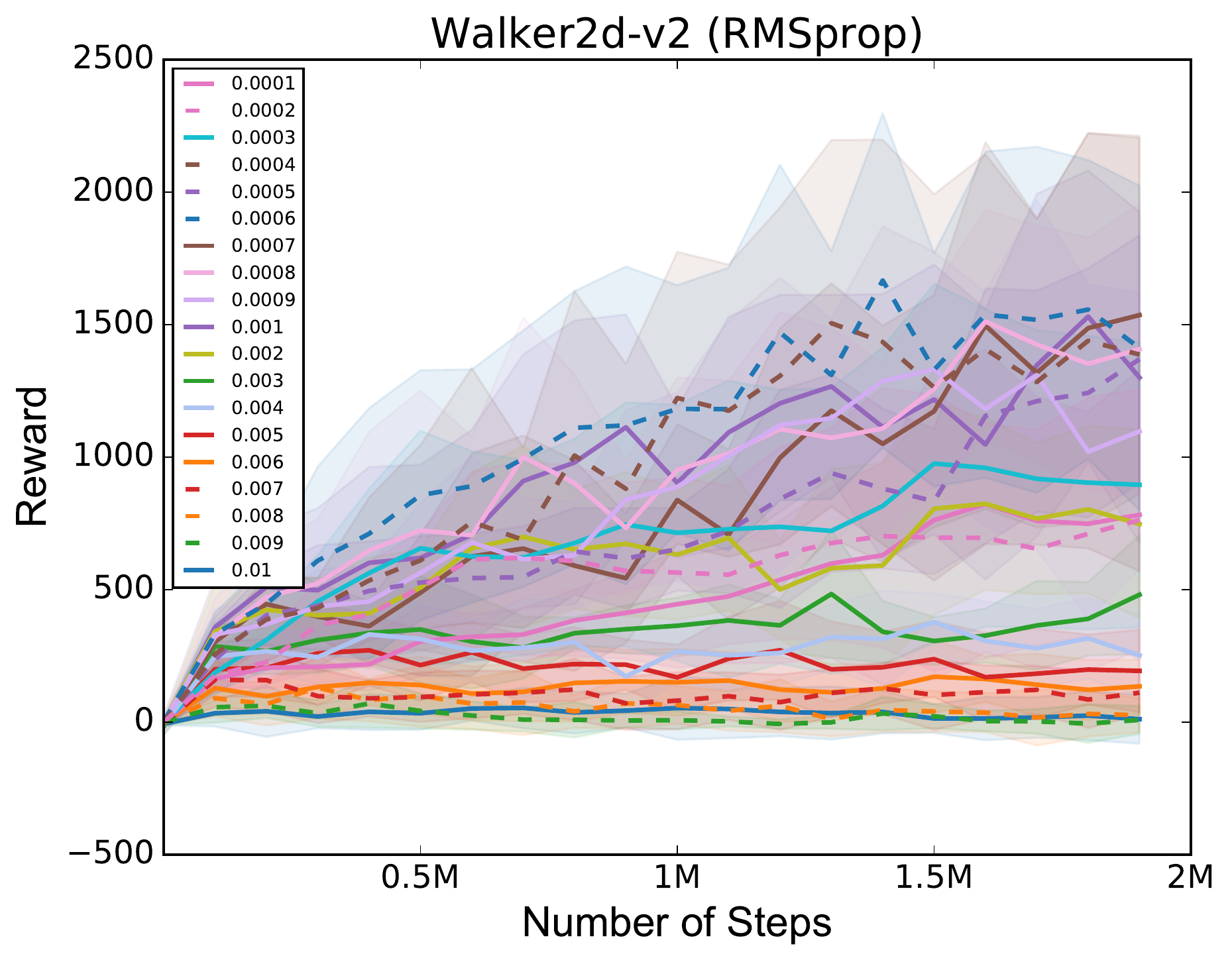}
    \includegraphics[width=.32\textwidth]{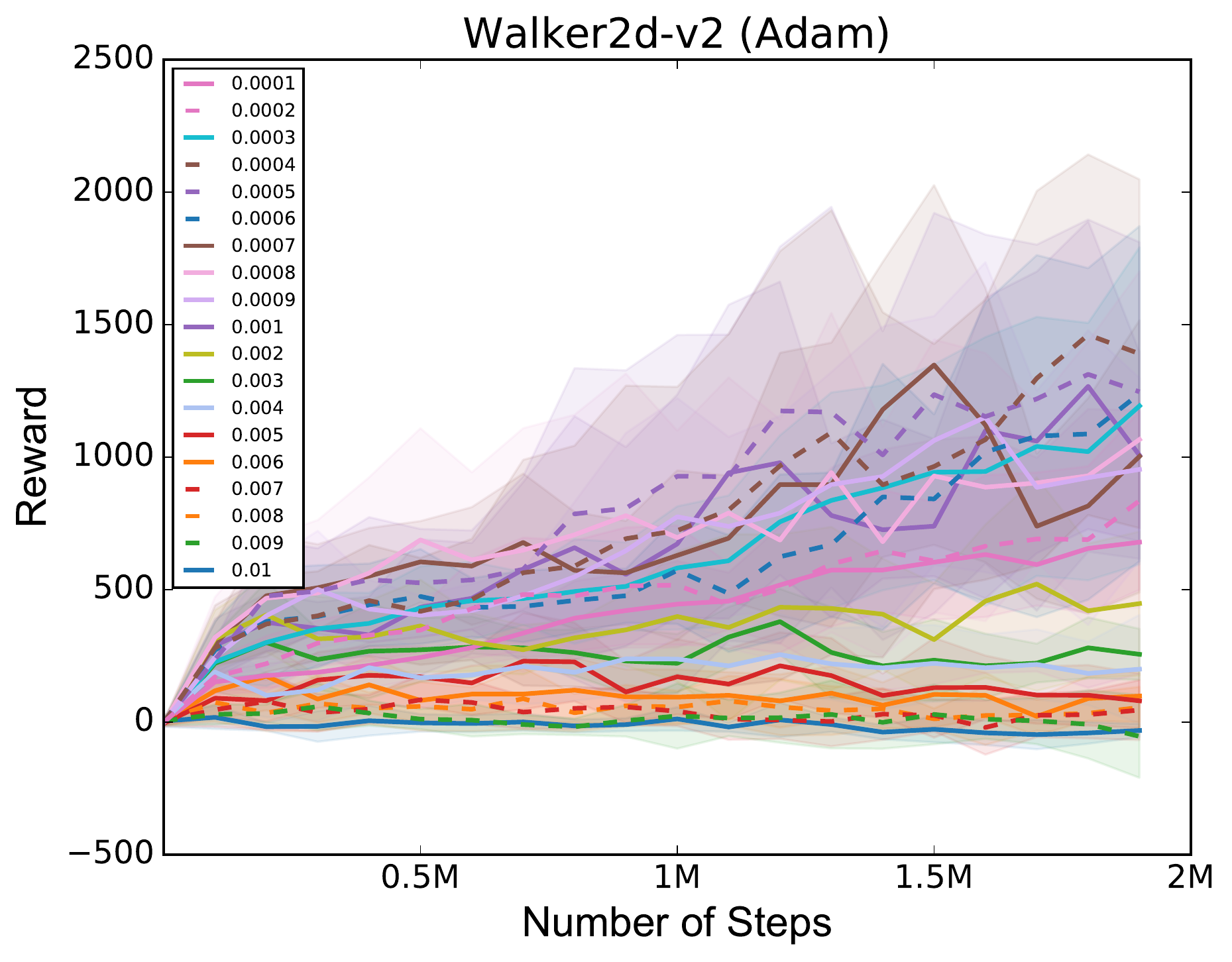}
    \includegraphics[width=.32\textwidth]{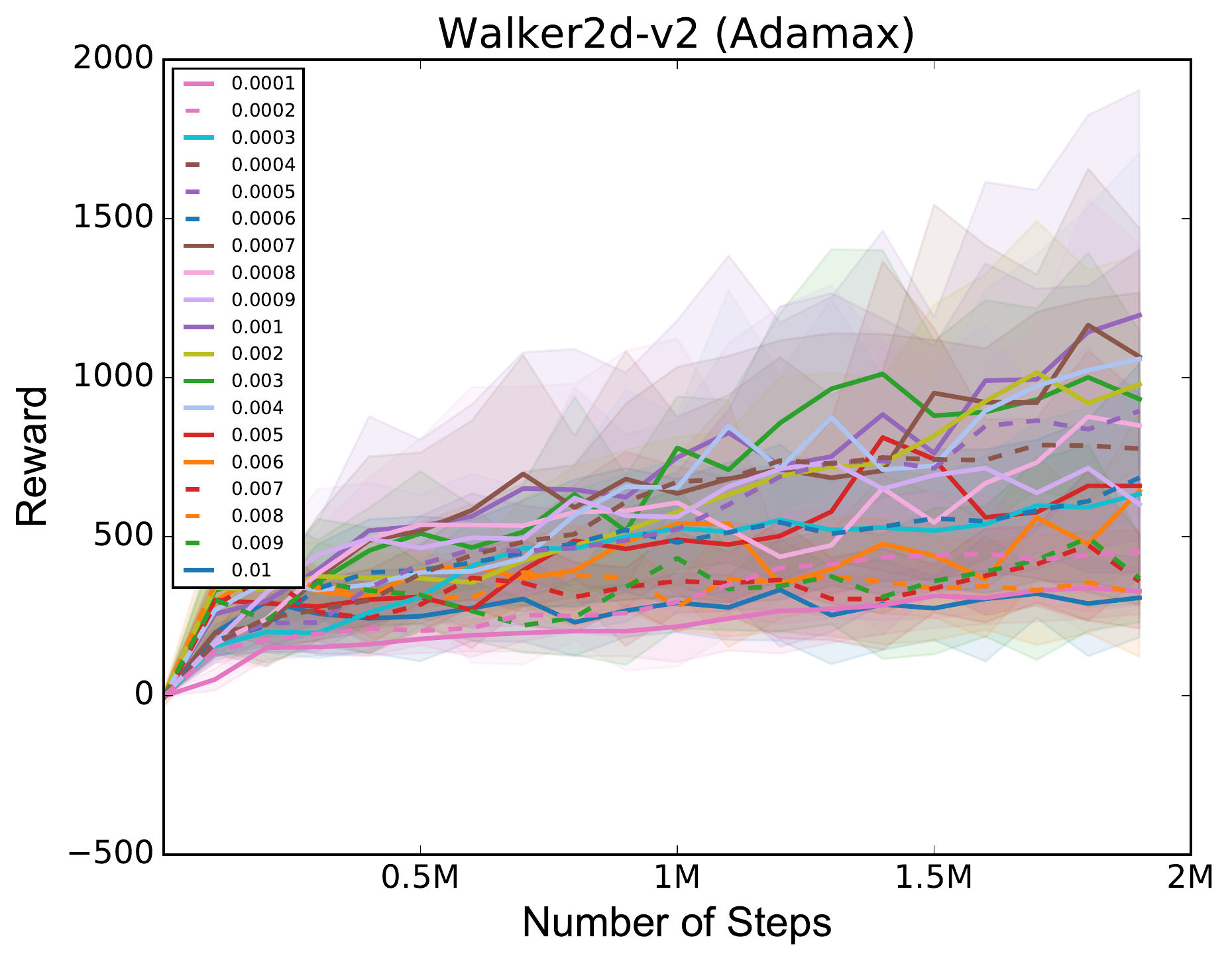}
    \includegraphics[width=.32\textwidth]{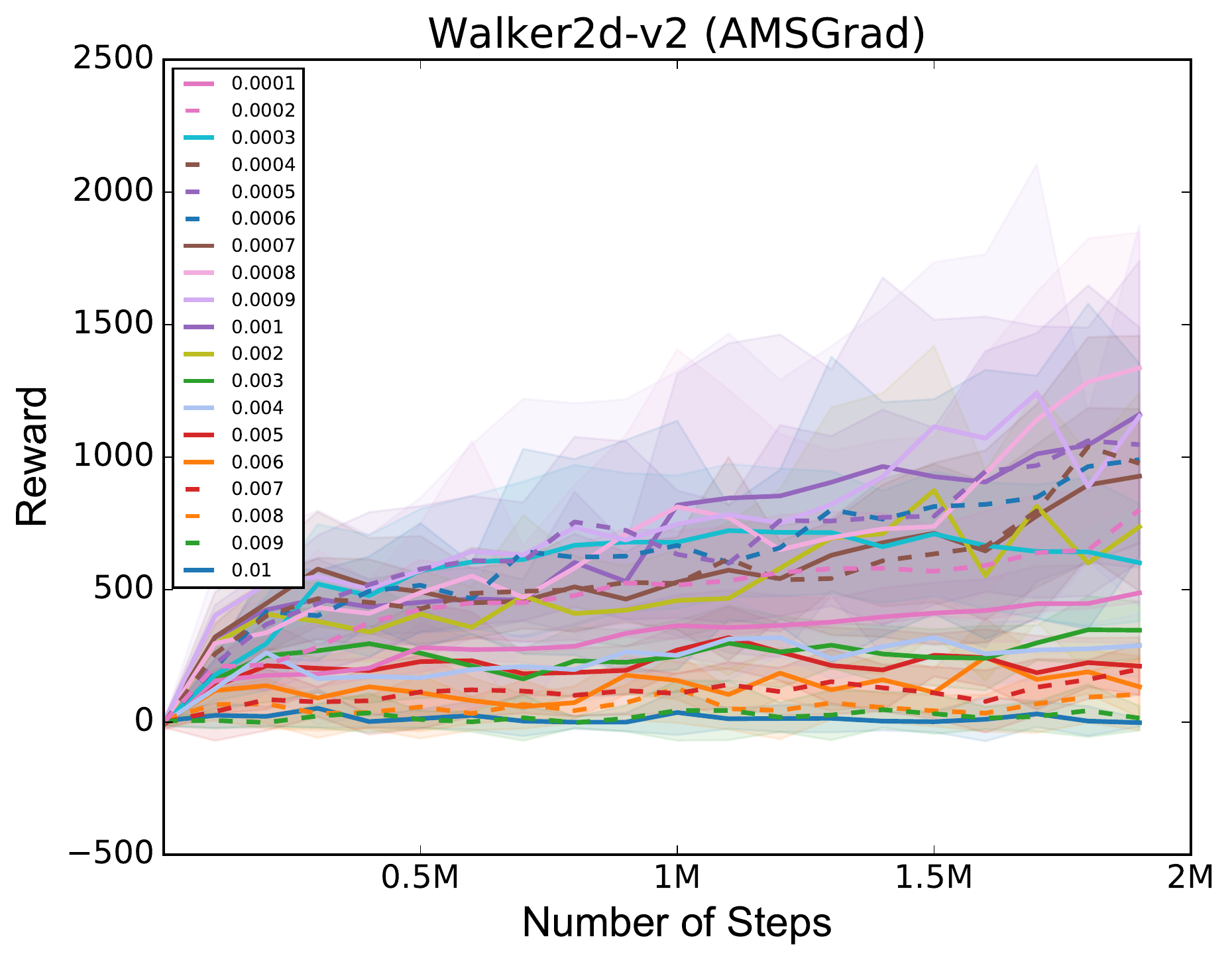}
    \includegraphics[width=.32\textwidth]{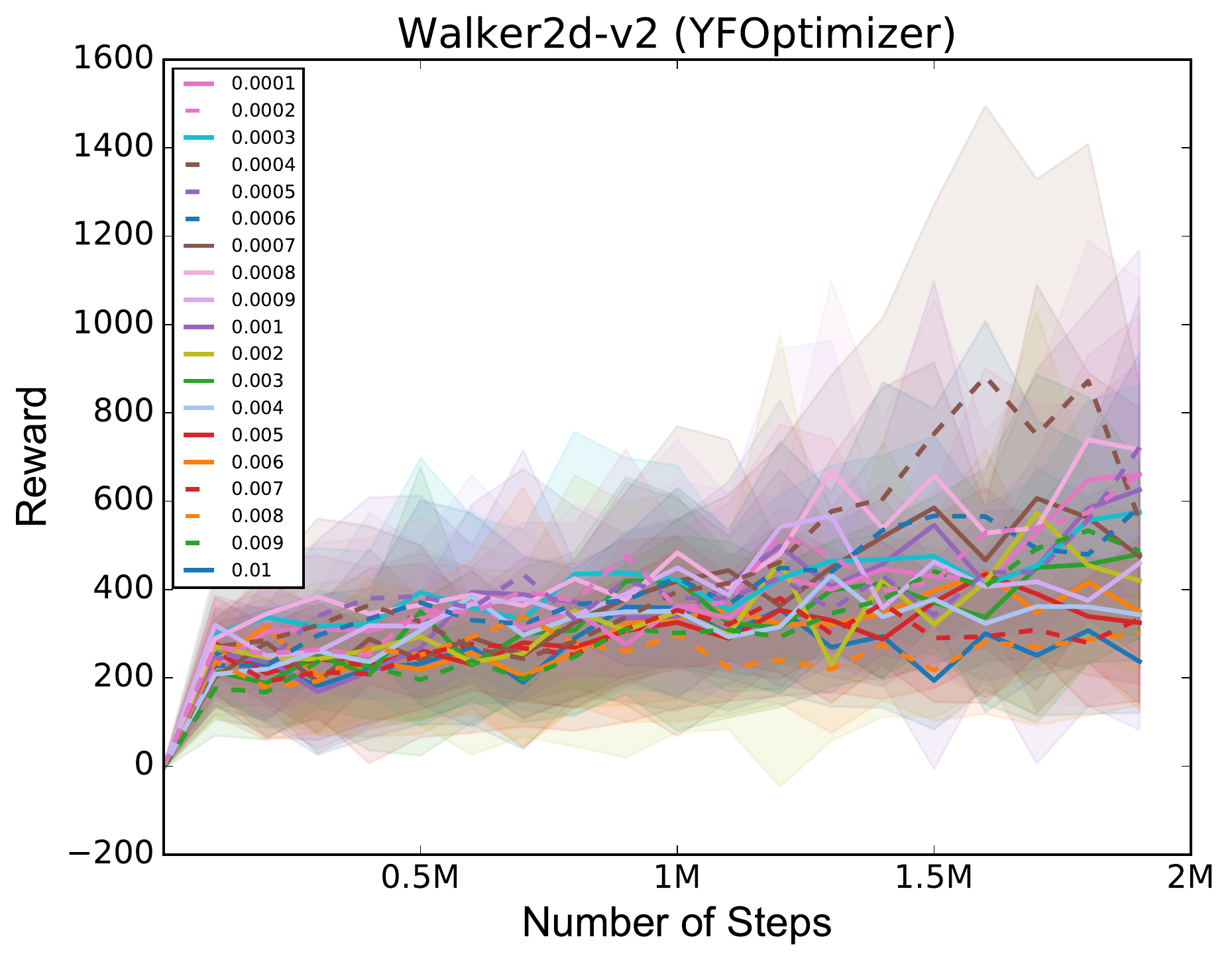}
    \caption{A2C performance across learning rates on the Walker2d environment.}
\end{figure}

\begin{figure}[H]
    \centering
    \includegraphics[width=.32\textwidth]{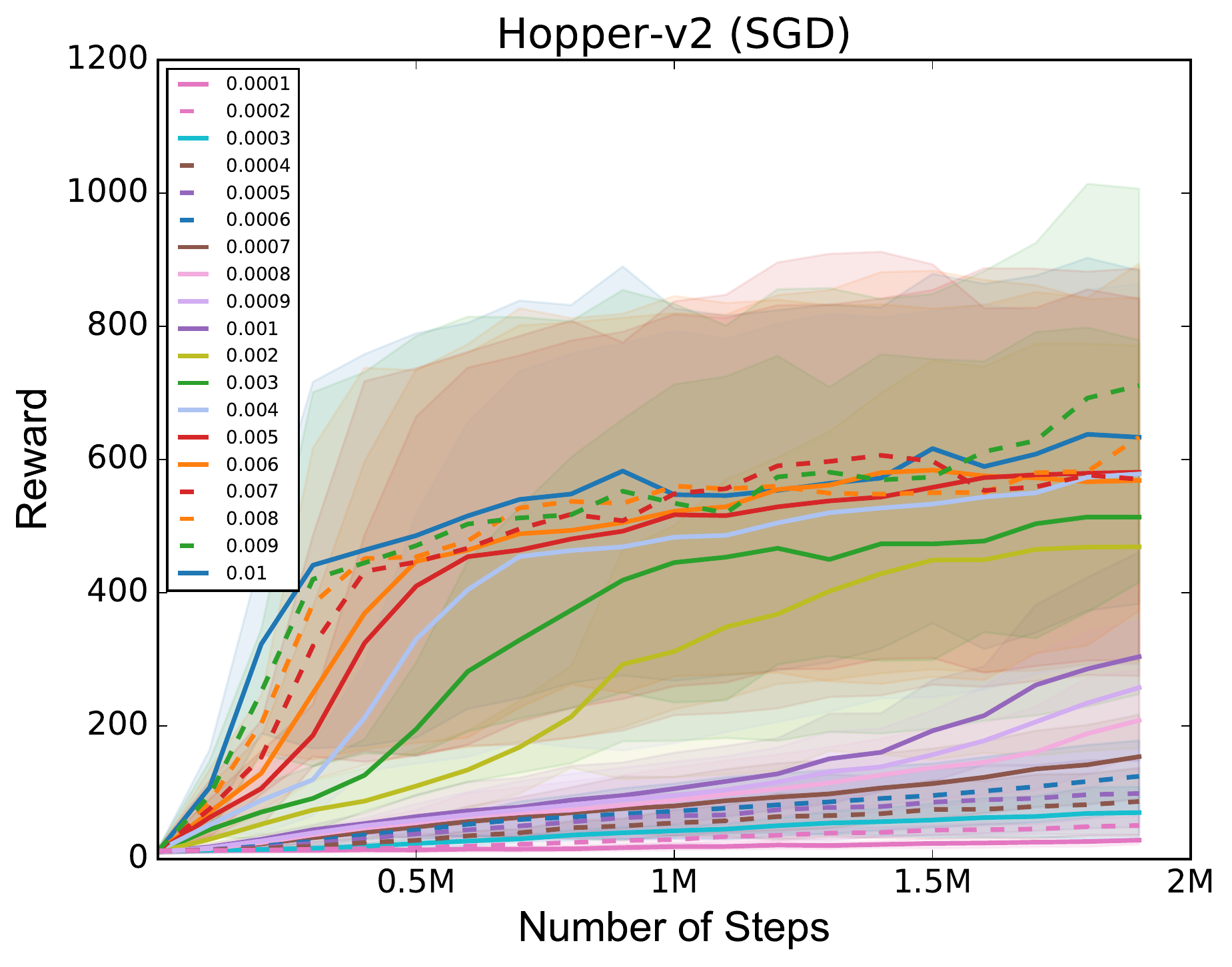}
    \includegraphics[width=.32\textwidth]{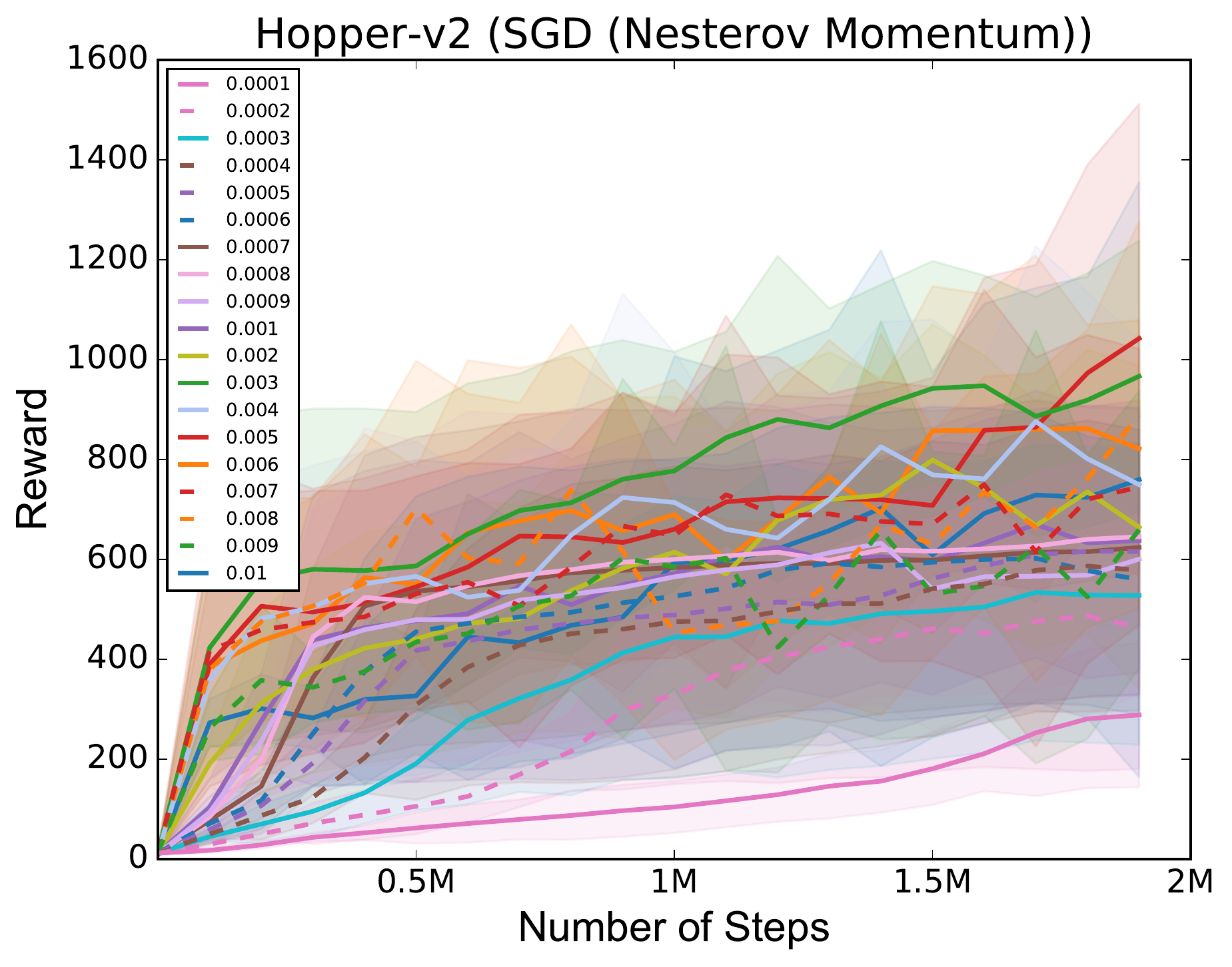}
    \includegraphics[width=.32\textwidth]{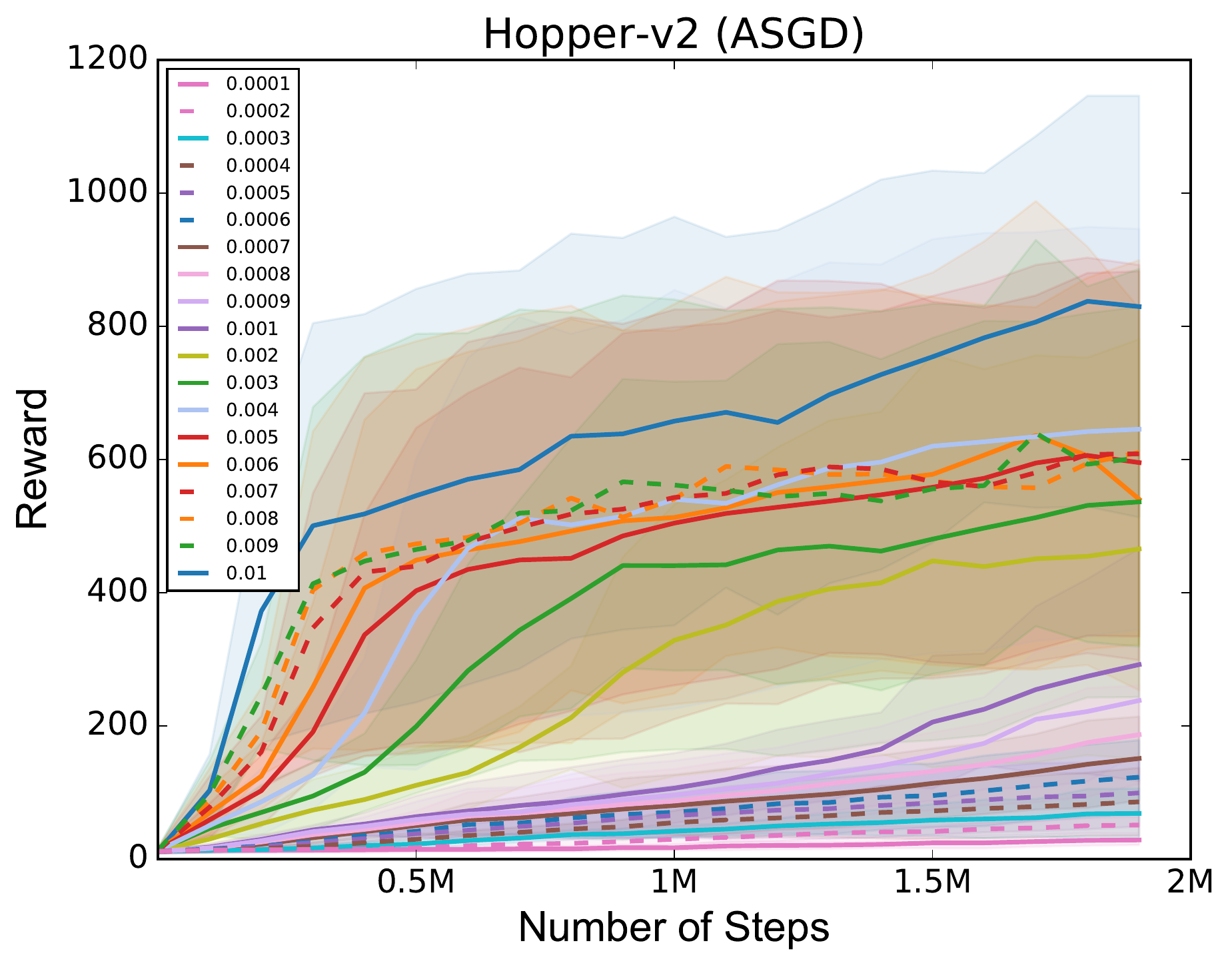}
    \includegraphics[width=.32\textwidth]{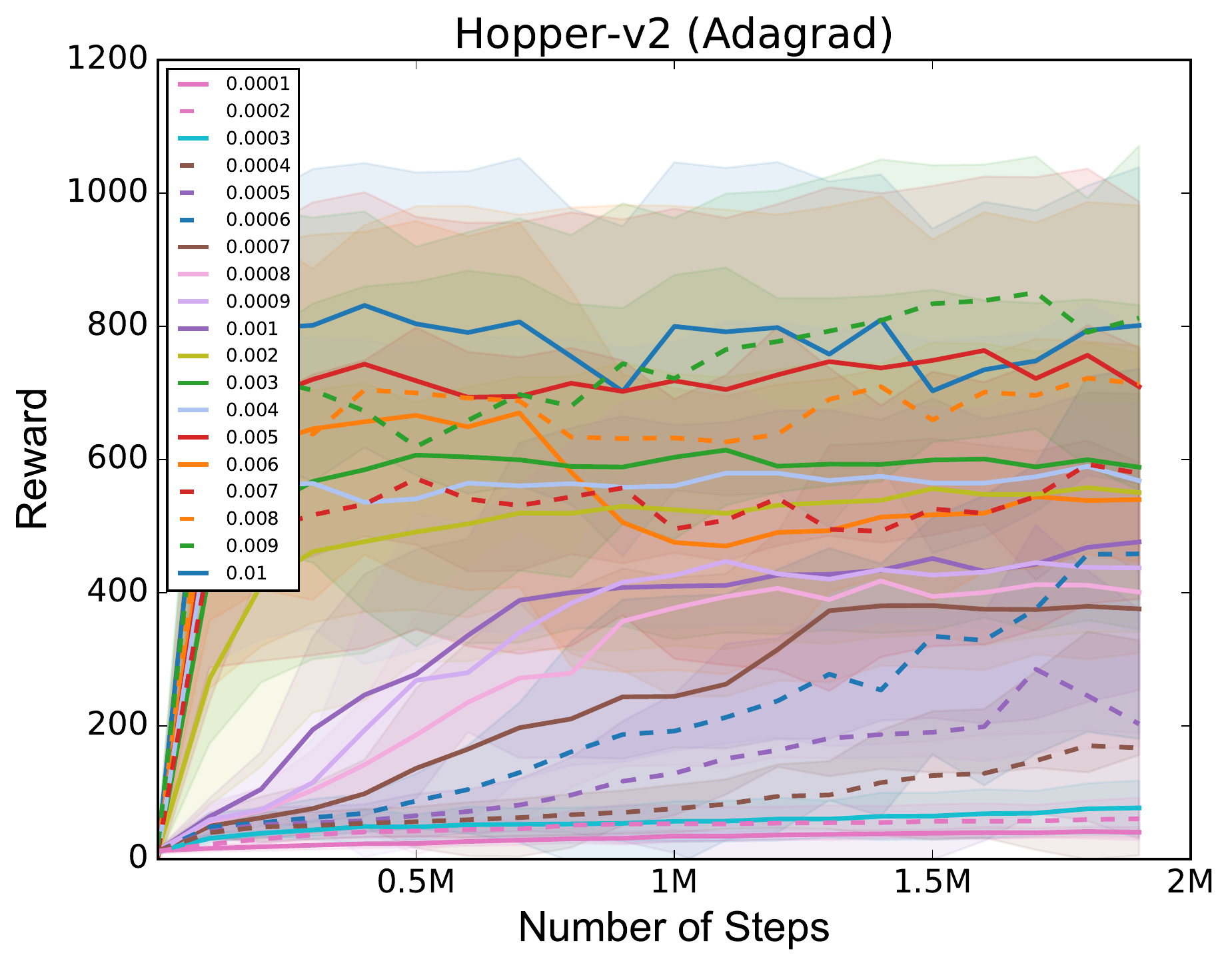}
    \includegraphics[width=.32\textwidth]{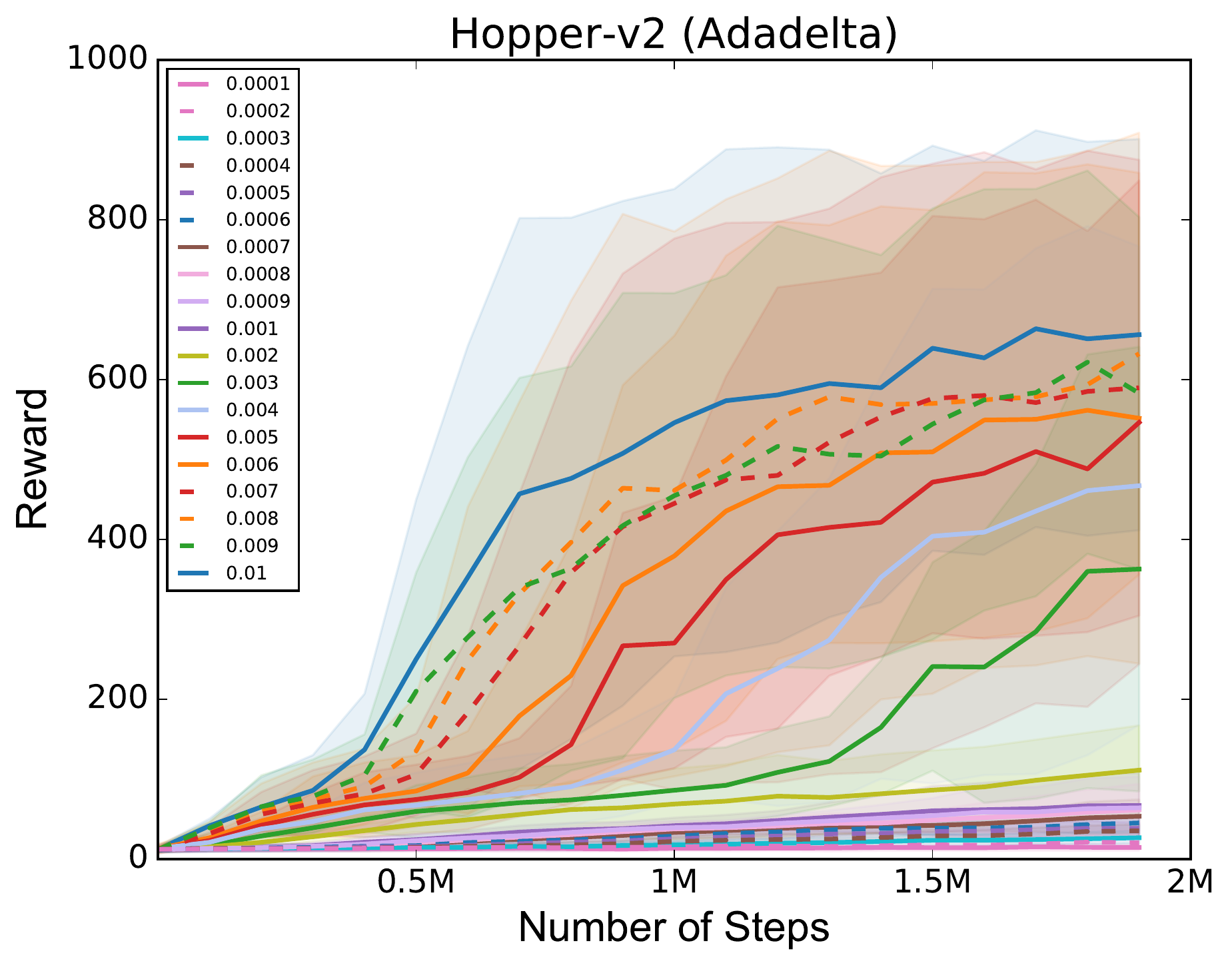}
    \includegraphics[width=.32\textwidth]{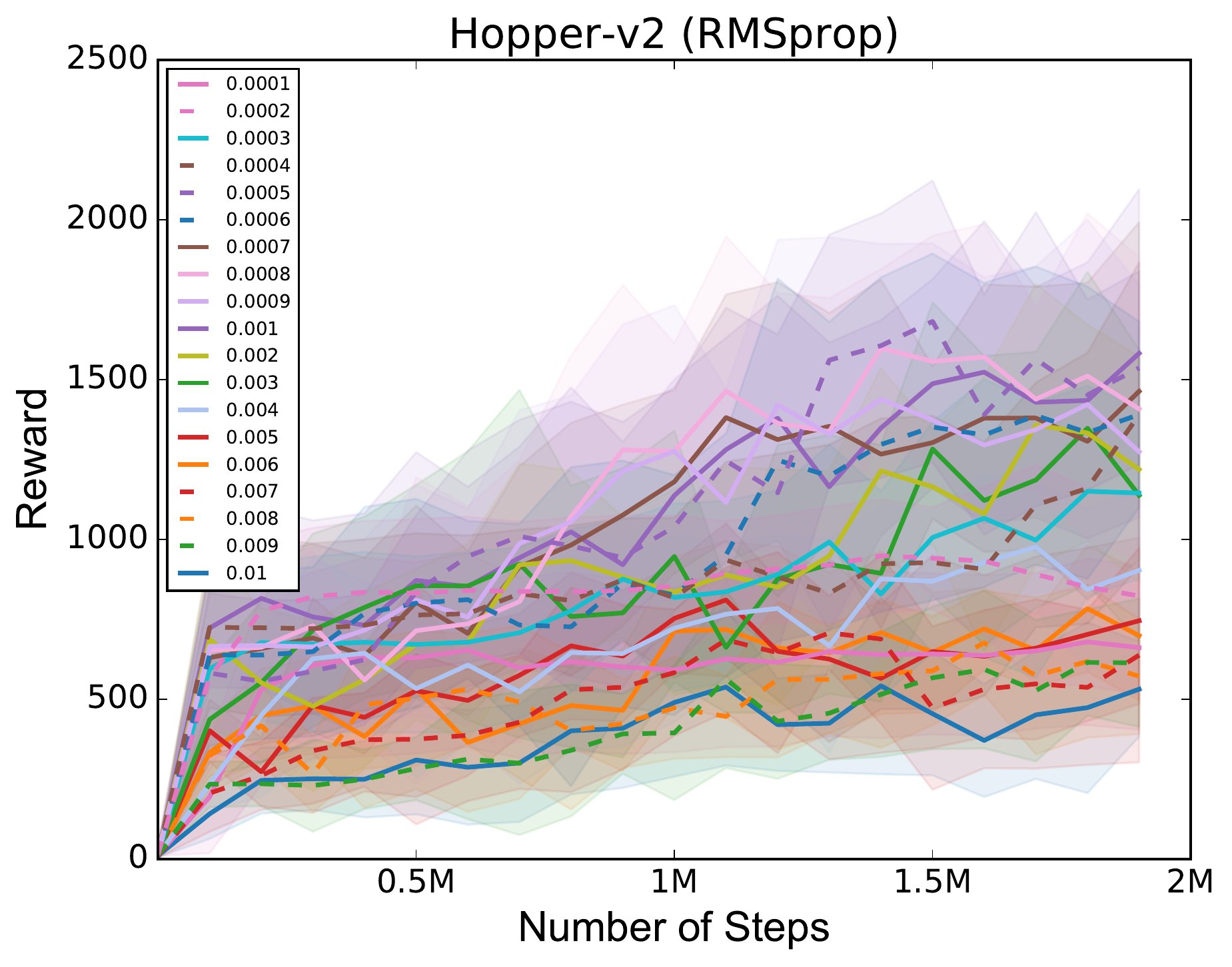}
    \includegraphics[width=.32\textwidth]{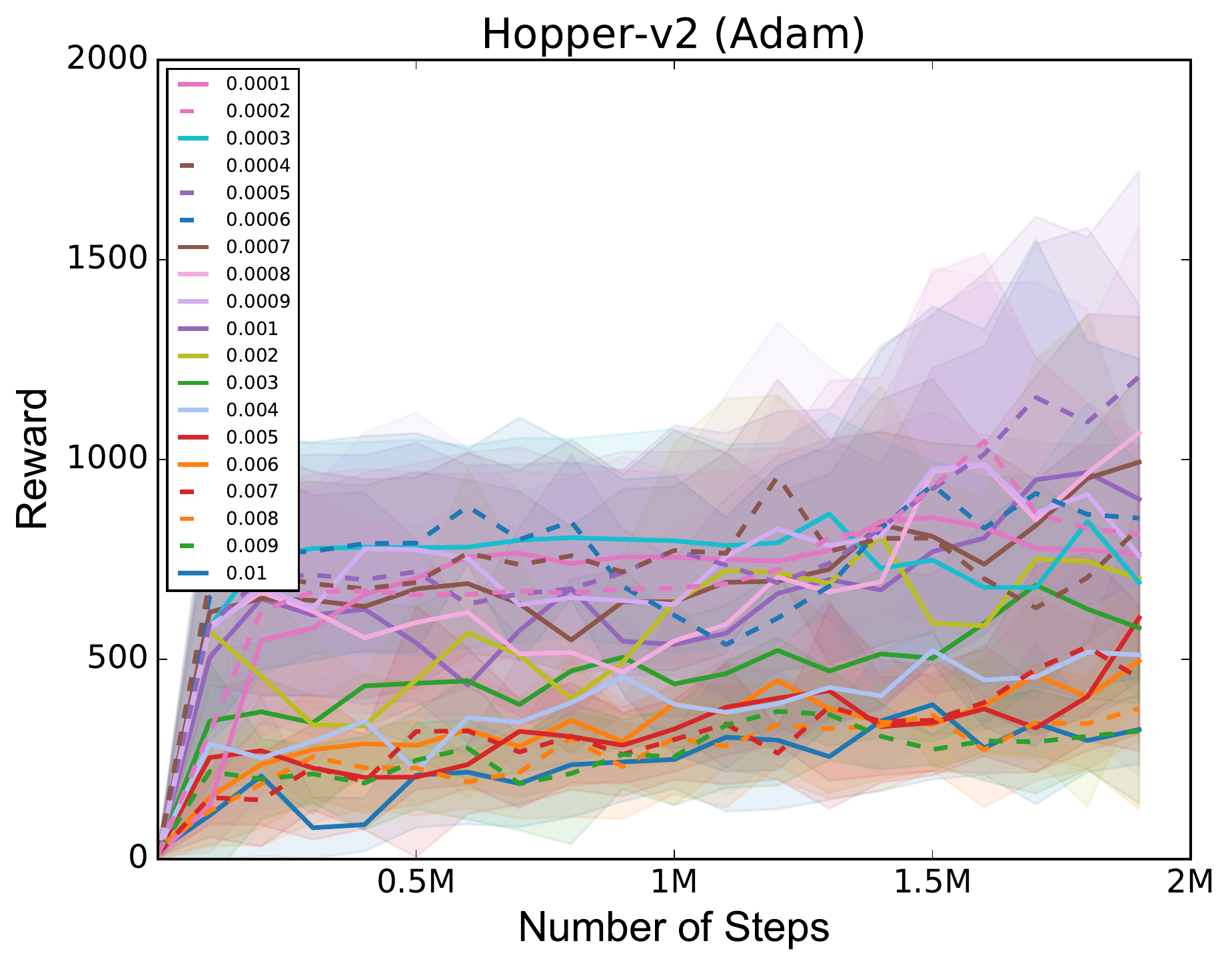}
    \includegraphics[width=.32\textwidth]{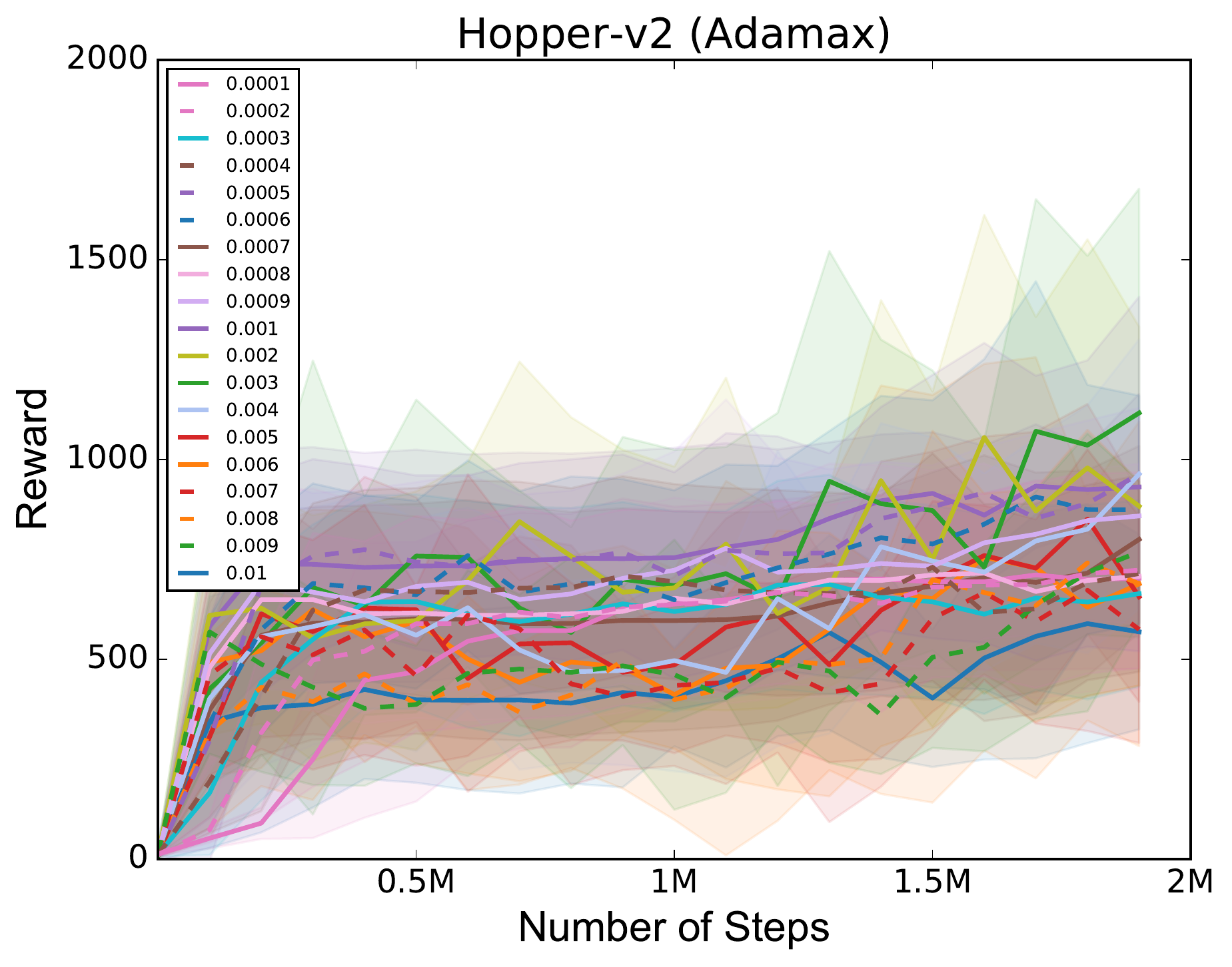}
    \includegraphics[width=.32\textwidth]{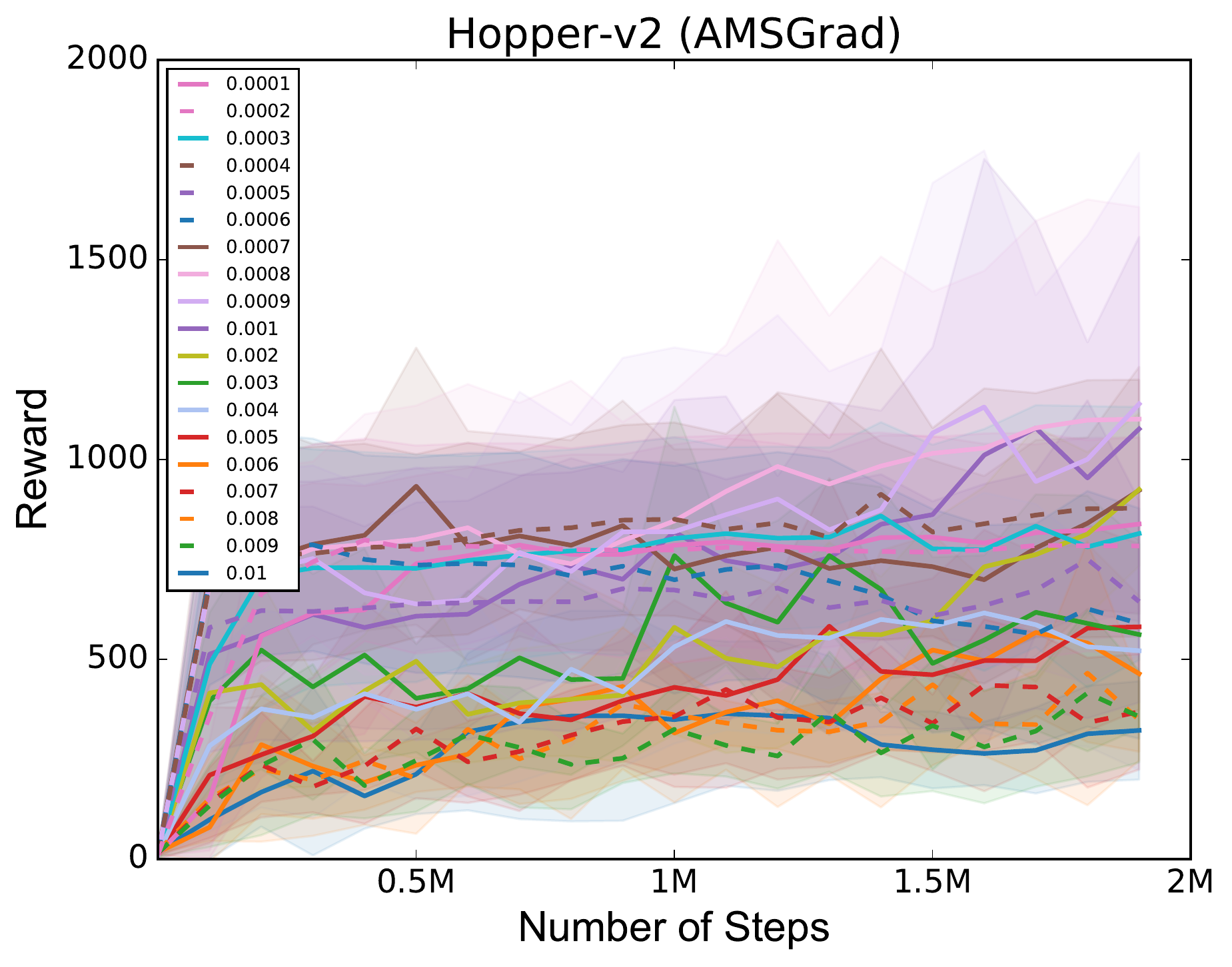}
    \includegraphics[width=.32\textwidth]{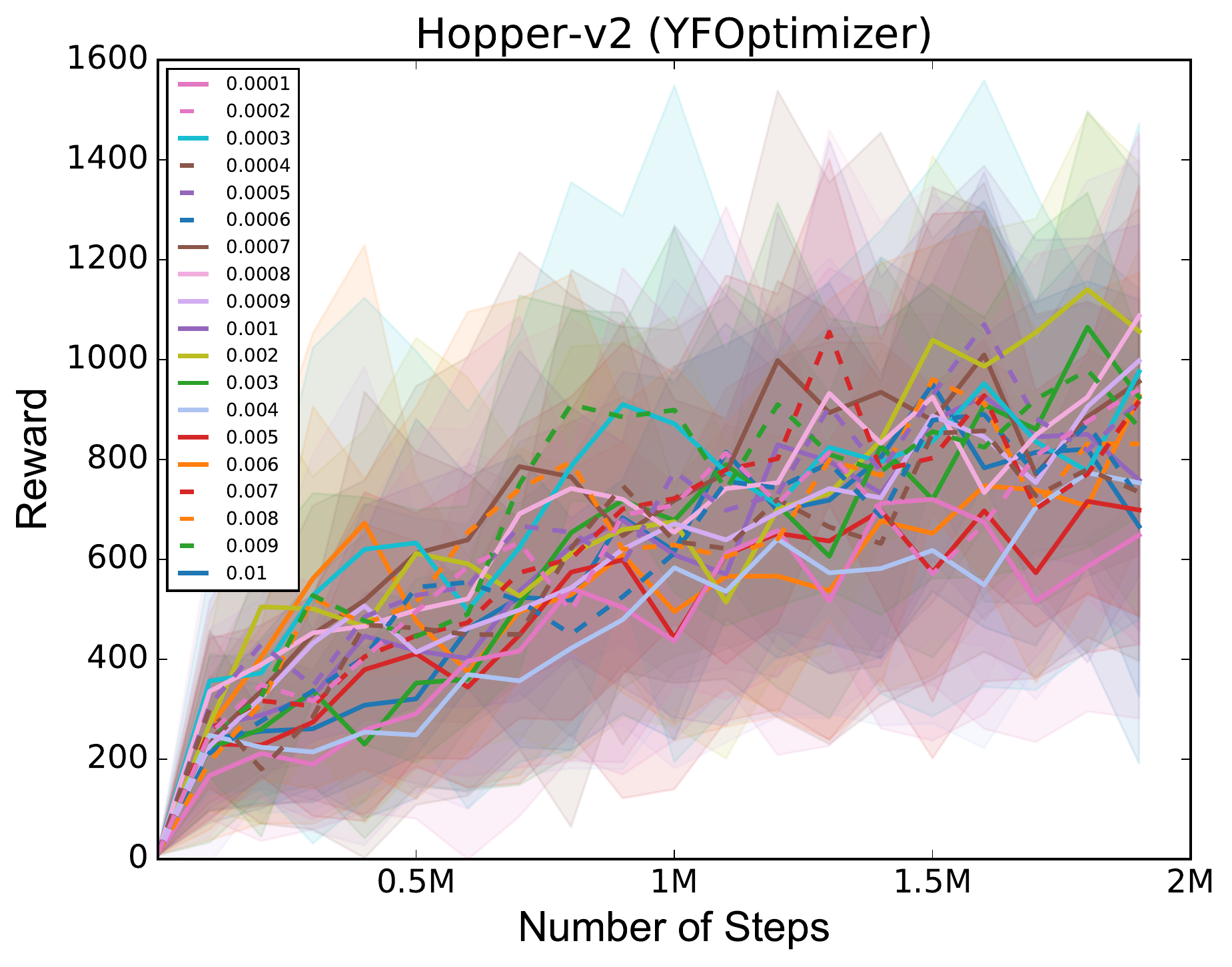}
    \caption{A2C performance across learning rates on the Hopper environment.}
\end{figure}

\begin{figure}[H]
    \centering
    \includegraphics[width=.32\textwidth]{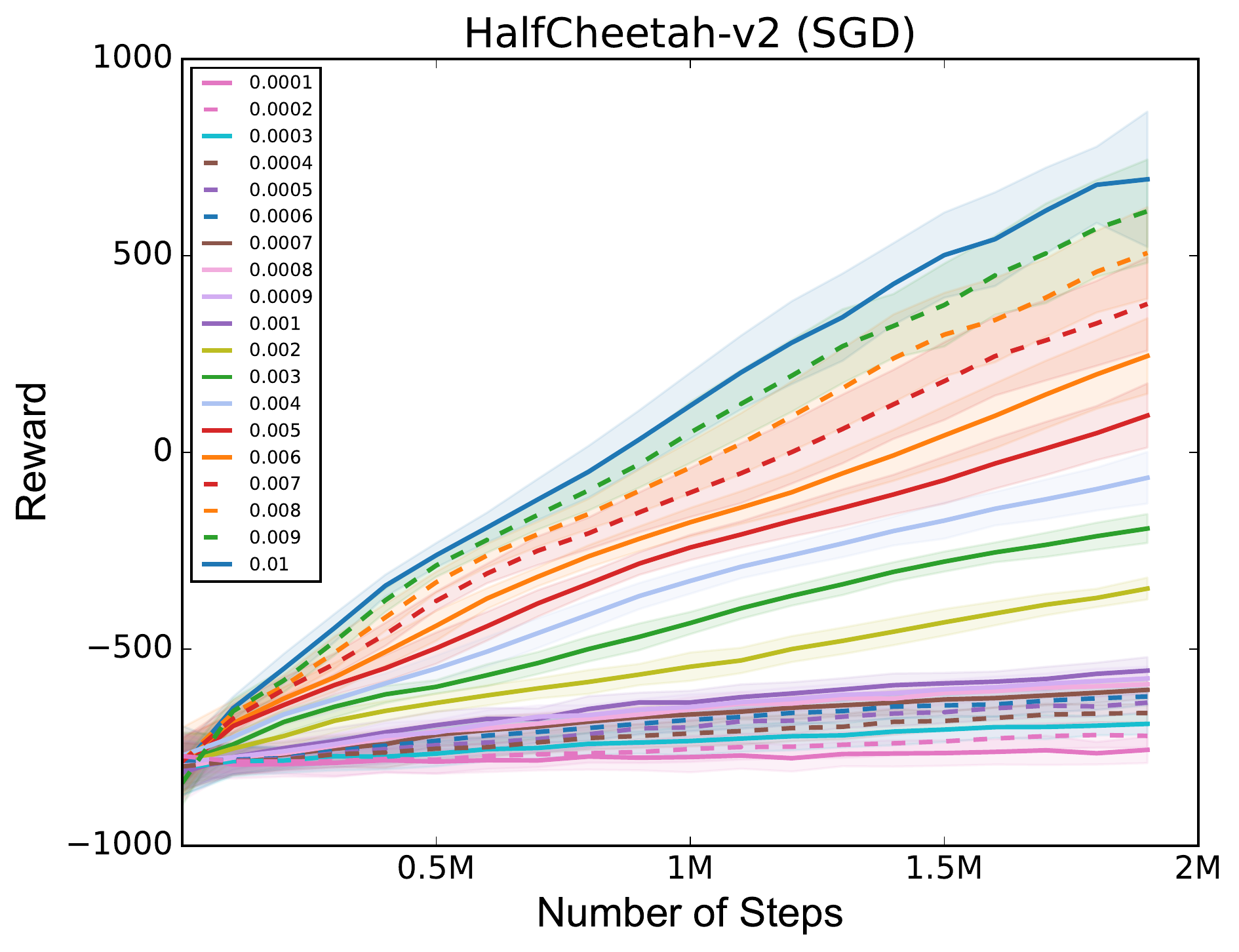}
    \includegraphics[width=.32\textwidth]{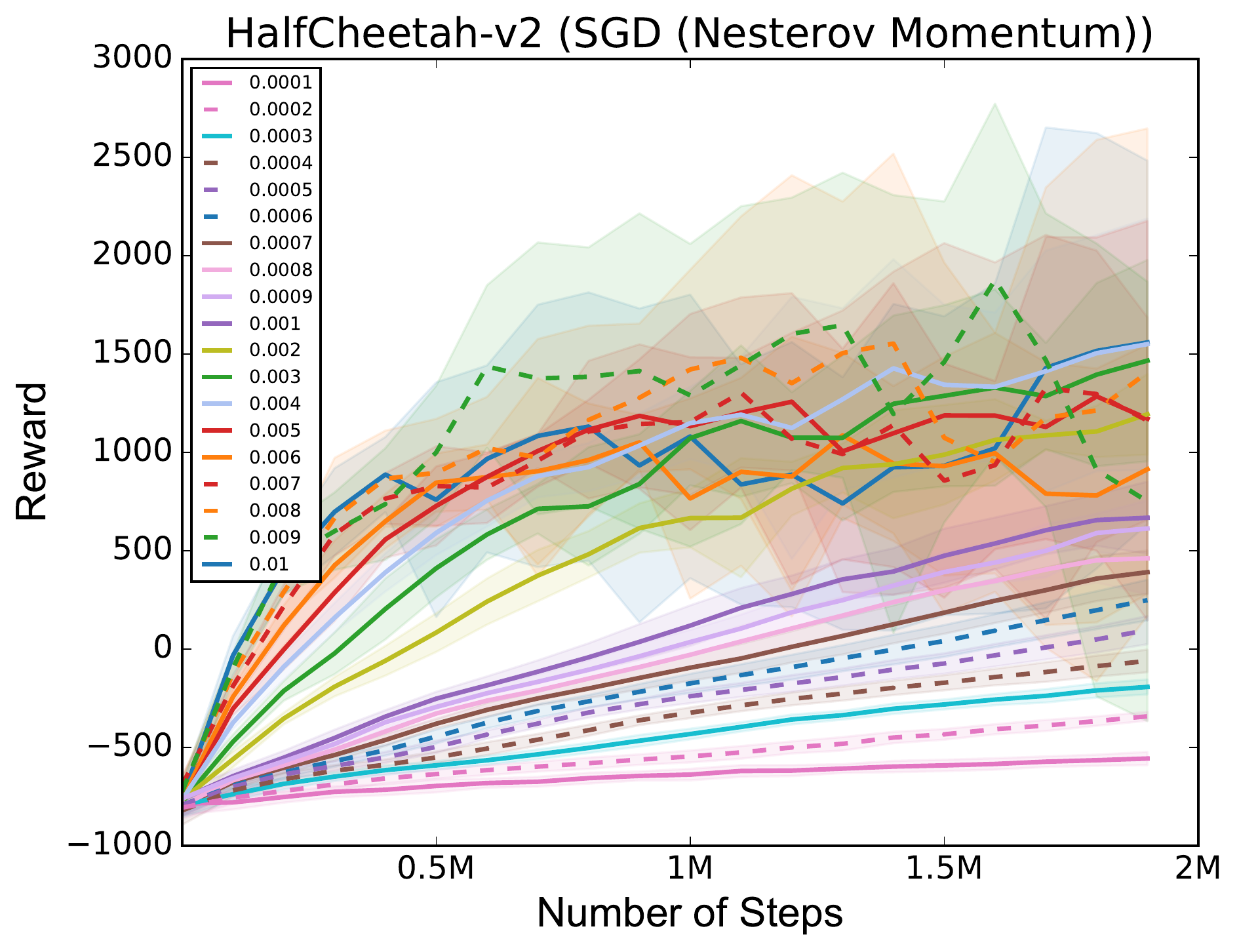}
    \includegraphics[width=.32\textwidth]{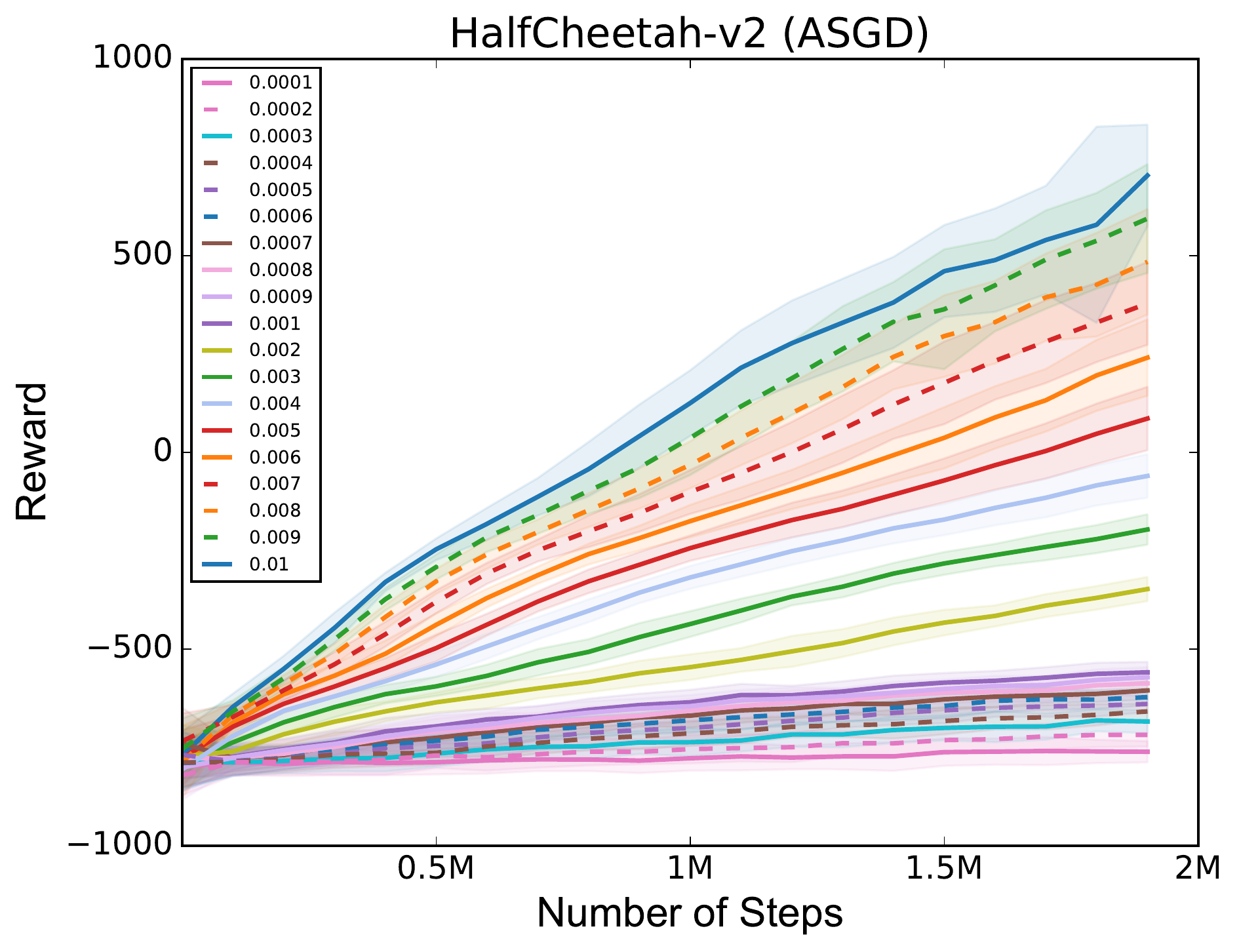}
    \includegraphics[width=.32\textwidth]{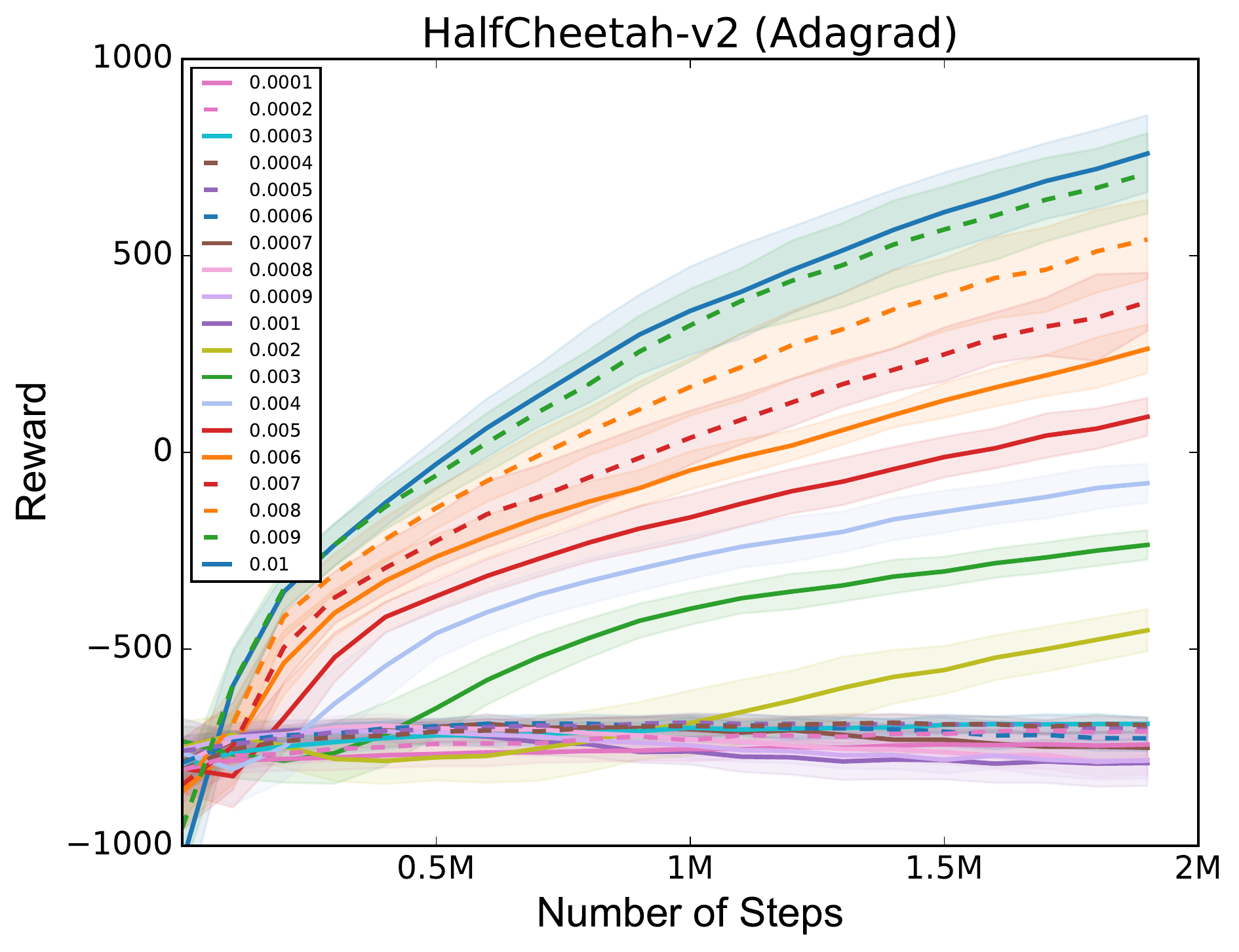}
    \includegraphics[width=.32\textwidth]{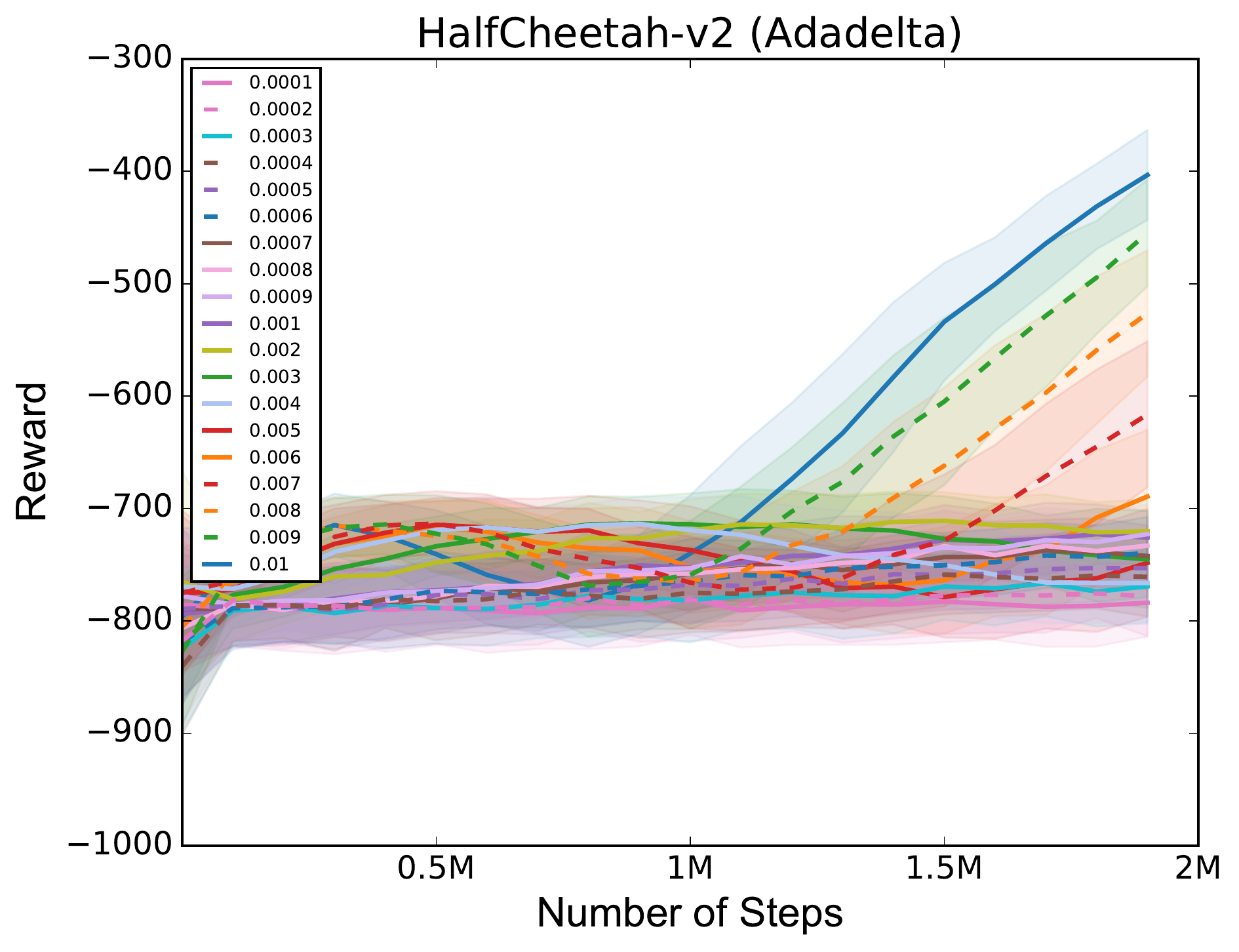}
    \includegraphics[width=.32\textwidth]{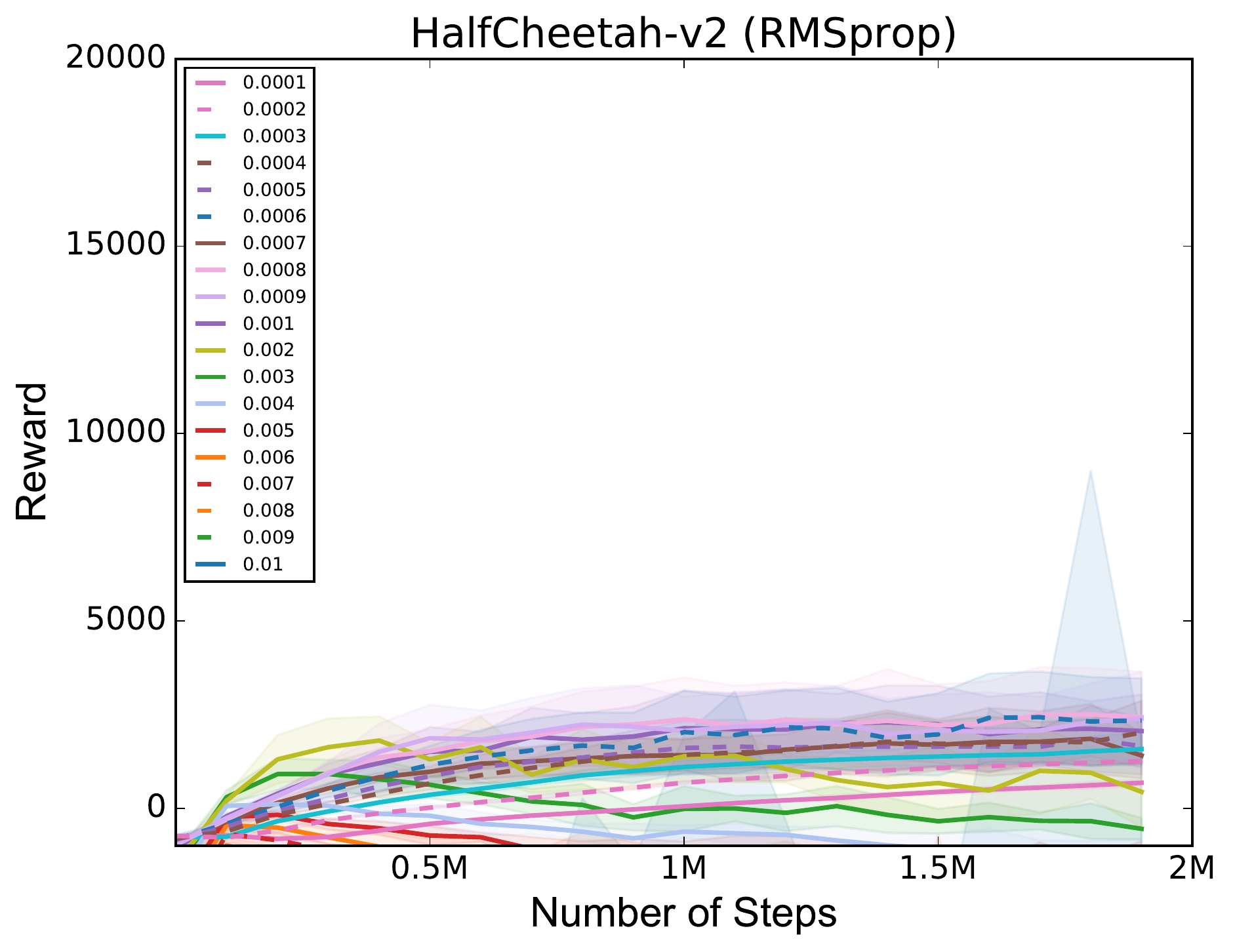}
    \includegraphics[width=.32\textwidth]{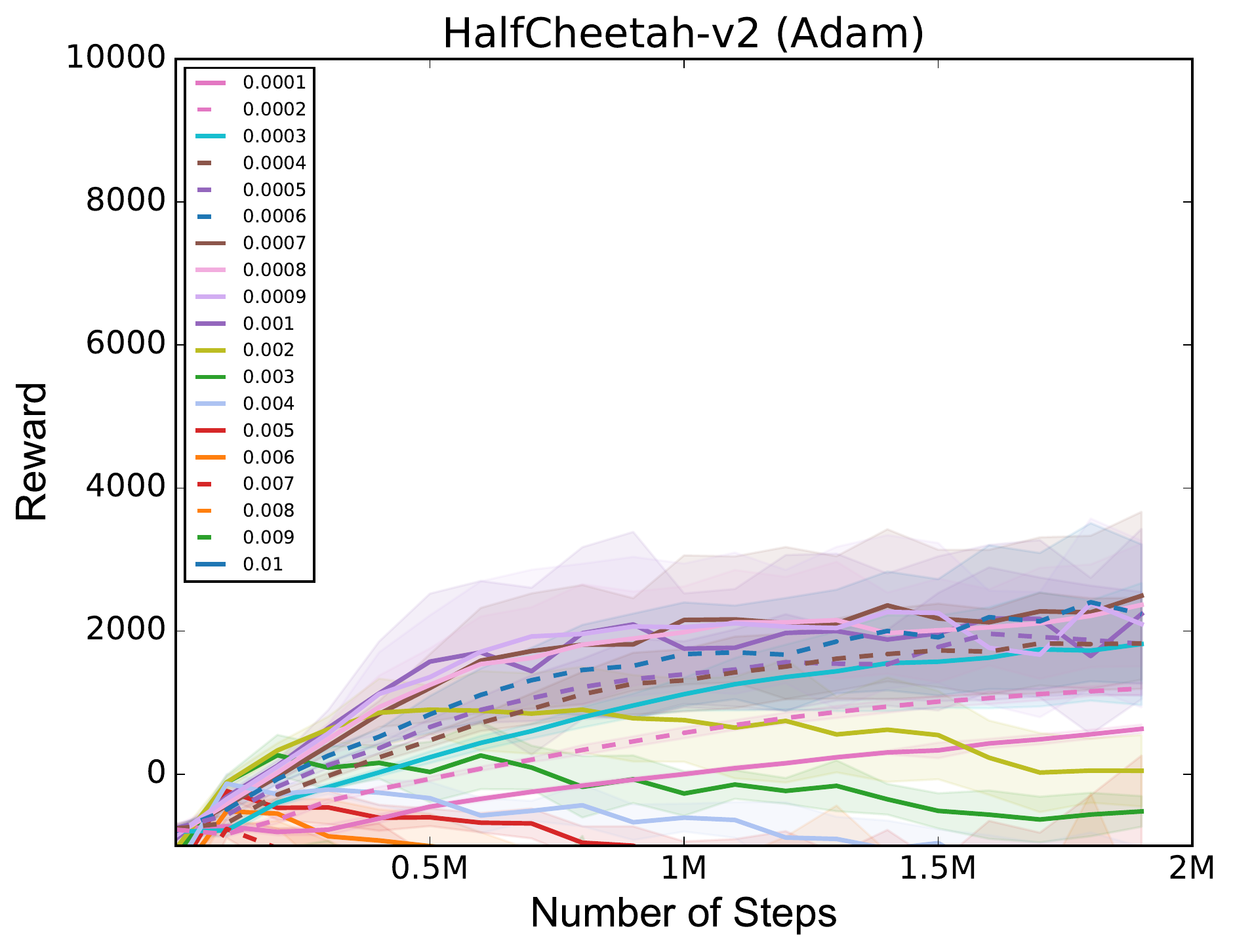}
    \includegraphics[width=.32\textwidth]{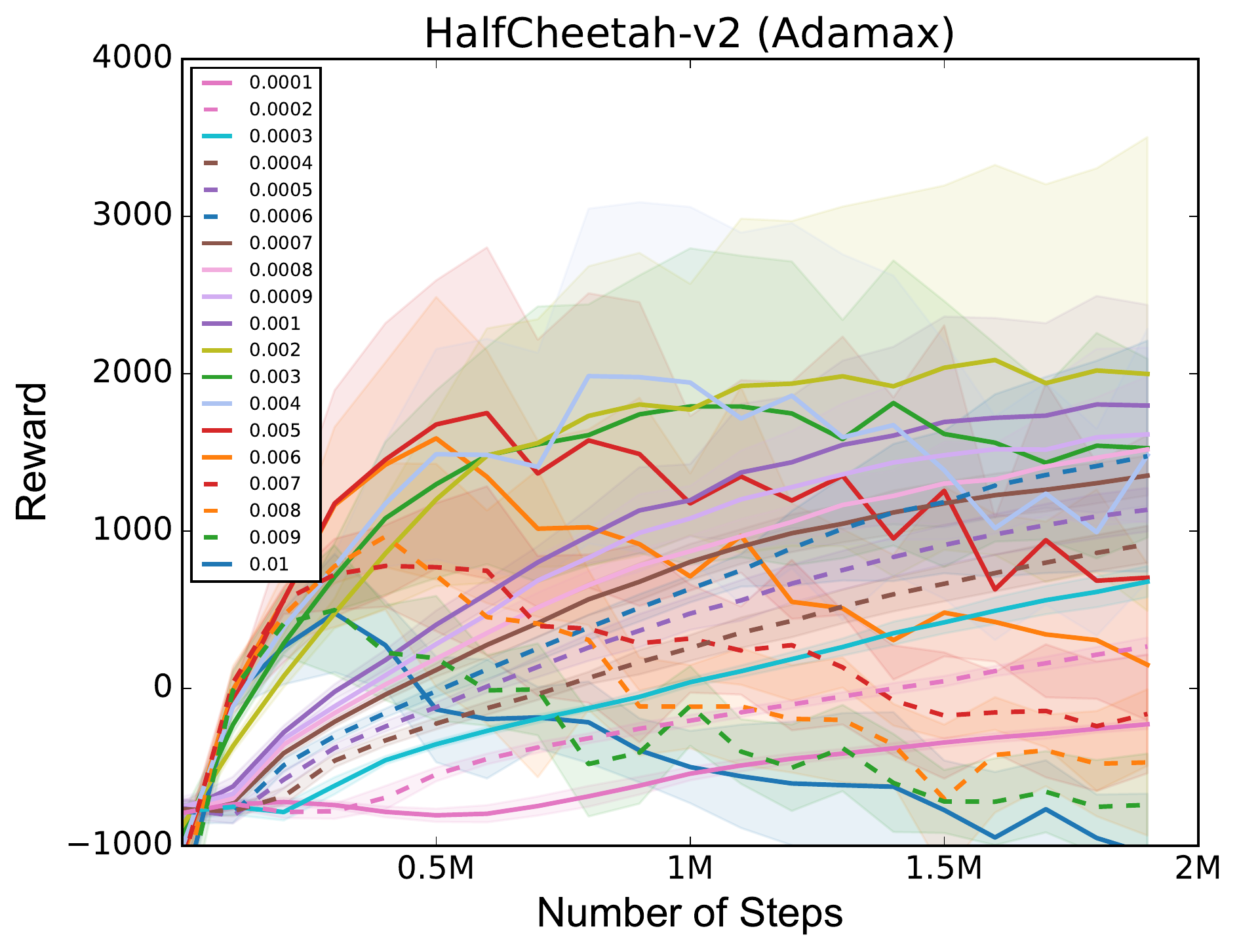}
    \includegraphics[width=.32\textwidth]{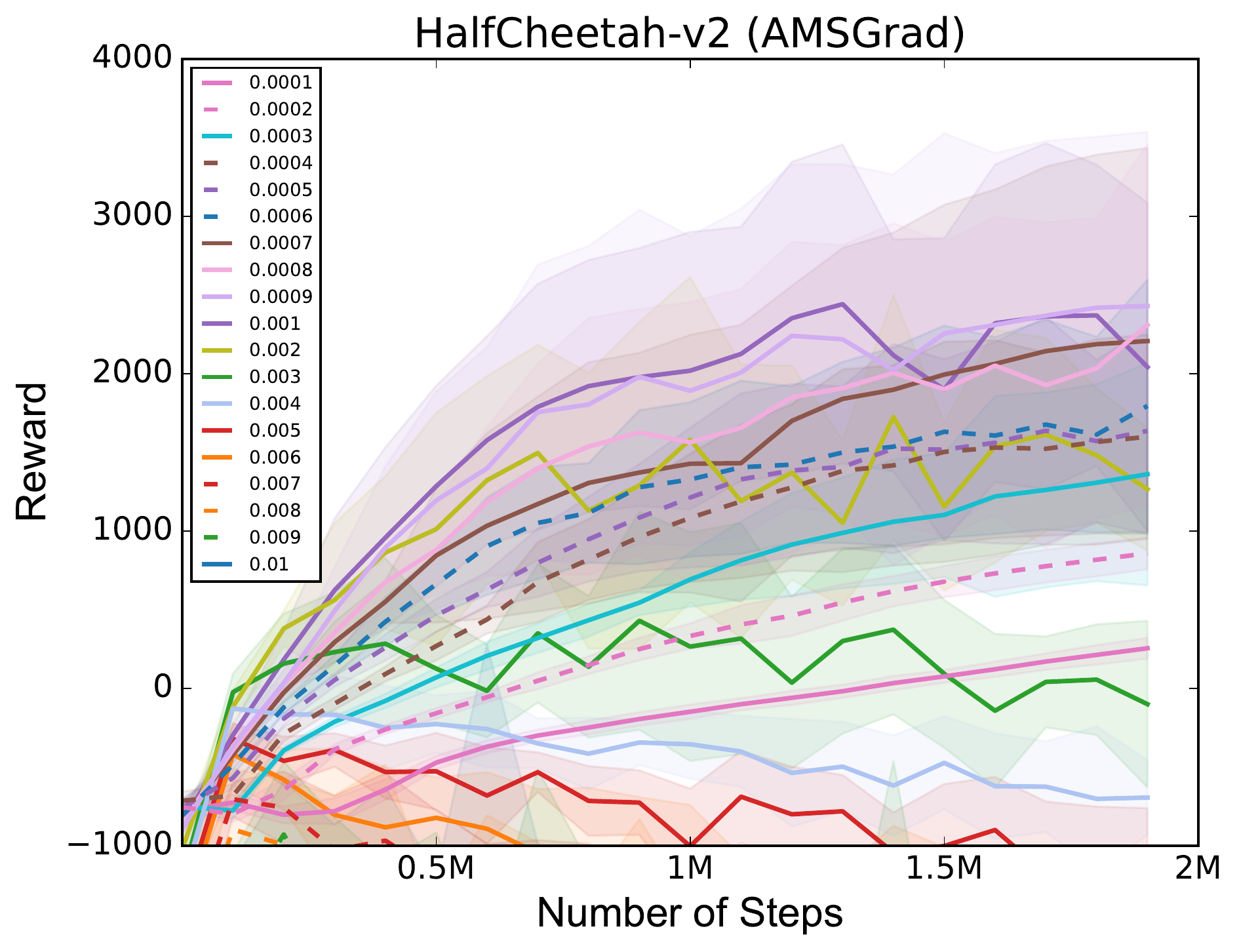}
    \includegraphics[width=.32\textwidth]{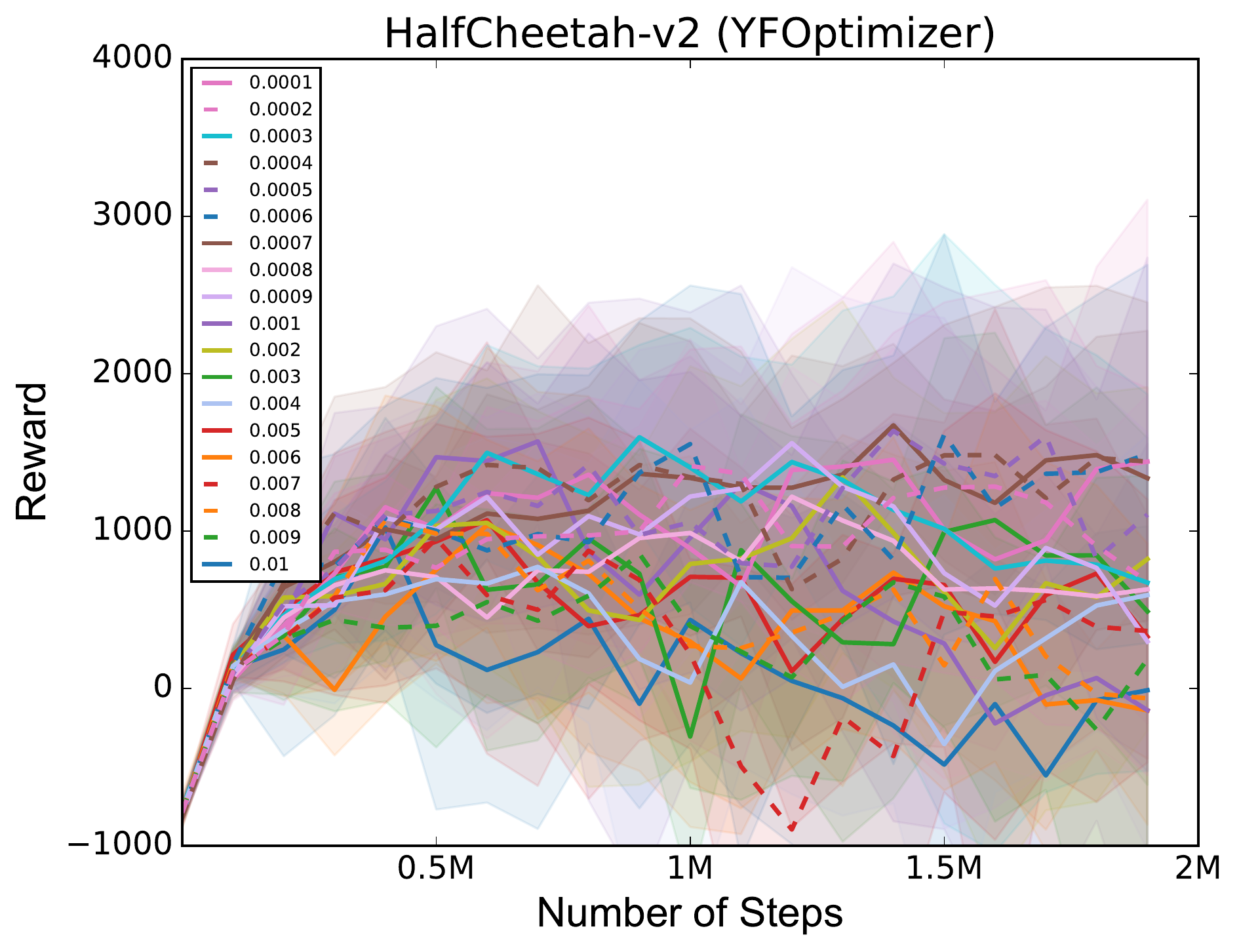}
    \caption{A2C performance across learning rates on the HalfCheetah environment.}
\end{figure}

\subsection{Learning Rate Alpha Plots}
\label{app:lrs_alpha}
\begin{figure}[H]
    \centering
        \includegraphics[width=.49\textwidth]{Ant-v2avg_results}
        \includegraphics[width=.49\textwidth]{HalfCheetah-v2avg_results}
        \includegraphics[width=.49\textwidth]{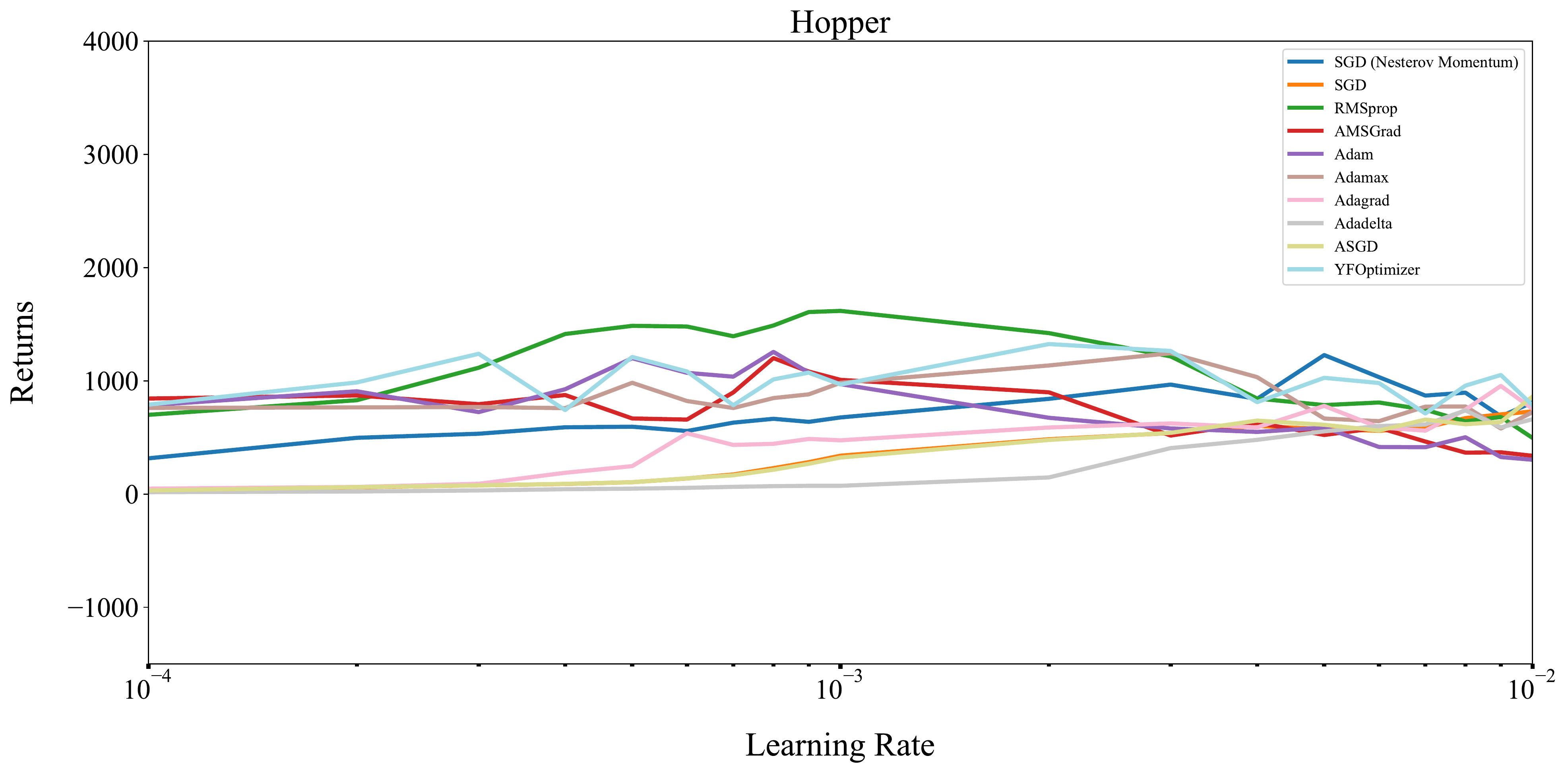}
        \includegraphics[width=.49\textwidth]{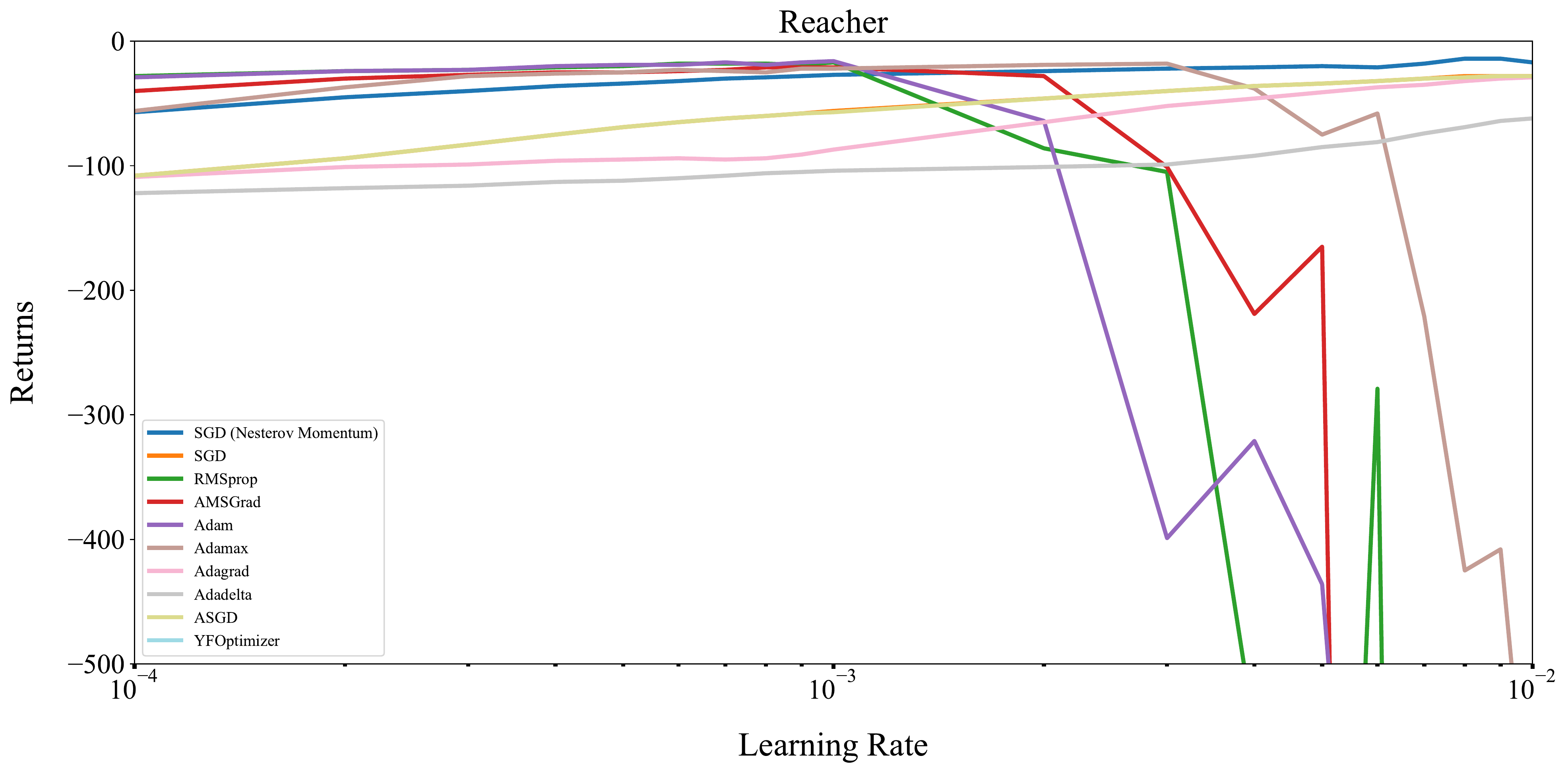}
        \includegraphics[width=.49\textwidth]{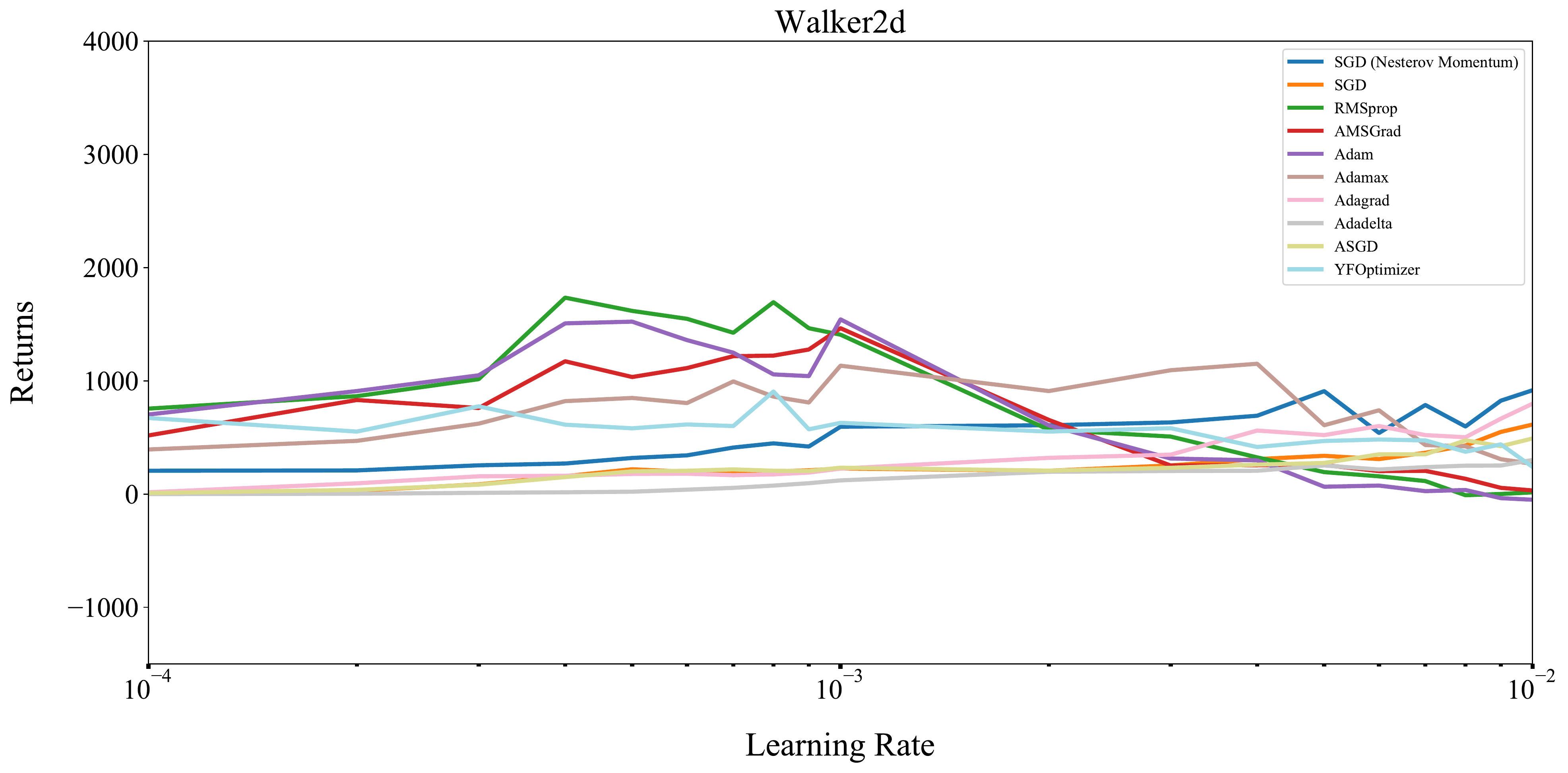}
    \caption{A2C asymptotic performance (averaged over last 50 episodes over 10 random seeds) at different learning rates.}
    \label{fig:alpha_asymptotic_a2c}
\end{figure}

\begin{figure}[H]
    \centering
        \includegraphics[width=.49\textwidth]{Ant-v2avg_results001}
        \includegraphics[width=.49\textwidth]{HalfCheetah-v2avg_results001}
        \includegraphics[width=.49\textwidth]{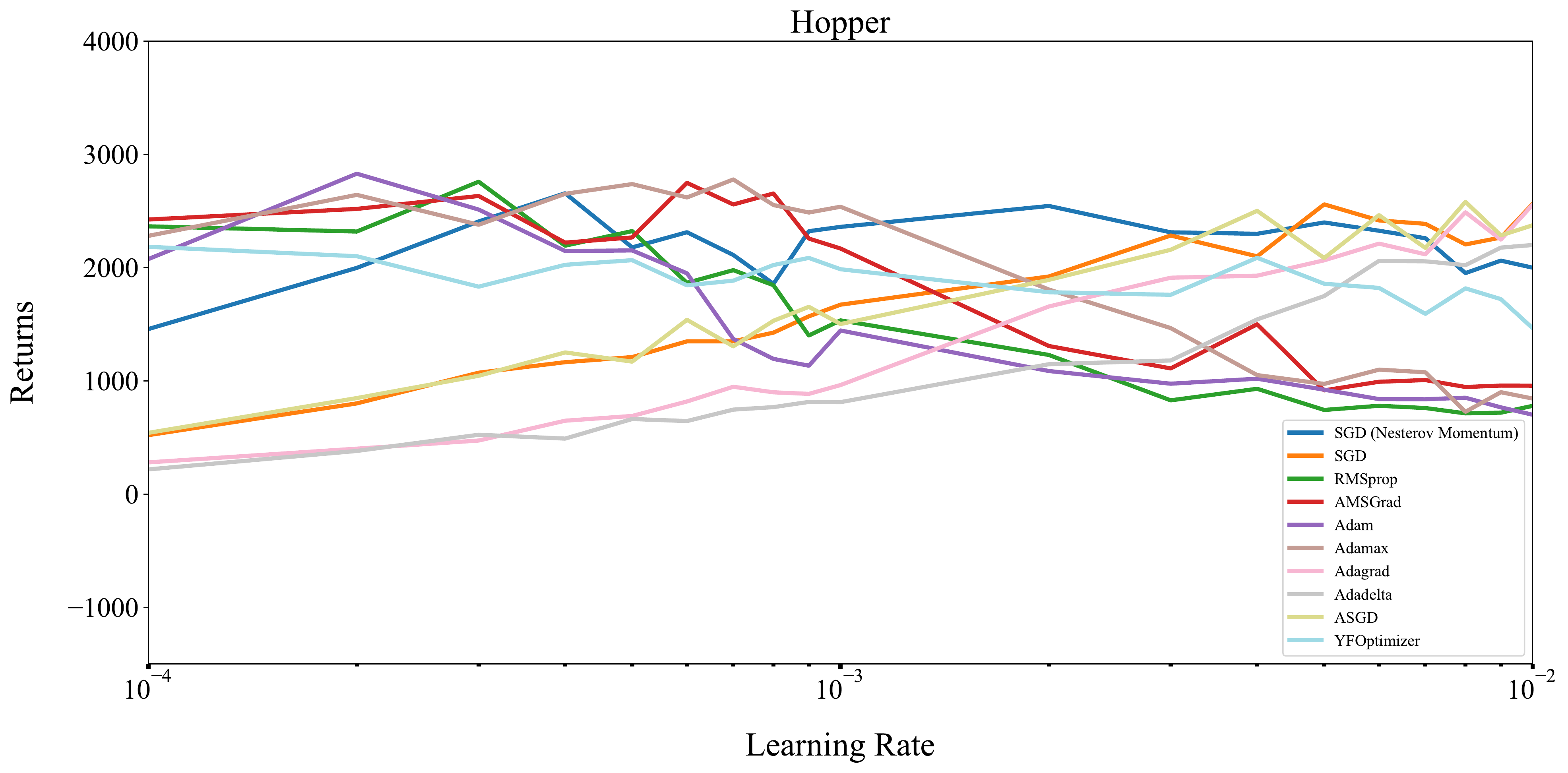}
        \includegraphics[width=.49\textwidth]{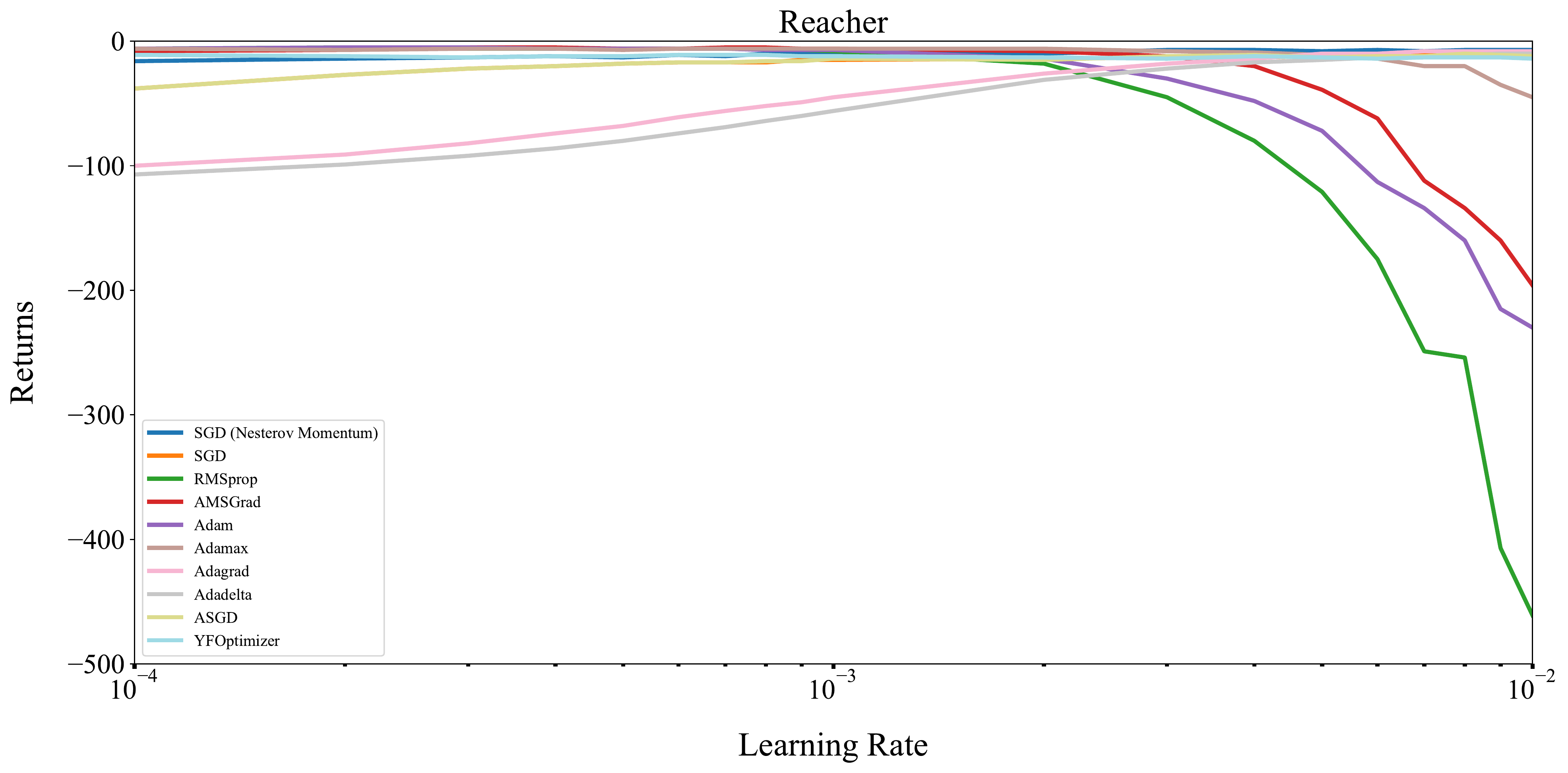}
        \includegraphics[width=.49\textwidth]{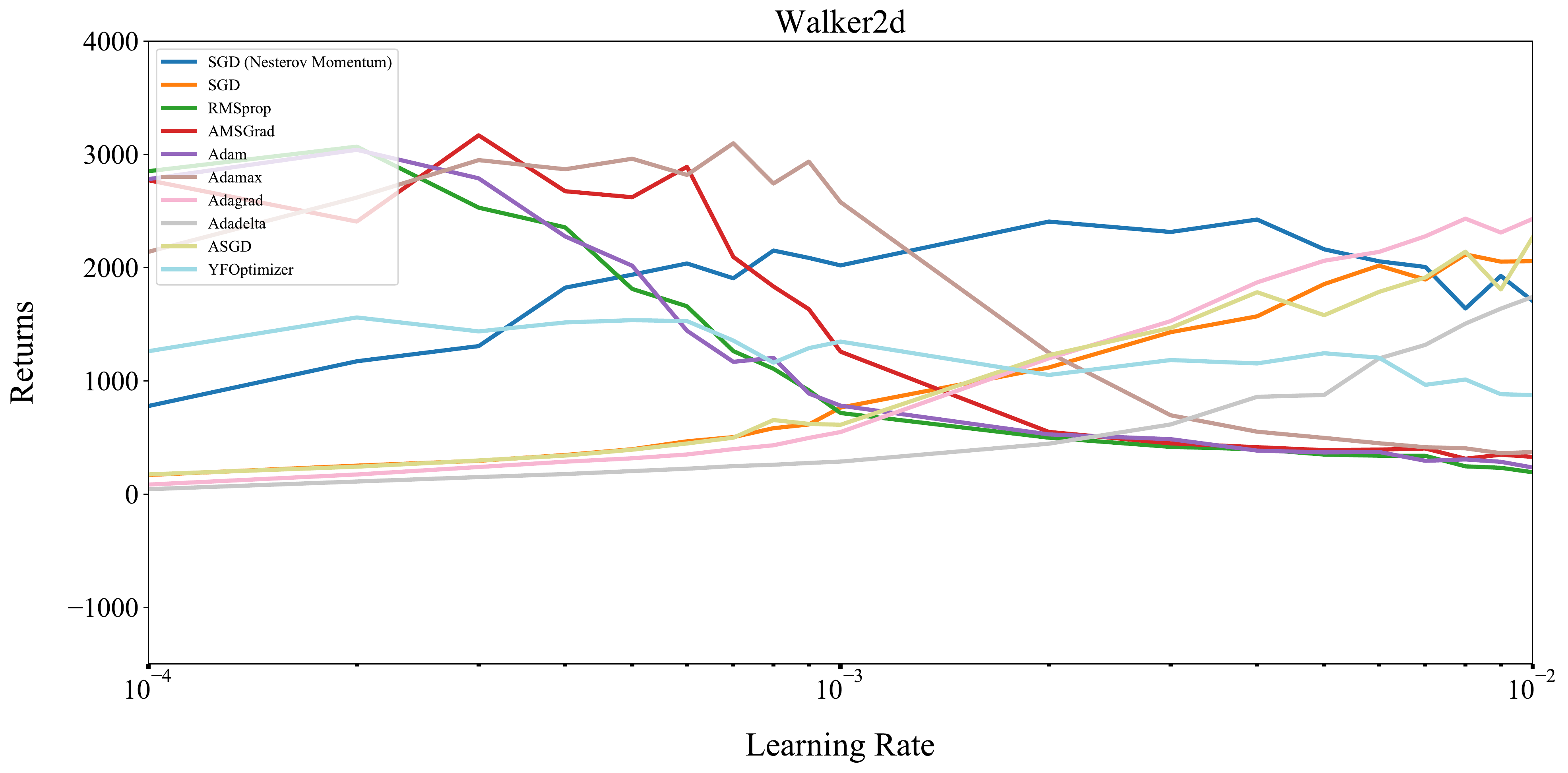}
    \caption{PPO asymptotic performance (averaged over last 50 episodes over 10 random seeds) at different learning rates.}
    \label{fig:alpha_asymptotic_ppo}
\end{figure}

\begin{figure}[H]
    \centering
        \includegraphics[width=.49\textwidth]{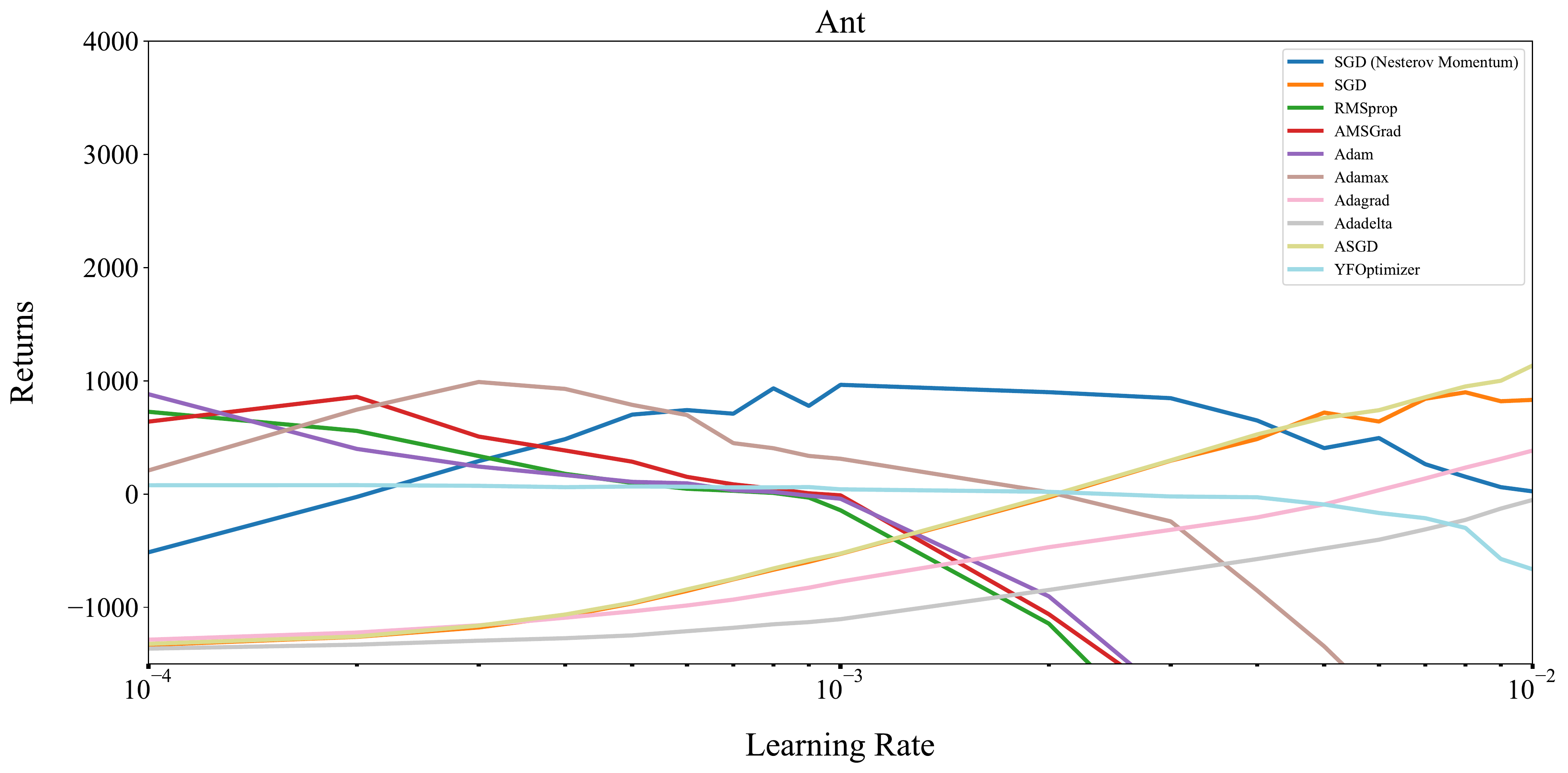}
        \includegraphics[width=.49\textwidth]{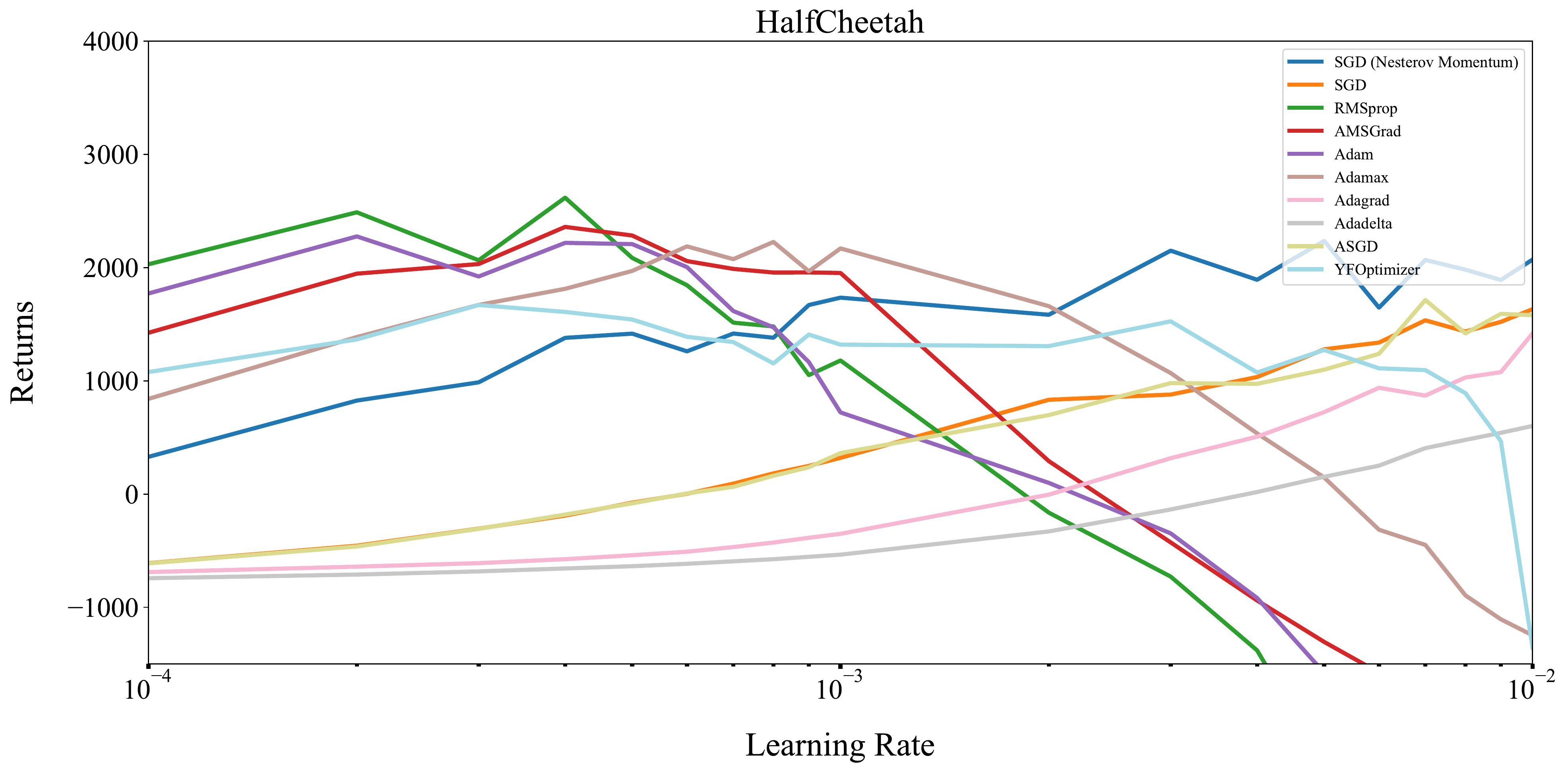}
        \includegraphics[width=.49\textwidth]{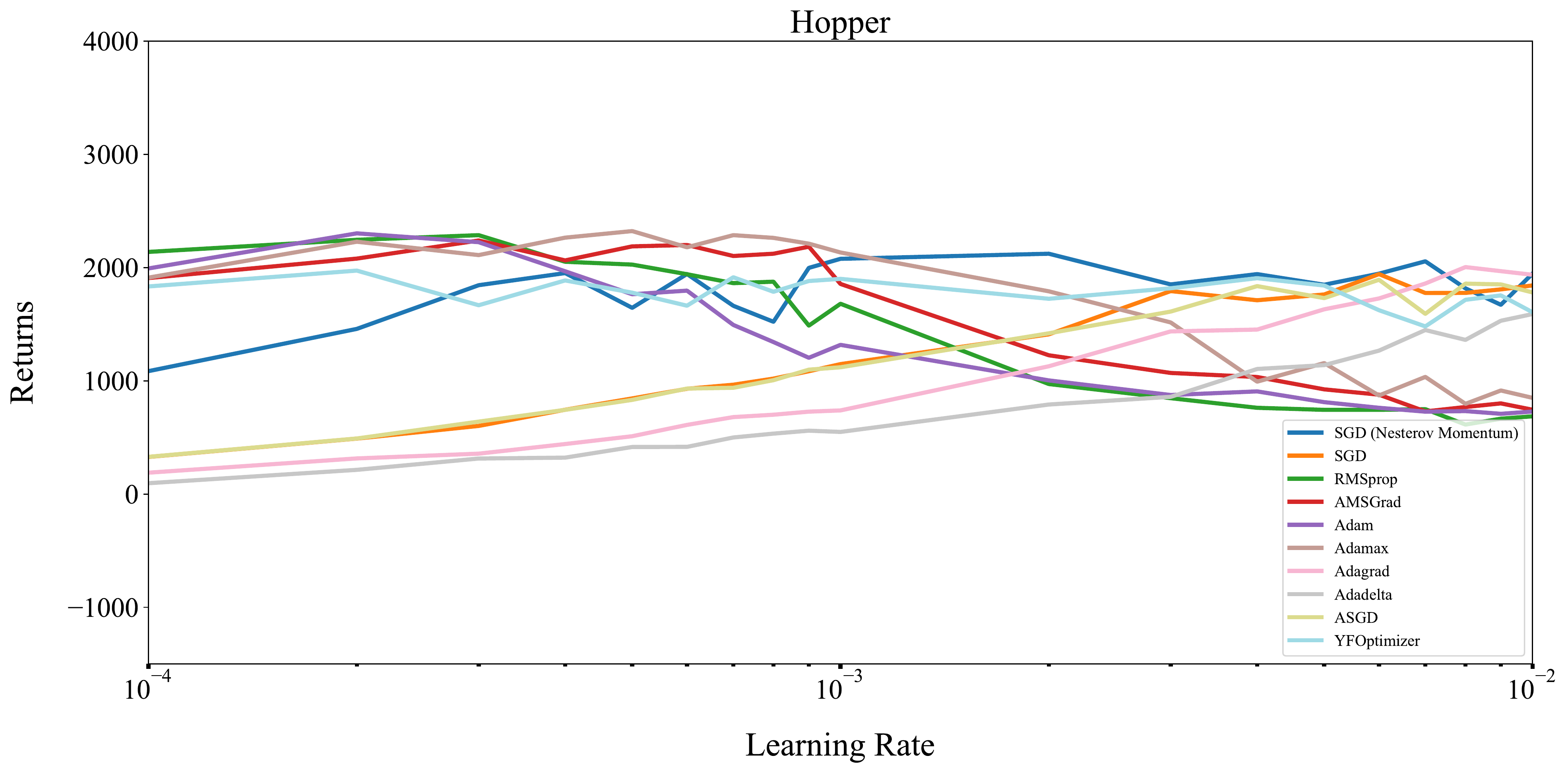}
        \includegraphics[width=.49\textwidth]{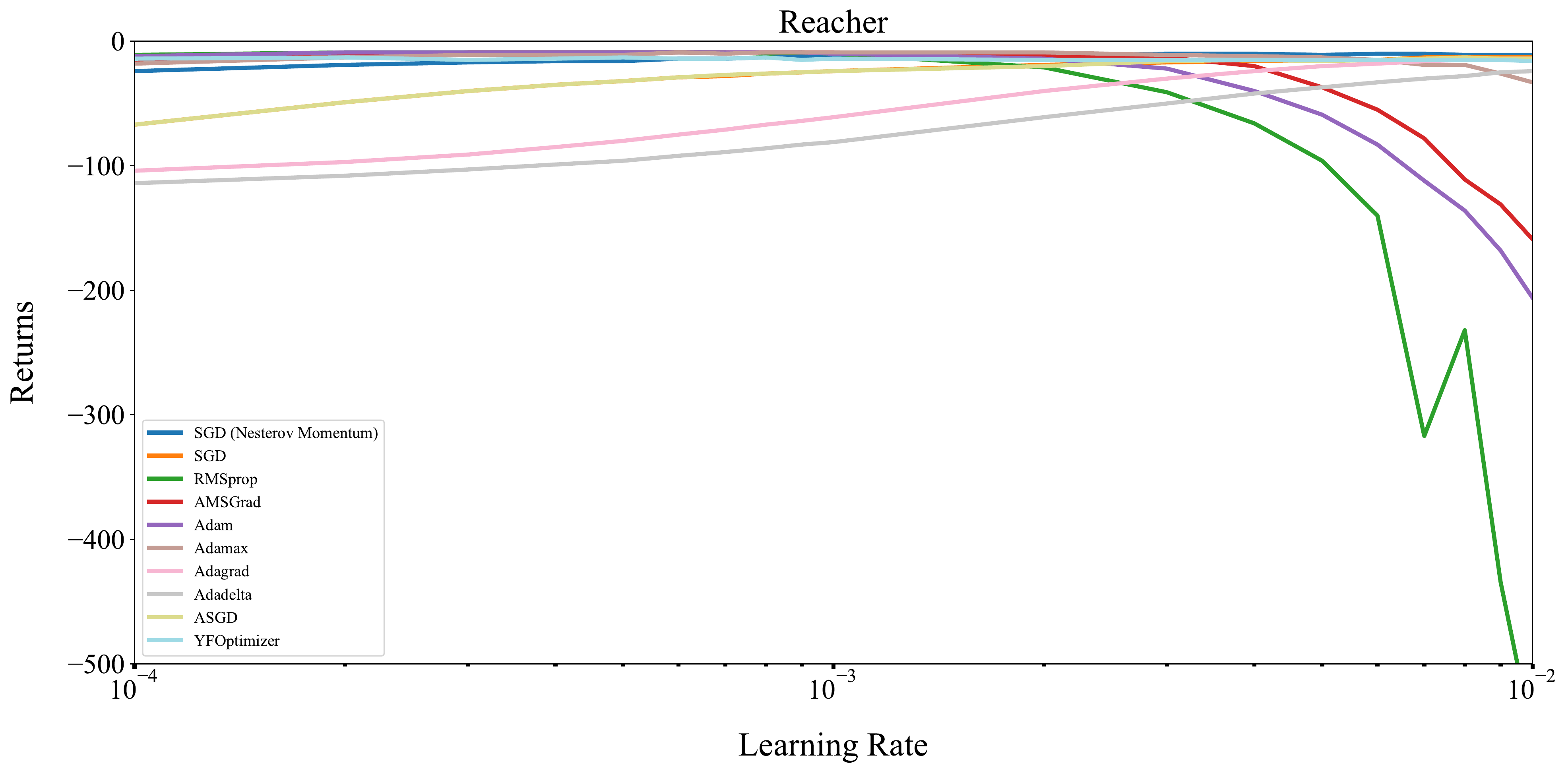}
        \includegraphics[width=.49\textwidth]{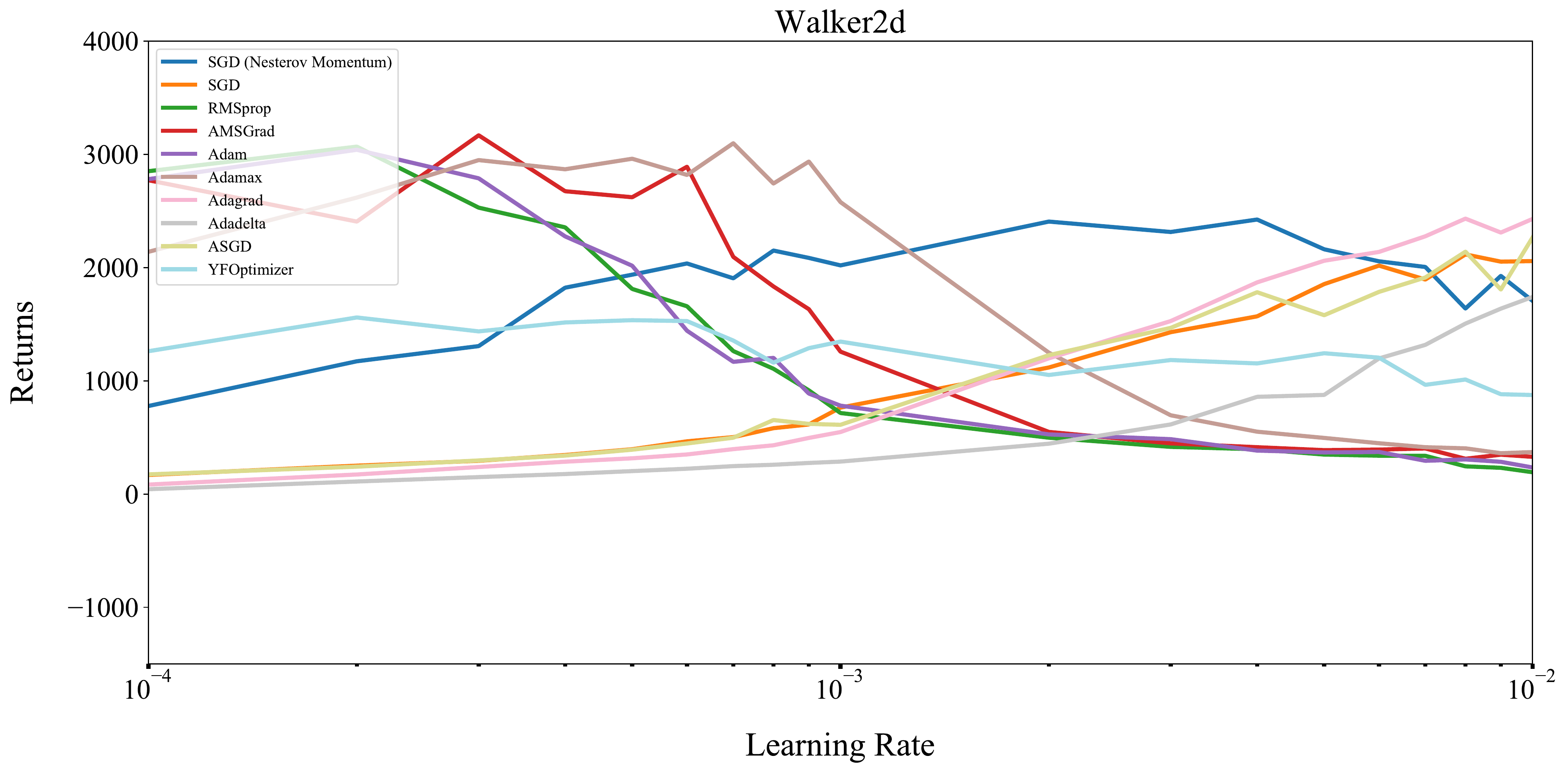}
    \caption{PPO average performance (averaged over last all episodes over 10 random seeds) at different learning rates.}
    \label{fig:alpha_average_ppo}
\end{figure}

\begin{figure}[H]
    \centering
        \includegraphics[width=.49\textwidth]{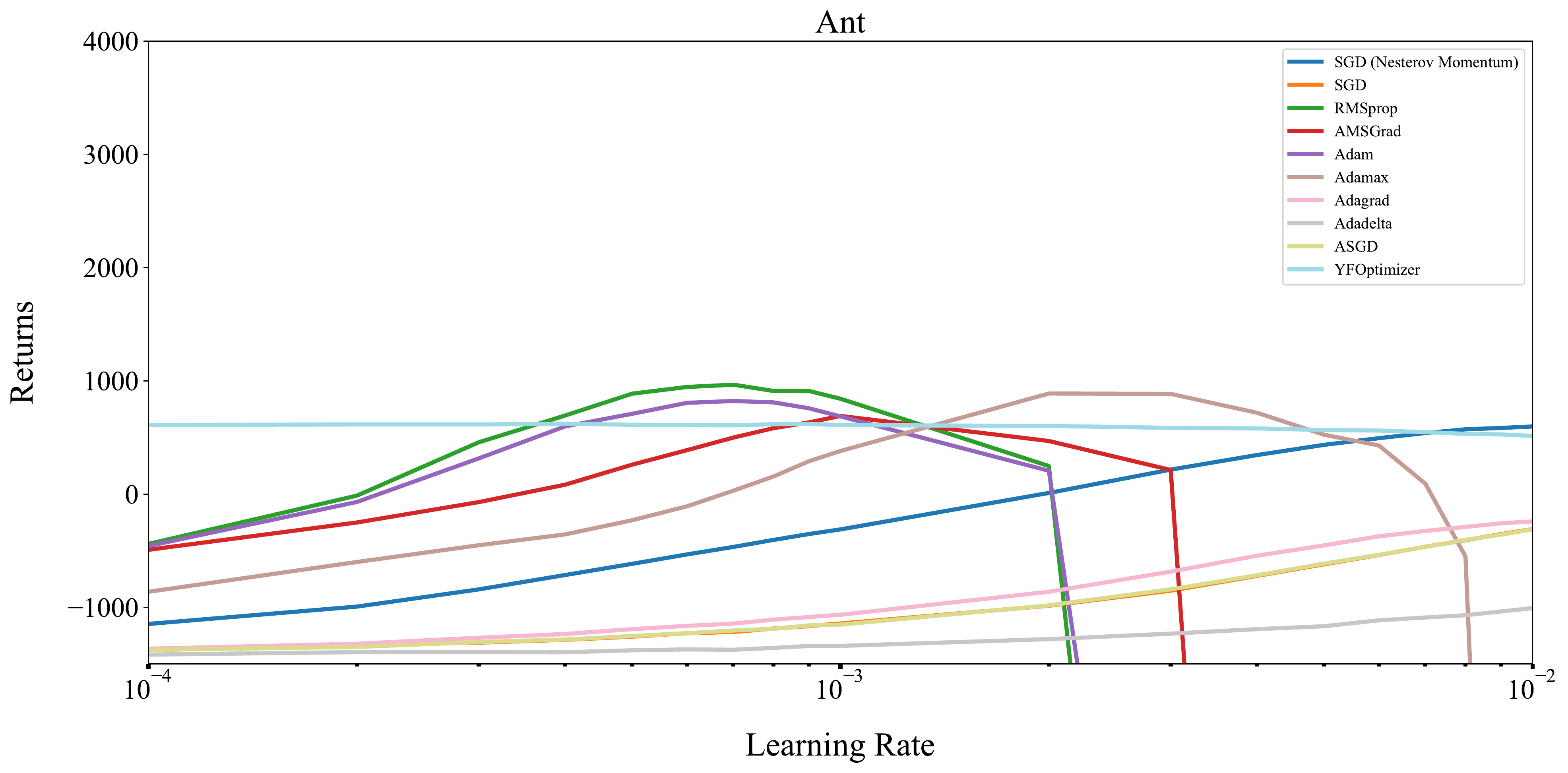}
        \includegraphics[width=.49\textwidth]{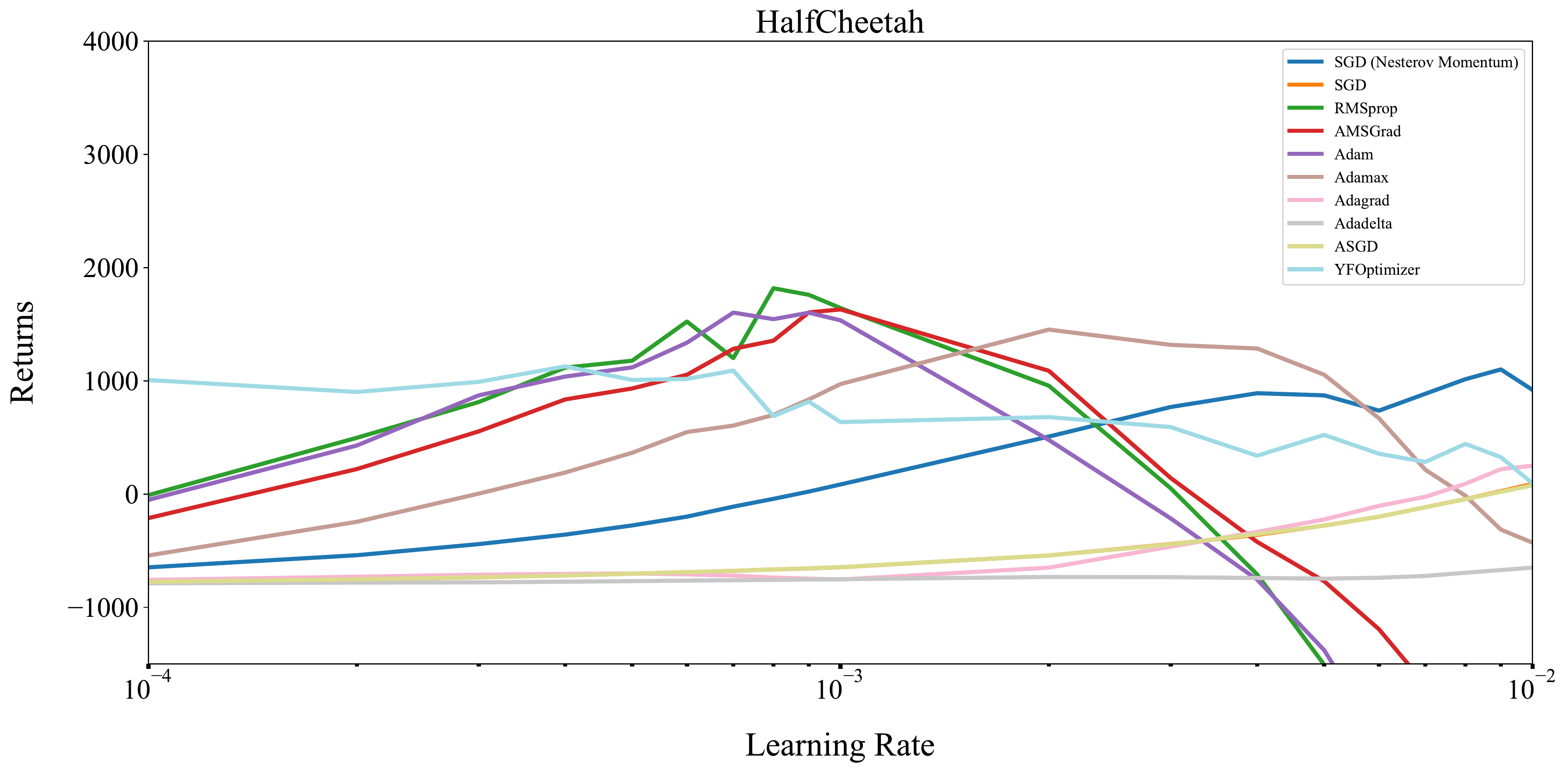}
        \includegraphics[width=.49\textwidth]{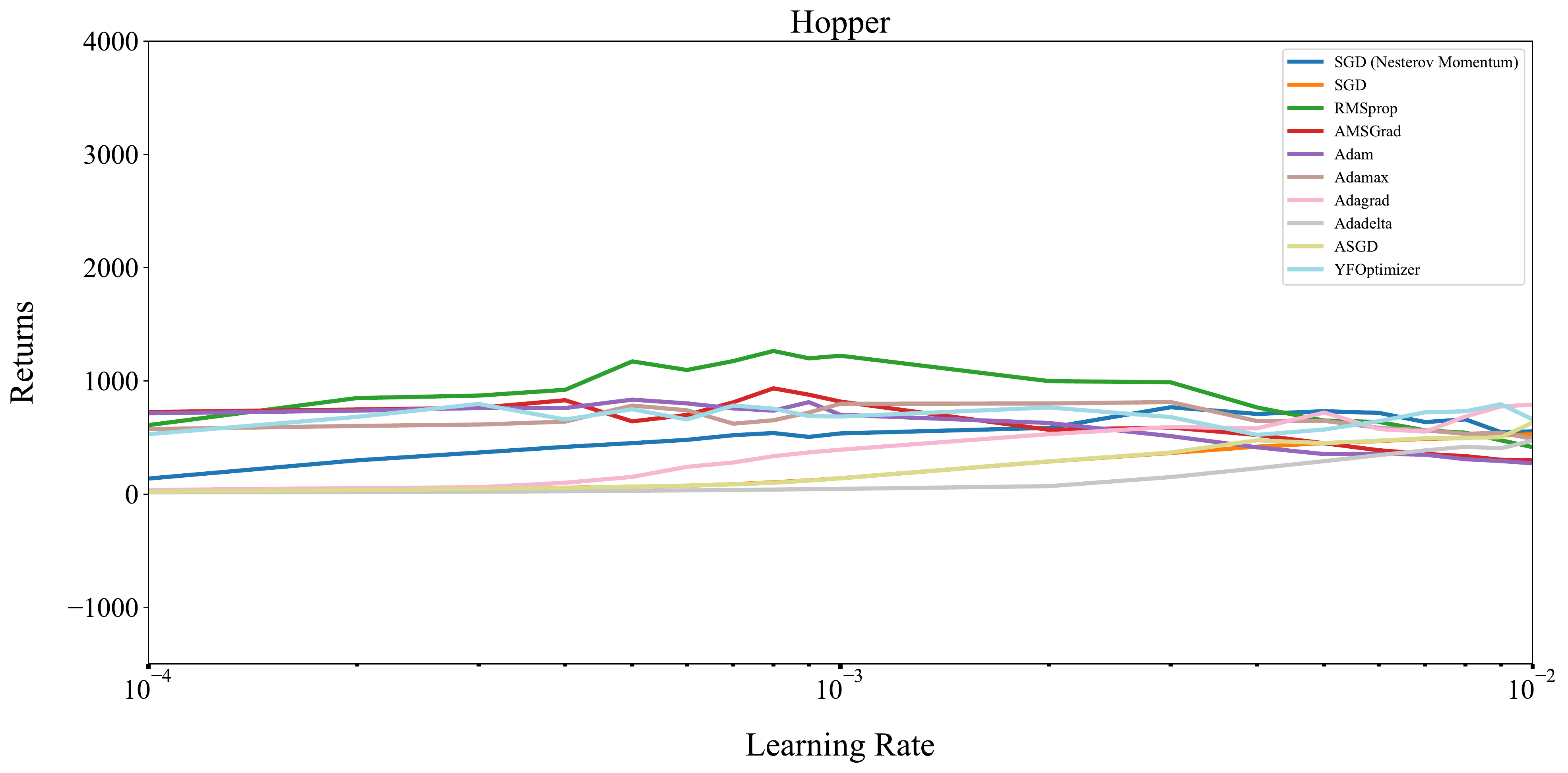}
        \includegraphics[width=.49\textwidth]{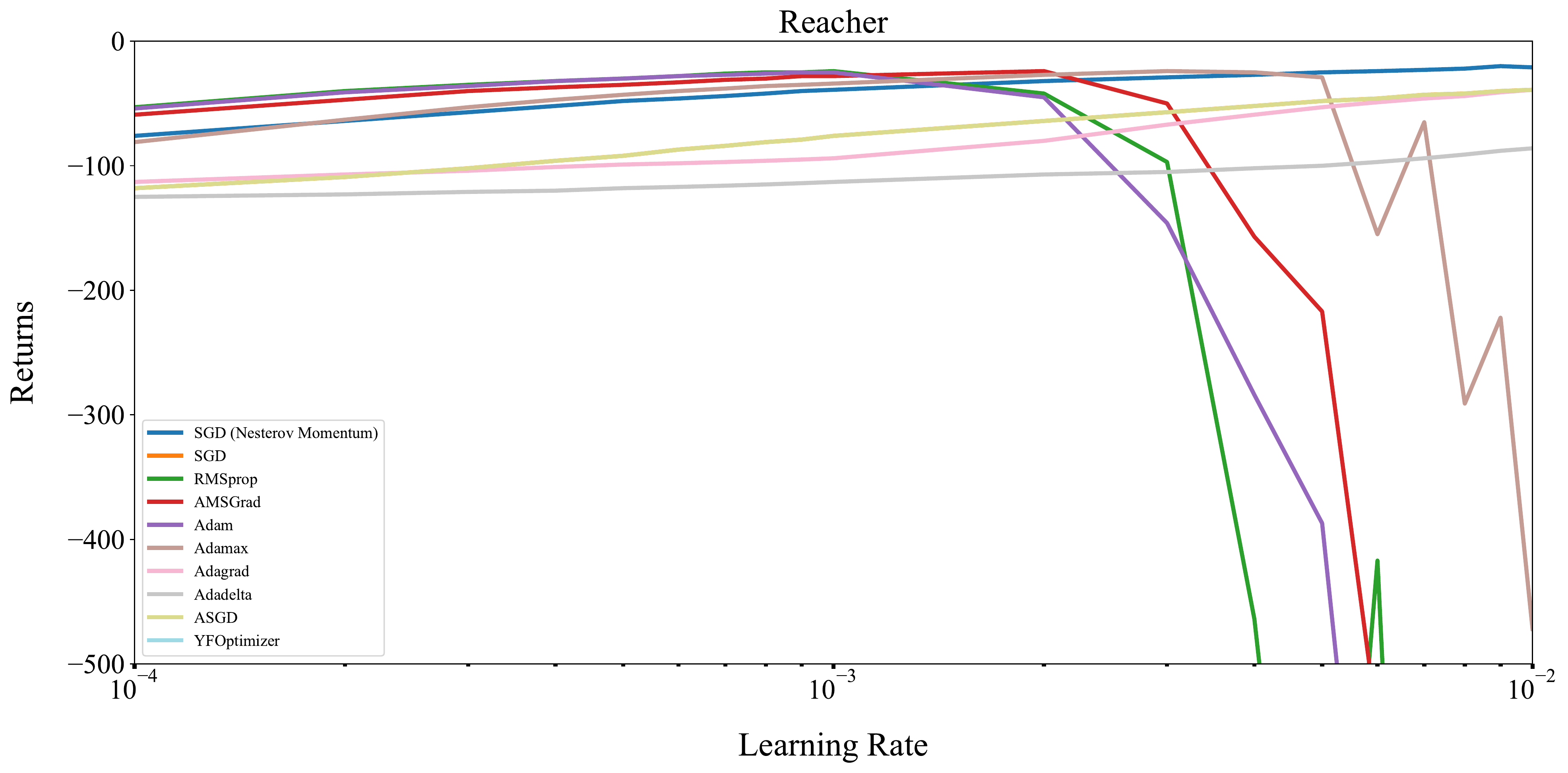}
        \includegraphics[width=.49\textwidth]{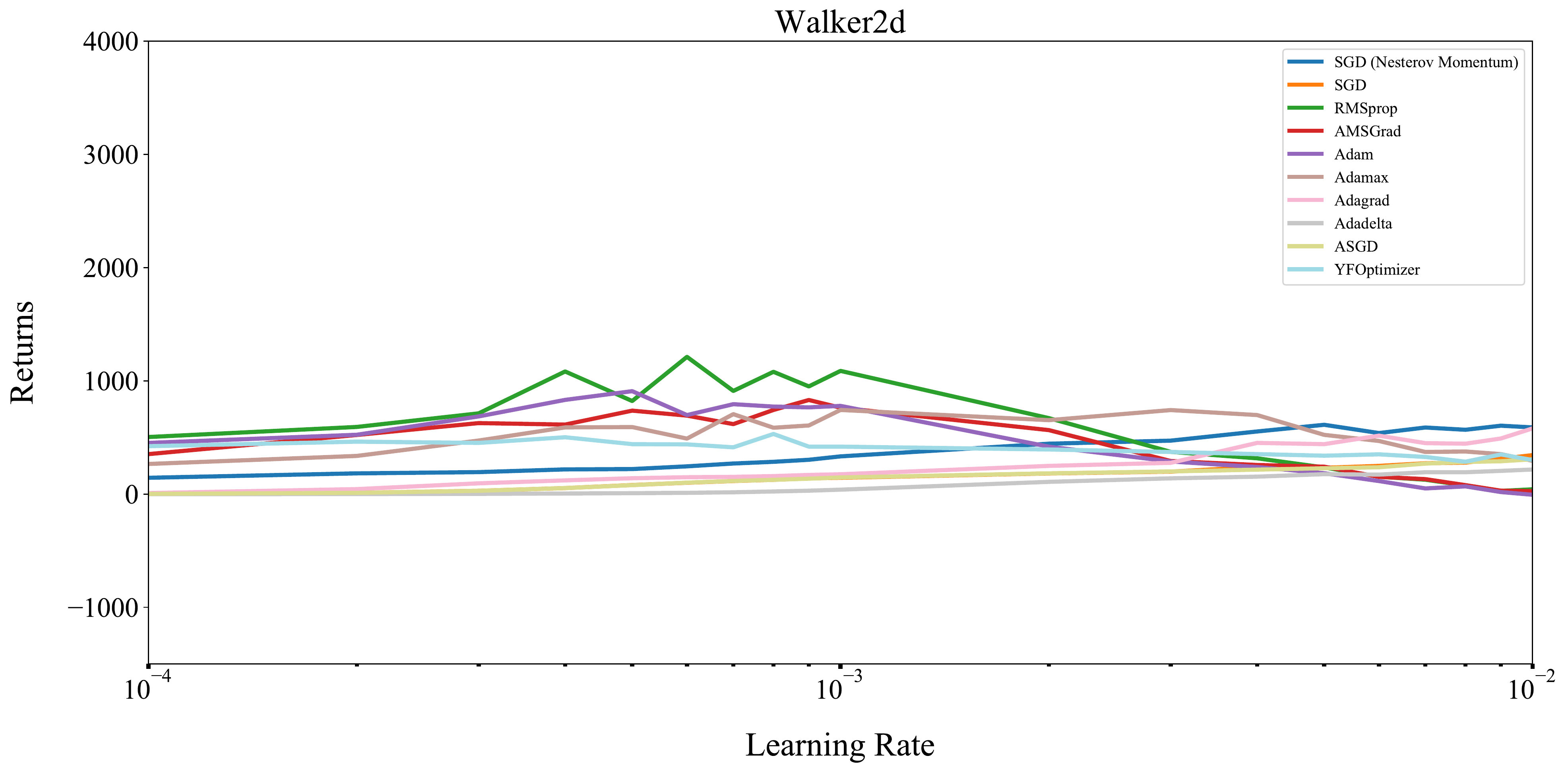}
    \caption{A2C average performance (averaged over last all episodes over 10 random seeds) at different learning rates.}
    \label{fig:alpha_average_a2c}
\end{figure}

Figures~\ref{fig:alpha_asymptotic_a2c} and~\ref{fig:alpha_asymptotic_ppo} show the asymptotic performance across different learning rates (over the last 50 episodes). Figures~\ref{fig:alpha_average_a2c} and~\ref{fig:alpha_average_ppo} show the average performance.


\subsection{Results for Learning Rate Experiments (Average Performance)}

\begin{table}[H]
    \centering
    \small{
    \begin{tabular}{|c|c|c|c|c|c|c|c|c|c|c|}
\hline
lr&SGDNM&SGD&RMS&AMS&A*m&A*max&A*grad&A*delta&ASGD&YF\\
\hline
0.0001&778&169&2852&2772&2782&2140&84&43&174&1262\\
0.0002&1173&251&3069&2407&3041&2619&173&111&242&1560\\
0.0003&1307&294&2530&3169&2789&2950&239&150&297&1437\\
0.0004&1823&345&2356&2675&2276&2869&286&177&340&1516\\
0.0005&1938&396&1812&2622&2017&2962&316&203&392&1536\\
0.0006&2037&466&1659&2891&1443&2819&350&224&447&1528\\
0.0007&1906&503&1263&2095&1168&3099&397&247&498&1356\\
0.0008&2151&582&1107&1834&1203&2743&431&259&654&1161\\
0.0009&2086&615&916&1633&888&2937&496&275&620&1289\\
0.001&2020&764&716&1258&781&2579&548&287&612&1347\\
0.002&2407&1118&498&549&527&1249&1199&445&1227&1052\\
0.003&2315&1430&417&448&485&696&1528&615&1468&1184\\
0.004&2425&1570&396&414&384&551&1870&859&1783&1154\\
0.005&2162&1855&349&388&368&496&2061&876&1580&1244\\
0.006&2056&2018&339&393&372&449&2139&1198&1786&1205\\
0.007&2006&1895&338&403&294&414&2277&1318&1913&965\\
0.008&1639&2117&245&312&306&404&2433&1506&2141&1012\\
0.009&1927&2053&232&348&285&361&2310&1638&1807&882\\
0.01&1707&2058&193&329&235&371&2430&1741&2266&875\\
\hline
    \end{tabular}}
    \caption{The average returns across all episodes over 10 random seeds for PPO on the Walker2d environment. }
    \label{tab:PPO_walker_lr}
\end{table}

\begin{table}[H]
    \centering
    \small{
    \begin{tabular}{|c|c|c|c|c|c|c|c|c|c|c|}
\hline
lr&SGDNM&SGD&RMS&AMS&A*m&A*max&A*grad&A*delta&ASGD&YF\\
\hline
0.0001&-24&-67&-11&-15&-12&-18&-104&-114&-67&-14\\
0.0002&-19&-49&-9&-12&-9&-13&-97&-108&-49&-13\\
0.0003&-17&-40&-9&-9&-9&-11&-91&-103&-40&-15\\
0.0004&-16&-35&-9&-9&-9&-11&-85&-99&-35&-14\\
0.0005&-16&-32&-9&-9&-9&-11&-80&-96&-32&-13\\
0.0006&-14&-29&-9&-9&-9&-9&-75&-92&-29&-14\\
0.0007&-14&-28&-9&-9&-9&-10&-71&-89&-27&-14\\
0.0008&-13&-26&-10&-9&-9&-9&-67&-86&-26&-13\\
0.0009&-12&-25&-9&-9&-9&-9&-64&-83&-25&-15\\
0.001&-12&-24&-10&-9&-10&-9&-61&-81&-24&-14\\
0.002&-12&-19&-21&-12&-14&-9&-40&-61&-20&-15\\
0.003&-10&-17&-41&-13&-22&-11&-30&-50&-16&-15\\
0.004&-10&-16&-66&-20&-40&-12&-24&-42&-15&-15\\
0.005&-11&-15&-96&-37&-59&-13&-20&-37&-16&-15\\
0.006&-10&-15&-140&-55&-83&-15&-18&-33&-15&-15\\
0.007&-10&-13&-317&-78&-112&-19&-16&-30&-14&-15\\
0.008&-11&-13&-232&-111&-136&-19&-15&-28&-13&-15\\
0.009&-11&-13&-434&-131&-168&-26&-14&-25&-14&-15\\
0.01&-11&-13&-559&-159&-206&-33&-14&-24&-14&-16\\
\hline
    \end{tabular}}
    \caption{The average returns across all episodes over 10 random seeds for PPO on the Reacher environment.}
    \label{tab:PPO_reacher_lr}
\end{table}

\begin{table}[H]
    \centering
    \small{
    \begin{tabular}{|c|c|c|c|c|c|c|c|c|c|c|}
\hline
lr&SGDNM&SGD&RMS&AMS&A*m&A*max&A*grad&A*delta&ASGD&YF\\
\hline
0.0001&1086&328&2139&1908&1993&1912&189&96&329&1834\\
0.0002&1459&490&2246&2080&2303&2229&315&214&491&1974\\
0.0003&1845&602&2287&2240&2225&2111&357&314&640&1668\\
0.0004&1952&745&2052&2064&1968&2265&442&322&744&1887\\
0.0005&1645&845&2027&2188&1765&2323&511&416&832&1779\\
0.0006&1944&930&1942&2200&1797&2180&611&417&932&1664\\
0.0007&1663&966&1863&2103&1494&2287&680&500&939&1915\\
0.0008&1522&1020&1876&2123&1343&2263&701&534&1006&1787\\
0.0009&1998&1085&1488&2185&1204&2212&728&560&1099&1882\\
0.001&2078&1148&1681&1856&1318&2134&739&549&1120&1901\\
0.002&2123&1411&972&1226&1004&1791&1129&791&1420&1725\\
0.003&1852&1793&847&1070&875&1516&1437&859&1612&1818\\
0.004&1943&1712&762&1034&907&993&1453&1105&1837&1907\\
0.005&1849&1763&744&925&812&1157&1632&1139&1731&1842\\
0.006&1947&1943&744&878&762&871&1728&1267&1893&1624\\
0.007&2056&1776&750&730&728&1035&1860&1449&1592&1481\\
0.008&1817&1777&614&769&733&797&2005&1362&1859&1716\\
0.009&1672&1808&666&801&709&915&1969&1531&1852&1755\\
0.01&1947&1843&687&745&729&850&1937&1590&1781&1606\\

\hline
    \end{tabular}}
    \caption{The average returns across all episodes over 10 random seeds for PPO on the Hopper environment.}
    \label{tab:PPO_hopper_lr}
\end{table}

\begin{table}[H]
    \centering
    \small{
    \begin{tabular}{|c|c|c|c|c|c|c|c|c|c|c|}
\hline
lr&SGDNM&SGD&RMS&AMS&A*m&A*max&A*grad&A*delta&ASGD&YF\\
\hline
0.0001&329&-610&2030&1425&1772&841&-689&-743&-612&1078\\
0.0002&826&-456&2489&1947&2276&1387&-641&-711&-463&1366\\
0.0003&987&-305&2065&2032&1922&1671&-610&-683&-307&1670\\
0.0004&1380&-190&2618&2360&2219&1813&-575&-657&-181&1609\\
0.0005&1417&-76&2086&2284&2208&1971&-540&-637&-82&1542\\
0.0006&1261&3&1845&2058&2005&2188&-509&-616&6&1389\\
0.0007&1418&92&1514&1989&1617&2075&-468&-594&64&1343\\
0.0008&1381&183&1481&1957&1473&2227&-428&-575&162&1154\\
0.0009&1670&249&1051&1958&1166&1968&-386&-554&237&1409\\
0.001&1735&320&1180&1953&721&2170&-351&-535&362&1320\\
0.002&1584&833&-164&291&99&1660&-5&-330&698&1307\\
0.003&2150&879&-729&-427&-347&1069&317&-136&980&1526\\
0.004&1893&1034&-1381&-940&-918&536&507&18&974&1074\\
0.005&2236&1279&-2384&-1306&-1579&146&724&153&1099&1273\\
0.006&1647&1338&-3762&-1577&-2432&-315&939&251&1239&1111\\
0.007&2068&1535&-4979&-2035&-3656&-449&870&406&1714&1095\\
0.008&1982&1436&-7160&-2574&-5136&-895&1030&479&1419&891\\
0.009&1891&1521&-9416&-3099&-6948&-1106&1077&541&1592&468\\
0.01&2071&1633&-12446&-3625&-8910&-1247&1419&602&1580&-1357\\
\hline
    \end{tabular}}
    \caption{The average returns across all episodes over 10 random seeds for PPO on the HalfCheetah environment. }
    \label{tab:PPO_hc_lr}
\end{table}

\begin{table}[H]
    \centering
    \small{
    \begin{tabular}{|c|c|c|c|c|c|c|c|c|c|c|}
\hline
lr&SGDNM&SGD&RMS&AMS&A*m&A*max&A*grad&A*delta&ASGD&YF\\
\hline
0.0001&-514&-1333&727&640&882&209&-1286&-1365&-1323&78\\
0.0002&-25&-1261&558&859&399&746&-1223&-1330&-1258&79\\
0.0003&291&-1178&336&508&243&990&-1159&-1295&-1162&73\\
0.0004&485&-1077&179&385&168&929&-1092&-1273&-1064&61\\
0.0005&702&-966&99&286&108&788&-1036&-1248&-959&67\\
0.0006&742&-855&47&152&95&698&-984&-1211&-842&65\\
0.0007&710&-753&29&86&34&450&-932&-1181&-749&59\\
0.0008&934&-668&10&47&19&405&-876&-1150&-657&59\\
0.0009&779&-598&-31&7&-14&337&-827&-1131&-583&62\\
0.001&965&-528&-143&-11&-41&312&-773&-1105&-525&42\\
0.002&900&-29&-1144&-1061&-903&17&-469&-847&-16&20\\
0.003&847&295&-2214&-1821&-1799&-241&-316&-687&298&-21\\
0.004&650&486&-3365&-2415&-3320&-852&-206&-574&527&-28\\
0.005&406&720&-5452&-3834&-4389&-1345&-91&-480&673&-92\\
0.006&495&641&-8507&-5143&-6757&-1816&34&-403&741&-167\\
0.007&264&842&-10322&-5452&-9529&-1933&138&-312&856&-213\\
0.008&153&899&-15851&-7874&-12138&-2589&234&-228&951&-299\\
0.009&61&820&-20024&-8682&-14339&-2728&311&-128&1001&-573\\
0.01&24&832&-172953&-11573&-18910&-3134&384&-52&1135&-664\\
\hline
    \end{tabular}}
    \caption{The average returns across all episodes over 10 random seeds for PPO on the Ant environment. }
    \label{tab:PPO_ant_lr}
\end{table}

\begin{table}[H]
    \centering
    \small{
    \begin{tabular}{|c|c|c|c|c|c|c|c|c|c|c|}
\hline
lr&SGDNM&SGD&RMS&AMS&A*m&A*max&A*grad&A*delta&ASGD&YF\\
\hline
0.0001&144&2&504&353&452&266&9&-1&2&423\\
0.0002&183&11&593&522&524&337&44&0&11&463\\
0.0003&194&29&713&627&686&473&95&2&29&452\\
0.0004&218&53&1083&614&832&589&121&5&53&502\\
0.0005&221&80&822&737&909&592&140&8&79&441\\
0.0006&245&100&1212&693&698&490&149&11&101&439\\
0.0007&270&115&911&618&794&706&152&16&116&413\\
0.0008&285&126&1080&743&773&586&159&23&126&531\\
0.0009&303&139&951&831&766&606&169&30&136&419\\
0.001&333&144&1088&763&779&743&175&39&147&419\\
0.002&445&182&671&565&419&653&249&108&183&394\\
0.003&472&198&373&292&288&742&276&139&200&372\\
0.004&553&239&316&257&239&698&452&154&216&353\\
0.005&612&232&235&242&184&523&441&176&228&339\\
0.006&539&249&153&152&115&469&516&172&237&351\\
0.007&588&272&127&132&50&372&450&193&269&328\\
0.008&568&278&76&79&68&376&445&193&284&286\\
0.009&604&312&28&30&18&353&491&205&290&350\\
0.01&589&344&42&26&-6&294&580&217&307&301\\
\hline
    \end{tabular}}
    \caption{The average returns across all episodes over 10 random seeds for A2C on the Walker2d environment. }
    \label{tab:a2c_walker_lr}
\end{table}

\begin{table}[H]
    \centering
    \small{
    \begin{tabular}{|c|c|c|c|c|c|c|c|c|c|c|}
\hline
lr&SGDNM&SGD&RMS&AMS&A*m&A*max&A*grad&A*delta&ASGD&YF\\
\hline
0.0001&137&22&609&725&713&574&36&16&22&529\\
0.0002&298&35&848&747&737&602&53&18&35&682\\
0.0003&367&46&870&762&760&614&60&21&46&796\\
0.0004&417&56&921&829&760&640&100&25&55&659\\
0.0005&450&65&1172&642&834&782&152&29&65&749\\
0.0006&479&74&1096&699&801&740&241&33&75&659\\
0.0007&520&87&1175&811&757&622&279&37&86&782\\
0.0008&539&104&1264&934&737&652&335&40&100&755\\
0.0009&505&121&1199&878&812&720&368&43&119&689\\
0.001&536&140&1222&817&700&798&392&46&140&685\\
0.002&585&288&998&568&628&800&529&70&289&765\\
0.003&767&363&987&588&512&813&593&150&367&680\\
0.004&708&419&765&522&413&645&581&228&475&523\\
0.005&732&450&649&449&353&647&720&291&450&570\\
0.006&717&468&637&387&356&585&574&344&472&644\\
0.007&636&488&563&355&348&564&552&386&493&723\\
0.008&660&495&542&335&308&532&685&418&497&731\\
0.009&547&518&475&303&292&542&774&404&502&794\\
0.01&554&528&415&300&272&488&790&472&631&660\\

\hline
    \end{tabular}}
    \caption{The average returns across all episodes over 10 random seeds for A2C on the Hopper environment. }
    \label{tab:a2c_hopper_lr}
\end{table}

\begin{table}[H]
    \centering
    \small{
    \begin{tabular}{|c|c|c|c|c|c|c|c|c|c|c|}
\hline
lr&SGDNM&SGD&RMS&AMS&A*m&A*max&A*grad&A*delta&ASGD&YF\\
\hline
0.0001&-647&-773&-9&-212&-51&-542&-758&-788&-775&1007\\
0.0002&-540&-754&497&221&429&-245&-730&-782&-753&902\\
0.0003&-442&-735&813&554&872&4&-712&-780&-734&991\\
0.0004&-358&-718&1117&836&1038&190&-705&-774&-719&1128\\
0.0005&-277&-703&1178&933&1119&365&-701&-769&-703&1008\\
0.0006&-199&-690&1524&1054&1337&548&-709&-764&-689&1017\\
0.0007&-111&-678&1201&1283&1603&605&-720&-761&-678&1091\\
0.0008&-42&-666&1818&1355&1545&699&-737&-758&-667&689\\
0.0009&22&-657&1760&1605&1602&836&-747&-754&-656&817\\
0.001&85&-646&1643&1631&1535&972&-755&-752&-645&636\\
0.002&508&-542&957&1089&478&1453&-650&-732&-542&680\\
0.003&769&-442&51&142&-215&1318&-464&-734&-445&592\\
0.004&891&-360&-710&-422&-755&1286&-334&-741&-353&339\\
0.005&872&-278&-1508&-770&-1379&1054&-224&-747&-279&523\\
0.006&737&-200&-2754&-1194&-2158&669&-104&-739&-200&356\\
0.007&886&-116&-4147&-1683&-3702&215&-24&-723&-116&284\\
0.008&1015&-44&-7891&-2284&-4062&-15&91&-695&-45&443\\
0.009&1101&25&-8413&-2856&-8724&-314&220&-671&20&326\\
0.01&920&90&-21693&-3233&-14145&-429&251&-649&77&95\\
\hline
    \end{tabular}}
    \caption{The average returns across all episodes over 10 random seeds for A2C on the HalfCheetah environment. }
    \label{tab:a2c_hc_lr}
\end{table}

\begin{table}[H]
    \centering
    \small{
    \begin{tabular}{|c|c|c|c|c|c|c|c|c|c|c|}
\hline
lr&SGDNM&SGD&RMS&AMS&A*m&A*max&A*grad&A*delta&ASGD&YF\\
\hline
0.0001&-76&-118&-53&-59&-54&-81&-113&-125&-118&-1193\\
0.0002&-64&-109&-40&-47&-41&-63&-107&-123&-109&-941\\
0.0003&-57&-102&-35&-40&-36&-53&-104&-121&-102&-1126\\
0.0004&-52&-96&-32&-37&-32&-47&-101&-120&-96&-2160\\
0.0005&-48&-92&-30&-35&-30&-43&-99&-118&-92&-2025\\
0.0006&-46&-87&-28&-33&-28&-40&-98&-117&-87&-2470\\
0.0007&-44&-84&-26&-31&-27&-38&-97&-116&-84&-1950\\
0.0008&-42&-81&-25&-30&-26&-36&-96&-115&-81&-2128\\
0.0009&-40&-79&-25&-28&-25&-35&-95&-114&-79&-3114\\
0.001&-39&-76&-24&-28&-25&-34&-94&-113&-76&-3124\\
0.002&-32&-64&-42&-24&-45&-27&-80&-107&-64&-1423\\
0.003&-29&-57&-97&-50&-146&-24&-67&-105&-57&-1825\\
0.004&-27&-52&-464&-157&-284&-25&-59&-102&-52&-4326\\
0.005&-25&-48&-950&-217&-387&-29&-53&-100&-48&-4095\\
0.006&-24&-46&-417&-549&-811&-155&-49&-97&-46&-3852\\
0.007&-23&-43&-1193&-513&-1828&-65&-46&-94&-43&-1745\\
0.008&-22&-42&-2953&-891&-2060&-291&-44&-91&-42&-2677\\
0.009&-20&-40&-3200&-2399&-8343&-222&-41&-88&-40&-5942\\
0.01&-21&-39&-2206&-3027&-7547&-472&-39&-86&-39&-6191\\
\hline
    \end{tabular}}
    \caption{The average returns across all episodes over 10 random seeds for A2C on the Reacher environment. }
    \label{tab:a2c_reacher_lr}
\end{table}
\begin{table}[H]
    \centering
    \small{
    \begin{tabular}{|c|c|c|c|c|c|c|c|c|c|c|}
\hline
lr&SGDNM&SGD&RMS&AMS&A*m&A*max&A*grad&A*delta&ASGD&YF\\
\hline
0.0001&-1147&-1378&-440&-492&-458&-863&-1366&-1418&-1376&610\\
0.0002&-994&-1332&-14&-251&-70&-600&-1322&-1396&-1350&615\\
0.0003&-842&-1312&457&-71&315&-452&-1269&-1395&-1307&615\\
0.0004&-715&-1287&694&83&598&-356&-1236&-1397&-1284&622\\
0.0005&-616&-1259&887&260&709&-231&-1193&-1380&-1254&612\\
0.0006&-533&-1229&946&388&806&-108&-1164&-1372&-1229&609\\
0.0007&-467&-1218&966&499&822&30&-1144&-1375&-1205&607\\
0.0008&-405&-1188&911&582&810&155&-1109&-1359&-1190&617\\
0.0009&-353&-1166&911&633&758&290&-1086&-1342&-1160&620\\
0.001&-312&-1143&842&690&686&381&-1066&-1341&-1153&609\\
0.002&10&-986&248&469&206&888&-863&-1281&-983&602\\
0.003&216&-853&-9613&213&-7060&885&-685&-1233&-841&585\\
0.004&345&-722&-19768&-10579&-17378&718&-544&-1193&-717&580\\
0.005&435&-620&-29898&-11825&-38297&524&-453&-1167&-614&566\\
0.006&494&-538&-34973&-9769&-50970&429&-373&-1115&-536&561\\
0.007&539&-465&-55208&-16308&-72371&92&-326&-1090&-466&547\\
0.008&572&-407&-198085&-22234&-113285&-550&-290&-1070&-404&532\\
0.009&584&-354&-120519&-36751&-91976&-7818&-258&-1037&-359&527\\
0.01&597&-310&-205884&-52366&-123956&-9693&-242&-1007&-312&514\\
\hline
    \end{tabular}}
    \caption{The average returns across all episodes over 10 random seeds for A2C on the Ant environment. }
    \label{tab:a2c_ant_lr}
\end{table}

\subsection{Results for Learning Rate Experiments (Asymptotic Performance)}
\begin{table}[H]
    \centering
    \small{
    \begin{tabular}{|c|c|c|c|c|c|c|c|c|c|c|}
\hline
lr&SGDNM&SGD&RMS&AMS&A*m&A*max&A*grad&A*delta&ASGD&YF\\
\hline
0.0001&206&8&755&518&703&394&16&0&9&672\\
0.0002&209&30&865&831&910&470&95&3&37&552\\
0.0003&254&86&1015&760&1049&622&157&11&83&775\\
0.0004&270&158&1735&1173&1508&821&162&16&150&613\\
0.0005&319&219&1618&1034&1523&849&178&21&201&581\\
0.0006&342&197&1548&1113&1360&804&180&39&206&615\\
0.0007&411&191&1425&1218&1249&995&167&55&218&601\\
0.0008&448&199&1695&1223&1057&861&174&75&206&906\\
0.0009&420&210&1465&1277&1042&809&190&96&204&572\\
0.001&595&226&1408&1465&1543&1134&225&121&233&629\\
0.002&608&206&568&655&607&910&320&197&207&552\\
0.003&633&253&508&252&314&1094&348&204&232&582\\
0.004&692&309&324&255&298&1151&561&206&260&416\\
0.005&909&338&194&257&65&608&521&253&275&469\\
0.006&540&309&157&203&75&740&601&217&351&482\\
0.007&787&365&115&205&26&433&521&238&351&474\\
0.008&597&429&-10&135&36&422&502&251&475&374\\
0.009&825&549&1&55&-36&308&666&252&423&439\\
0.01&917&614&15&32&-50&267&798&301&491&240\\
\hline
    \end{tabular}}
    \caption{The asymptotic returns averaged across last 50 episodes over 10 random seeds for A2C on the Walker2d environment. }
    \label{tab:a2c_asymptotic_lr_walker}
\end{table}

\begin{table}[H]
    \centering
    \small{
    \begin{tabular}{|c|c|c|c|c|c|c|c|c|c|c|}
\hline
lr&SGDNM&SGD&RMS&AMS&A*m&A*max&A*grad&A*delta&ASGD&YF\\
\hline
0.0001&-57&-108&-28&-40&-29&-56&-109&-122&-108&-1962\\
0.0002&-45&-94&-24&-30&-24&-37&-101&-118&-94&-817\\
0.0003&-40&-83&-23&-27&-23&-28&-99&-116&-83&-3610\\
0.0004&-36&-75&-21&-25&-20&-26&-96&-113&-75&-2800\\
0.0005&-34&-69&-20&-25&-19&-25&-95&-112&-69&-2566\\
0.0006&-32&-65&-18&-24&-19&-23&-94&-110&-65&-3479\\
0.0007&-30&-62&-18&-23&-17&-24&-95&-108&-62&-2084\\
0.0008&-29&-60&-18&-21&-19&-25&-94&-106&-60&-1244\\
0.0009&-28&-58&-19&-21&-17&-22&-91&-105&-58&-1461\\
0.001&-27&-56&-17&-21&-16&-22&-87&-104&-57&-5347\\
0.002&-24&-46&-86&-28&-64&-19&-65&-101&-46&-2808\\
0.003&-22&-40&-105&-101&-399&-18&-52&-99&-40&-3885\\
0.004&-21&-36&-554&-219&-321&-38&-46&-92&-36&-9103\\
0.005&-20&-34&-1299&-165&-436&-75&-41&-85&-34&-11072\\
0.006&-21&-32&-279&-2616&-989&-58&-37&-81&-32&-3465\\
0.007&-18&-30&-2645&-1062&-1773&-221&-35&-74&-30&-1385\\
0.008&-14&-28&-7486&-548&-2375&-425&-32&-69&-29&-6912\\
0.009&-14&-28&-3615&-2389&-7495&-408&-30&-64&-28&-9465\\
0.01&-17&-28&-2681&-7829&-9271&-676&-29&-62&-28&-14031\\
\hline
    \end{tabular}}
    \caption{The asymptotic returns averaged across last 50 episodes over 10 random seeds for A2C on the Reacher environment. }
    \label{tab:a2c_asymptotic_lr_reacher}
\end{table}

\begin{table}[H]
    \centering
    \small{
    \begin{tabular}{|c|c|c|c|c|c|c|c|c|c|c|}
\hline
lr&SGDNM&SGD&RMS&AMS&A*m&A*max&A*grad&A*delta&ASGD&YF\\
\hline
0.0001&316&33&700&844&784&761&48&17&32&790\\
0.0002&497&60&829&871&908&766&63&23&62&986\\
0.0003&533&77&1116&794&724&769&91&32&77&1239\\
0.0004&590&89&1414&875&926&759&188&43&90&742\\
0.0005&595&105&1486&668&1200&983&247&48&105&1211\\
0.0006&557&138&1480&658&1071&822&537&55&139&1083\\
0.0007&631&173&1394&898&1037&760&434&64&166&785\\
0.0008&665&229&1489&1201&1256&848&444&70&215&1013\\
0.0009&638&282&1608&1082&1073&881&487&73&268&1073\\
0.001&677&339&1618&1009&971&983&475&73&323&968\\
0.002&841&484&1422&899&674&1136&588&147&478&1325\\
0.003&967&539&1216&517&580&1244&624&406&541&1264\\
0.004&841&602&844&631&549&1033&589&479&649&812\\
0.005&1228&574&786&522&588&667&779&554&612&1027\\
0.006&1033&559&809&586&416&644&589&601&561&981\\
0.007&870&587&745&464&414&772&560&614&658&712\\
0.008&896&671&642&366&502&774&746&739&617&958\\
0.009&687&704&681&369&327&579&955&589&637&1052\\
0.01&810&730&496&338&303&724&753&661&861&784\\
\hline
    \end{tabular}}
    \caption{The asymptotic returns averaged across last 50 episodes over 10 random seeds for A2C on the Hopper environment. }
    \label{tab:a2c_asymptotic_lr_hopper}
\end{table}

\begin{table}[H]
    \centering
    \small{
    \begin{tabular}{|c|c|c|c|c|c|c|c|c|c|c|}
\hline
lr&SGDNM&SGD&RMS&AMS&A*m&A*max&A*grad&A*delta&ASGD&YF\\
\hline
0.0001&-550&-753&733&288&735&-211&-739&-773&-747&1149\\
0.0002&-328&-724&1280&902&1244&303&-707&-776&-714&852\\
0.0003&-163&-682&1685&1437&1673&743&-689&-751&-664&1003\\
0.0004&-16&-649&1825&1715&1875&957&-704&-744&-653&1337\\
0.0005&141&-611&1934&1757&1665&1190&-710&-757&-634&580\\
0.0006&292&-614&2425&1919&1976&1470&-723&-752&-611&1227\\
0.0007&413&-587&1692&2267&2710&1367&-762&-736&-587&946\\
0.0008&557&-572&2846&2391&2404&1449&-791&-720&-584&868\\
0.0009&672&-554&2524&2558&2187&1662&-787&-717&-553&-40\\
0.001&745&-542&2039&1988&2114&1732&-783&-714&-548&161\\
0.002&1190&-325&461&1204&130&2098&-452&-714&-330&920\\
0.003&1456&-174&-615&349&-582&1798&-217&-742&-177&655\\
0.004&1319&-20&-1685&-718&-1184&1666&-48&-774&-20&223\\
0.005&1442&133&-2544&-976&-3267&736&126&-736&144&347\\
0.006&1038&276&-6034&-1828&-4411&184&272&-668&278&199\\
0.007&1281&417&-9289&-2948&-10275&-298&420&-585&396&365\\
0.008&1011&556&-16336&-3173&-7959&-334&574&-481&565&308\\
0.009&745&684&-19835&-4108&-21408&-961&727&-442&651&323\\
0.01&1775&792&-53399&-5579&-39233&-1402&785&-384&744&-278\\
\hline
    \end{tabular}}
    \caption{The asymptotic returns averaged across last 50 episodes over 10 random seeds for A2C on the HalfCheetah environment. }
    \label{tab:a2c_asymptotic_lr_hc}
\end{table}

\begin{table}[H]
    \centering
    \small{
    \begin{tabular}{|c|c|c|c|c|c|c|c|c|c|c|}
\hline
lr&SGDNM&SGD&RMS&AMS&A*m&A*max&A*grad&A*delta&ASGD&YF\\
\hline
0.0001&-845&-1392&-52&-99&-64&-399&-1329&-1358&-1438&706\\
0.0002&-676&-1171&1074&95&1004&-203&-1125&-1206&-1165&874\\
0.0003&-460&-1209&1821&502&1576&-72&-1061&-1334&-1154&847\\
0.0004&-244&-1219&1979&790&2088&105&-1168&-1306&-951&875\\
0.0005&-149&-1080&1989&1219&1967&453&-1080&-1151&-1003&885\\
0.0006&-84&-1020&2222&1510&2158&795&-933&-1228&-986&783\\
0.0007&-35&-1030&2035&1483&1727&1075&-862&-1306&-1033&800\\
0.0008&19&-935&1968&1566&1579&1192&-819&-1222&-850&820\\
0.0009&63&-875&1644&1502&1579&1378&-890&-1248&-837&899\\
0.001&104&-919&1399&1519&984&1560&-887&-1091&-1001&809\\
0.002&527&-699&70&541&-43&1723&-689&-1258&-609&779\\
0.003&726&-361&-33184&-1132&-20378&1220&-335&-950&-382&773\\
0.004&797&-256&-74476&-33021&-29906&690&-290&-894&-250&753\\
0.005&832&-132&-84827&-39256&-71094&404&-210&-810&-144&740\\
0.006&827&-76&-74078&-13535&-100350&160&-159&-944&-74&795\\
0.007&857&-22&-87454&-15977&-92454&-1729&-131&-668&-26&675\\
0.008&892&15&-277787&-13865&-238074&-5196&-104&-738&17&703\\
0.009&842&55&-288298&-49565&-201878&-47343&-59&-715&52&726\\
0.01&764&108&-274927&-78027&-199549&-56241&-42&-772&94&751\\
\hline
    \end{tabular}}
    \caption{The asymptotic returns averaged across last 50 episodes over 10 random seeds for A2C on the Ant environment. }
    \label{tab:a2c_asymptotic_lr_ant}
\end{table}

\begin{table}[H]
    \centering
    \small{
    \begin{tabular}{|c|c|c|c|c|c|c|c|c|c|c|}
\hline
lr&SGDNM&SGD&RMS&AMS&A*m&A*max&A*grad&A*delta&ASGD&YF\\
\hline
0.0001&1387&267&4149&4159&3754&3412&135&119&271&1248\\
0.0002&1927&362&4024&3162&3732&3736&254&216&364&1451\\
0.0003&2252&417&2756&4058&3543&4090&309&237&442&1332\\
0.0004&2993&566&2367&3253&2367&3986&367&285&510&1515\\
0.0005&3402&697&1554&3193&1926&3518&386&313&686&1459\\
0.0006&3157&921&1646&3453&1396&3766&432&357&759&1295\\
0.0007&3245&929&1301&2219&955&3852&533&358&931&1253\\
0.0008&3292&1151&931&1844&746&3062&655&367&1284&945\\
0.0009&3265&1132&788&1394&789&3704&819&408&1081&986\\
0.001&3099&1509&746&954&612&3081&843&427&1073&1326\\
0.002&3104&1788&539&554&516&1242&1886&895&1983&1070\\
0.003&3027&2269&420&418&503&591&2287&1151&2448&1320\\
0.004&3264&2724&446&432&352&454&2720&1657&2687&1186\\
0.005&2518&2865&355&387&375&410&2870&1672&2444&1159\\
0.006&2465&3272&328&430&329&416&2978&2267&2818&1195\\
0.007&2382&3009&327&475&298&405&3392&2406&3145&889\\
0.008&1776&3478&268&332&305&448&3582&2483&3049&1012\\
0.009&2050&3150&274&375&279&350&3259&2674&2917&761\\
0.01&1673&2823&180&337&248&374&3548&2855&3550&698\\
\hline
    \end{tabular}}
    \caption{The asymptotic returns averaged across last 50 episodes over 10 random seeds for PPO on the Walker2d environment. }
    \label{tab:ppo_asymptotic_lr_walker}
\end{table}

\begin{table}[H]
    \centering
    \small{
    \begin{tabular}{|c|c|c|c|c|c|c|c|c|c|c|}
\hline
lr&SGDNM&SGD&RMS&AMS&A*m&A*max&A*grad&A*delta&ASGD&YF\\
\hline
0.0001&-16&-38&-6&-8&-6&-6&-100&-107&-38&-11\\
0.0002&-14&-27&-5&-7&-5&-7&-91&-99&-27&-12\\
0.0003&-13&-22&-6&-5&-5&-6&-82&-92&-22&-13\\
0.0004&-12&-20&-6&-5&-6&-6&-74&-86&-20&-12\\
0.0005&-13&-18&-6&-6&-6&-7&-68&-80&-18&-12\\
0.0006&-11&-17&-6&-6&-6&-6&-61&-74&-17&-11\\
0.0007&-12&-17&-6&-5&-6&-6&-56&-69&-17&-11\\
0.0008&-10&-17&-6&-5&-7&-6&-52&-64&-16&-11\\
0.0009&-9&-15&-7&-6&-7&-6&-49&-60&-16&-12\\
0.001&-9&-15&-8&-6&-7&-6&-45&-56&-14&-12\\
0.002&-10&-14&-18&-8&-14&-6&-26&-31&-15&-13\\
0.003&-7&-13&-45&-12&-30&-8&-18&-22&-12&-14\\
0.004&-7&-13&-80&-20&-48&-9&-14&-17&-11&-12\\
0.005&-8&-12&-121&-39&-72&-11&-10&-15&-13&-13\\
0.006&-7&-12&-175&-62&-113&-14&-10&-13&-12&-14\\
0.007&-8&-10&-249&-112&-134&-20&-8&-12&-12&-13\\
0.008&-7&-10&-254&-134&-160&-20&-8&-11&-10&-13\\
0.009&-7&-10&-407&-160&-215&-35&-8&-10&-11&-13\\
0.01&-7&-10&-461&-196&-230&-45&-8&-10&-12&-14\\
\hline
    \end{tabular}}
    \caption{The asymptotic returns averaged across last 50 episodes over 10 random seeds for PPO on the Reacher environment. }
    \label{tab:ppo_asymptotic_lr_reacher}
\end{table}

\begin{table}[H]
    \centering
    \small{
    \begin{tabular}{|c|c|c|c|c|c|c|c|c|c|c|}
\hline
lr&SGDNM&SGD&RMS&AMS&A*m&A*max&A*grad&A*delta&ASGD&YF\\
\hline
0.0001&1458&521&2365&2425&2076&2281&280&218&541&2184\\
0.0002&1998&801&2319&2519&2830&2643&401&381&848&2101\\
0.0003&2408&1072&2759&2633&2513&2379&473&524&1045&1832\\
0.0004&2657&1165&2194&2221&2147&2652&648&490&1251&2025\\
0.0005&2179&1210&2323&2267&2152&2738&690&663&1170&2066\\
0.0006&2313&1349&1866&2749&1950&2620&818&645&1539&1844\\
0.0007&2112&1349&1978&2558&1370&2779&948&746&1305&1885\\
0.0008&1857&1426&1845&2655&1194&2553&899&768&1531&2023\\
0.0009&2322&1570&1401&2257&1134&2487&885&814&1654&2086\\
0.001&2360&1673&1534&2169&1445&2538&963&812&1502&1986\\
0.002&2545&1922&1229&1308&1087&1807&1657&1147&1892&1783\\
0.003&2312&2284&828&1110&975&1466&1911&1180&2158&1760\\
0.004&2299&2100&930&1500&1019&1051&1929&1543&2502&2087\\
0.005&2399&2559&743&916&922&973&2065&1749&2085&1858\\
0.006&2326&2417&780&992&839&1099&2212&2060&2464&1821\\
0.007&2260&2387&760&1007&838&1076&2118&2056&2174&1592\\
0.008&1952&2205&713&946&851&727&2489&2022&2581&1817\\
0.009&2062&2266&719&958&766&900&2248&2177&2285&1722\\
0.01&2000&2562&779&957&701&844&2555&2200&2374&1466\\

\hline
    \end{tabular}}
    \caption{The asymptotic returns averaged across last 50 episodes over 10 random seeds for PPO on the Hopper environment. }
    \label{tab:ppo_asymptotic_lr_hopper}
\end{table}

\begin{table}[H]
    \centering
    \small{
    \begin{tabular}{|c|c|c|c|c|c|c|c|c|c|c|}
\hline
lr&SGDNM&SGD&RMS&AMS&A*m&A*max&A*grad&A*delta&ASGD&YF\\
\hline
0.0001&1015&-451&3301&2407&2887&1615&-645&-690&-445&1668\\
0.0002&1569&-148&3674&2861&3327&2384&-610&-646&-179&1567\\
0.0003&1757&168&2848&2695&2599&2736&-543&-595&90&1935\\
0.0004&2165&204&3411&3212&2656&2652&-502&-558&289&1909\\
0.0005&2396&416&2559&3221&2982&2842&-443&-528&424&1766\\
0.0006&2039&489&2422&2523&2327&3104&-393&-467&487&1780\\
0.0007&2179&665&1752&2548&1878&2892&-350&-437&596&1455\\
0.0008&2162&752&1585&2218&1198&2962&-263&-413&725&905\\
0.0009&2381&977&384&2471&1074&2474&-193&-368&940&1626\\
0.001&2640&956&1045&2267&608&2933&-169&-331&1205&1506\\
0.002&2254&1576&-241&22&105&1938&366&88&1300&1328\\
0.003&3004&1555&-915&-517&-664&893&731&402&1632&1948\\
0.004&2634&1639&-1777&-1018&-1259&298&999&598&1498&851\\
0.005&2857&2116&-3003&-1517&-2151&-581&1251&823&1768&1633\\
0.006&2323&1967&-4417&-2059&-3141&-551&1537&963&1882&922\\
0.007&2733&2303&-6030&-2598&-4876&-719&1455&1183&2585&1188\\
0.008&2624&2158&-9124&-3054&-6106&-1367&1622&1179&1984&576\\
0.009&2514&2442&-11003&-3780&-9071&-1613&1734&1185&2473&323\\
0.01&2615&2516&-14279&-4326&-10546&-1808&2265&1328&2347&-585\\
\hline
    \end{tabular}}
    \caption{The asymptotic returns averaged across last 50 episodes over 10 random seeds for PPO on the HalfCheetah environment. SGDNM stands for Nesterov Momentum. A* is the Ada family of algorithms. RMS indicates RMSProp and AMS indicates AMSGrad.}
    \label{tab:ppo_asymptotic_lr_hc}
\end{table}

\begin{table}[H]
    \centering
    \small{
    \begin{tabular}{|c|c|c|c|c|c|c|c|c|c|c|}
\hline
lr&SGDNM&SGD&RMS&AMS&A*m&A*max&A*grad&A*delta&ASGD&YF\\
\hline
0.0001&68&-1278&1882&2046&2238&1371&-1044&-1300&-1198&27\\
0.0002&678&-968&912&2020&694&2266&-1094&-1118&-1010&52\\
0.0003&1116&-750&568&1161&317&2662&-1073&-1067&-634&23\\
0.0004&1626&-669&217&681&229&2195&-962&-1153&-507&21\\
0.0005&1974&-292&132&353&170&1520&-817&-970&-332&54\\
0.0006&1970&-162&63&157&114&1413&-603&-994&-244&31\\
0.0007&1905&-167&50&86&56&828&-618&-840&-138&17\\
0.0008&2215&-41&19&64&65&590&-514&-777&-50&34\\
0.0009&1784&10&14&26&-23&459&-355&-889&12&26\\
0.001&2322&62&-91&12&34&400&-417&-830&65&14\\
0.002&1814&775&-1258&-1245&-1207&-7&-289&-290&734&14\\
0.003&1564&1251&-3132&-1580&-2412&-423&-52&-289&1135&-35\\
0.004&1251&1459&-3209&-2243&-5075&-1440&100&-149&1469&-188\\
0.005&636&1971&-6566&-4004&-6578&-1846&431&-24&1859&-321\\
0.006&705&1635&-7386&-6256&-6071&-2774&617&92&1915&-802\\
0.007&273&2175&-12323&-6014&-9988&-2370&875&287&1852&-701\\
0.008&38&2126&-19267&-11176&-13108&-3342&940&531&2505&-1186\\
0.009&51&2051&-16438&-10500&-17834&-3188&1058&697&2422&-1667\\
0.01&-37&1936&-120820&-9117&-49749&-3730&1229&794&2485&-1821\\
\hline
    \end{tabular}}
    \caption{The asymptotic returns averaged across last 50 episodes over 10 random seeds for PPO on the Ant environment. }
    \label{tab:ppo_asymptotic_lr_ant}
\end{table}

\subsection{Variance Between Random Seeds (Average)}

\begin{table}[H]
    \centering
    \small{
    \begin{tabular}{|c|c|c|c|c|c|c|c|c|c|c|}
\hline
lr&SGDNM&SGD&RMS&AMS&A*m&A*max&A*grad&A*delta&ASGD&YF\\
\hline
0.0001&375&80&818&894&868&820&71&46&88&761\\
0.0002&555&95&935&1081&989&843&90&72&94&854\\
0.0003&573&100&1027&967&1199&970&101&78&96&745\\
0.0004&712&112&1075&1042&1133&856&107&82&101&886\\
0.0005&672&131&987&1196&1020&944&100&85&124&1048\\
0.0006&830&158&983&1104&928&921&94&85&180&884\\
0.0007&667&171&854&1255&739&913&122&90&189&786\\
0.0008&780&201&617&927&708&1055&122&88&287&630\\
0.0009&814&218&475&930&518&1175&145&92&206&706\\
0.001&864&349&332&783&428&1070&162&92&195&811\\
0.002&924&519&241&218&223&783&458&161&631&570\\
0.003&1070&623&221&170&210&277&579&275&612&691\\
0.004&1038&729&151&170&160&270&807&411&738&735\\
0.005&1057&720&195&168&171&195&777&314&786&633\\
0.006&1159&666&194&163&187&180&771&483&693&725\\
0.007&1069&823&162&187&161&178&759&531&724&538\\
0.008&906&952&137&138&140&156&862&625&797&591\\
0.009&1220&866&115&158&141&167&876&623&670&432\\
0.01&1084&1004&111&163&139&165&873&625&813&464\\
\hline
    \end{tabular}}
    \caption{The average standard deviation of returns between 10 different random seeds for PPO on the Walker environment.}
    \label{tab:ppo_average_variance_walker}
\end{table}

\begin{table}[H]
    \centering
    \small{
    \begin{tabular}{|c|c|c|c|c|c|c|c|c|c|c|}
\hline
lr&SGDNM&SGD&RMS&AMS&A*m&A*max&A*grad&A*delta&ASGD&YF\\
\hline
0.0001&3&8&3&4&4&4&12&14&8&4\\
0.0002&3&6&3&4&3&4&11&13&6&4\\
0.0003&4&5&3&3&3&3&11&12&5&4\\
0.0004&4&5&3&3&3&4&10&12&4&4\\
0.0005&3&4&3&3&3&4&9&11&4&4\\
0.0006&4&4&3&3&3&3&9&11&4&4\\
0.0007&4&4&3&3&3&3&8&11&4&4\\
0.0008&4&4&3&3&3&3&8&10&4&4\\
0.0009&4&3&3&3&3&3&8&10&3&4\\
0.001&4&3&3&3&3&3&7&10&3&4\\
0.002&4&3&14&4&4&3&5&7&3&4\\
0.003&4&3&17&4&7&4&4&6&4&4\\
0.004&3&4&25&6&12&4&4&5&4&4\\
0.005&4&4&36&12&18&4&4&5&3&4\\
0.006&3&3&47&18&26&4&4&5&4&4\\
0.007&4&4&309&25&35&6&4&5&4&4\\
0.008&4&4&79&34&43&5&4&5&4&4\\
0.009&3&4&351&41&51&8&4&5&4&4\\
0.01&4&4&475&56&67&13&4&5&4&4\\

\hline
    \end{tabular}}
    \caption{The average standard deviation of returns between 10 different random seeds for PPO on the Reacher environment. }
    \label{tab:ppo_average_variance_reacher}
\end{table}

\begin{table}[H]
    \centering
    \small{
    \begin{tabular}{|c|c|c|c|c|c|c|c|c|c|c|}
\hline
lr&SGDNM&SGD&RMS&AMS&A*m&A*max&A*grad&A*delta&ASGD&YF\\
\hline
0.0002&514&214&709&709&692&645&175&116&220&751\\
0.0003&568&214&738&713&722&698&134&184&218&731\\
0.0004&609&241&764&791&859&683&152&159&229&777\\
0.0005&751&249&782&753&758&681&190&211&236&725\\
0.0006&699&280&758&691&856&776&221&188&299&742\\
0.0007&684&275&751&795&769&769&190&230&249&731\\
0.0008&838&282&783&739&716&707&235&235&288&747\\
0.0009&804&424&755&789&671&840&228&241&353&780\\
0.001&646&410&745&788&648&738&248&219&349&744\\
0.002&642&515&529&559&569&815&477&227&532&743\\
0.003&824&567&401&509&439&807&467&243&574&815\\
0.004&786&619&349&518&437&514&489&446&570&755\\
0.005&732&642&356&435&373&621&550&398&598&766\\
0.006&802&625&305&429&330&417&564&452&691&735\\
0.007&796&685&333&309&268&550&587&566&815&715\\
0.008&878&705&263&344&290&420&620&517&678&773\\
0.009&746&777&254&359&309&504&631&499&753&739\\
0.01&722&775&308&359&307&394&681&524&699&699\\
\hline
    \end{tabular}}
    \caption{The average standard deviation of returns between 10 different random seeds for PPO on the Hopper environment.}
    \label{tab:ppo_average_variance_hopper}
\end{table}

\begin{table}[H]
    \centering
    \small{
    \begin{tabular}{|c|c|c|c|c|c|c|c|c|c|c|}
\hline
lr&SGDNM&SGD&RMS&AMS&A*m&A*max&A*grad&A*delta&ASGD&YF\\
\hline
0.0001&329&67&876&677&722&351&62&67&67&632\\
0.0002&436&75&1266&971&1244&576&60&64&74&820\\
0.0003&401&103&1245&1100&998&777&56&61&98&921\\
0.0004&592&147&1299&1220&1119&1022&59&60&136&1031\\
0.0005&695&186&1044&1204&1063&919&65&59&184&998\\
0.0006&489&223&951&965&1241&1089&66&57&206&894\\
0.0007&549&239&802&1037&1016&994&82&58&237&822\\
0.0008&490&249&899&963&845&1206&84&58&260&864\\
0.0009&728&288&656&1027&629&972&91&59&287&879\\
0.001&713&279&845&975&561&1060&98&62&326&963\\
0.002&573&528&597&641&360&868&147&75&389&874\\
0.003&1027&447&398&518&296&644&209&149&392&909\\
0.004&968&368&313&292&282&468&281&137&378&921\\
0.005&1222&522&359&236&268&700&406&152&473&915\\
0.006&729&516&604&215&329&610&485&172&473&913\\
0.007&905&752&955&252&543&521&332&279&711&806\\
0.008&1112&633&1305&274&707&472&483&240&613&844\\
0.009&926&624&2048&376&1054&372&461&218&743&830\\
0.01&911&617&2486&399&1256&429&681&277&674&5480\\
\hline
    \end{tabular}}
    \caption{The average standard deviation of returns between 10 different random seeds for PPO on the HalfCheetah environment. }
    \label{tab:ppo_average_variance_hc}
\end{table}

\begin{table}[H]
    \centering
    \small{
    \begin{tabular}{|c|c|c|c|c|c|c|c|c|c|c|}
\hline
lr&SGDNM&SGD&RMS&AMS&A*m&A*max&A*grad&A*delta&ASGD&YF\\
\hline
0.0001&462&1083&465&427&504&356&1072&1115&1084&143\\
0.0002&315&1042&401&481&269&500&1033&1094&1037&140\\
0.0003&311&980&283&374&210&554&1005&1078&969&139\\
0.0004&347&907&177&293&177&548&974&1061&895&139\\
0.0005&400&823&162&217&157&461&945&1048&814&138\\
0.0006&427&737&145&178&159&398&917&1027&724&146\\
0.0007&384&654&132&157&150&304&892&1004&650&130\\
0.0008&510&581&136&133&132&281&861&991&571&131\\
0.0009&415&524&131&130&143&246&830&977&513&137\\
0.001&498&471&234&133&137&228&799&958&466&142\\
0.002&448&314&961&991&749&136&536&810&318&128\\
0.003&387&324&2260&1775&1755&267&357&690&317&129\\
0.004&432&357&3902&2733&3416&776&251&586&362&166\\
0.005&281&411&6579&4356&5327&1258&224&491&426&190\\
0.006&389&384&10549&5938&8093&1682&229&419&407&265\\
0.007&244&489&14688&7015&12211&2030&253&373&442&353\\
0.008&181&507&22949&9746&16269&2559&269&356&445&389\\
0.009&152&417&28009&11626&19736&2964&291&339&485&650\\
0.01&152&422&451006&15373&26733&3339&303&331&533&691\\
\hline
    \end{tabular}}
    \caption{The average standard deviation of returns between 10 different random seeds for PPO on the Ant environment.}
    \label{tab:ppo_average_variance_ant}
\end{table}

\begin{table}[H]
    \centering
    \small{
    \begin{tabular}{|c|c|c|c|c|c|c|c|c|c|c|}
\hline
lr&SGDNM&SGD&RMS&AMS&A*m&A*max&A*grad&A*delta&ASGD&YF\\
\hline
0.0001&87&9&306&161&287&153&20&5&9&223\\
0.0002&103&24&408&244&254&163&54&6&24&306\\
0.0003&116&41&487&295&355&241&78&9&41&312\\
0.0004&124&56&667&335&576&332&87&15&56&355\\
0.0005&132&65&420&395&547&338&90&19&65&259\\
0.0006&134&71&656&426&402&245&91&24&71&284\\
0.0007&147&75&536&296&450&399&90&30&76&287\\
0.0008&152&78&591&465&533&430&96&36&79&353\\
0.0009&169&87&556&590&462&376&102&43&83&288\\
0.001&215&87&622&457&510&507&115&49&90&327\\
0.002&235&104&408&391&295&339&182&76&107&325\\
0.003&324&116&205&188&186&389&144&88&122&241\\
0.004&340&161&201&159&197&411&250&92&129&207\\
0.005&375&138&185&174&192&321&237&110&131&231\\
0.006&317&147&151&149&144&254&233&106&135&213\\
0.007&344&158&143&152&117&191&184&124&150&197\\
0.008&357&152&126&123&124&211&212&122&160&206\\
0.009&369&186&88&89&115&225&245&134&153&251\\
0.01&358&220&107&95&69&185&315&146&180&208\\

\hline
    \end{tabular}}
    \caption{The average standard deviation of returns between 10 different random seeds for A2C on the Walker2d environment.}
    \label{tab:a2c_average_variance_walker}
\end{table}

\begin{table}[H]
    \centering
    \small{
    \begin{tabular}{|c|c|c|c|c|c|c|c|c|c|c|}
\hline
lr&SGDNM&SGD&RMS&AMS&A*m&A*max&A*grad&A*delta&ASGD&YF\\
\hline
0.0001&10&15&6&6&6&11&14&16&15&1603\\
0.0002&8&14&4&5&5&7&13&16&14&1131\\
0.0003&7&13&4&4&4&6&12&15&13&1533\\
0.0004&6&12&4&4&5&5&12&15&12&3266\\
0.0005&5&11&4&4&5&5&12&15&11&3296\\
0.0006&5&11&4&4&4&5&12&14&11&3315\\
0.0007&5&10&4&4&4&4&12&14&10&2888\\
0.0008&4&10&4&4&5&4&12&14&10&3119\\
0.0009&4&10&4&4&5&4&12&14&10&5136\\
0.001&4&10&4&4&5&4&12&14&10&5289\\
0.002&4&8&38&7&41&4&11&13&8&1675\\
0.003&4&7&109&60&191&4&9&13&7&2197\\
0.004&3&6&714&264&398&8&7&13&6&7717\\
0.005&3&5&1819&328&460&17&6&13&5&5829\\
0.006&4&5&410&790&1247&373&5&13&5&6589\\
0.007&4&5&1579&711&2369&73&5&13&5&2163\\
0.008&5&4&4108&1251&2024&490&4&12&4&3281\\
0.009&6&4&4030&3725&18586&288&4&12&4&8131\\
0.01&6&4&2480&4247&8828&954&4&11&4&9244\\
\hline
    \end{tabular}}
    \caption{The average standard deviation of returns between 10 different random seeds for A2C on the Reacher environment.}
    \label{tab:a2c_average_variance_reacher}
\end{table}

\begin{table}[H]
    \centering
    \small{
    \begin{tabular}{|c|c|c|c|c|c|c|c|c|c|c|}
\hline
lr&SGDNM&SGD&RMS&AMS&A*m&A*max&A*grad&A*delta&ASGD&YF\\
\hline
0.0001&78&14&293&265&283&267&19&11&14&367\\
0.0002&190&19&250&284&349&271&32&12&19&481\\
0.0003&232&28&361&297&291&282&37&14&27&562\\
0.0004&250&34&397&307&323&276&85&15&34&443\\
0.0005&261&39&525&324&446&336&149&16&39&457\\
0.0006&265&42&567&292&402&350&198&18&42&388\\
0.0007&283&49&539&338&379&290&226&21&48&477\\
0.0008&282&56&587&494&440&291&253&24&56&451\\
0.0009&280&69&595&508&491&303&263&27&65&491\\
0.001&286&79&581&445&443&259&269&30&81&390\\
0.002&298&191&479&272&419&479&268&45&195&461\\
0.003&310&226&567&375&292&493&286&116&234&413\\
0.004&373&248&404&330&241&353&262&178&270&321\\
0.005&368&265&351&297&206&375&287&213&260&391\\
0.006&369&269&333&248&203&404&288&238&277&409\\
0.007&320&273&337&220&219&343&267&251&265&465\\
0.008&379&269&300&199&207&324&312&260&271&445\\
0.009&318&285&281&185&175&325&290&249&278&424\\
0.01&382&279&256&186&185&305&263&264&298&394\\
\hline
    \end{tabular}}
    \caption{The average standard deviation of returns between 10 different random seeds for A2C on the Hopper environment. }
    \label{tab:a2c_average_variance_hopper}
\end{table}

\begin{table}[H]
    \centering
    \small{
    \begin{tabular}{|c|c|c|c|c|c|c|c|c|c|c|}
\hline
lr&SGDNM&SGD&RMS&AMS&A*m&A*max&A*grad&A*delta&ASGD&YF\\
\hline
0.0001&59&68&72&71&73&61&65&69&67&1122\\
0.0002&53&66&58&96&82&64&61&68&67&957\\
0.0003&51&65&205&288&390&77&58&67&66&998\\
0.0004&56&64&474&421&455&94&57&66&64&1057\\
0.0005&58&64&537&421&531&108&56&66&63&960\\
0.0006&71&63&872&503&694&270&57&66&62&1121\\
0.0007&84&62&564&754&865&109&59&66&61&808\\
0.0008&89&61&992&754&779&179&61&65&61&902\\
0.0009&88&60&1010&963&955&306&63&64&60&1067\\
0.001&104&60&906&883&993&377&69&65&58&1161\\
0.002&180&55&905&838&661&880&83&61&55&1024\\
0.003&299&52&563&622&401&720&68&61&53&1004\\
0.004&395&57&485&374&414&891&78&63&55&980\\
0.005&448&61&809&388&888&904&74&64&61&1114\\
0.006&435&66&1449&534&1301&795&75&67&67&918\\
0.007&502&81&2172&707&2144&505&91&69&78&1006\\
0.008&781&86&4932&1118&2047&542&95&70&84&981\\
0.009&786&91&4040&1466&5419&405&102&71&96&907\\
0.01&819&95&22045&1885&8588&429&103&68&107&1012\\
\hline
    \end{tabular}}
    \caption{The average standard deviation of returns between 10 different random seeds for A2C on the HalfCheetah environment.}
    \label{tab:a2c_average_variance_hc}
\end{table}

\begin{table}[H]
    \centering
    \small{
    \begin{tabular}{|c|c|c|c|c|c|c|c|c|c|c|}
\hline
lr&SGDNM&SGD&RMS&AMS&A*m&A*max&A*grad&A*delta&ASGD&YF\\
\hline
0.0001&929&1123&469&523&482&818&1111&1150&1120&170\\
0.0002&806&1089&337&292&331&627&1073&1141&1093&176\\
0.0003&685&1064&392&239&363&487&1045&1133&1065&164\\
0.0004&587&1035&446&245&421&387&1016&1135&1036&164\\
0.0005&501&1014&486&282&429&335&996&1120&1011&165\\
0.0006&433&993&502&304&459&326&973&1116&993&166\\
0.0007&381&978&508&322&459&324&961&1116&975&158\\
0.0008&335&958&483&346&443&335&944&1109&959&166\\
0.0009&301&943&487&354&429&345&932&1101&938&171\\
0.001&277&929&454&381&391&361&918&1095&927&161\\
0.002&186&799&197&232&200&477&819&1058&799&163\\
0.003&146&690&11428&392&9154&468&715&1024&686&163\\
0.004&134&588&23612&17418&21257&400&611&996&584&159\\
0.005&131&508&33625&16325&46092&285&522&976&498&164\\
0.006&143&440&43881&14908&64667&231&446&952&434&159\\
0.007&143&377&65953&18838&93479&586&386&934&380&164\\
0.008&151&335&420013&29690&168984&1061&338&912&336&169\\
0.009&152&300&141179&49452&107119&11963&301&899&300&170\\
0.01&160&277&304400&90664&141829&11515&280&878&277&173\\
\hline
    \end{tabular}}
    \caption{The average standard deviation of returns between 10 different random seeds for A2C on the Ant environment. }
    \label{tab:a2c_average_variance_ant}
\end{table}

\section{Momentum Experiments}
\label{app:momentum}

\subsection{Learning Curves}

\begin{figure}[H]
    \centering
    \includegraphics[width=.32\textwidth]{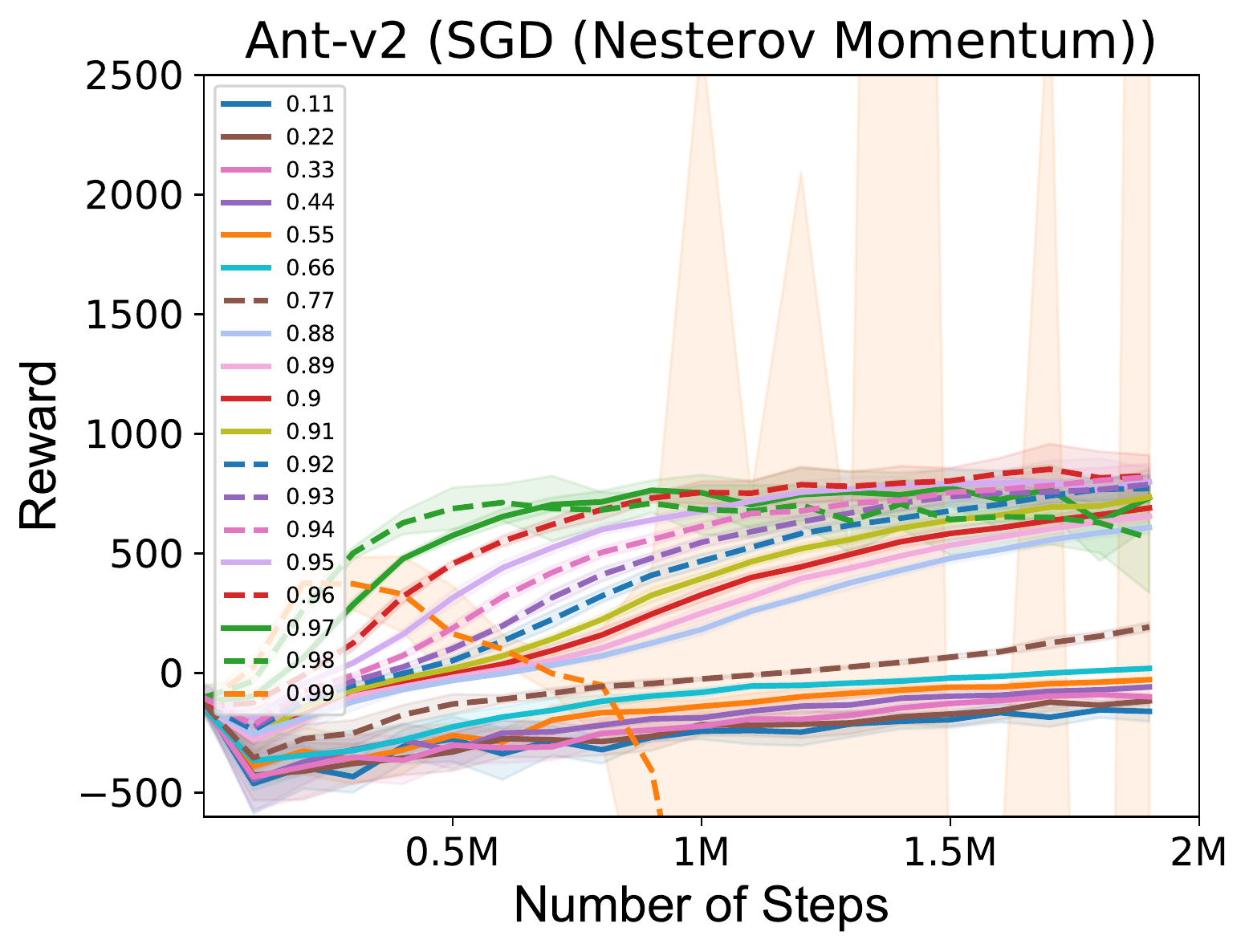}
    \includegraphics[width=.32\textwidth]{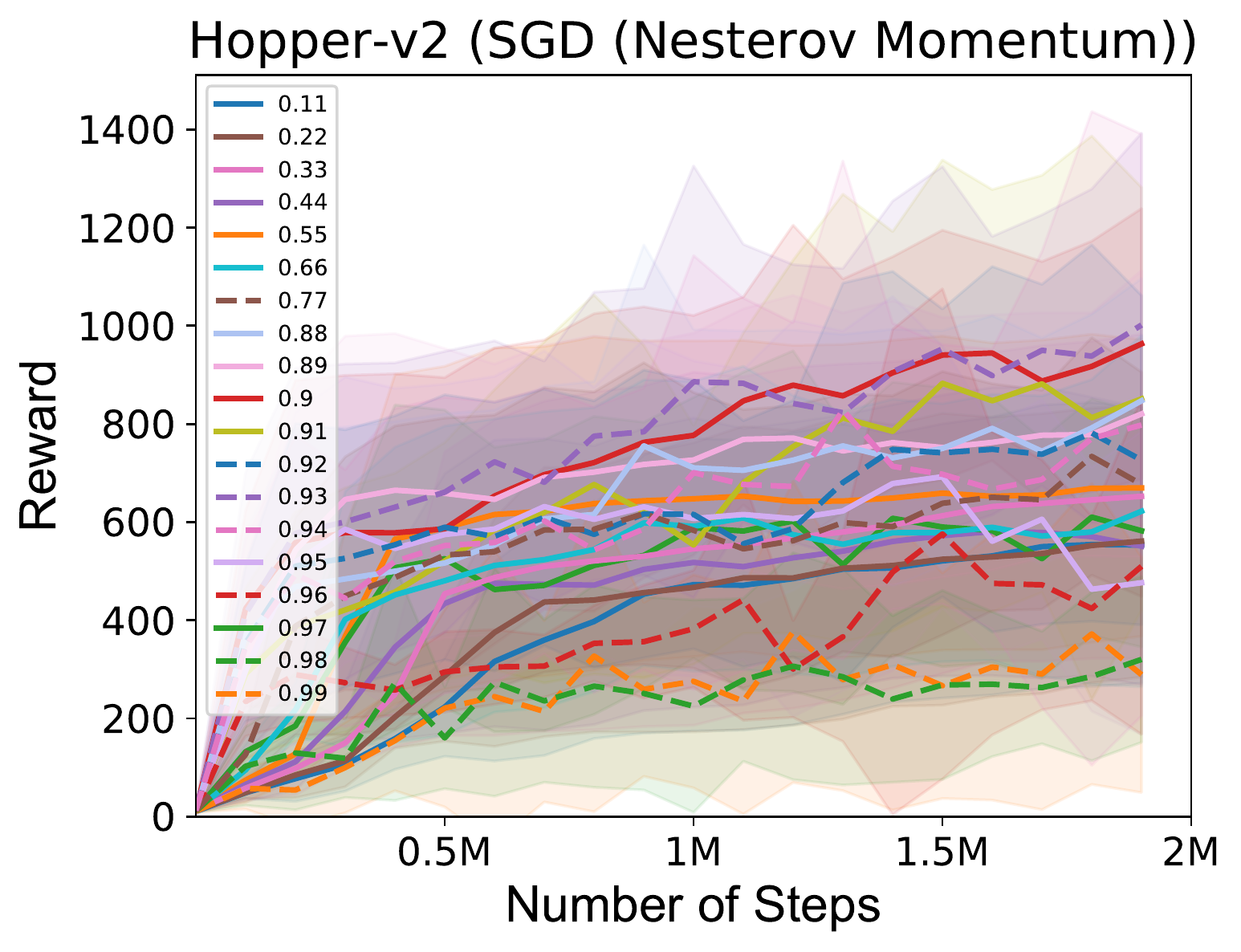}
    \includegraphics[width=.32\textwidth]{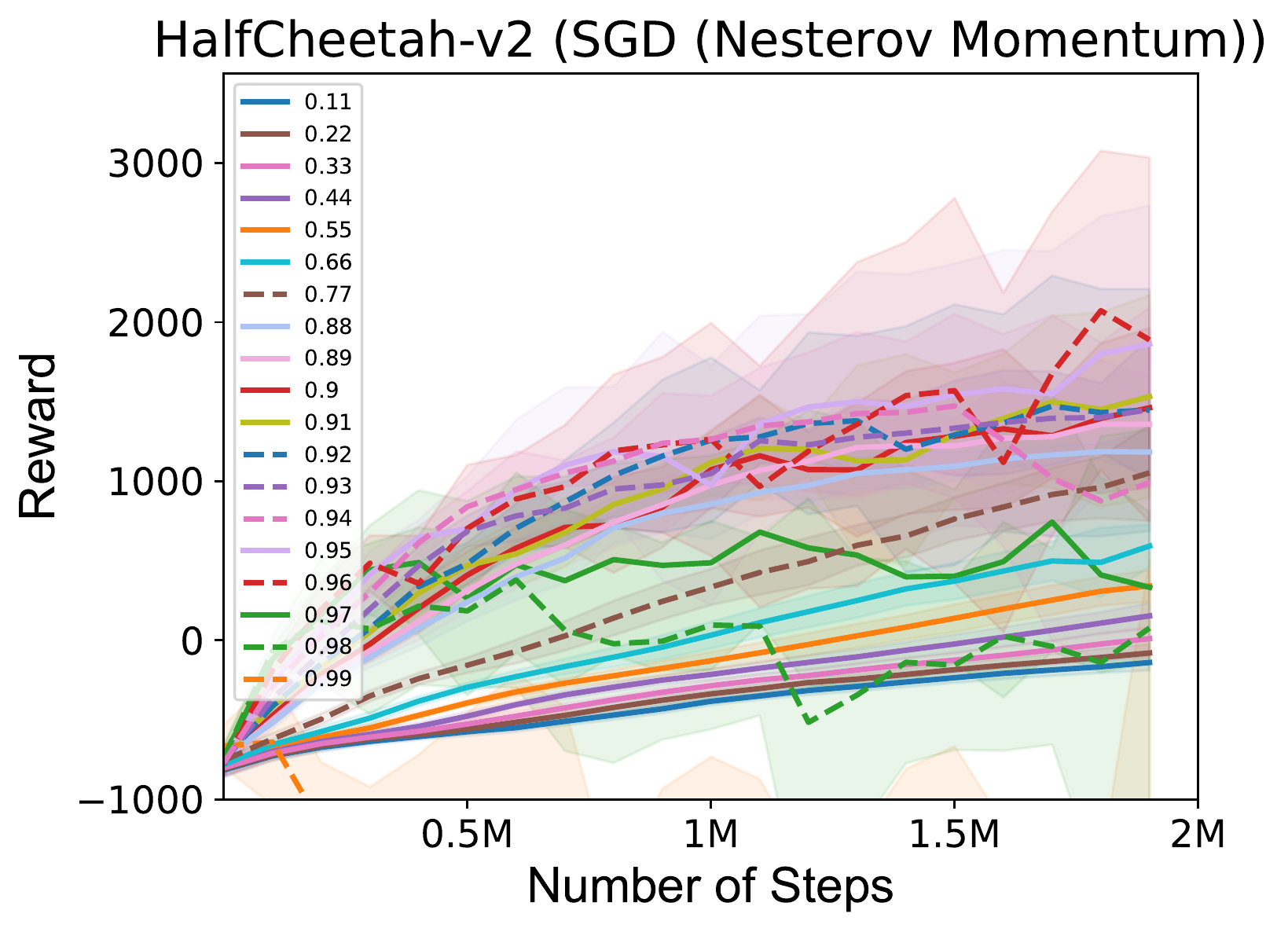}
    \includegraphics[width=.32\textwidth]{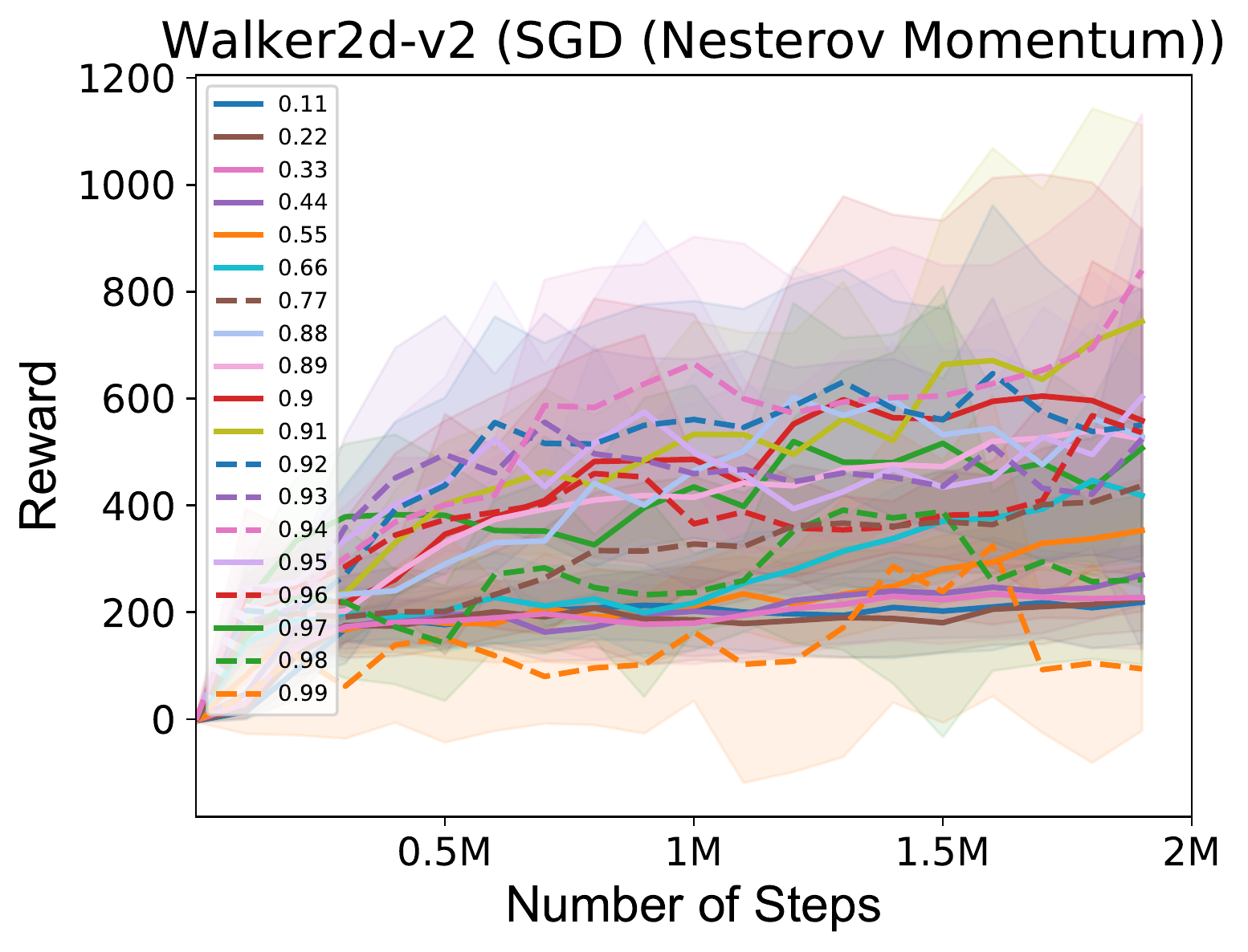}
    \includegraphics[width=.32\textwidth]{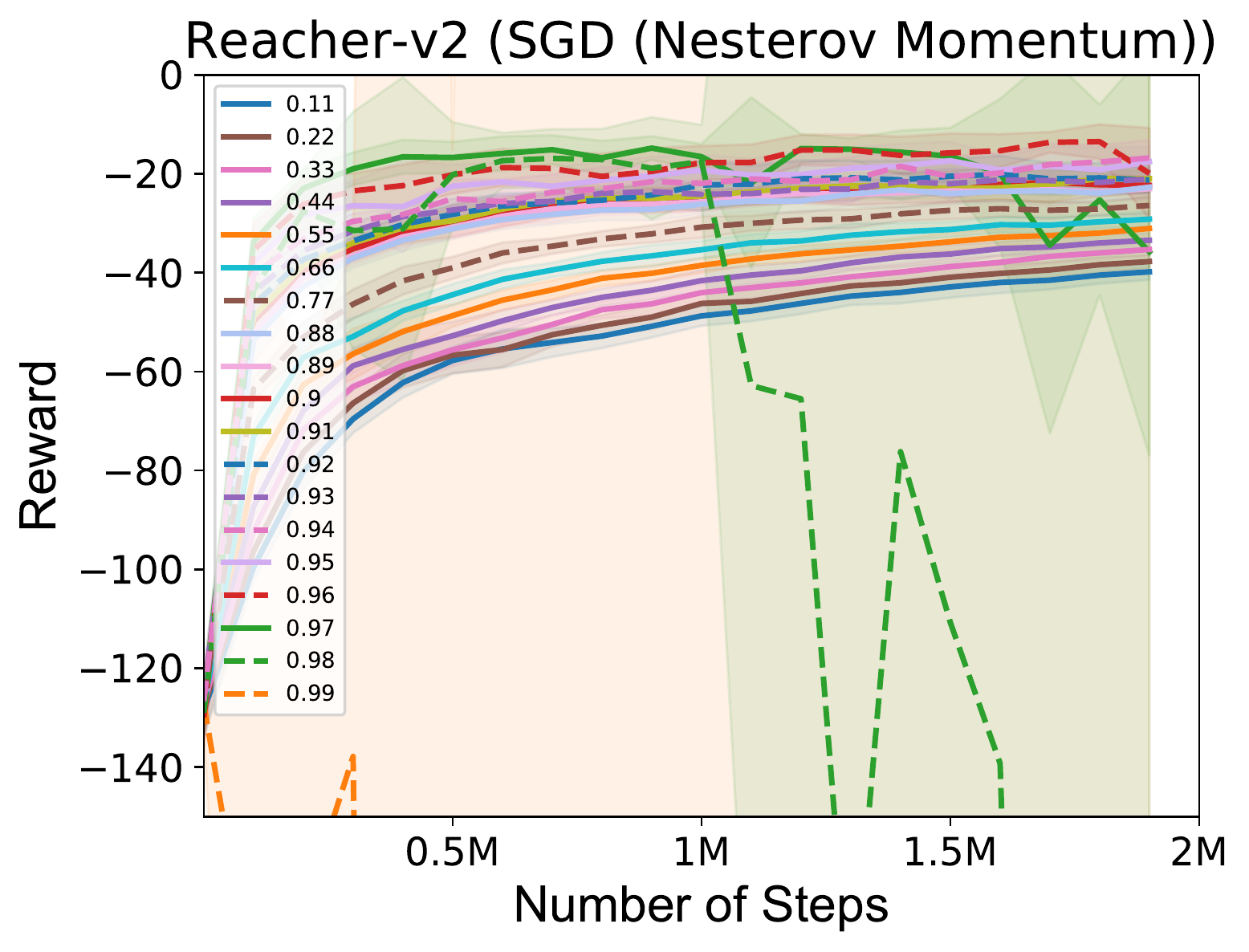}
    \caption{A2C performance across momentum values using SGD with nesterov momentum.}
\end{figure}

\begin{figure}[H]
    \centering
    \includegraphics[width=.32\textwidth]{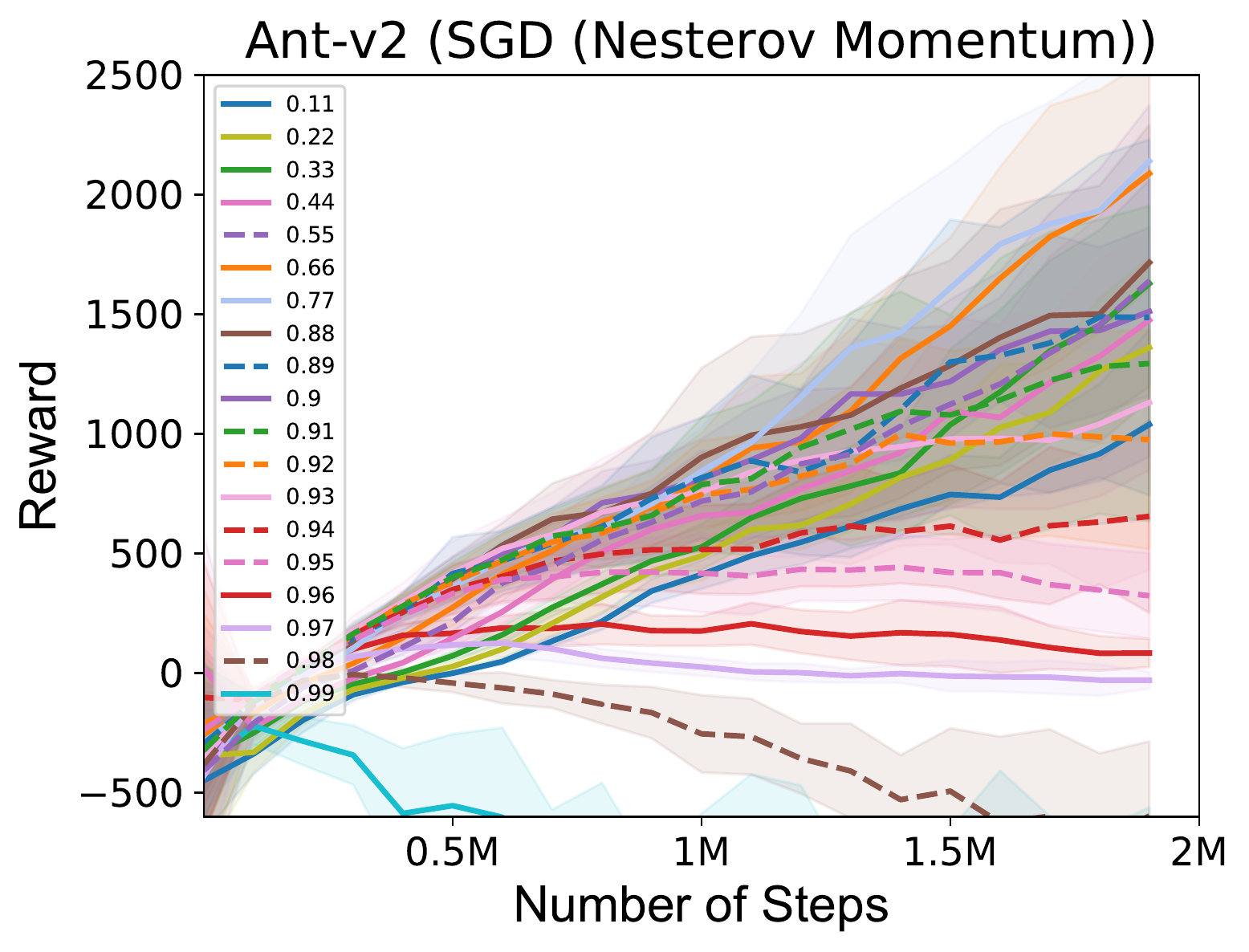}
    \includegraphics[width=.32\textwidth]{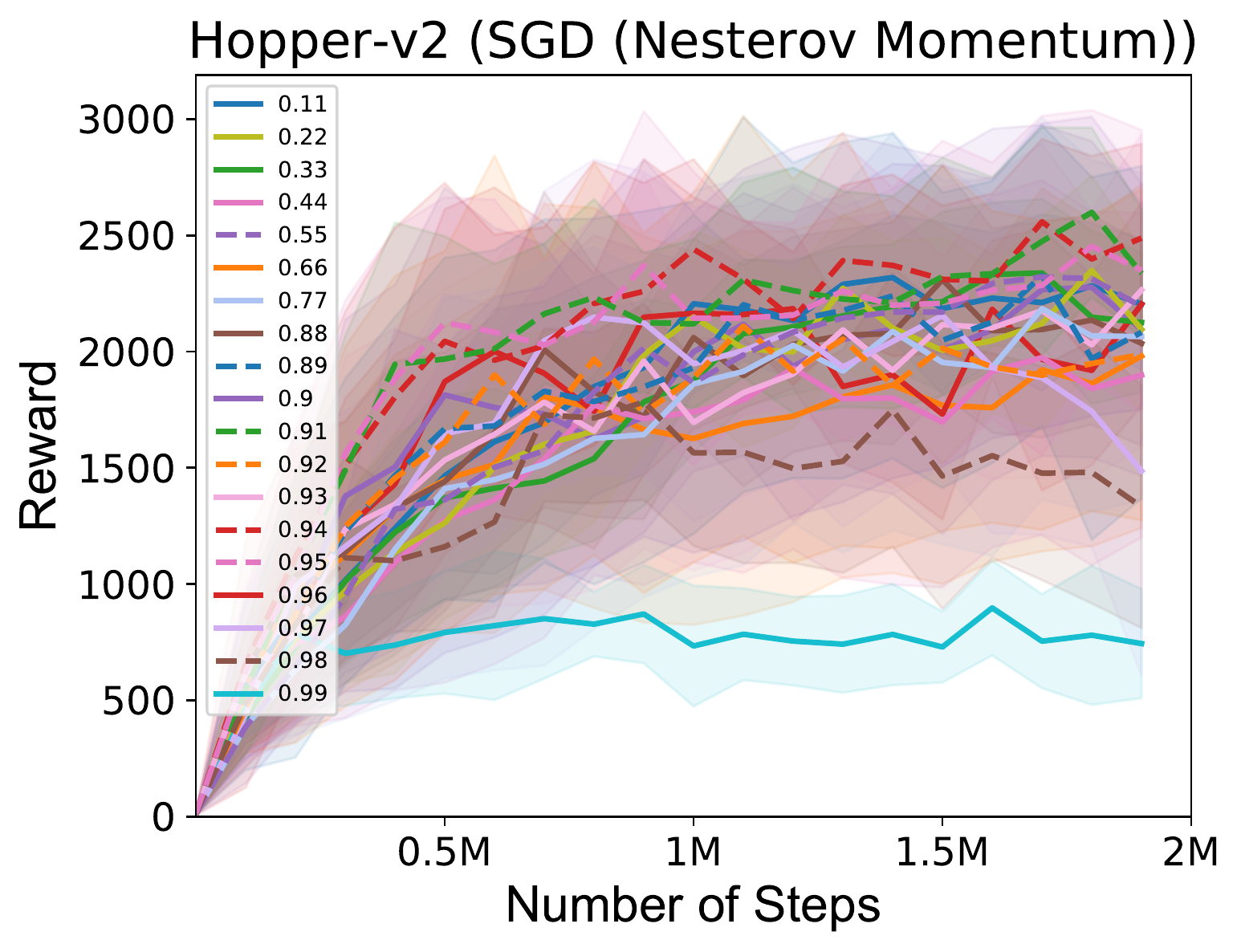}
    \includegraphics[width=.32\textwidth]{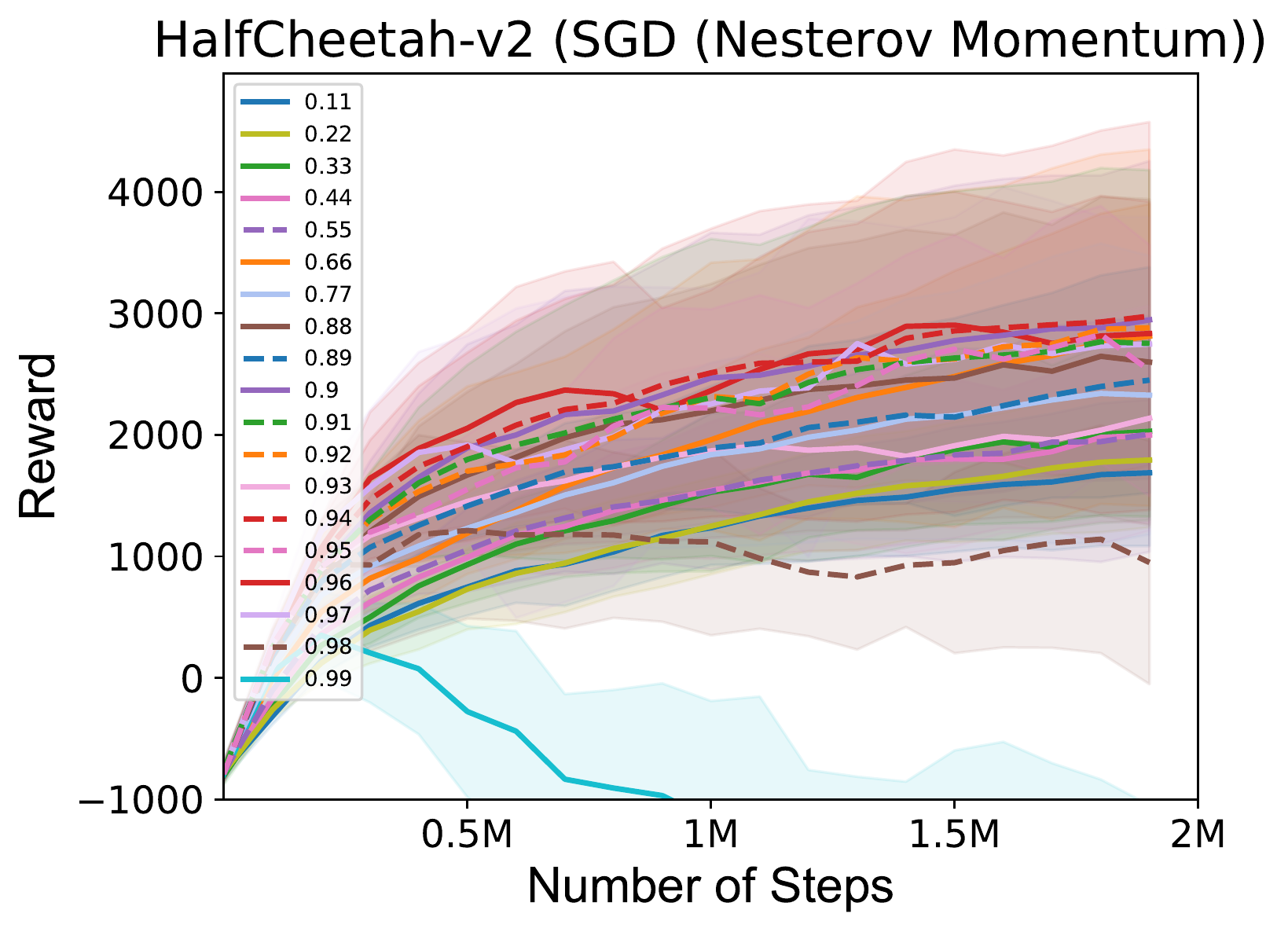}
    \includegraphics[width=.32\textwidth]{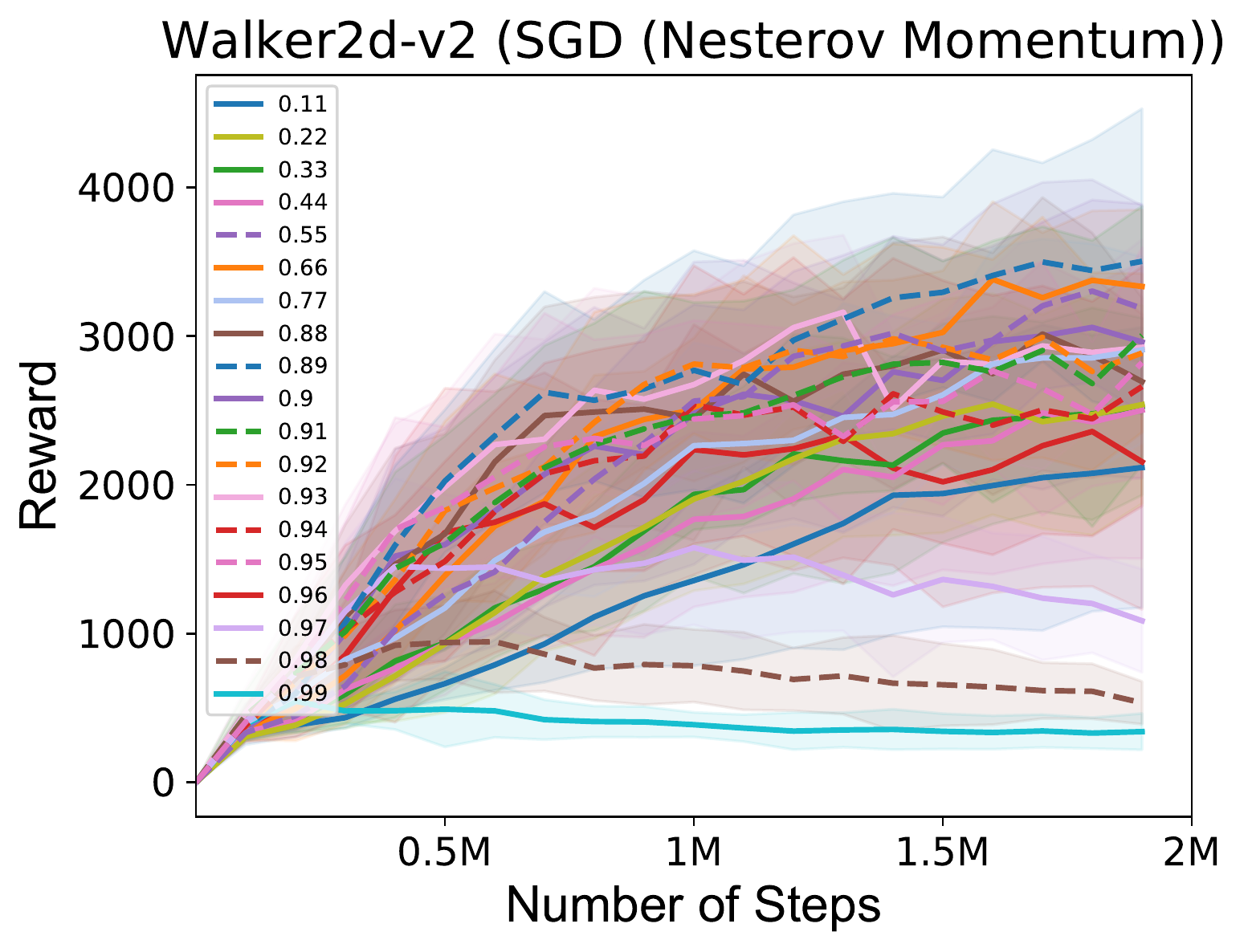}
    \includegraphics[width=.32\textwidth]{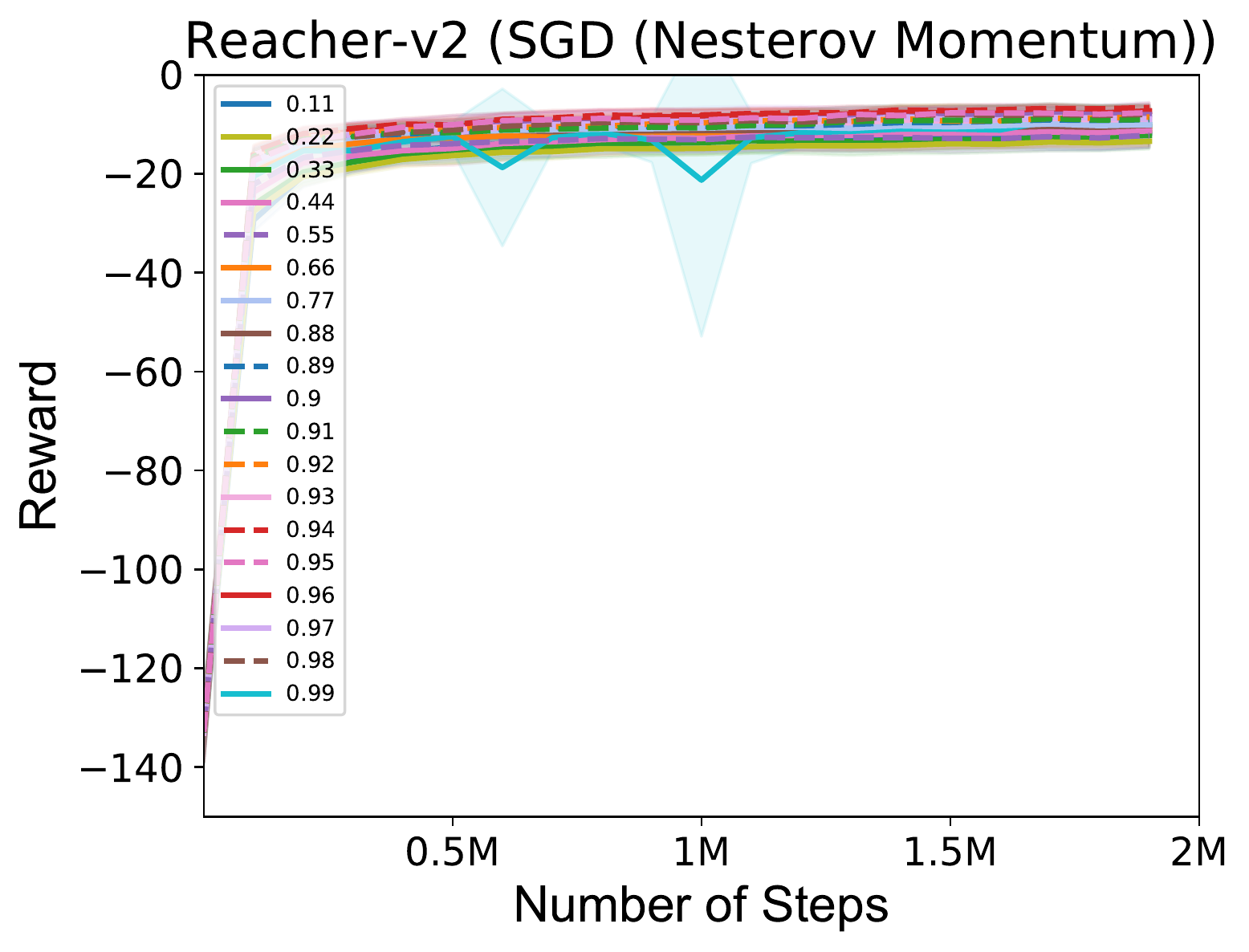}
    \caption{PPO performance across momentum values using SGD with nesterov momentum.}
\end{figure}

\subsection{Average and Asymptotic performance}
\label{app:momentumavperf}

\begin{table}[H]
\centering
\footnotesize{
\begin{tabular}{|c|c|c|c|c|c|}
\hline
 Momentum &      Ant &  HalfCheetah &  Hopper &  Reacher &  Walker2d \\
\hline
     0.11 &   -356&       -109&   566&    -39&     252\\
     0.22 &   -276&        -48&   573&    -37&     250\\
     0.33 &   -220&         42&   644&    -35&     280\\
     0.44 &   -141&        193&   557&    -32&     297\\
     0.55 &    -38&        388&   645&    -30&     379\\
     0.66 &     51&        624&   659&    -29&     453\\
     0.77 &    248&       1113&   743&    -26&     472\\
     0.88 &    642&       1234&   901&    -23&     538\\
     0.89 &    668&       1456&   801&    -23&     596\\
     0.90 &    725&       1480&  1000&    -22&     636\\
     0.91 &    773&       1535&   935&    -20&     731\\
     0.92 &    801&       1278&   827&    -21&     597\\
     0.93 &    798&       1111&  1121&    -19&     433\\
     0.94 &    815&        917&   810&    -16&     727\\
     0.95 &    833&       1629&   497&    -17&     623\\
     0.96 &    815&       1727&   458&    -14&     639\\
     0.97 &    807&        472&   642&    -22&     389\\
     0.98 &    569&       -419&   330&   -401&     279\\
     0.99 & -57391&      -3910&   292&  -3590&     151\\
\hline
\end{tabular}

}
\caption{A2C asymptotic performance across various momentum values with SGD and Nesterov Momentum. Average returns over the final 50 episodes across 10 random seeds.}\label{table:a2c-asymptotic-avg}
\end{table}

\begin{table}[H]
\centering
\small{
\begin{tabular}{|c|c|c|c|c|c|}
\hline
 Momentum &      Ant &  HalfCheetah &  Hopper &  Reacher &  Walker2d \\
\hline
     0.11 &    290&         55&   293&      3&     133\\
     0.22 &    216&         80&   302&      3&     135\\
     0.33 &    172&         87&   326&      3&     162\\
     0.44 &     97&         94&   278&      3&     117\\
     0.55 &     39&        102&   314&      3&     154\\
     0.66 &     46&        170&   399&      2&     179\\
     0.77 &     95&        420&   247&      2&     220\\
     0.88 &     56&         45&   419&      3&     225\\
     0.89 &     57&        507&   312&      2&     287\\
     0.90 &     76&        580&   387&      3&     643\\
     0.91 &     92&        690&   579&      4&     489\\
     0.92 &    121&       1043&   537&      3&     251\\
     0.93 &    106&        697&   529&      4&     180\\
     0.94 &    163&       1408&   497&      5&     306\\
     0.95 &    183&        992&   330&      5&     553\\
     0.96 &    237&       1569&   223&      5&     331\\
     0.97 &    173&        650&   300&     13&     199\\
     0.98 &    195&       1602&   347&    845&     153\\
     0.99 &  65640&       2695&   231&   6740&     133\\
\hline
\end{tabular}

}
\caption{A2C asymptotic performance across various momentum values with SGD and Nesterov Momentum. Standard Deviation returns over the final 50 episodes across 10 random seeds.}\label{table:a2c-asymptotic-std}
\end{table}

\begin{table}[H]
\centering
\small{
\begin{tabular}{|c|c|c|c|c|c|}
\hline
 Momentum &     Ant &  HalfCheetah &  Hopper &  Reacher &  Walker2d \\
\hline
     0.11 &  1152&       1737&  2434&    -12&    2284\\
     0.22 &  1758&       1844&  2276&    -13&    2691\\
     0.33 &  1699&       2102&  2170&    -12&    2878\\
     0.44 &  1546&       2119&  2120&    -11&    2609\\
     0.55 &  1746&       2026&  2422&    -12&    3340\\
     0.66 &  2297&       2970&  2114&    -11&    3525\\
     0.77 &  2509&       2436&  2173&    -10&    3176\\
     0.88 &  1923&       2730&  2218&    -10&    3236\\
     0.89 &  1669&       2439&  2186&     -9&    3492\\
     0.90 &  1564&       3004&  2312&     -7&    3027\\
     0.91 &  1414&       2907&  2568&     -9&    3203\\
     0.92 &   994&       2550&  2148&     -8&    3407\\
     0.93 &  1322&       2149&  1977&     -9&    3154\\
     0.94 &   602&       3202&  2607&     -6&    3024\\
     0.95 &   359&       2571&  2333&     -8&    2789\\
     0.96 &    81&       2830&  2327&     -7&    2057\\
     0.97 &   -70&       2593&  1680&     -8&    1283\\
     0.98 & -2159&       1286&  1627&     -8&     668\\
     0.99 & -4648&      -2468&   835&    -10&     376\\
\hline
\end{tabular}

}
\caption{PPO asymptotic performance across various momentum values with SGD and Nesterov Momentum. Average returns over the final 50 training episodes across 10 random seeds.}\label{table:ppo-asymptotic-avg}
\end{table}

\begin{table}[H]
\centering
\small{
\begin{tabular}{|c|c|c|c|c|c|}
\hline
 Momentum &     Ant &  HalfCheetah &  Hopper &  Reacher &  Walker2d \\
\hline
     0.11 &   498&        701&   712&      4&    1108\\
     0.22 &   779&        700&   772&      2&    1059\\
     0.33 &   737&        898&   684&      4&    1038\\
     0.44 &   871&        899&   867&      4&    1216\\
     0.55 &   931&        937&   811&      4&     885\\
     0.66 &   933&       1268&   855&      4&     963\\
     0.77 &  1059&       1233&   904&      4&    1096\\
     0.88 &   712&       1363&   888&      4&    1073\\
     0.89 &   877&        952&  1002&      4&    1379\\
     0.90 &   646&       1402&   963&      3&    1301\\
     0.91 &   805&       1542&   741&      4&    1297\\
     0.92 &   574&       1522&   966&      3&    1220\\
     0.93 &   792&        933&   895&      4&    1103\\
     0.94 &   404&       1648&   707&      2&    1293\\
     0.95 &   240&       1190&   885&      4&    1340\\
     0.96 &   159&       1188&   771&      3&    1178\\
     0.97 &   139&       1369&  1113&      5&     617\\
     0.98 &  1757&       1428&   879&      4&     323\\
     0.99 &  4672&       1162&   370&      4&     184\\
\hline
\end{tabular}

}
\caption{PPO asymptotic performance across various momentum values with SGD and Nesterov Momentum. Standard Deviation returns over the final 50 training episodes across 10 random seeds.}\label{table:ppo-asymptotic-std}
\end{table}

\begin{table}[H]
\centering
\small{
\begin{tabular}{|c|c|c|c|c|c|}
\hline
 Momentum &     Ant &  HalfCheetah &  Hopper &  Reacher &  Walker2d \\
\hline
     0.11 &   343&       1058&  1883&    -15&    1344\\
     0.22 &   489&       1088&  1801&    -17&    1764\\
     0.33 &   610&       1293&  1762&    -16&    1740\\
     0.44 &   622&       1323&  1587&    -15&    1639\\
     0.55 &   718&       1348&  1820&    -15&    2226\\
     0.66 &   923&       1769&  1658&    -13&    2363\\
     0.77 &  1021&       1612&  1680&    -12&    2050\\
     0.88 &   899&       1951&  1868&    -12&    2346\\
     0.89 &   817&       1708&  1879&    -11&    2607\\
     0.90 &   847&       2150&  1852&    -10&    2315\\
     0.91 &   773&       2040&  2135&    -11&    2327\\
     0.92 &   681&       2028&  1795&    -11&    2454\\
     0.93 &   713&       1603&  1807&    -11&    2514\\
     0.94 &   466&       2212&  2164&     -9&    2155\\
     0.95 &   348&       1951&  2119&    -10&    2318\\
     0.96 &   127&       2232&  1936&    -10&    1968\\
     0.97 &   -15&       2077&  1818&    -11&    1403\\
     0.98 &  -893&        970&  1519&    -11&     771\\
     0.99 & -3444&      -1096&   812&    -14&     422\\
\hline
\end{tabular}

}
\caption{PPO average performance across various momentum values with SGD and Nesterov Momentum. Average returns over all training episodes over 10 random seeds.}\label{table:ppo-average-avg}
\end{table}

\begin{table}[H]
\centering
\small{
\begin{tabular}{|c|c|c|c|c|c|}
\hline
 Momentum &     Ant &  HalfCheetah &  Hopper &  Reacher &  Walker2d \\
\hline
     0.11 &   332&        422&   597&      4&     645\\
     0.22 &   360&        464&   629&      3&     722\\
     0.33 &   400&        591&   547&      4&     687\\
     0.44 &   407&        556&   795&      4&     720\\
     0.55 &   442&        650&   789&      4&     713\\
     0.66 &   448&        651&   885&      4&     720\\
     0.77 &   484&        724&   844&      4&     764\\
     0.88 &   470&        955&   746&      4&     933\\
     0.89 &   461&        616&   881&      4&     922\\
     0.90 &   387&       1027&   824&      4&    1070\\
     0.91 &   468&       1077&   725&      4&     971\\
     0.92 &   383&       1054&   932&      4&     907\\
     0.93 &   453&        688&   845&      4&     955\\
     0.94 &   310&       1108&   753&      3&    1017\\
     0.95 &   241&        805&   785&      4&     964\\
     0.96 &   177&        981&   774&      3&    1061\\
     0.97 &   142&       1080&   859&      4&     743\\
     0.98 &   808&        795&   680&      4&     394\\
     0.99 &  3105&        933&   356&      8&     191\\
\hline
\end{tabular}

}
\caption{PPO average performance across various momentum values with SGD and Nesterov Momentum. Standard Deviation returns over all training episodes over 10 random seeds.}\label{table:ppo-average-std}
\end{table}

\begin{table}[H]
\centering
\small{
\begin{tabular}{|c|c|c|c|c|c|}
\hline
 Momentum &      Ant &  HalfCheetah &  Hopper &  Reacher &  Walker2d \\
\hline
     0.11 &   -792&       -408&   389&    -55&     205\\
     0.22 &   -734&       -368&   406&    -53&     208\\
     0.33 &   -666&       -321&   480&    -50&     214\\
     0.44 &   -577&       -244&   456&    -47&     224\\
     0.55 &   -483&       -139&   566&    -44&     253\\
     0.66 &   -366&         13&   514&    -41&     296\\
     0.77 &   -199&        284&   561&    -36&     334\\
     0.88 &    127&        638&   666&    -30&     441\\
     0.89 &    173&        734&   704&    -29&     423\\
     0.90 &    216&        769&   766&    -29&     472\\
     0.91 &    271&        834&   710&    -28&     519\\
     0.92 &    326&        885&   648&    -26&     524\\
     0.93 &    381&        885&   820&    -27&     458\\
     0.94 &    438&        920&   655&    -24&     565\\
     0.95 &    499&       1083&   600&    -23&     469\\
     0.96 &    565&       1057&   405&    -20&     402\\
     0.97 &    590&        411&   513&    -21&     428\\
     0.98 &    582&        -28&   241&   -111&     286\\
     0.99 & -14693&      -1920&   266&  -1981&     166\\
\hline
\end{tabular}

}
\caption{A2C average performance across various momentum values with SGD and Nesterov Momentum. Average returns over all training episodes over 10 random seeds.}\label{table:a2c-average-avg}
\end{table}

\begin{table}[H]
\centering
\small{
\begin{tabular}{|c|c|c|c|c|c|}
\hline
 Momentum &      Ant &  HalfCheetah &  Hopper &  Reacher &  Walker2d \\
\hline
     0.11 &    648&         52&   234&      6&     119\\
     0.22 &    598&         56&   242&      6&     128\\
     0.33 &    541&         60&   284&      6&     123\\
     0.44 &    472&         61&   258&      5&     122\\
     0.55 &    395&         73&   294&      5&     134\\
     0.66 &    307&        100&   286&      4&     144\\
     0.77 &    236&        135&   275&      4&     167\\
     0.88 &    162&        120&   307&      3&     208\\
     0.89 &    150&        213&   296&      3&     217\\
     0.90 &    147&        298&   309&      4&     323\\
     0.91 &    138&        335&   449&      4&     273\\
     0.92 &    134&        481&   347&      4&     266\\
     0.93 &    135&        233&   394&      4&     262\\
     0.94 &    135&        464&   406&      4&     267\\
     0.95 &    136&        622&   373&      5&     264\\
     0.96 &    142&        798&   258&      5&     196\\
     0.97 &    144&        610&   298&     10&     232\\
     0.98 &    155&        777&   189&    240&     211\\
     0.99 &  18727&       1244&   265&   3936&     215\\
\hline
\end{tabular}

}
\caption{A2C average performance across various momentum values with SGD and Nesterov Momentum. Standard Deviation returns over all training episodes over 10 random seeds.}\label{table:a2c-average-std}
\end{table}

\section{Step Experiments}
\label{app:steps}

To probe the effects of step size and number of workers -- that is, the number of separate environment instance using the same policy to take steps in parallel -- we setup several experiments which use a grid of momentum values to decrease number of steps while increasing workers such that the total batch size per update remains the same as the default used in other experiments. For PPO we use: 2048 steps, 1 worker (the default); 1024 steps, 2 workers; 256 steps, 8 workers; 64 steps, 32 workers. For A2C we run: 80 steps, 1 worker; 40 steps, 2 workers; 5 steps, 16 workers (the default); 2 steps; 40 workers.

\subsection{A2C}

\begin{table}[H]
\centering
\small{
\begin{tabular}{|c|c|c|c|c|c|}
\hline
 Momentum &      Ant &  HalfCheetah &  Hopper &  Reacher &  Walker2d \\
\hline
     0.11 &   -632&       -502&   381&    -48&     176\\
     0.22 &   -588&       -471&   395&    -46&     181\\
     0.33 &   -531&       -429&   496&    -44&     185\\
     0.44 &   -465&       -395&   464&    -42&     185\\
     0.55 &   -378&       -333&   556&    -39&     227\\
     0.66 &   -277&       -259&   589&    -36&     232\\
     0.77 &   -119&        -99&   569&    -32&     318\\
     0.88 &    187&        -46&   682&    -26&     483\\
     0.99 & -13585&      -1922&    92&   -422&      65\\
\hline
\end{tabular}

}
\caption{A2C average performance across various momentum values with SGD and Nesterov Momentum. Average returns over all training episodes over 10 random seeds. 2 Steps, 40 workers. }\label{table:a2c-average-avg-steps-1}
\end{table}

\begin{table}[H]
\centering
\small{
\begin{tabular}{|c|c|c|c|c|c|}
\hline
 Momentum &      Ant &  HalfCheetah &  Hopper &  Reacher &  Walker2d \\
\hline
     0.11 &  -1142&       -451&   331&    -67&     220\\
     0.22 &  -1107&       -401&   348&    -63&     236\\
     0.33 &  -1054&       -328&   420&    -59&     249\\
     0.44 &   -966&       -242&   413&    -54&     264\\
     0.55 &   -874&       -147&   483&    -49&     297\\
     0.66 &   -712&         56&   555&    -42&     345\\
     0.77 &   -486&        332&   749&    -36&     466\\
     0.88 &    -94&        916&   954&    -29&     764\\
     0.99 & -20523&      -2876&   375&  -5930&     167\\
\hline
\end{tabular}

}
\caption{A2C average performance across various momentum values with SGD and Nesterov Momentum. Average returns over all training episodes over 10 random seeds. 20 steps. 4 workers.}\label{table:a2c-average-avg-steps-2}
\end{table}

\begin{table}[H]
\centering
\small{
\begin{tabular}{|c|c|c|c|c|c|}
\hline
 Momentum &     Ant &  HalfCheetah &  Hopper &  Reacher &  Walker2d \\
\hline
     0.11 & -1228&       -515&   300&    -67&     206\\
     0.22 & -1190&       -476&   315&    -62&     222\\
     0.33 & -1147&       -417&   341&    -58&     239\\
     0.44 & -1073&       -346&   365&    -52&     255\\
     0.55 &  -975&       -245&   421&    -46&     284\\
     0.66 &  -819&        -92&   505&    -39&     322\\
     0.77 &  -586&        148&   669&    -31&     414\\
     0.88 &  -189&        662&   938&    -21&     659\\
     0.99 & -9980&      -3867&   554& -12501&     194\\
\hline
\end{tabular}

}
\caption{A2C average performance across various momentum values with SGD and Nesterov Momentum. Average returns over all training episodes over 10 random seeds. 40 steps. 2 workers.}\label{table:a2c-average-avg-steps-3}
\end{table}

\begin{table}[H]
\centering
\small{
\begin{tabular}{|c|c|c|c|c|c|}
\hline
 Momentum &     Ant &  HalfCheetah &  Hopper &  Reacher &  Walker2d \\
\hline
     0.11 & -1237&       -581&   275&    -68&     201\\
     0.22 & -1205&       -545&   291&    -64&     220\\
     0.33 & -1147&       -503&   309&    -58&     233\\
     0.44 & -1088&       -446&   334&    -53&     261\\
     0.55 &  -977&       -352&   384&    -47&     281\\
     0.66 &  -819&       -224&   459&    -39&     313\\
     0.77 &  -593&         11&   597&    -31&     371\\
     0.88 &  -218&        496&   832&    -21&     519\\
     0.99 & -9418&      -5054&   620& -10784&     212\\
\hline
\end{tabular}

}
\caption{A2C average performance across various momentum values with SGD and Nesterov Momentum. Average returns over all training episodes over 10 random seeds. 80 steps. 1 worker.}\label{table:a2c-average-avg-steps-4}
\end{table}

\begin{table}[H]
\centering
\small{
\begin{tabular}{|c|c|c|c|c|c|}
\hline
 Momentum &      Ant &  HalfCheetah &  Hopper &  Reacher &  Walker2d \\
\hline
     0.11 &    485&         89&   250&      6&     111\\
     0.22 &    451&         89&   264&      6&     112\\
     0.33 &    402&         93&   298&      5&     114\\
     0.44 &    351&         94&   275&      5&     115\\
     0.55 &    288&        105&   281&      5&     170\\
     0.66 &    231&        143&   277&      4&     150\\
     0.77 &    187&        271&   285&      4&     174\\
     0.88 &    120&        227&   309&      3&     257\\
     0.99 &  17529&       1796&   110&    689&     116\\
\hline
\end{tabular}

}
\caption{A2C average performance across various momentum values with SGD and Nesterov Momentum. Standard Deviation returns over all training episodes over 10 random seeds. 2 steps. 40 workers. }\label{table:a2c-average-std-steps-1}
\end{table}

\begin{table}[H]
\centering
\small{
\begin{tabular}{|c|c|c|c|c|c|}
\hline
 Momentum &      Ant &  HalfCheetah &  Hopper &  Reacher &  Walker2d \\
\hline
     0.11 &    948&         66&   171&      8&      90\\
     0.22 &    914&         70&   181&      8&      91\\
     0.33 &    881&         82&   223&      7&      92\\
     0.44 &    817&         84&   202&      6&      93\\
     0.55 &    740&        132&   216&      6&      96\\
     0.66 &    609&        141&   217&      5&     115\\
     0.77 &    425&        151&   280&      4&     231\\
     0.88 &    289&        407&   326&      3&     350\\
     0.99 &  27919&       1984&   273&  11481&     162\\
\hline
\end{tabular}

}
\caption{A2C average performance across various momentum values with SGD and Nesterov Momentum. Standard Deviation returns over all training episodes over 10 random seeds. 20 steps. 4 workers.}\label{table:a2c-average-std-steps-2}
\end{table}

\begin{table}[H]
\centering
\small{
\begin{tabular}{|c|c|c|c|c|c|}
\hline
 Momentum &      Ant &  HalfCheetah &  Hopper &  Reacher &  Walker2d \\
\hline
     0.11 &   1007&         71&   146&      8&      81\\
     0.22 &    979&         75&   150&      8&      85\\
     0.33 &    947&         86&   161&      7&      85\\
     0.44 &    891&        100&   164&      6&      83\\
     0.55 &    813&        122&   173&      6&      91\\
     0.66 &    682&        165&   164&      5&      96\\
     0.77 &    498&        226&   228&      4&     138\\
     0.88 &    316&        420&   306&      4&     253\\
     0.99 &  12159&       4360&   381&  18190&     147\\
\hline
\end{tabular}

}
\caption{A2C average performance across various momentum values with SGD and Nesterov Momentum. Standard Deviation returns over all training episodes over 10 random seeds. 40 steps. 2 workers.}\label{table:a2c-average-std-steps-3}
\end{table}

\begin{table}[H]
\centering
\small{
\begin{tabular}{|c|c|c|c|c|c|}
\hline
 Momentum &      Ant &  HalfCheetah &  Hopper &  Reacher &  Walker2d \\
\hline
     0.11 &   1018&         72&   110&      8&      82\\
     0.22 &    987&         72&   114&      8&      84\\
     0.33 &    946&         78&   117&      7&      84\\
     0.44 &    888&         82&   126&      6&      86\\
     0.55 &    800&         98&   135&      6&      83\\
     0.66 &    678&        134&   150&      5&      87\\
     0.77 &    491&        181&   204&      4&      95\\
     0.88 &    305&        282&   291&      4&     129\\
     0.99 &  12423&       3515&   409&  15905&     171\\
\hline
\end{tabular}

}
\caption{A2C average performance across various momentum values with SGD and Nesterov Momentum. Standard Deviation returns over all training episodes over 10 random seeds. 80 steps. 1 worker.}\label{table:a2c-average-std-steps-4}
\end{table}

\subsection{PPO}

\begin{table}[H]
\centering
\small{
\begin{tabular}{|c|c|c|c|c|c|}
\hline
 Momentum &     Ant &  HalfCheetah &  Hopper &  Reacher &  Walker2d \\
\hline
     0.11 &   422&       1011&  1675&    -17&    1627\\
     0.22 &   471&       1195&  1805&    -16&    1652\\
     0.33 &   563&       1187&  1700&    -17&    1838\\
     0.44 &   689&       1428&  1857&    -16&    1786\\
     0.55 &   772&       1505&  1955&    -15&    1846\\
     0.66 &   746&       1453&  1814&    -15&    2213\\
     0.77 &   993&       1842&  1920&    -12&    2468\\
     0.88 &   880&       1659&  1760&    -12&    2335\\
     0.99 & -3157&       -930&   910&    -13&     400\\
\hline
\end{tabular}

}
\caption{PPO average performance across various momentum values with SGD and Nesterov Momentum. Average returns over all training episodes over 10 random seeds. 1024 steps. 2 workers.}\label{table:ppo-average-avg-steps-1}
\end{table}

\begin{table}[H]
\centering
\small{
\begin{tabular}{|c|c|c|c|c|c|}
\hline
 Momentum &     Ant &  HalfCheetah &  Hopper &  Reacher &  Walker2d \\
\hline
     0.11 &   348&       1195&  1733&    -18&    1590\\
     0.22 &   493&       1089&  1736&    -18&    1612\\
     0.33 &   550&       1326&  1844&    -17&    1594\\
     0.44 &   701&       1374&  1919&    -17&    1919\\
     0.55 &   824&       1526&  1751&    -16&    2158\\
     0.66 &   765&       1840&  1728&    -15&    1917\\
     0.77 &   844&       2210&  1828&    -14&    2407\\
     0.88 &   895&       2113&  1755&    -13&    2132\\
     0.99 & -2694&      -1069&   912&    -14&     416\\
\hline
\end{tabular}

}
\caption{PPO average performance across various momentum values with SGD and Nesterov Momentum. Average returns over all training episodes over 10 random seeds. 256 steps. 8 workers.}\label{table:ppo-average-avg-steps-2}
\end{table}

\begin{table}[H]
\centering
\small{
\begin{tabular}{|c|c|c|c|c|c|}
\hline
 Momentum &     Ant &  HalfCheetah &  Hopper &  Reacher &  Walker2d \\
\hline
     0.11 &   379&       1249&  1436&    -19&    1294\\
     0.22 &   408&       1280&  1374&    -19&    1603\\
     0.33 &   484&       1353&  1559&    -18&    1499\\
     0.44 &   613&       1244&  1408&    -17&    1898\\
     0.55 &   738&       1225&  1687&    -16&    1827\\
     0.66 &   801&       1313&  1767&    -16&    1856\\
     0.77 &   752&       1824&  1288&    -15&    2072\\
     0.88 &   766&       2128&  1356&    -14&    2372\\
     0.99 & -3164&       -739&   882&    -15&     452\\
\hline
\end{tabular}

}
\caption{PPO average performance across various momentum values with SGD and Nesterov Momentum. Average returns over all training episodes over 10 random seeds. 64 steps. 32 workers.}\label{table:ppo-average-avg-steps-3}
\end{table}

\begin{table}[H]
\centering
\small{
\begin{tabular}{|c|c|c|c|c|c|}
\hline
 Momentum &     Ant &  HalfCheetah &  Hopper &  Reacher &  Walker2d \\
\hline
     0.11 &   348&        471&   619&      3&     625\\
     0.22 &   354&        621&   655&      3&     718\\
     0.33 &   364&        462&   710&      3&     755\\
     0.44 &   373&        721&   696&      4&     732\\
     0.55 &   423&        670&   776&      3&     677\\
     0.66 &   378&        643&   857&      3&     799\\
     0.77 &   508&        910&   767&      4&     843\\
     0.88 &   409&        985&   780&      4&     920\\
     0.99 &  3065&        933&   530&      7&     157\\
\hline
\end{tabular}

}
\caption{PPO average performance across various momentum values with SGD and Nesterov Momentum. Standard Deviation returns over all training episodes over 10 random seeds. 1024 steps. 2 workers.}\label{table:ppo-average-std-steps-1}
\end{table}

\begin{table}[H]
\centering
\small{
\begin{tabular}{|c|c|c|c|c|c|}
\hline
 Momentum &     Ant &  HalfCheetah &  Hopper &  Reacher &  Walker2d \\
\hline
     0.11 &   290&        576&   630&      3&     726\\
     0.22 &   348&        398&   711&      3&     675\\
     0.33 &   381&        566&   651&      3&     756\\
     0.44 &   413&        535&   709&      2&     787\\
     0.55 &   434&        669&   835&      3&     728\\
     0.66 &   399&        833&   774&      4&    1049\\
     0.77 &   425&       1004&   798&      4&     820\\
     0.88 &   487&        903&   666&      4&     970\\
     0.99 &  2642&        612&   479&      7&     171\\
\hline
\end{tabular}

}
\caption{PPO average performance across various momentum values with SGD and Nesterov Momentum. Standard Deviation returns over all training episodes over 10 random seeds. 256 steps. 8 workers.}\label{table:ppo-average-std-steps-2}
\end{table}

\begin{table}[H]
\centering
\small{
\begin{tabular}{|c|c|c|c|c|c|}
\hline
 Momentum &     Ant &  HalfCheetah &  Hopper &  Reacher &  Walker2d \\
\hline
     0.11 &   275&        522&   589&      3&     551\\
     0.22 &   277&        537&   781&      3&     689\\
     0.33 &   299&        618&   812&      3&     734\\
     0.44 &   366&        538&   710&      4&     723\\
     0.55 &   409&        503&   803&      4&     736\\
     0.66 &   377&        511&   862&      4&     900\\
     0.77 &   413&        824&   783&      4&     856\\
     0.88 &   379&        980&   821&      4&     868\\
     0.99 &  2833&        497&   455&      8&     209\\
\hline
\end{tabular}

}
\caption{PPO average performance across various momentum values with SGD and Nesterov Momentum. Standard Deviation returns over all training episodes over 10 random seeds. 64 steps. 32 workers.}\label{table:ppo-average-std-steps-3}
\end{table}

\subsection{Normalized Plots}
\label{app:momentum-step-graphs}
\begin{figure}[H]
    \centering
        \includegraphics[width=.49\textwidth]{ppofull}
        \includegraphics[width=.49\textwidth]{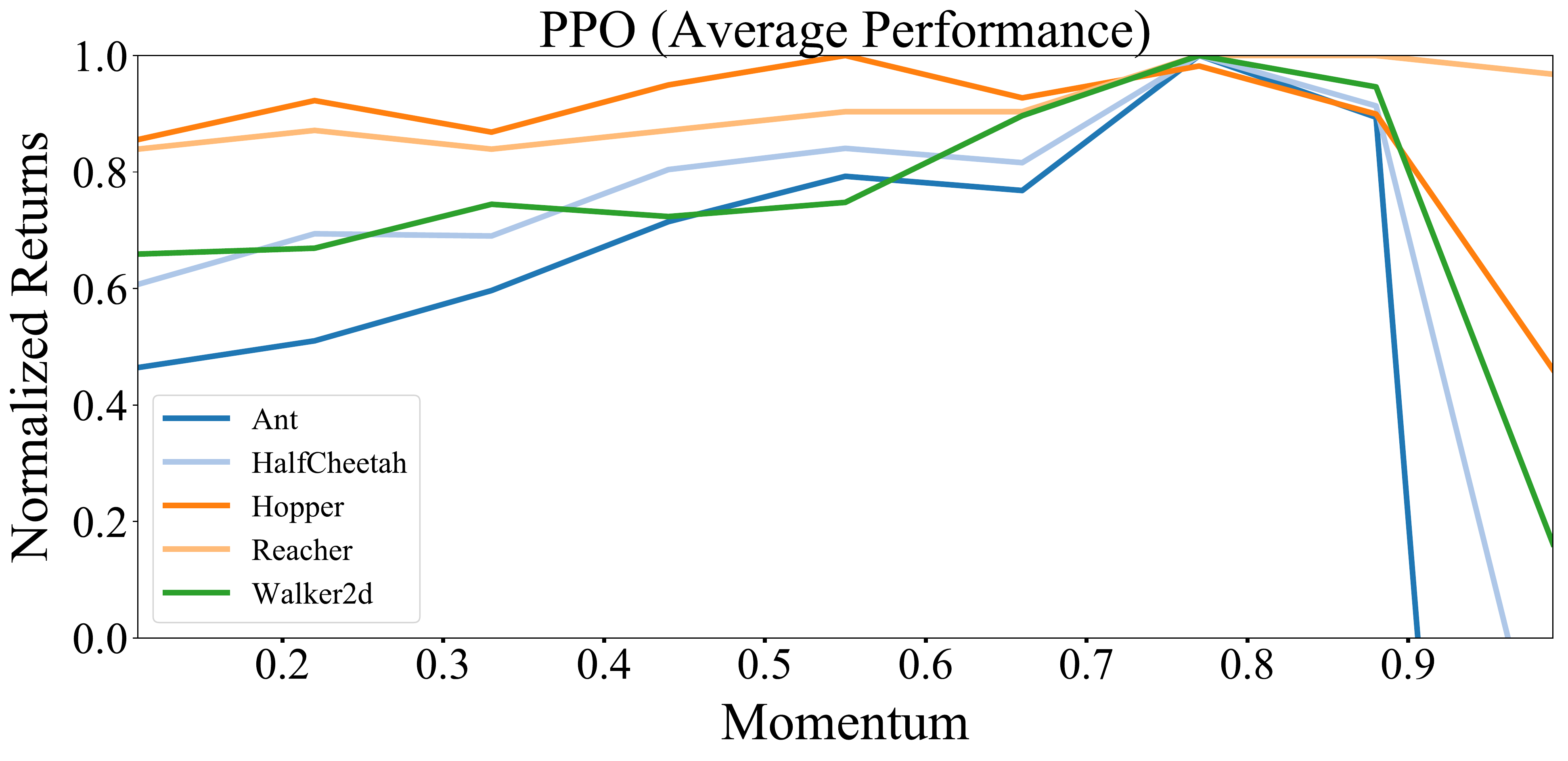}
        \includegraphics[width=.49\textwidth]{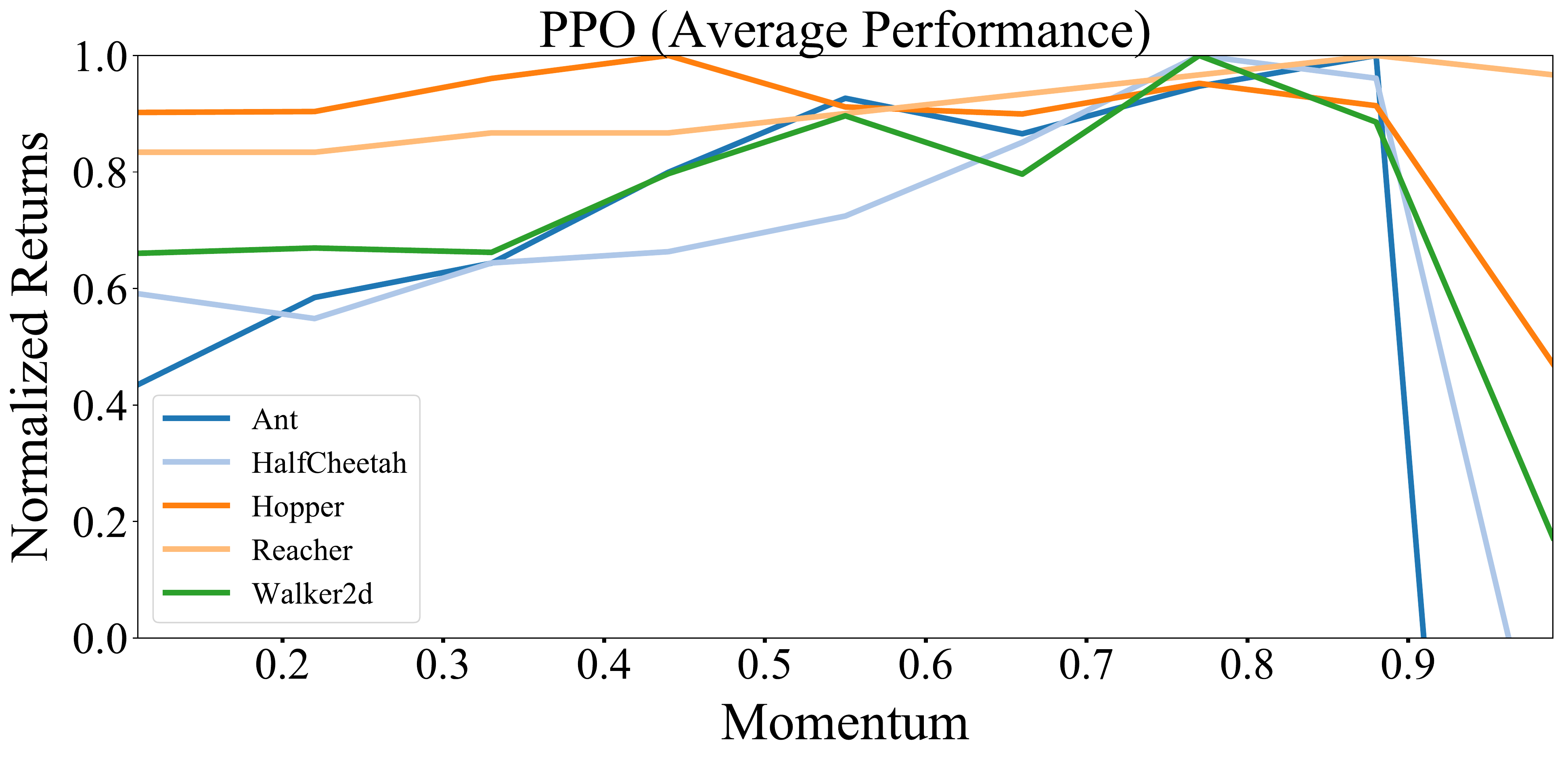}
        \includegraphics[width=.49\textwidth]{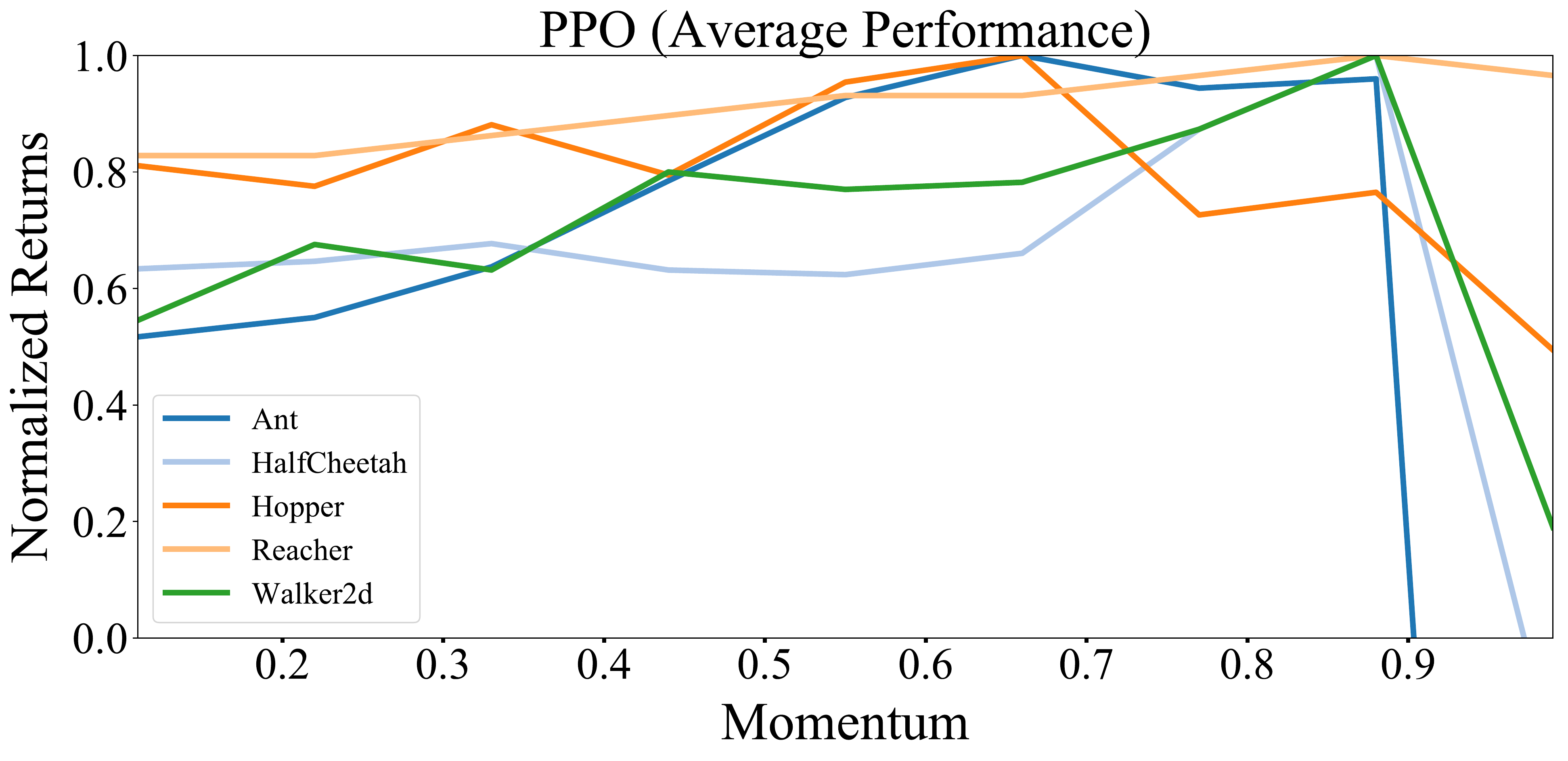}
    \caption{Normalized performance of PPO across momentum factors in different environments. Normalization is per environment using a random agent policy (see Appendix~\ref{app:random}) such that the Normalized Return corresponds to $\frac{\text{Average Return} - \text{Random Agent}}{\text{Best Average Return} - \text{Random Agent}}$. 1 worker, 2048 steps (top-left). 1 worker, 2048 steps (top-left). 2 workers, 1024 steps (top-right). 8 workers, 256 steps (bottom-left). 32 workers, 64 steps (bottom-right).}
    \label{fig:momentums-steps-ppo}
\end{figure}

\begin{figure}[H]
    \centering
        \includegraphics[width=.49\textwidth]{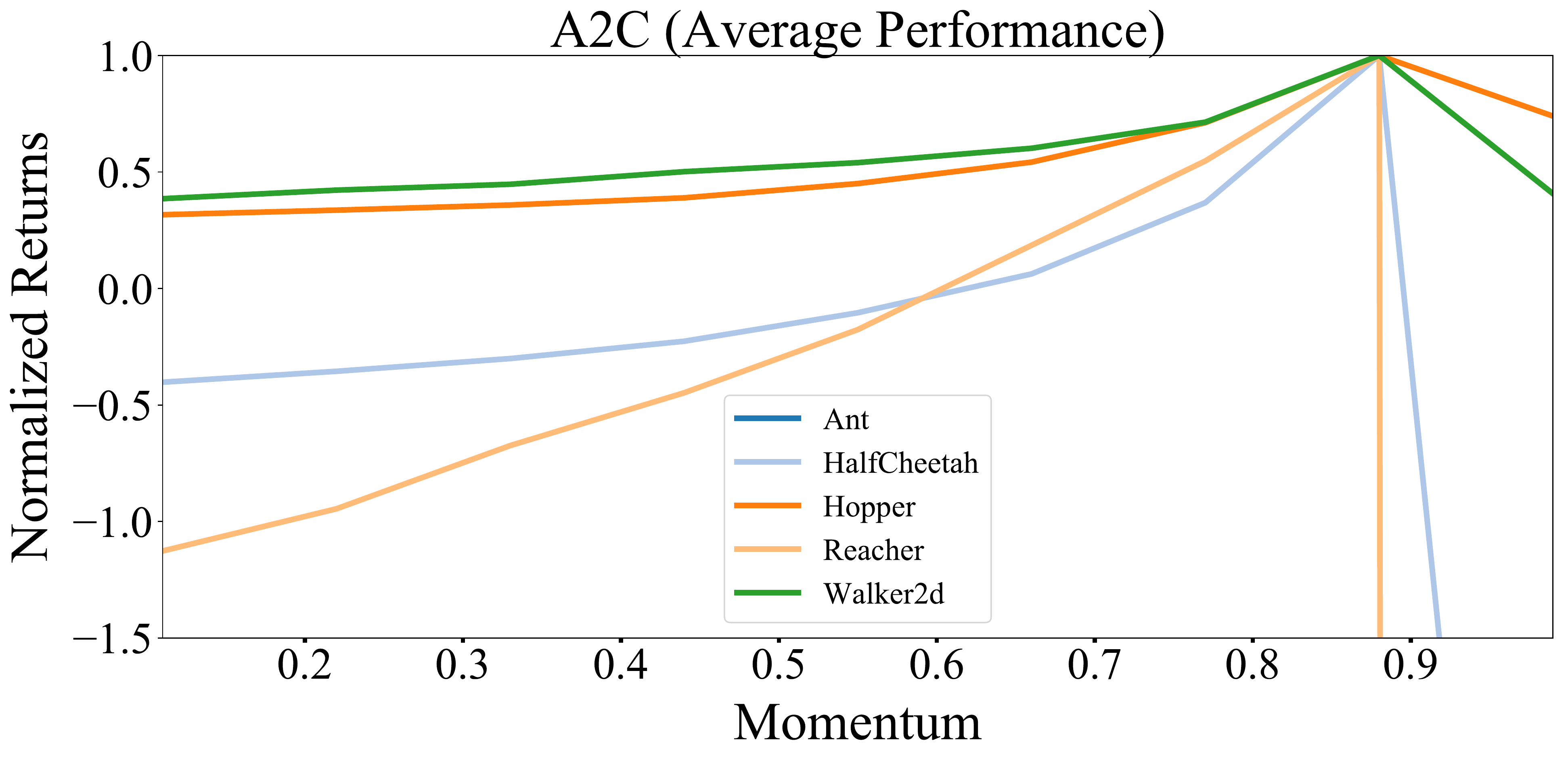}
        \includegraphics[width=.49\textwidth]{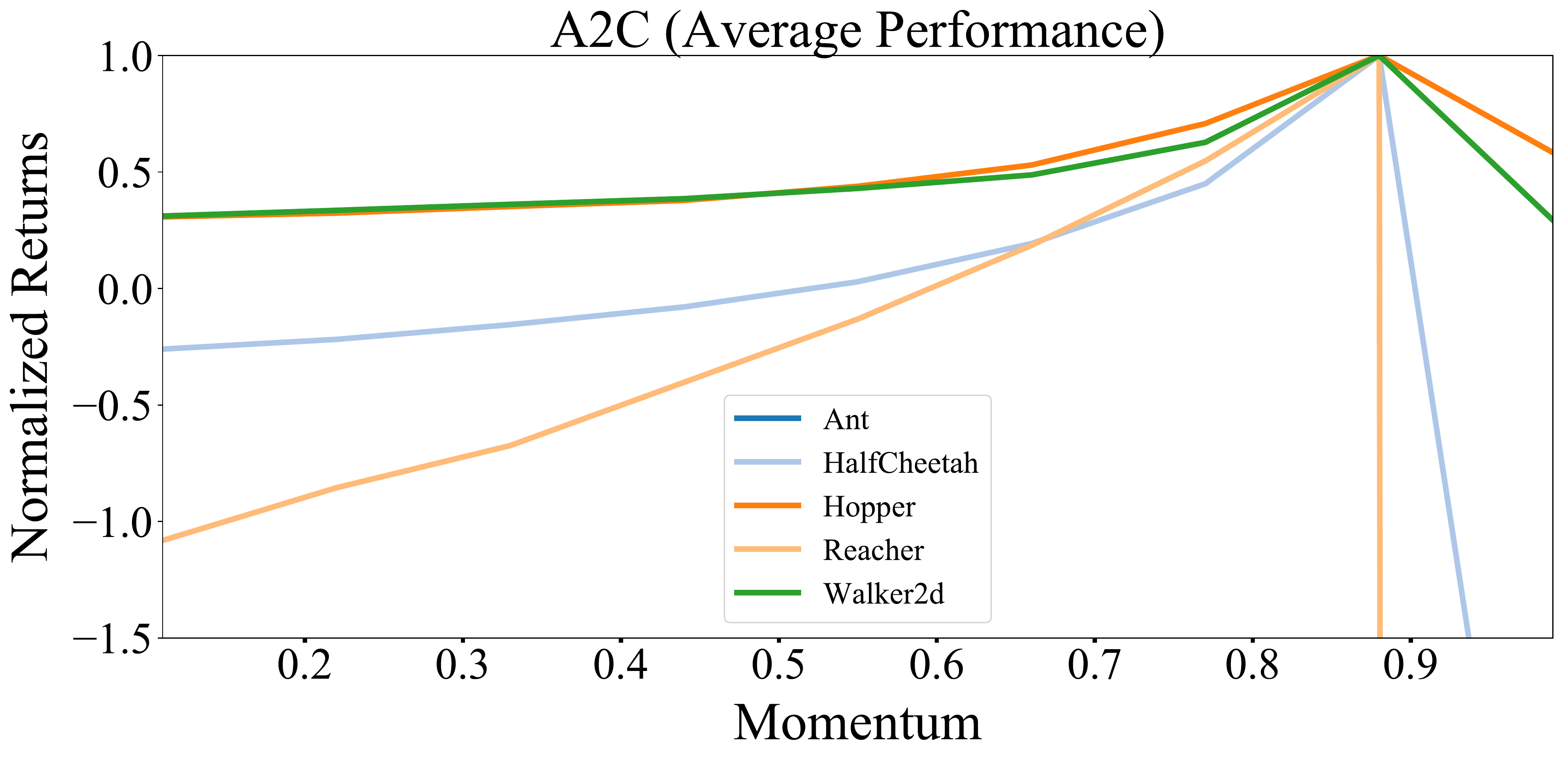}
        \includegraphics[width=.49\textwidth]{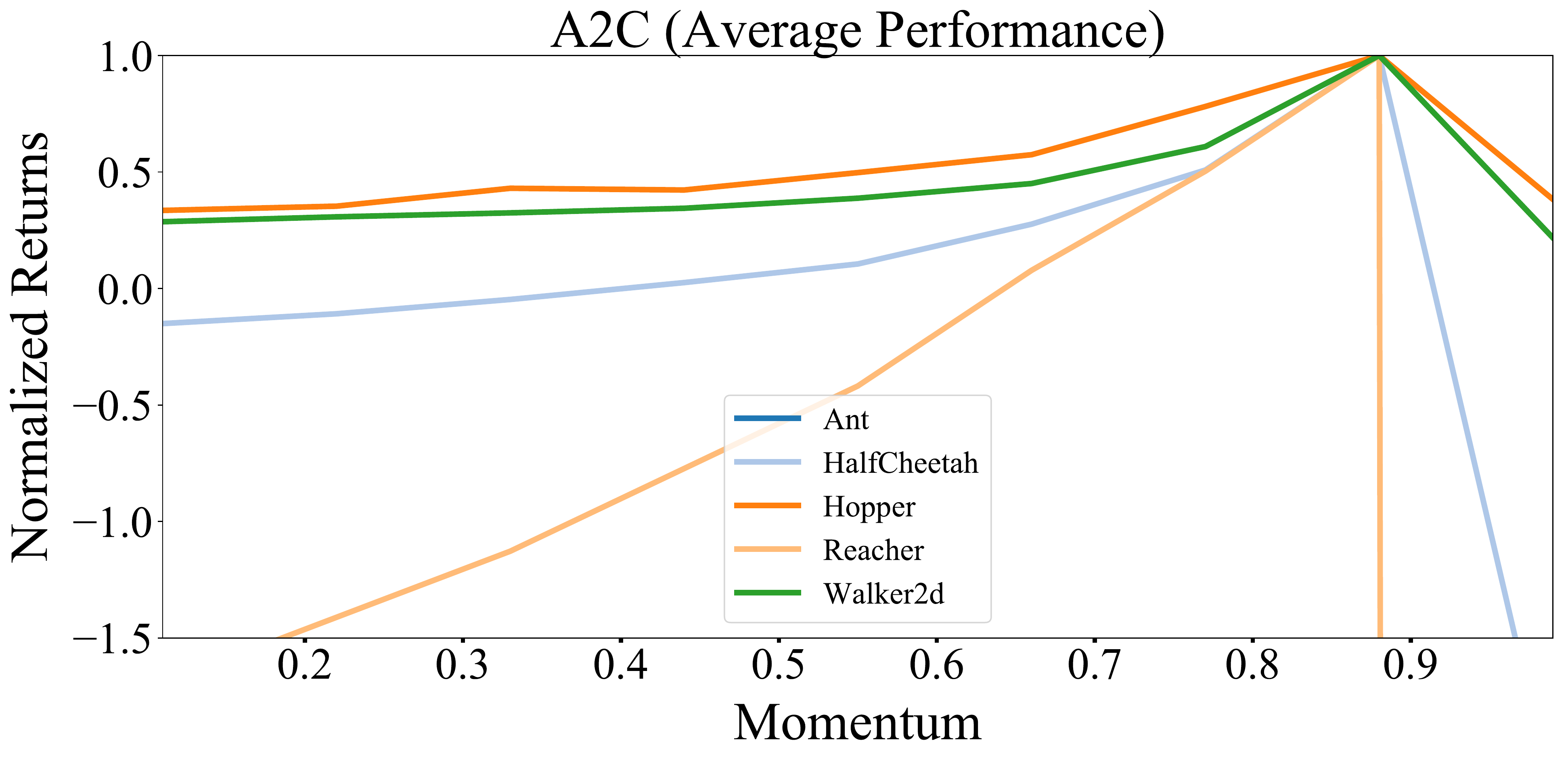}
        \includegraphics[width=.49\textwidth]{a2cfull}
        \includegraphics[width=.49\textwidth]{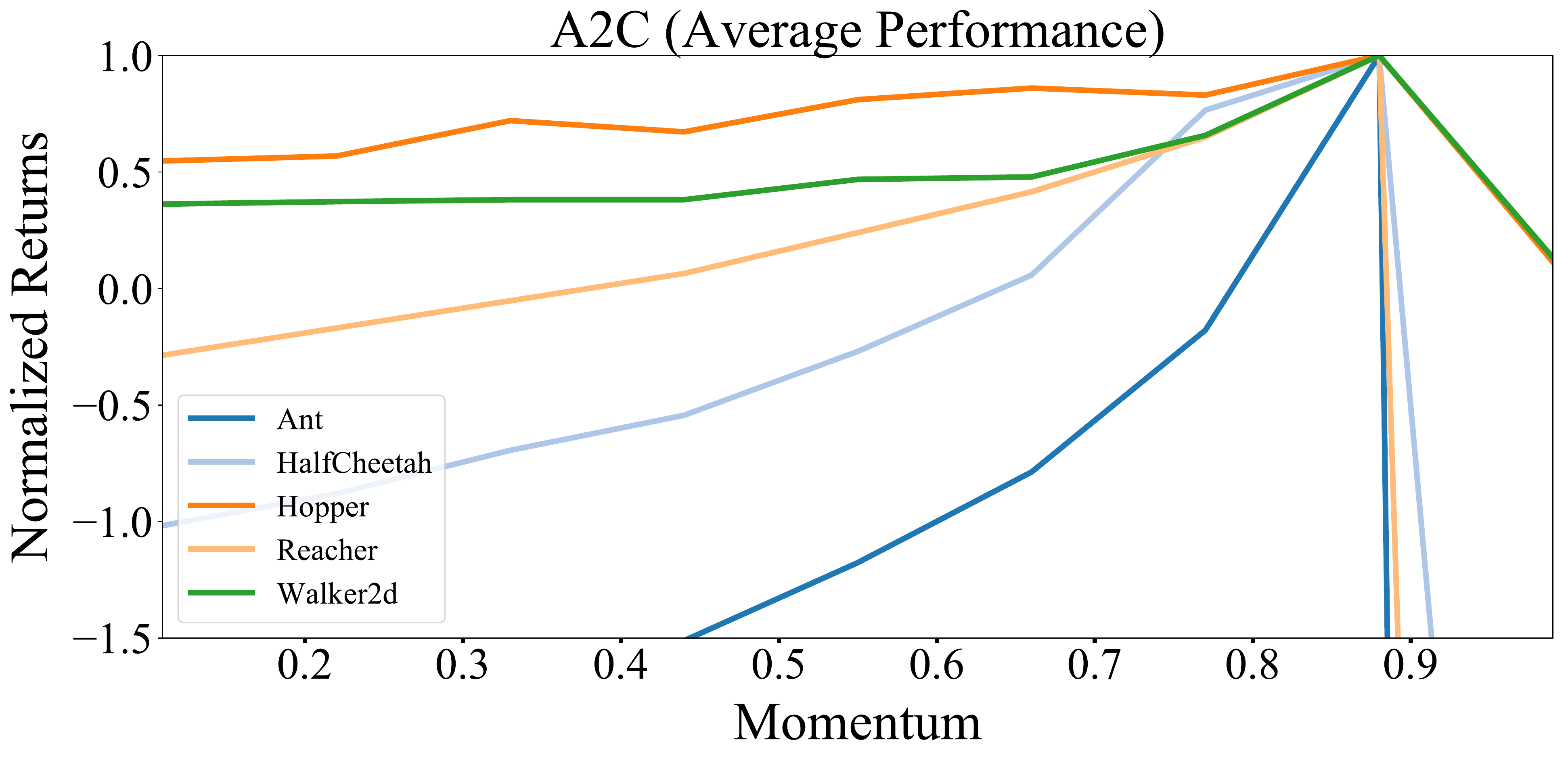}

    \caption{Normalized performance of A2C across momentum factors in different environments. Normalization is per environment using a random agent policy (see Appendix~\ref{app:random}) such that the Normalized Return corresponds to $\frac{\text{Average Return} - \text{Random Agent}}{\text{Best Average Return} - \text{Random Agent}}$. 1 worker, 80 steps (top-left). 2 workers 40 steps. (top-right). 4 workers, 20 steps (middle-left). 16 workers, 5 steps (middle-right). 40 workers, 2 steps (bottom).}
    \label{fig:momentums-steps-a2c}
\end{figure}

From using multiple workers we posit that there may be a similar implicit momentum as in asynchronous settings~\cite{mitliagkas2016asynchrony}. 
In a full Monte Carlo setting with one synchronous worker, the rollout is biased based on the policy. Any momentum will use prior policies across the range of timesteps. Reducing the number of steps across many workers will likely bias the sampling toward a smaller window of time (smaller number of states). Therefore there may be an implicit momentum based on the smaller window of timesteps such that the prior gradient is biased towards policies updated for smaller windows of states. This effect may go away in settings where there are shortened episodes (possibly due to failure). In such a case, the restarts may cause workers to see wider ranges of timesteps.
Figure~\ref{fig:momentums-steps-ppo} and~\ref{fig:momentums-steps-a2c} show the normalized average return across different worker to n-step ratios. There does appear to be a (noisy) trend on some environments such that lower momentum values perform better at higher worker-to-step ratios. This may imply that there are some notions of implicit momentum happening from using parallel workers (even synchronously) and only in some environments. However, this is with the caveat that this trend is noisy.

\section{Other Possible Unexamined Factors}
\label{app:otherfactors}
There are several other possible affecting factors which we do not discuss in the main text. For example, policy gradient methods essentially scale the gradient by the value function. A larger per-step average reward will yield larger gradients and may further affect the performance of adaptive gradient methods. However, this is not necessarily true of the value function loss. Perhaps different optimizers should be used for the value function and policy loss in such cases. We also do not examine learning rate schedules here. We avoid this for two reasons. First, this adds another layer of hyperparameters to optimize (which we want to avoid for complex settings). Second, in online settings, it is unclear how a schedule would work given that an agent must continuously learn. However, this may be a possible factor to examine in future work.


\begin{thebibliography}{27}
\providecommand{\natexlab}[1]{#1}
\providecommand{\url}[1]{\texttt{#1}}
\expandafter\ifx\csname urlstyle\endcsname\relax
  \providecommand{\doi}[1]{doi: #1}\else
  \providecommand{\doi}{doi: \begingroup \urlstyle{rm}\Url}\fi

\bibitem[Barth-Maron et~al.(2018)Barth-Maron, Hoffman, Budden, Dabney, Horgan,
  Muldal, Heess, and Lillicrap]{barth2018distributed}
Gabriel Barth-Maron, Matthew~W Hoffman, David Budden, Will Dabney, Dan Horgan,
  Alistair Muldal, Nicolas Heess, and Timothy Lillicrap.
\newblock Distributed distributional deterministic policy gradients.
\newblock \emph{arXiv preprint arXiv:1804.08617}, 2018.

\bibitem[Brockman et~al.(2016)Brockman, Cheung, Pettersson, Schneider,
  Schulman, Tang, and Zaremba]{brockman2016openai}
Greg Brockman, Vicki Cheung, Ludwig Pettersson, Jonas Schneider, John Schulman,
  Jie Tang, and Wojciech Zaremba.
\newblock Openai gym.
\newblock \emph{arXiv preprint arXiv:1606.01540}, 2016.

\bibitem[Cohen et~al.(2018)Cohen, Jordan, and Croft]{cohen2018distributed}
Daniel Cohen, Scott~M Jordan, and W~Bruce Croft.
\newblock Distributed evaluations: Ending neural point metrics.
\newblock \emph{arXiv preprint arXiv:1806.03790}, 2018.

\bibitem[Duchi et~al.(2011)Duchi, Hazan, and Singer]{duchi2011adaptive}
John Duchi, Elad Hazan, and Yoram Singer.
\newblock Adaptive subgradient methods for online learning and stochastic
  optimization.
\newblock \emph{Journal of Machine Learning Research}, 12\penalty0
  (Jul):\penalty0 2121--2159, 2011.

\bibitem[Heess et~al.(2017)Heess, Sriram, Lemmon, Merel, Wayne, Tassa, Erez,
  Wang, Eslami, Riedmiller, et~al.]{heess2017emergence}
Nicolas Heess, Srinivasan Sriram, Jay Lemmon, Josh Merel, Greg Wayne, Yuval
  Tassa, Tom Erez, Ziyu Wang, Ali Eslami, Martin Riedmiller, et~al.
\newblock Emergence of locomotion behaviours in rich environments.
\newblock \emph{arXiv preprint arXiv:1707.02286}, 2017.

\bibitem[Henderson et~al.(2017)Henderson, Islam, Bachman, Pineau, Precup, and
  Meger]{henderson2017deep}
Peter Henderson, Riashat Islam, Philip Bachman, Joelle Pineau, Doina Precup,
  and David Meger.
\newblock Deep reinforcement learning that matters.
\newblock \emph{arXiv preprint arXiv:1709.06560}, 2017.

\bibitem[Heusel et~al.(2017)Heusel, Ramsauer, Unterthiner, Nessler, and
  Hochreiter]{heusel2017gans}
Martin Heusel, Hubert Ramsauer, Thomas Unterthiner, Bernhard Nessler, and Sepp
  Hochreiter.
\newblock Gans trained by a two time-scale update rule converge to a local nash
  equilibrium.
\newblock In \emph{Advances in Neural Information Processing Systems}, pages
  6629--6640, 2017.

\bibitem[Hinton et~al.(2012)Hinton, Srivastava, and Swersky]{RMSProp}
Geoffrey Hinton, Nitish Srivastava, and Kevin Swersky.
\newblock Neural networks for machine learning lecture 6a overview of
  mini-batch gradient descent, 2012.

\bibitem[Islam et~al.(2017)Islam, Henderson, Gomrokchi, and
  Precup]{islam2017reproducibility}
Riashat Islam, Peter Henderson, Maziar Gomrokchi, and Doina Precup.
\newblock Reproducibility of benchmarked deep reinforcement learning tasks for
  continuous control.
\newblock \emph{arXiv preprint arXiv:1708.04133}, 2017.

\bibitem[Kingma and Ba(2014)]{kingma2014adam}
Diederik~P Kingma and Jimmy Ba.
\newblock Adam: A method for stochastic optimization.
\newblock \emph{arXiv preprint arXiv:1412.6980}, 2014.

\bibitem[Kostrikov(2018)]{pytorchrl}
Ilya Kostrikov.
\newblock Pytorch implementations of reinforcement learning algorithms.
\newblock \url{https://github.com/ikostrikov/pytorch-a2c-ppo-acktr}, 2018.

\bibitem[Mitliagkas et~al.(2016)Mitliagkas, Zhang, Hadjis, and
  R{\'e}]{mitliagkas2016asynchrony}
Ioannis Mitliagkas, Ce~Zhang, Stefan Hadjis, and Christopher R{\'e}.
\newblock Asynchrony begets momentum, with an application to deep learning.
\newblock In \emph{Communication, Control, and Computing (Allerton), 2016 54th
  Annual Allerton Conference on}, pages 997--1004. IEEE, 2016.

\bibitem[Mnih et~al.(2015)Mnih, Kavukcuoglu, Silver, Rusu, Veness, Bellemare,
  Graves, Riedmiller, Fidjeland, Ostrovski, et~al.]{mnih2015human}
Volodymyr Mnih, Koray Kavukcuoglu, David Silver, Andrei~A Rusu, Joel Veness,
  Marc~G Bellemare, Alex Graves, Martin Riedmiller, Andreas~K Fidjeland, Georg
  Ostrovski, et~al.
\newblock Human-level control through deep reinforcement learning.
\newblock \emph{Nature}, 518\penalty0 (7540):\penalty0 529, 2015.

\bibitem[Mnih et~al.(2016)Mnih, Badia, Mirza, Graves, Lillicrap, Harley,
  Silver, and Kavukcuoglu]{a3c}
Volodymyr Mnih, Adria~Puigdomenech Badia, Mehdi Mirza, Alex Graves, Timothy
  Lillicrap, Tim Harley, David Silver, and Koray Kavukcuoglu.
\newblock Asynchronous methods for deep reinforcement learning.
\newblock In \emph{International Conference on Machine Learning}, pages
  1928--1937, 2016.

\bibitem[Nesterov(1983)]{nesterov1983method}
Yurii Nesterov.
\newblock A method for unconstrained convex minimization problem with the rate
  of convergence o (1/k\^{} 2).
\newblock In \emph{Doklady AN USSR}, volume 269, pages 543--547, 1983.

\bibitem[Polyak and Juditsky(1992)]{polyak1992acceleration}
Boris~T Polyak and Anatoli~B Juditsky.
\newblock Acceleration of stochastic approximation by averaging.
\newblock \emph{SIAM Journal on Control and Optimization}, 30\penalty0
  (4):\penalty0 838--855, 1992.

\bibitem[Reddi et~al.(2018)Reddi, Kale, and Kumar]{amsgrad}
Sashank~J Reddi, Satyen Kale, and Sanjiv Kumar.
\newblock On the convergence of adam and beyond.
\newblock In \emph{International Conference on Learning Representations}, 2018.

\bibitem[Ruder(2016)]{ruder2016overview}
Sebastian Ruder.
\newblock An overview of gradient descent optimization algorithms.
\newblock \emph{arXiv preprint arXiv:1609.04747}, 2016.

\bibitem[Schulman et~al.(2015)Schulman, Moritz, Levine, Jordan, and
  Abbeel]{schulman2015high}
John Schulman, Philipp Moritz, Sergey Levine, Michael Jordan, and Pieter
  Abbeel.
\newblock High-dimensional continuous control using generalized advantage
  estimation.
\newblock \emph{arXiv preprint arXiv:1506.02438}, 2015.

\bibitem[Schulman et~al.(2017)Schulman, Wolski, Dhariwal, Radford, and
  Klimov]{PPO}
John Schulman, Filip Wolski, Prafulla Dhariwal, Alec Radford, and Oleg Klimov.
\newblock Proximal policy optimization algorithms.
\newblock \emph{arXiv preprint arXiv:1707.06347}, 2017.

\bibitem[Sutton(1988)]{sutton1988learning}
Richard~S Sutton.
\newblock Learning to predict by the methods of temporal differences.
\newblock \emph{Machine learning}, 3\penalty0 (1):\penalty0 9--44, 1988.

\bibitem[Sutton and Barto(1998)]{sutton1998reinforcement}
Richard~S Sutton and Andrew~G Barto.
\newblock \emph{Reinforcement learning: An introduction}, volume~1.
\newblock MIT press Cambridge, 1998.

\bibitem[Sutton et~al.(2000)Sutton, McAllester, Singh, and
  Mansour]{sutton2000policy}
Richard~S Sutton, David~A McAllester, Satinder~P Singh, and Yishay Mansour.
\newblock Policy gradient methods for reinforcement learning with function
  approximation.
\newblock In \emph{Advances in neural information processing systems}, pages
  1057--1063, 2000.

\bibitem[Williams(1992)]{williams1992simple}
Ronald~J Williams.
\newblock Simple statistical gradient-following algorithms for connectionist
  reinforcement learning.
\newblock In \emph{Reinforcement Learning}, pages 5--32. Springer, 1992.

\bibitem[Wilson et~al.(2017)Wilson, Roelofs, Stern, Srebro, and
  Recht]{wilson2017marginal}
Ashia~C Wilson, Rebecca Roelofs, Mitchell Stern, Nati Srebro, and Benjamin
  Recht.
\newblock The marginal value of adaptive gradient methods in machine learning.
\newblock In \emph{Advances in Neural Information Processing Systems}, pages
  4151--4161, 2017.

\bibitem[Zeiler(2012)]{zeiler2012adadelta}
Matthew~D Zeiler.
\newblock Adadelta: an adaptive learning rate method.
\newblock \emph{arXiv preprint arXiv:1212.5701}, 2012.

\bibitem[Zhang et~al.(2017)Zhang, Mitliagkas, and R{\'e}]{zhang2017yellowfin}
Jian Zhang, Ioannis Mitliagkas, and Christopher R{\'e}.
\newblock Yellowfin and the art of momentum tuning.
\newblock \emph{arXiv preprint arXiv:1706.03471}, 2017.

\end{thebibliography}


\vskip 0.2in
\end{document}